\documentclass[a4paper,10pt,twoside]{report}

\usepackage{thesis}
\pdfoutput=1
\newcommand{\shortdoctitle}{Master's Thesis}
\newcommand{\doctitle}{Predictability of Machine Learning Algorithms and Related Feature Extraction Techniques}
\newcommand{\docsubtitle}{Master Thesis}

\newcommand{\me}{Yunbo Dong}
\newcommand{\keywords}{Machine Learning, Feature Extraction, Predictability}
\newcommand{\version}{Third Version}
\newcommand{\monthYear}{\today}

\newcommand{\firstCommitteeMember}{Morteza Monemizadeh}
\newcommand{\secondCommitteeMember}{Meng Fang}
\newcommand{\thirdCommitteeMember}{Wouter Meulemans}

\author{\me}

\hypersetup
{
    pdfauthor={\me},
    pdftitle={\shortdoctitle},
    pdfsubject={\doctitle},
    pdfkeywords={\keywords}
}

\begin{document}

%use this include for PDF and distribution versions
\pagenumbering{roman}
\begin{titlepage}
\begin{center}
\includegraphics[height=2cm]{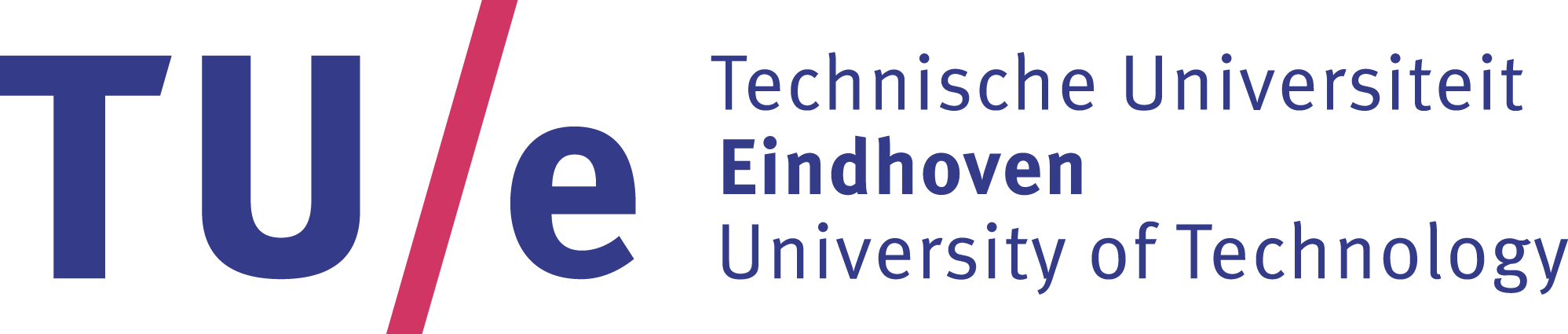}\\
%\LARGE
%Eindhoven University of Technology \\
\large
Department of Mathematics and Computer Science  \\
Algorithms, Geometry, and Applications

\vspace*{10cm}

\setlength{\TPHorizModule}{1mm}
\setlength{\TPVertModule}{\TPHorizModule}
% Set the Paragraph Indent to zero, so the first line is not Indented
% Back-up the current value so it can be put back at the end of the title page
\newlength{\backupparindent}
\setlength{\backupparindent}{\parindent}
\setlength{\parindent}{0mm}			
% Begins a textbox at 72 mm from the left of the edge of the paper and 89 mm from the top
% The width of the textbox is 95 mm (167 - 72 mm)
% The height of the box cannot be defined, so it is your task to keep the text not too long
\begin{textblock}{95}(62,89)
    \vspace*{1mm}
    \huge
    \textbf{\doctitle \\}
    \Large
    \vspace*{5mm}
    \textit{\docsubtitle}\\
    \vspace*{10mm}
    \Large
    \me\\
\end{textblock}

\large
Supervisors:\\
\begin{tabular}{rl}
    \firstCommitteeMember\\
    \secondCommitteeMember\\
    \thirdCommitteeMember\\
\end{tabular}

\vfill
\version

\vfill
%\docdate \\
\large
Eindhoven, \monthYear\\

% Put the Paragraph Indent back to its original value
\setlength{\parindent}{\backupparindent}
\end{center}
\end{titlepage} 

\normalsize

\clearemptydoublepage

%Sometimes line numbers are nice, uncomment the next line to enable:
% \linenumbers

%It could be handy to have a list of todos and brainstorms in your thesis
%\chapter*{*General todos*}\todo{remove this chapter}
%\input{chapters/general_todos}

%\chapter*{*Brainstorm results*}\todo{remove this chapter}
%\input{chapters/brainstorm_results}

\chapter*{Abstract}
\label{chapter:abstract}
To implement machine learning, it is essential to first determine an appropriate algorithm for the dataset. 
Different algorithms may produce a large number of different models with different hyperparameter configurations, and it usually takes a lot of time to run the model on a large dataset when the model is relatively complex. 
Therefore, how to predict the performance of a model on a dataset is an fundamental problem to be solved. 
This thesis designs a prediction system based on matrix factorization to predict the classification accuracy of a specific model on a particular dataset.

In this thesis, we conduct a comprehensive empirical research on more than fifty datasets that 
we collected from the openml web site. 
We study the performance prediciton of three fundamental machine learning algorithms, namely, 
\emph{random forest, XGBoost, and MultiLayer Perceptron(MLP)}. 
In particular, we obtain the following results:

\begin{itemize}
    \item \textbf{Predictability of fine-tuned models using coarse-tuned variants:} 
    
    Usually, training and testing complex machine learning models are time-consuming. 
    Thus, we hope to predict the complicate models by their simple ones. 
    Three machine learning algorithms are compared in experiments.
    We find that random forest and XGBoost have good predictability on most datasets that is, 
    as the model becomes more complex, the performance of the model becomes better, 
    and thus the accuracy of the complex model can be foreseen directly from its simple model. 
    Hence, we can decide efficiently which algorithm to utilize by comparing simple models. 
    
    \item \textbf{Predictability of MLP using feature extraction techniques:} 
    
    Often, real datasets have quite numerous features, from a few hundred to a few thousand features. 
    Training models on fully-featured datasets is a very time-consuming task. 
    We explore the idea of training a model $\mathcal{D}$ on datasets that are projected on a few features and use this as a hint to 
    predict the performance of the model $\mathcal{D}$ when we consider all features of the dataset. 
    % Of course the main issue would be to find a feature extraction method that finds prioritize features. 
    We try different feature extraction techniques including techniques based on permutation importance, 
    gain-based feature importance, hierarchical clustering based on Spearman correlation, and principal component analysis.
    We study the performance of techniques on the multilayer perceptron (MLP) model and observe that feature extraction with 
    permutation importance and hierarchical clustering based on Spearman correlation has better performance. 
    That is, on most datasets, the accuracy of the MLP 
    % that is trained on prefixes on features that are prioritize using these techniques 
    improves as the number of features extracted by these techniques increases.
%     Indeed, we first prioritize features using each feature extraction technique and then train   
%     For random forest and XGBoost, they have good predictability on most datasets that is, as the model becomes more complex, the performance of the model becomes better, and thus the accuracy of the complex model can be foreseen directly from its simple model. 
% With this, it is efficient to decide which algorithm to utilize by comparing simple models. 
% Whereas for MLP this approach cannot be enforced on most of the data sets because it is observed to be less predictable.
% Then, the study is directed with the idea of data dimensionality reduction, hoping to be able to predict the accuracy of an algorithm on the full dataset by its performance on a small dataset. 
% Four different techniques are used,  

    % \item \textbf{Predictability of models using implicit feedbacks:} 
    \item \textbf{Predict model performance using implicit feedback:} 
    
    After researching the predictability of three algorithms, our goal is to discover a method that can be used to predict the specific model performance on a particular dataset.
    In order to predict the classification accuracy of different algorithms on different datasets with different hyperparameters, 
    a prediction system with matrix factorization is built to predict the performance of different models on different datasets. 
    With this system, the input accuracy's can be seen as implicit feedback because there is no more information about the cause of these performance.
    This system can best achieve an mean absolute error of only $6.7\%$ in the experiment.
\end{itemize}

% Four different techniques are used, including technology based on permutation importance, gain-based feature importance, hierarchical clustering with Spearman correlations, and principal component analysis. 
% It was eventually found that feature extraction with permutation importance and hierarchical clustering have better predictability, namely on most datasets, the accuracy of the same model improves as the number of features extracted by the technique increases.
% Finally, in order to predict the classification accuracy of different algorithms on different datasets with different hyperparameters, this prediction system with matrix factorization is built to predict the performance of different models on different datasets. 
% This system can best achieve an mean absolute error of only $6.7\%$ in the experiment.

% \clearemptydoublepage
\newpage

%An executive summary if you want:
%\chapter*{Executive summary}\label{chapter:executive_summary}
%\input{chapters/executive_summary}

%\clearemptydoublepage

\chapter*{Acknowledgement}
\label{chapter:preface}
% Before you start to read this dissertation \textit{Predictability of Machine Learning Algorithms and Related Feature Extraction Techniques}, I hope to introduce the background briefly and make an acknowledgement. 
% It has been written to fulfill the graduation requirements of the Master for Data Science in Engineering at Eindhoven University of Technology(TU/e). 
% I was engaged in researching and writing this dissertation from April to October 2021.

% The idea of this project was originally come from my supervisor Morteza Monemizadeh. 
% The project firstly aimed to discover the pattern of model variation so that solve a practical problem that is to predict the model performance. 
% It became a research about predictability after continuous improvement.
% The study was difficult but conducting extensive empirical investigation allowed me to make contribution to give my answer to the practical questions. 
% And fortunately, Morteza is always available and willing to answer my queries and give me detailed instructions.

I would like to thank my supervisor Morteza for his excellent guidance and support during this process. 
Thanks to my friends and families as well. 
You kept me motivated. 
And My parents deserve a particular note of appreciation for always encouraging me. 
Last but not the least, I am grateful to professor Meng Fang and Wouter Meulemans to be the committee members for me. 
I hope you enjoy your reading.

\clearemptydoublepage

\tableofcontents

\clearemptydoublepage

% \listoffigures

% \clearemptydoublepage

% \listoftables

% \clearemptydoublepage

% \lstlistoflistings

% \clearemptydoublepage

\chapter{Introduction}
\label{chap:intro}
\setcounter{page}{0}
\pagenumbering{arabic}
%from here on, start the 'real' page numbering, from 1, with normal digits
Machine learning (ML) is one of the major fields in computer science that studies self-improving algorithms with the use of data~\cite{mitchell1997machine}. 
Machine learning is everywhere to avoid explicit programming~\cite{gero2012artificial}. 
Engineers use machine learning and related techniques in learning normal and abnormal behavior of social networks~\cite{benchettara2010supervised}, computer vision~\cite{forsyth2011computer}, speech recognition~\cite{yu2016automatic}, medicine~\cite{deo2015machine}, and etc., where conventional algorithms are difficult or impossible to develop to achieve the needed goals~\cite{hu2020voronoi}.
The focus of this thesis is on classification algorithms, where a labeled dataset is given and the goal is to train a machine learning model with it such that new data without labels can be classified. 
The implicit assumption here is if the distribution of the labeled data that is given to the machine model is a good approximation or proxy of the distribution of the unseen data, we may feed the labeled data to the model that in turn starts training itself to improve its performance. 
 
Concentrating on predicting by computers, the most area of machine learning(ML) is closely related to computational statistics, even if not all machine learning(ML) is statistical learning~\cite{el2015machine}. 
It involves computers learning from data provided so that they can carry out certain tasks. 
For simple assignments, it is possible for a human to manually create the program required to solve the problem, which however can be challenging for advanced tasks to instruct the machine how to execute all needed steps explicitly~\cite{alpaydin2020introduction}. 

Hence, in most cases of machine learning(ML), one approach is to label some of the correct answers and then use them as training data for the computer to improve the algorithms determining correct answers, which is known as supervised learning (SL)~\cite{alpaydin2020introduction}. 
In other words, supervised learning is based on example input-output pairs to infer a function that can be used for mapping new inputs~\cite{russell2002artificial}. 
There are tremendous number of supervised machine learning(ML) models used in the experiments of the thesis, including the primary linear learner/linear model~\cite{zhang2001text}, multi-layer perceptron (MLP)~\cite{schmidhuber2015deep}, random forest~\cite{quinlan1986induction}, and another seminal model XGBoost~\cite{chen2016xgboost}, all of which are introduced in chapter \ref{chap:method_model}.

With ML models at hand, there are several potential questions raised by practitioners and engineers:
\begin{enumerate}[\textbf{Question} 1]
    \item \label{que:best_model} Which one of the known learning models performs the best? 
    Though with respect to the accuracy, loss, or other learning measurements, one or more models may have the highest value, it is the best model that only needs to meet the expected criteria, like accuracy, and trains with less time. 
    For example, a random forest with some particular parameters performs best on a specific dataset with $0.80$ accuracy and $0.2$ seconds training time though an MLP with some particular parameters hits $0.99$ accuracy but 2 hours training time on the same dataset.
    \item \label{que:predict_from_simple} Is it possible to predict the performance of advanced ensemble or boosted models using simple learners? 
    Since learning tasks are often time-consuming~\cite{langley1995applications}, starting with simple learners or shallow networks for a few rounds is a good choice if the performance of the advanced counterpart can be predicted. 
    For example, if the prediction shows that the complex networks cannot improve the accuracy prominently comparing with the simple ones, then there is no need to train complex ones.
    \item[]
    \item \label{que:extract_fea} How to efficiently extract important features of a dataset? 
    Often datasets have quite a lot features, thousands or even millions of features, thus training the machine learning models 
    using all the features of a dataset might be very time-consuming. 
    An interesting research direction is to employ efficient feature extraction algorithms~\cite{sarangi2020optimization}
    that can find the features that are critical for learning and discovering the concealed correlations between features. 
    In this way, we might be able to train the models using the most important features of the dataset and 
    that could potentially speed up the process of learning. 
    \item \label{que:predict_perfomance} Is it possible to predict the performance of a specific model on a specific dataset with some known results? 
    For instance, suppose that there are reported accuracies for a few known machine learning models for the dataset 
    that we would like to analyze. 
    Then, for a new model, we can save time if it is possible to approximately predict the accuracy of the dataset because the model will not be required training on the dataset if we ensure that the predicted performance of the new model cannot meet the expected accuracy that we look for. 
    Similarly, is the performance of a model trained for different datasets predictable upon a new dataset?
\end{enumerate}
To answer these questions, a platform for empirical experiments is developed to study the popular machine learning algorithms including linear model, multi-layer perceptron (MLP), random forest, and XGBoost with various datasets. 
To this end, more than 50 datasets are collected, which have different numbers of samples, features, and target labels and are trained with the four popular classification algorithms that have been repeatedly reported to achieve good performance. 

\section{Datasets}

54 datasets from openml web site are collected~\cite{OpenML2013}. 
The number of instances in these datasets are from a few hundred to hundred thousands. 
The number of features are from 3 features to a couple hundred features among them the one with the most number of features is Fashion-MNIST dataset whose features are black and white images consisting of $28 \times 28$ pixels, that are 784 features. 

Among all these datasets, 32 of datasets are binary classification tasks, 8 of them have 3 labels and the other 14 datasets are multi-class classification tasks with more than 3 classifications. 
The categorical features of the datasets are all one-hot encoded. 
The complete information about these datasets are given in the table \ref{tab:binary_datasets} and \ref{tab:multi_datasets}, divided based on the number of labels. 

In two tables, the id of a dataset is the data id used in OpenML for download; the name is the data name annotated in OpenML~\cite{OpenML2013}; 
the size is the number of samples in the dataset; 
features and labels are the number of features and labels of corresponding datasets respectively. 
The table is first arranged in ascending order by the number of labels, then by the number of features if the labels are the same, and finally by the size. 
In the table \ref{tab:binary_datasets}, the division is according to datasets with less than 40 features and more than 100 features. 
In the table \ref{tab:multi_datasets}, the division is according to datasets with no more than 3 labels, more than 3 labels but less than 10 labels, and more than 10 labels. 

\begin{table}[htbp]
 \centering
 \caption{The attributes of datasets with binary classification}
    \begin{tabular}{l l l l l}
    \toprule
    ID    & Name  & Size  & Features & Labels \\
    \hline \hline
    41025 & random-teste & 29    & 3     & 2 \\
    4534~\cite{Dua:2019}  & PhishingWebsites & 11055 & 30    & 2 \\
    4154  & CreditCardSubset & 14240 & 30    & 2 \\
    833   & bank32nh & 8192  & 32    & 2 \\
    1452  & PieChart2 & 745   & 36    & 2 \\
    1069~\cite{shirabad2005promise}  & pc2   & 5589  & 36    & 2 \\
    1443  & PizzaCutter1 & 661   & 37    & 2 \\
    1451  & PieChart1 & 705   & 37    & 2 \\
    1444  & PizzaCutter3 & 1043  & 37    & 2 \\
    1453  & PieChart3 & 1077  & 37    & 2 \\
    1049~\cite{shirabad2005promise}  & pc4   & 1458  & 37    & 2 \\
    1050~\cite{shirabad2005promise}  & pc3   & 1563  & 37    & 2 \\
    1056~\cite{shirabad2005promise}  & mc1   & 9466  & 38    & 2 \\
    \hline
    979   & waveform-5000 & 5000  & 40    & 2 \\
    734   & ailerons & 13750 & 40    & 2 \\
    1494~\cite{mansouri2013quantitative}  & qsar-biodeg & 1055  & 41    & 2 \\
    40705 & tokyo1 & 959   & 44    & 2 \\
    40999~\cite{van2014endgame} & jungle\_chess\_2pcs\_endgame\_elephant\_elephant & 2351  & 46    & 2 \\
    41007~\cite{van2014endgame} & jungle\_chess\_2pcs\_endgame\_lion\_lion & 2352  & 46    & 2 \\
    41005~\cite{van2014endgame} & jungle\_chess\_2pcs\_endgame\_rat\_rat & 3660  & 46    & 2 \\
    904   & fri\_c0\_1000\_50 & 1000  & 50    & 2 \\
    44~\cite{Dua:2019}    & spambase & 4601  & 57    & 2 \\
    1487~\cite{zhang2008forecasting}  & ozone-level-8hr & 2534  & 72    & 2 \\
    \hline
    718   & fri\_c4\_1000\_100 & 1000  & 100   & 2 \\
    1479~\cite{Dua:2019}  & hill-valley & 1212  & 100   & 2 \\
    316~\cite{dietterich2002advances}   & yeast\_ml8 & 2417  & 116   & 2 \\
    41048 & ex1-features-holiday-person-es-201610 & 1809  & 132   & 2 \\
    41049 & ex1-features-holiday-person-en-201610 & 2350  & 132   & 2 \\
    41050 & ex2-features-holiday-person-event-es-201610 & 1809  & 144   & 2 \\
    41051 & ex2-features-holiday-person-event-en-201610 & 2350  & 144   & 2 \\
    41052 & ex3-features-holiday-all-es-201610 & 1809  & 299   & 2 \\
    41053 & ex3-features-holiday-all-en-201610 & 2350  & 299   & 2 \\
    \toprule
    \end{tabular}%
 \label{tab:binary_datasets}%
\end{table}%

\begin{table}[htbp]
 \caption{The attributes of datasets with multi classification}
 \centering
    \begin{tabular}{l l l l l}
    \toprule
    ID    & Name  & Size  & Features & Labels \\
    \hline \hline
    41027~\cite{van2014endgame} & jungle\_chess\_2pcs\_raw\_endgame\_complete & 44819 & 6     & 3 \\
    60~\cite{Dua:2019}    & waveform-5000 & 5000  & 40    & 3 \\
    40668~\cite{Dua:2019} & connect-4 & 67557 & 42    & 3 \\
    41004~\cite{van2014endgame} & jungle\_chess\_2pcs\_endgame\_lion\_elephant & 4704  & 46    & 3 \\
    40997~\cite{van2014endgame} & jungle\_chess\_2pcs\_endgame\_panther\_lion & 4704  & 46    & 3 \\
    41000~\cite{van2014endgame} & jungle\_chess\_2pcs\_endgame\_panther\_elephant & 4704  & 46    & 3 \\
    1548  & autoUniv-au4-2500 & 2500  & 100   & 3 \\
    40670~\cite{Dua:2019} & dna   & 3186  & 180   & 3 \\
    \hline
    4538~\cite{Dua:2019, madeo2013gesture, wagner2014gesture, wagner2013segmentacc}  & GesturePhaseSegmentationProcessed & 9873  & 32    & 5 \\
    182~\cite{Dua:2019}   & satimage & 6430  & 36    & 6 \\
    180   & covertype & 110393 & 54    & 7 \\
    1549  & autoUniv-au6-750 & 750   & 40    & 8 \\
    1555  & autoUniv-au6-1000 & 1000  & 40    & 8 \\
    \hline
    22~\cite{Dua:2019}    & mfeat-zernike & 2000  & 47    & 10 \\
    14~\cite{Dua:2019}    & mfeat-fourier & 2000  & 76    & 10 \\
    1501~\cite{Dua:2019}  & semeion & 1593  & 256   & 10 \\
    40996~\cite{xiao2017fashion} & Fashion-MNIST & 70000 & 784   & 10 \\
    \hline
    313   & spectrometer & 531   & 101   & 44 \\
    41014 & qType\_2prev & 188290 & 4     & 55 \\
    1493~\cite{mallah2013plant}  & one-hundred-plants-texture & 1599  & 64    & 100 \\
    1492~\cite{mallah2013plant}  & one-hundred-plants-shape & 1600  & 64    & 100 \\
    1491~\cite{mallah2013plant}  & one-hundred-plants-margin & 1600  & 64    & 100 \\
    \toprule
    \end{tabular}%
 \label{tab:multi_datasets}%
\end{table}%

\section{Contribution}

With an extensive empirical study, a comprehensive result of their performance is given, such as when these models perform well and what is the suitable feature extraction technique for these models. 
The work sheds light upon the performance and applicability of these algorithms for various datasets. 
In particular, the experiments reveal:
\begin{enumerate}[(1)]
    \item \label{finding:mlp_best11} When accuracy is taken as the standard of performance evaluation, multi-layer perceptron (MLP) and XGboost outperforms random forest for majority of datasets in experiments.
    \item \label{finding:rf_fastest}The time consumption of random forest is significantly less than MLP and XGboost though its performance is often worse than XGboost and MLP.
    \item \label{finding:shallow_network} With deep networks of MLP, most classification tasks can be achieved decently. However, for a large number of datasets, shallow networks can accomplished the accuracy that is very close to the highest known accuracy as well. As an example, a network with only 1 layer but 512 nodes can achieve the best performance for the Fashion MNIST dataset.
    \item \label{finding:pattern} For most models of random forest and XGBoost, there are obviously increasing patterns of accuracy with the increase of the number of trees and the max depth of each tree respectively, whereas such approximately monotonic patterns are not frequently seen in the model of the MLP. That is no matter increasing the number of layers or nodes, the accuracy of the MLP model cannot be always guaranteed to improve.
    \item \label{finding:feature_extract} The features of data can be sorted by random forest and XGBoost because the feature importance can be given by the models, which can be utilized for the function of feature extraction and/or dimension reduction. In the perspective of predictability, with the performance of principal component analysis (PCA) as the baseline, the performance of permutation feature importance in random forest is more predictable than PCA, worse than which gain-based feature importance of XGBoost performs. Moreover, hierarchical clustering of the features based on Spearman correlations has the best result.
\end{enumerate}

Beyond these observations about the basis of experiments, there are further innovations to predict the accuracy of these algorithms. 
Especially, according to the finding (\ref{finding:pattern}), it is difficult to predict the accuracy of MLP models because of the lack of patterns within the algorithm, which emphasizes the value of performance prediction. 
What is accomplished is that, in accordance with the idea of matrix factorization, a complete accuracy matrix can arise from a primal matrix with a number of entries left to be predicted, whose rows are specific models, columns are different datasets. 
Each entry $(i,\ j)$ in the matrix corresponds to the accuracy of the model $i$ for the dataset $j$. 
Hence, if the accuracy of the model $i$ for the dataset $j$ is unknown, the entry $(i,\ j)$ will be missing. 
With the final complete matrix, all missing entries, namely unknown accuracy, are predicted. 
The forecast results are so low that they are acceptable, which are measured by root mean square error (RMSE) between all predictions and actual values.

Overall, in this thesis, through a series of experiments, not only the predictability of several traditional machine learning algorithms and related feature extraction techniques are investigated, but a specific prediction method for the accuracy of the algorithms is also proposed. 
Chapter \ref{chap:previous} introduces the previous related work to show how research has evolved and what is the innovation in the paper. 
It is in chapter \ref{chap:method_model} that the detail of machine learning models, feature extraction techniques, and matrix factorization is illustrated. 
The specific experimental results and the intuitive explanation are shown in chapter \ref{chap:result}. chapter \ref{chap:discussion} \nameref{chap:discussion} includes further analysis of results, limitations and the direction of future work. 
Finally, chapter \ref{chap:conclu} \nameref{chap:conclu} concludes all of achievements and answers questions.

\newpage

% \chapter{Preliminaries}
% \label{chapter:preliminaries}
% \input{chapters/preliminaries}

% \clearemptydoublepage

\chapter{Previous Work}
\label{chap:previous}
% this chapter works as preliminaries
In the area of machine learning, there were extensive works comparing neural networks with decision trees in specific application domains. 
E.Brown, Corruble, and Pittard~\cite{brown1993comparison} compared neural networks using backpropagation with decision trees to study emitter classification and digit recognition, which were multi-modal problems. 
Chang~\cite{chang2011comparative} made a comparison between models produced by different methods, such as neural networks(NN), decision trees, and the hybrid method, for the forecast of stock price, which were applied on 10 different stocks in 320 data sets empirically, and found that NN achieved a more reliable result. 
Khemphila and Boonjing~\cite{khemphila2010comparing} studied the difference between classification techniques, including logistic regression, decision tree, and neural network (NN), in the perspective of predicting the patients with heart disease, which also draws the conclusion that NN was the best technique. 

In addition to these studies on data sets in the specific area, there were researches on some general or even randomly generated datasets, which could draw some more general conclusions. 
Kim ~\cite{kim2008comparison}compared output of different techniques like NN, decision trees and linear regression on different dataset with varying independent variables and sample size, 
during which he found that linear regression had less error on pure continuous independent variables no matter how many samples while NN improved with increasing the number of categorical variables and had the lowest error when there were no less than 2 categorical variables. 
Dr. Berkman Sahiner et al.~\cite{sahiner2005comparison} compared the performance of decision tree with linear discriminant analysis and NN under different conditions influenced by training sample size, class distributions, and feature space dimension, in which a bagging method was investigated to improve the accuracy of decision tree as well.

Furthermore, some researchers broadened experiments into large-scale empirical studies which were conducted to draw more general results. 
To present a benchmark, Balaji and Allen~\cite{balaji2018benchmarking} assessed on the most mature open source solutions including dense random forests, generalized linear models, and basic deep learning models in auto ML systems, for 57 classification tasks and 30 regression tasks of diverse datasets. 
Fern{\'a}ndez-Delgado, Cernadas, Barro, and Amorim~\cite{fernandez2014we} evaluated 179 classifiers including neural networks, support vector machines(SVM), random forests, and other methods over 121 datasets to achieve significant conclusions about the classifier behavior that random forest was likely to be the best among all classifiers and SVM with Gaussian kernel was the second best with an insignificant difference. 
Leite, Brazdil, and Vanschoren~\cite{leite2012selecting} proposed a technique to select finite cross-validation tests for the best algorithm on a specific dataset and evaluated through 292 classification algorithm-parameter combinations on 76 datasets, which showed that the model output by the technique was very close to the optimum. 

Some of these studies, especially comparing within large-scale experiments, suggested that tree ensembles were the best method to use in many classification tasks. 
However, neural network models in most of these investigations were considered without recent advances in training and architectural decisions, such as ReLU activation, Adam optimization, batch normalization, dropout regularization, and etc. 
Besides, XGBoost was supposed to show superiority as well according to a number of machine learning and data mining challenges. 
For example, among the 29 published top solutions in challenges in 2015 hosted by the Kaggle, a machine learning competition website, there 17 solution using XGBoost~\cite{chen2016xgboost}. 
Hence, the achievable performance among recently popular and advanced algorithms may be understated and the relative performance of random forest was overestimated.

Moreover, these studies mainly focused on the best performance, which was only researched in a static idea namely lack of hyperparameter tuning and tendency prediction. 
Filling this gap, Rijn and Hutter~\cite{van2018hyperparameter} experimented on hyperparameter tuning with 100 datasets to propose a method for selecting hyperparameters. 
Unfortunately, the chosen algorithms only included random forest, support vector machine(SVM), and Adaboost, namely lacked recently advanced algorithms like MLP and XGBoost. 
In addition, like other studies through static thinking, the work of Rijn et al. only expounded the importance of some hyperparameters by the correlation between them and model performance but lacked the prediction of particular model performance with changing hyperparameters.

There were explorations for further prediction of model performance. 
Denil et al.~\cite{denil2013predicting} demonstrated several different architectures to predict model parameters/weights trained by a number of weights and were able to predict more than $95\%$ of the weights keeping most performance. 
They kept the performance unchanged as much as possible and predicted the weights in models during experiments, which is followed some flaws. 
On the one hand, there was a lack of research on the changes of models performance. 
As mentioned in question \ref{que:best_model} of section \nameref{chap:intro}, what is necessary most of the time for accuracy is to reach an acceptable value. 
On the other hand, engineers pay more attention to hyperparameters than to specific weights inside models. 
In addition, their experiment scope is relatively small, only for 3 kinds of NNs, including MLP, convolutional network, and independent component analysis on 5 data sets.

Beyond these papers, Hestness et al.~\cite{hestness2017deep} empirically presented characterization of generalization error and model size growth with increasing training sets and introduced a methodology to measure, through which power-law generalization error scaling was discovered across various factors. 
But what the prediction gave is only the scaling of accuracy caused by model size or data size because they were only state-of-the-art architectures at that time that they used to experiment in different domains separately. 
Based on the same reason, their research also lacked hyperparameters tuning.

Considering the results of these prior works, this paper integrates advantages and conducts large-scale and general research on different kinds of algorithms from a new perspective that is to examine the predictability of machine learning algorithms and construct a framework to predict the performance of algorithms on different datasets, which can also give the algorithms suitable for a specific dataset, and the related hyperparameter tuning. 
Meanwhile, experiments also regard the scaling the data but in a different way from the work of Hestness et al.~\cite{hestness2017deep} that related feature extraction techniques of different algorithms are used and compared.

\newpage
\chapter{Models and Methodology}
\label{chap:method_model}
In this section, firstly, four models selected for experiments are introduced in detail, namely linear classifier, multilayer perceptron (MLP), random forest and XGBoost. 
Afterwards, feature extraction methods used with different algorithms are explained including permutation feature importance measured by random forest, 
gain-based feature importance measured by XGBoost, 
hierarchical clustering based on spearman correlations, 
and principal component analysis (PCA). 
Finally, the matrix factorization is covered, which is the tool used to predict the performance of models on different datasets.

\section{Linear Classifier}
It is the goal of classification to use the features of an object to identify its category. 
Linear classifiers have achieved considerable success in the field of classification though it is one of the simplest models in machine learning~\cite{yang1994example, zhang2001text, anzai2012pattern, hart2000pattern, yuan2012recent}. 
A linear classifier classifies data into labels based on a linear combination of input features~\cite{joachims1998text}.

Linear classifiers work well for practical problems such as document classification, and even for problems with many variables/features, they can result accuracy comparable to non-linear classifiers, however with less time consumption to train and use~\cite{yuan2012recent}. 
Hence, a linear classifier is often used in situations where the speed of classification is an issue, especially when input is sparse or the number of its dimensions is large. 

In experiments, linear model from TensorFlow library is used~\cite{tensorflow2015-whitepaper}. The model approximates the function in which there is the output $y$ based on $N$ input features $x_i$ : $$ y = \beta + \sum_{i=1}^{N}w_i * x_i$$ where $\beta$ is the bias and $w_i$ is the weight for each feature. 

\begin{figure}[ht]
    \centering
    \includegraphics[width=0.4\linewidth]{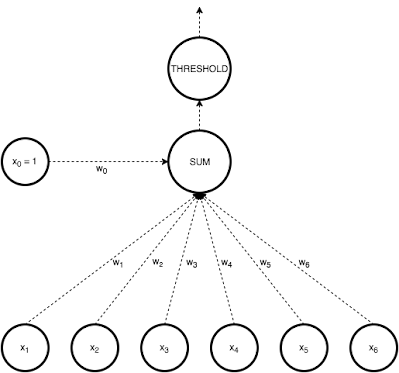}
    \caption{Schematic of Perceptron}
    \label{fig:linear_classifier}
\end{figure}

It can be seen as a single-layer perceptron, of which functioning is similar to the functioning of a single neuron in our brain~\cite{pal1992multilayer}. 
Figure \ref{fig:linear_classifier} is a schematic for the perceptron. 
It takes weighted linear combination of input features and passes it through a thresholding or activating function, which usually outputs 1 or 0 for binary classification. 

As shown in figure \ref{fig:linear_classifier}, for binary classification, the activation function is: $$ output = \left\{
\begin{aligned}
1 & , & y \geq 0 \\
0 & , & y < 0
\end{aligned}
\right.
$$ 
Since then, the problem is to find a set of weights fitting the most of training data into their correct labels, which is an optimization problem. 

Perceptron, the algorithm implemented as a linear classifier, was invented in 1958 by Frank Rosenblatt~\cite{rosenblatt1957perceptron}. 
There were many years for researching and improving before it became the basis of the neural network, especially of Multilayer Perceptron (MLP). 
It is recognized that feedforward neural networks with two or more layers, namely multilayer perceptrons, have greater processing power~\cite{mohri2013perceptron}.

%----------------------------------------------------------------------------
%----------------------------------------------------------------------------
%----------------------------------------------------------------------------
\section{Multilayer Perceptron}

\begin{figure}[ht]
    \centering
    \includegraphics[width=0.4\linewidth]{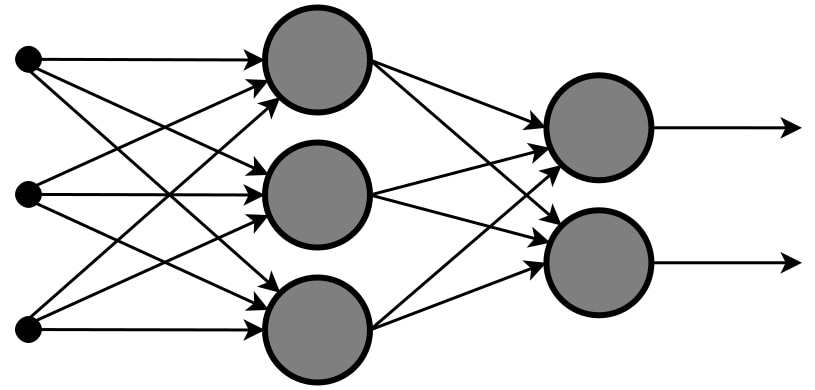}
    \caption{Schematic of MultiLayer Perceptron}
    \label{fig:mlp}
\end{figure}

A multilayer perceptron (MLP) is a class of feedforward artificial neural network(ANN), which is the first and simplest type of artificial neural network devised~\cite{schmidhuber2015deep}. 
In this paper, MLP is used strictly to refer to networks composed of multiple layers of perceptrons with threshold activation. 
An MLP consists of at least three layers of nodes: an input layer, at least one hidden layer and an output layer. 
And these layers are interconnected in a feed-forward way, that is, of each layers, nodes are connected to all outputs of the previous layer and all inputs for nodes in the next layer. 
This means that all hidden layers implemented are dense layers, also called fully-connected layers. 
Figure \ref{fig:mlp} shows a simple sample model structure with one input layer, one hidden layer and one output layer.

MLP uses back-propagation as the learning technique, where the output is compared with the correct answer to gain the error based on a predefined function. 
With the error feeding back, the weights of each connection will be updated to reduce the error. 
This process can be repeated until the training of MLP is finished, which usually means that the model converges to some state. 

The activation function of a single perceptron, namely the linear classifier, is usually discontinuous due to its thresholding function. 
However, according to the learning technique, it should be necessary to use a continuous differentiable activation function. 
Then, Rectified Linear Unit (RELU) is used as activation function for each layer, except the output layer, for which softmax or sigmoid is used. 
RELU is defined as the non-negative part of its argument: $$f(x) = max(0, x)$$ 
\\
It works empirically well and is very popular because it has strong biological motivations and mathematical justifications and enables better training of deeper networks~\cite{hahnloser2000digital, hahnloser2003permitted, glorot2011deep, lecun2012efficient} though it is not mathematical differentiable at $0$. 
In experiments, MLP is implemented by building sequential model of TensorFlow library~\cite{tensorflow2015-whitepaper}. 
In the library, to solve the problem of indifferentiable activation function RELU, the derivative when $x = 0$ is set to 0 as default. 

As for the output layer of MLP, the activation function is set to softmax function for multi-classification or sigmoid function for binary classification. 
The softmax function is used to normalize input into a probability distribution, that is a function $\sigma: \mathbb{R}^K \to [0, 1]^K$ defined by formula: $$\sigma(z)_i = \frac{e^{z_i}}{\sum_{j = 1}^{K}e^{z_j}}$$ for $i = 1, 2, ..., K$ and $z = (z_1, z_2, ..., z_K) \in \mathbb{R}^K$. 
Through this normalization, the sum of the output vectors $\sigma(z)_i$ are ensured to be 1. 
Therefore, the output can be seen as possibilities of different classes, according to which the classification can be decided. 
Another is sigmoid function, which has a characteristic "S"-shaped curve, used for binary classification. 
It is defined as by: $$S(x) = \frac{1}{1+e^{-x}} = \frac{e^{x}}{1+e^{x}}$$ 
Due to its characteristic "S"-shaped curve, it can flatten low and high values of inputs, which suits many natural processes that exhibit a progression from small beginning to a climax over time but may cause the gradient vanishing problem if it is used in hidden layers. 
The problem will prevent the weight from changing its value and make model training very slow~\cite{hochreiter1991untersuchungen, hochreiter2001gradient}.

%----------------------------------------------------------------------------
%----------------------------------------------------------------------------
%----------------------------------------------------------------------------
\section{Random Forest}
\label{subsec:model_rf}
To solve regression and classification problems, random forest model is a frequently-used machine learning technique. 
A technique named ensemble learning is utilized in the model, which combines many classifiers to provide solutions for complex problems~\cite{quinlan1986induction}. 
Hence, a random forest algorithm consists of many decision trees. 

As an ensemble learning method for classification, regression and other tasks, random forest operates by constructing a multitude of decision trees at training time. 
Namely, the ‘forest’ generated by the random forest algorithm is trained through bagging or bootstrap aggregating~\cite{quinlan1990decision}. 
Bagging is an ensemble meta-algorithm that improves the accuracy of machine learning algorithms~\cite{breiman1996bagging}. 
The algorithm establishes the outcome based on the predictions of the decision trees. 
By taking the average or mean, it outputs the prediction from various trees.

For classification tasks, the class that is selected by most trees becomes the output of the random forest~\cite{ho1998random}. 
For regression tasks, the model returns the mean or average prediction of all individual trees~\cite{ho1995random}. 
Random decision forests correct for decision trees' habit of overfitting to their training set~\cite{friedman2001elements}.

The general technique of bootstrap aggregating or bagging is applied to tree learners in the training algorithm for random forests. 
Given a training set $X = x_1, x_2, ..., x_n$, with responses $Y = y_1, y_2, ..., y_n$, bagging repeatedly selects a random sample with replacement of the training set $B$ times and fits trees to these samples: 
for $b = 1, 2, .., B$
\begin{enumerate}
    \item Sample, with replacement, $n$ training examples from $X$, $Y$, call these $X_b$, $Y_b$
    \item Train a classification or regression tree $f_b$ on $X_b$, $Y_b$.
\end{enumerate}
 
\begin{figure}[ht]
    \centering
    \includegraphics[width=0.6\linewidth]{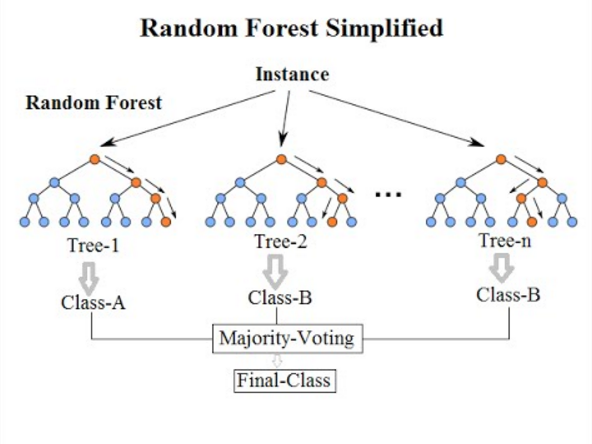}
    \caption{Schematic of Random Forest~\cite{RandomForestTemplate}}
    \label{fig:rf}
\end{figure}

After training, predictions for unseen samples $x'$ can be made by averaging the predictions from all the individual regression trees on $x'$: 
$$\hat{f} = \frac{1}{B} \sum_{b=1}^{B}f_b(x')$$ 
or by taking the majority vote in the case of classification trees, as shown in figure \ref{fig:rf}~\cite{RandomForestTemplate}.

With this bootstrapping procedure, there could be better model performance because it decreases the variance of the model, even though without increasing the bias. 
This means that, as long as the trees are not correlated, the average of many trees is not sensitive to noise in its training set while the predictions of a single tree are highly sensitive. 
To avoid strongly correlated trees, which are given by simply training many trees on a single training set or even training the same tree many times with the deterministic training algorithm, it is necessary to use bootstrap sampling to decorrelate the trees by showing them different training sets.

For trees, the original bagging algorithm can be described by above procedure. 
There is another type of bagging scheme for random forests as well: 
there can be an improved tree learning algorithm that selects a random subset of features at each candidate split in the learning process. 
This process is sometimes referred to as "feature bagging". 
The reason for this is the correlation of the tree in the common bootstrap sample: 
if one or several features are very strong predictors of the response variable, namely target output, the algorithm will select these features among many B-trees, making them relevant.

%----------------------------------------------------------------------------
%----------------------------------------------------------------------------
%----------------------------------------------------------------------------
\section{Gradient Boosted Decision Trees}
\label{subsec:model_xgb}
Gradient boosting is a machine learning technique for tasks such as classification and regression, which produces a prediction model in the form of an ensemble of weak prediction models, typically decision trees~\cite{piryonesi2020data}. 
The algorithm is called gradient boosted trees when a decision tree is the weak learner, of which the performance is usually superior to random forest~\cite{hastie2009boosting}. 
The model is built by the algorithm in a stage-wise fashion like other boosting methods do, and such models can be generalized by allowing optimizing an arbitrary differentiable loss function~\cite{madeh2021using}.

Like other boosting methods, weak "learners" are combined iteratively into a single strong learner in the gradient boosting algorithm~\cite{li2016gentle}. 
In the least-squares regression setting, the goal is to "teach" a model $F$ to predict values of the form $$\hat{y} = F(x)$$ by minimizing the mean squared error 
$$\frac{1}{n}\sum_{i}(\hat{y_i}-y_i)^2$$ 
where $i$ indexes over some training set of size $n$ of actual values of the output variable $y$, $\hat{y_i}$ is the predicted value of $F(x)$, $y_i$ is the observed value and $n$ is the number of samples in $y$.

For a gradient boosting algorithm with $M$ stages, at each stage $m$ ($1\leq m\leq M$) of gradient boosting, suppose a imperfect model $F_m$. 
In order to improve $F_m$, the algorithm should add some new estimator $h(m)$ such that 
$$F_{m+1}(x) = F_m(x) + h_m(x) = y$$ 
namely, 
$$h_m(x) = y-F_m(x)$$ 
Then, with gradient boosting, $h$ will be fitted to this residual.

In the field of learning to rank, gradient boosting is very popular~\cite{cossock2008statistical}. 
It can be also utilized for data analysis in High Energy Physics~\cite{lalchand2020extracting}. 
However, it somehow sacrifices intelligibility and interpretability while boosting can increase the accuracy of a base learner, such as a decision tree~\cite{wu2008top}. 
Furthermore, due to the higher computational demand, it may be more difficult to implement the algorithm.

\subsection{XGBoost}

Among all variants of gradient boosting tree algorithm, XGBoost is a scalable end-to-end tree boosting system, which data scientists is using widely on many machine learning challenges to achieve state-of-the-art results~\cite{chen2016xgboost}. 
In the XGBoost model, a given dataset with $n$ samples and $m$ features is defined as 
$$\mathcal{D} = \{ (x_i, y_i)\}$$ 
for which $|\mathcal{D}| = n$, $x_i\in \mathbb{R}^m$ and $y_i\in \mathbb{R}$. 
Then, with $K$ additive functions, a tree ensemble model is used to predict the output 
$$\hat{y}_i =\phi(x_i) = \sum_{k=1}^K f(x_i), f_k\in \mathcal{F}$$ 
where $\mathcal{F} = \{ f(x) = w_{q(x)}|\ q:\mathcal{R}^m\to T,\ w\in \mathcal{R}^T\}$ is the space of trees. 
The structure of each tree is represented by $q$, which maps  an example to the corresponding leaf index. $T$ is the number of leaves in the tree. 
Each $f_k$ corresponds to an independent tree structure $q$ and leaf weights $w$. 
To learn the set of functions used in the model, the regularized objective function can be minimized
$$\mathcal{L}(\phi) = \sum_i l(y_i,\ \hat{y}_i) + \sum_k\Omega(f_k)$$ 
where $\Omega(f_k) = \gamma T + \frac{1}{2}\lambda \| w\|^2$. 
A differentiable convex loss function is represented by $l$ that measures the difference between the target $y_i$ and the prediction $\hat{y}_i$. 
As for the complexity of the model, it is penalized by the second term $\Omega$. 

To guarantee traditional optimization methods in Euclidean space, the model is trained in an additive manner that is 
$$\mathcal{L}^{(t)} = \sum_{i=1}^nl(y_i,\ \hat{y}_i^{(t-1)} + f_t(x_i)) + \Omega(f_t)$$ 
where $\hat{y}_i^{(t)}$ is the prediction of the $i$-th instance at the $t$-th iteration.

After mathematical deviation, for which readers can check the citation~\cite{chen2016xgboost}, what determines the objective function is 
$$\tilde{\mathcal{L}}^{(t)}(q) = -\frac{1}{2}\sum_{j=i}^T\frac{(\sum_{i\in I_j} g_i)^2}{\sum_{i \in I_j} h_j + \lambda} + \gamma T$$ 
where $I_j = \{ i|\ q(x_i)=j\}$ defined as the instance set of leaf j, $g_i = \partial_{\hat{y}^{(t-1)}}l(y_i,\ \hat{y}^{(t-1)})$ and $h_i = \partial_{\hat{y}^{(t-1)}}^2 l(y_i,\ \hat{y}^{(t-1)})$ are first and second order gradient statistics on the loss function. 
With this function, calculation of the corresponding optimal value can be seemed as the measurement of the quality of a tree structure $q$.

Afterwards, a greedy algorithm that starts from a single leaf and iteratively adds branches to the tree can be used instead of enumerating all the possible tree structures $q$, which is almost impossible in practice. 
Using $I_L$ and $I_R$ to represent the instance sets of left and right nodes after the split, the loss reduction after the split can be given by 
$$\mathcal{L}_{split} = \frac{1}{2} \left[ \frac{(\sum_{i\in I_L} g_i)^2}{\sum_{i \in I_L} h_j + \lambda} + \frac{(\sum_{i\in I_R} g_i)^2}{\sum_{i \in I_R} h_j + \lambda} - \frac{(\sum_{i\in I} g_i)^2}{\sum_{i \in I} h_j + \lambda} \right] - \gamma$$ 
With this, the algorithm can evaluate the split candidates, find the best one and therefore decide the model.

%----------------------------------------------------------------------------
%----------------------------------------------------------------------------
%----------------------------------------------------------------------------
\section{Feature Extraction}

Not only in the area of machine learning, feature extraction, also known as feature/variable selection in pattern recognition and image processing, starts with an initial dataset to build derived values or features, which are designed to provide information and simplicity. 
With this procedure facilitating the subsequent learning and generalization steps, it is always easier to interpret for human. 
On this basis, feature extraction is related to dimensionality reduction~\cite{sarangi2020optimization}.

\begin{figure}[ht]
    \centering
    \includegraphics[width=0.6\linewidth]{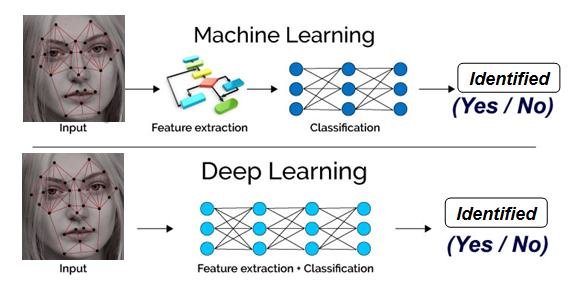}
    \caption{Example of utilizing feature extraction~\cite{najm2019effective}}
    \label{fig:feature_extraction}
\end{figure}

When some input data is too large to process for an algorithm, and it is suspected to be redundant, it can be converted into a subset with extracted features. 
For example, the input data includes the same measurement in both feet and meters, then the feature extraction will drop the redundant information like the data measured in feet. 
The process of determining such a subset is called feature extraction/selection~\cite{alpaydin2020introduction}. 
The extracted features should contain relevant information from the initial dataset so that this simplified representation can be used as the alteration of the full original dataset to perform the required task shown as figure \ref{fig:feature_extraction}~\cite{najm2019effective}.

The main task involved by feature extraction is to reduce the resource which are necessary to describe large amounts of data. 
When analyzing or learning with massive complex data, which usually requires a lot of memory and computational power, one of the main problems are caused by the number of variables involved. 
Besides, with the original dataset, a classification algorithm or model is inclined to overfit training samples and may be generalized poorly afterwards. 
In general, feature extraction is a generic term for methods combining features or variables to solve these problems with performing sufficient accuracy comparing with the initial dataset. 
It is commonly believed for many machine learning researchers that the key to effectively construct a model is properly optimized feature extraction~\cite{bastien2012theano}.

In the experiments, except for the classic dimensionality reduction technique Principal Component Analysis (PCA), several feature extraction methods are used such as permutation importance,feature importance and hierarchical clustering based on Spearman correlations.

\subsection{Permutation Importance}

Permutation feature importance measures the increase in the prediction error of the model after we permuted the feature’s values, which breaks the relationship between the feature and the true outcome. 
In other words, the importance of a feature is measured by calculating the increase in the model’s prediction error after permuting the feature values. 
If shuffling the values of that feature increases the model error, the feature will be recognised as \textit{important} because this phenomenon means that the model is relying on that feature to predict~\cite{breiman2001random}.

The permutation feature importance measurement is introduced for and also experimented based on random forests, which has some advantages. 
Permutation importance has a rather decent interpretation that feature importance is the increase in model error when the information of the feature is destroyed which provides global insight for the model behavior. 
Moreover, feature importance considers all interactions with other features automatically because shuffling the feature values will also destroy the interactions between that feature with other features. 

However, considering all interactions with other features is a disadvantage because, when features are correlated, permuting one feature will also influence the performance in an adding-up way with other correlated features. 
Then the sum of drop is much larger when it should be for that exact feature. 
Meanwhile, a correlated feature may decrease the permutation importance of the aim feature because it may weaken the influence caused by the aim feature. 
For example, the temperature at 8:00 AM of the day may be the important feature to predict the probability of rain until another strongly correlated feature, like the temperature at 9:00 AM, is added. 
After introducing the temperature at 9:00 AM, one of the possibilities is that the temperature at 8:00 AM may become less important because the model can also rely on the temperature at 9:00 AM now. 
These disadvantages may make the result difficult to interpret. 

\subsection{Gain-Based Feature Importance}

An XGBoost model for a classification problem can produce the importance for features, which is based on the relative contribution of the corresponding feature to the model calculated by the contribution of each feature for each tree in the model that is the \textit{gain} of a feature. 
A higher value of this metric implies more importance of a feature for generating the prediction. 

The idea of measuring the \textit{gain} of a feature is to observe how much model performance improved when adding a new split based on this feature. 
Hence, gain-based feature importance is the most relevant interpretation of each feature in the XGBoost model~\cite{shi2019feature}.

\subsection{Hierarchical Clustering Based on Spearman Correlations}

Hierarchical clustering is widely used when the actual number of clusters is unknown or the relationships between objects are interesting for researchers, which is also the reason why it is preferred to non-hierarchical clustering. 
The unweighted pair group method using arithmetic average (UPGMA) method is the most popular hierarchical clustering method, which starts from single element clusters which are successively fused together until there is only one single cluster~\cite{kaufman2009finding, massart1983interpretation}. 

In detail, the first step of the clustering is to construct a similarity matrix based on the Spearman correlations between each pair of features in the experiments because it is difficult to use traditional method, like Euclidean distance, to measure the similarity between features. 
The Spearman correlation can be calculated by the formula 
$$\frac{\sum_i(x_i-\Bar{x})(y_i-\Bar{y})}{\sqrt{\sum_i(x_i-\Bar{x})^2 \sum_i(y_y-\Bar{y})^2}}$$ 
where $x_i$ and $y_i$ are observations, $\Bar{x}$ and $\Bar{y}$ are the mean value, and $i=$ paired score. 
The formula is utilized when there are tied ranks.

\begin{figure}[H]
    \centering
    \includegraphics[width=0.6\linewidth]{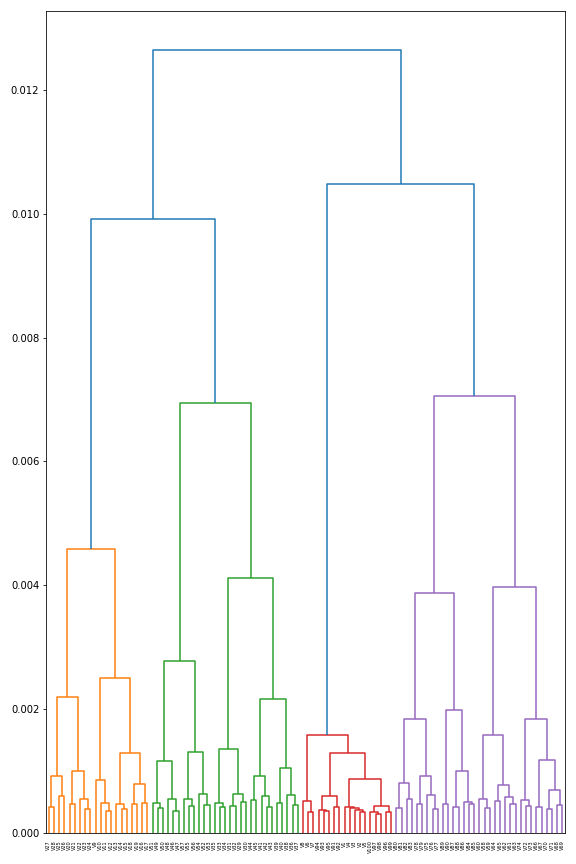}
    \caption{An example dendrogram for dataset 1479}
    \label{fig:example_dendrogram_1479}
\end{figure}

Then the most correlated features are selected and joined in one new cluster. 
This process is repeated until there is only one cluster including all objects. 
Hence, a dendrogram can always be used to represent the hierarchy that is a nested set of clusters. 
Here is an instance result in experiments shown in the figure \ref{fig:example_dendrogram_1479}, where, based on different height, different number of clusters can be decided.

With the preservation of the hierarchical cluster, selected feature subset can eliminate redundant features while the complete structure that represent the correlation between features is unaltered. 
Therefore, hierarchical clustering based on Spearman correlations between features is easy to perform feature extraction by choosing one feature per cluster~\cite{krier2007feature}.

\subsection{Principal Component Analysis}
\label{subsec:model_pca_intro}

In experiments, a method using singular value decomposition (SVD) of the data for linear dimensionality reduction, is called principal component analysis (PCA) used to project the data to a lower dimensional space, which can be considered as feature extraction as well. 
In real coordinate space, a sequence of $p$ unit vectors is the principal components of the set of data points, where the $i$-th vector is the line direction orthogonal to the first $i-1$ vectors and best fits the data, which means that the average squared distance from the points to the line is minimized. 
An orthonormal basis can be constructed from these directions so that there is no linear correlation between different individual dimensions of the data~\cite{markopoulos2017efficient}. 
It is PCA that calculates the principal components and uses them to perform a change of basis on the data that can be seen as dimensionality reduction or feature extraction.

%----------------------------------------------------------------------------
%----------------------------------------------------------------------------
%----------------------------------------------------------------------------
\section{Matrix Factorization}

In the mathematical discipline, a matrix factorization, also known as matrix decomposition, means that a matrix is factorized into a product of matrices. 
There are many different useful matrix decomposition methods available for a specific class of problems.
To introduce the matrix factorization model, taking The example of recommendation systems, especially for user-item recommendation, is much clear and easy to understand.

For the recommendation system, matrix factorization algorithms, which can be seen as a class of collaborative filtering algorithms, decompose the user-item interaction matrix into the product of two lower dimensionality rectangular matrices~\cite{koren2009matrix}. 
In terms of recommendation systems, it is a common task to improve customer experience by personalized recommendations, which have to be based on prior implicit feedback. 
Hence, to model user preferences, different sorts of user behavior, such as browsing activity, watching habits and purchase history, can be passively track by these systems. 
After the effectiveness of matrix factorization approach is reported by Simon Funk in a 2006 blog post, it became widely known~\cite{funk2006netflix}. 
However, unlike the explicit feedback, which is researched much more broadly, there is no direct input from the users about their preferences. 
It is proposed that the data is treated as an indication of positive and negative preference associated with vastly different confidence level, which leads to a factor model which is specifically tailored for implicit feedback recommenders~\cite{hu2008collaborative}.

Different types of input are relied on by recommender systems. 
Most convenient is the high quality explicit feedback, which includes explicit input by users regarding their interest in products. 
However, explicit feedback is not always available. 
Thus, recommenders can infer user preferences from the more abundant implicit feedback, which indirectly reflect opinion through observing user behavior~\cite{oard1998implicit}. 

The task of matrix factorization is to approximate the true original matrix $R$ by 
$$\hat{R} = W \cdot H^T$$ 
namely the approximated matrix is the product of two feature matrices $W$ and $H$. 
Take recommender system for online customer as an example, $W$: $|U| \times k$ means that the the $i$-th row $w_i$ of $W$ contains $k$ features that describe the $i$-th user, and $H$: $|I| \times k$ means the $j$-th row $h_j$ of $H$ contains k corresponding features for the $j$-th item. 
Hence, $\hat{R} = W \cdot H^T$ is equivalent to the following 
$$\hat{r}_{(i, j)} = \langle w_i,\ h_j \rangle = \sum_{f=1}^k w_{(i, f)} \cdot h_{(j, f)}$$ 
In order to center the approximation, a bias term is added to guarantee that only residuals are learned: 
$$\hat{r}_{(i, j)} = \sum_{f=1}^k w_{(i, f)} \cdot h_{(j, f)} + b_{(i, j)}$$ 
Normally, $b_{(i, j)}$, the bias term, is the global average, the user average, or the item average, but the result of another prediction algorithm could be an alternation as well~\cite{rendle2008online}.

\subsection{Implicit model}

Once the user gives approval to collect usage data, an implicit model requires no additional explicit feedback, such as ratings, which, from users' view, is much more convenient than to feedback explicitly. 
The recommender system with matrix factorization model used in experiments can be seen as implicit model because it was designed almost with no explicit feedback. 
Here are the two prime characteristics of inputs for the current model~\cite{hu2008collaborative}:
\begin{enumerate}
    \item All inputs are non-negative. In the experiments, the input matrix is the accuracy of each algorithms on different datasets. Hence, it is impossible with the model to utilize any negative inputs.
    \item There is inherent noise in the inputs. For example, some algorithms may produce overfitting results, which does not necessarily indicate a better performance for the algorithm. 
\end{enumerate}

Usually, the implicit-feedback system needs appropriate measurements for evaluation. 
However, for the experimented matrix decomposition model, clear metrics such as mean squared error are available and easy to interpret just like in the traditional setting.

\subsection{Distance between different algorithms and datasets}

\begin{figure}[ht]
    \centering
    \includegraphics[width=0.6\linewidth]{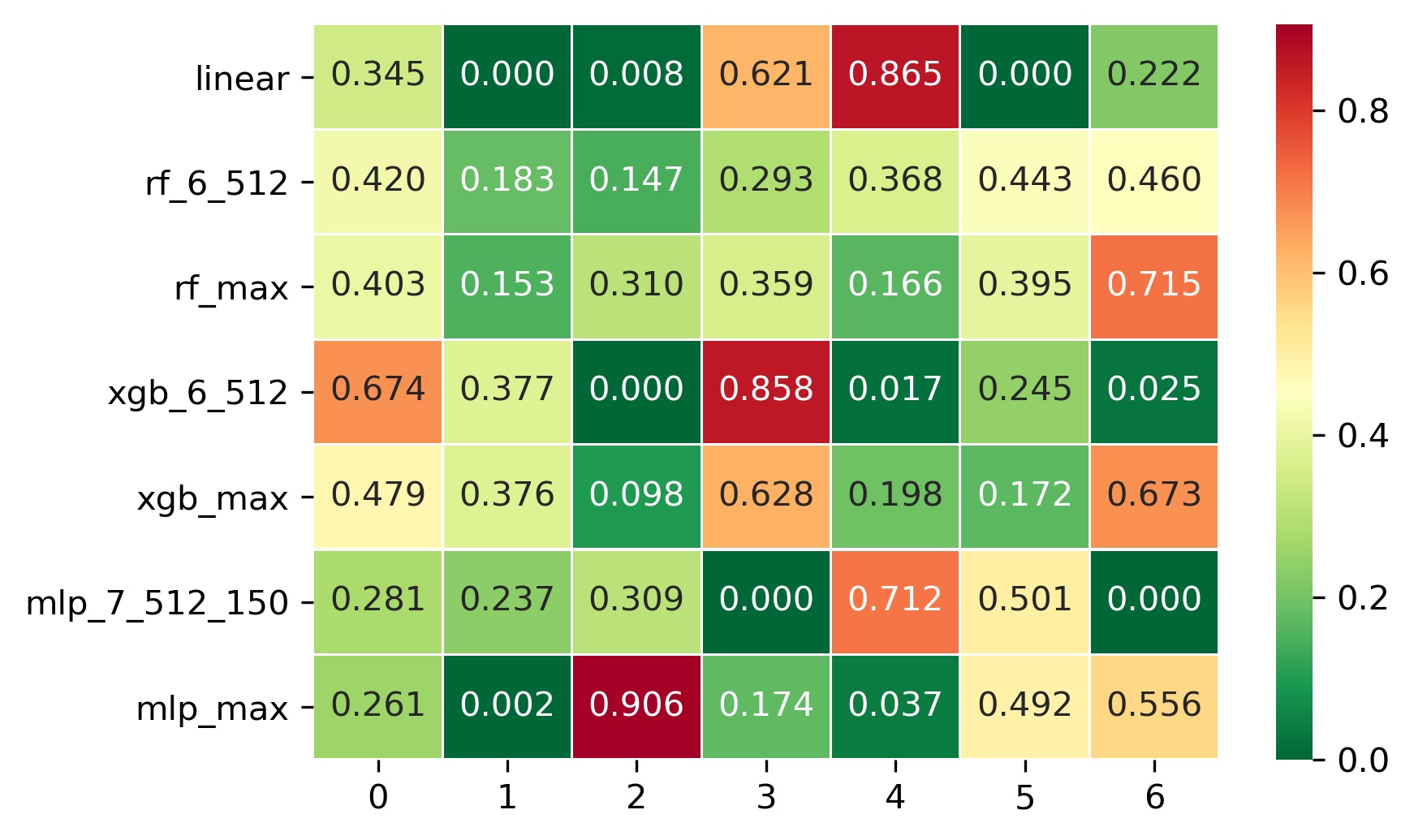}
    \caption{The visualization for coordinates of different algorithms}
    \label{fig:alg_coordinate}
\end{figure}

After factorizing the input matrix with $\hat{R} = W \cdot H^T$, it is possible to split each row for both $W$ and $H$ as $w_i$ and $h_j$. 
The $w_i$ can be seen as the coordination of the $i$-th algorithm. 
In the same way, the $h_j$ can be seen as the coordination of the $j$-th dataset. 
The figure \ref{fig:alg_coordinate} shows the coordination of different algorithms. 

As shown in the figure, the x-axis is the dimension of coordinates, which is originally equal to the number of models we use, 
and the y-axis is the name of models. 
We use abbreviation in the figure that RF\_max, XGB\_max, and MLP\_max mean the best performance of model random forest, XGBoost, and MLP, respectively. 
Besides, RF\_6\_512 and XGB\_6\_512 symbolize the accuracy of model random forest and XGBoost with 512 trees and max depth 6 for each tree in the model, which can be seen as the most complex model. 
Similarly, MLP\_7\_512\_150 stands for the accuracy of the model MLP trained for 150 epochs with 7 layers and 512 nodes per layer. 
Also, with the heatmap, we can visualize the value of the coordination.

In this figure, we can see that the coordinate of linear model is $(0.345,\ 0.000,\ 0.008,\ 0.621,\ 0.865,\ 0.000,\ 0.222)$. 
That is $(0.420,\ 0.183,\ 0.147,\ 0.293,\ 0.368,\ 0.443,\ 0.460)$ 
for the model of random forest with $512$ trees and each tree with max depth equal to 6. 
As for the model random forest with the best performance, that is $(0.403,\ 0.153,\ 0.310,\ 0.359,\ 0.166,\ 0.395,\ 0.715)$.
And we can read other coordinates similarly.

Based on the coordination, the similarity of algorithms and datasets can be calculated. 
The results are shown in the figure \ref{fig:algorithm_dist} in the section \ref{subsec:dist_alg_data}. 
We will describe it in detail in that part.

% \todo{explain more and mention the result of distance}

\newpage
\chapter{Results of Experiments}
\label{chap:result}
In this section, the whole process and results of experiments are stated. 
First of all, it is necessary to check whether each algorithm has a clear, stable, and reasonable trend of change due to the change of hyperparameters. 
What is expected for a selected algorithm is to observe the monotonically increasing accuracy in most datasets, as the model becomes more complex because of hyperparameter tuning. 
If such a pattern exists, a prediction framework for algorithms can be built simply according to this pattern. 
Therefore, a set of heatmaps for each algorithm are presented to intuitively show whether there is a certain change pattern in the performance of different algorithms with the change of hyperparameters.

While experimenting with each algorithm, it is also found possible to scale the data by changing the dimension of the data, namely the features. 
In particular, random forest and XGBoost both have a function to rank features, in accordance with which feature extraction can be done. 
Considering the lack of experiments with feature extraction in previous studies, various techniques are empirically studied and the related patterns changing with the number of features are compared. 
The ultimate goal is to test the predictability of these technologies as well.

Finally, the framework supported by matrix factorization is built to forecast the accuracy of an algorithm on a given dataset. 
With this framework, it is feasible to analyze the similarity among algorithms or datasets.
 
\subsection{Heatmap}
Heatmaps are used for the experiments that are with one kind of model alone. 
In specific, for random forest models and XGBoost models, the experiments traverse all combinations with max depth of each tree from 1 to 6 and trees number of each model equal to $2^n$ where n is from 1 to 9. 
As for multi-layer perceptron, the experiments go through all combinations with layer numbers from 1 to 6 and nodes number of each layer equal to $2^n$ where n is from 1 to 9. 
Meanwhile, the density of the color shows the level of accuracy of the model on that dataset, where dark blue is for the low accuracy and dark red is for the high accuracy. 
Moreover, unless otherwise stated, the training set ratio is $0.33$, which means $33\%$ of dataset is utilized to train the model.

Using heatmaps can give an instant overview of all results and latent patterns of how accuracy changes with different combinations of hyperparameters. 
For readability, only two groups of representative heatmaps are selected for each algorithm. 
The results of other pictures can be found in the appendix.

Since linear classifier does not need a lot of hyperparameters tuning like the other three types of models, we start from random forest to describe the results obtained and give out summery of all results at the end of this subsection. 

\subsubsection{Random Forest}
\label{subsec:heatmap_rf}
\begin{figure}[H]
    \centering
    \includegraphics[width=0.3\linewidth]{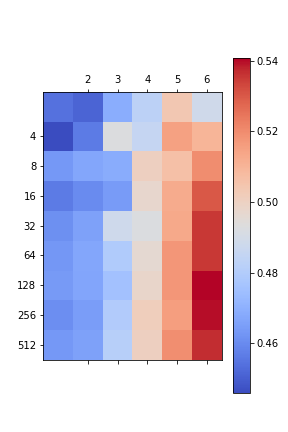}
    \caption{The performance of Random Forest on dataset 4538}
    \label{fig:forest_4538}
\end{figure}

Firstly, consider dataset 4538 for example, which is for gesture phase segmentation\cite{Dua:2019, wagner2014gesture, madeo2013gesture, madeo2016gesture, wagner2013segmentacc} and related attributes of the dataset are given in the table\ref{tab:multi_datasets}. 
The overall pattern can be seen from the heatmap \ref{fig:forest_4538}, where the x-axis represents the max depth of each tree and the y-axis represents the number of trees. 
According to the result, the performance of random forest models is usually affected more by the max depth of each tree used in the model than the number of trees because it can be observed that a vertical line divides the heatmap into left and right parts. 
In particular, for a random forest model with only two trees, when the max depth is 4, the performance is $0.483$, which is better than that of the model with 512 trees but the max depth only is equal to 1, which is $0.463$.

The experiments are using the function \texttt{sklearn.ensemble.RandomForestClassifier} from the library scikit-learn, where other hyperparameters are set as default that is only \texttt{n\_estimators}, representing the number of trees, and \texttt{max\_depth}, namely the max depth of each tree, are changed during implementation\cite{sklearn_api, scikit-learn}. 
As for the detail of other hyperparameters, please check the reference and \hyperlink{https://scikit-learn.org/stable/modules/generated/sklearn.ensemble.RandomForestClassifier.html}{this link}.

\begin{figure}[H]
    \centering
    \begin{subfigure}[b]{0.24\textwidth} 
        \centering
        \includegraphics[width = \linewidth]{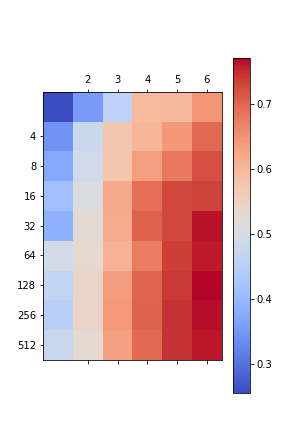}
        \caption{22}
    \end{subfigure}
    \begin{subfigure}[b]{0.24\textwidth}
        \centering
        \includegraphics[width = \linewidth]{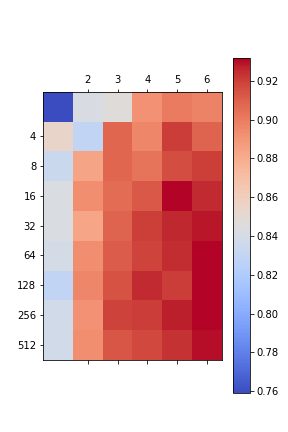}
        \caption{44}
    \end{subfigure}
    \begin{subfigure}[b]{0.24\textwidth} 
        \centering
        \includegraphics[width = \linewidth]{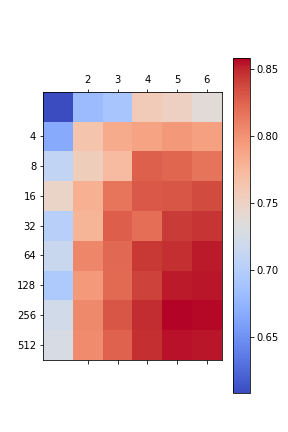}
        \caption{60}
    \end{subfigure}
    \begin{subfigure}[b]{0.24\textwidth} 
        \centering
        \includegraphics[width = \linewidth]{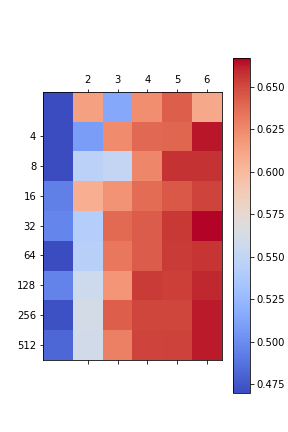}
        \caption{180}
    \end{subfigure}
    \caption{Part of results of random forest}
    \label{fig:heatmap_forest1}
\end{figure}

\begin{figure}[H]
    \centering
    \begin{subfigure}[b]{0.24\textwidth} 
        \centering
        \includegraphics[width = \linewidth]{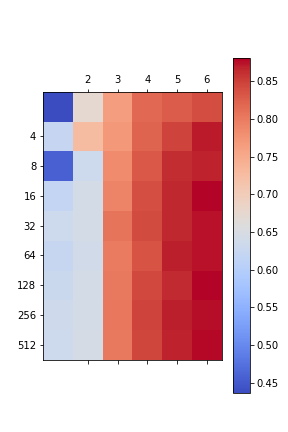}
        \caption{182}
    \end{subfigure}
    \begin{subfigure}[b]{0.24\textwidth} 
        \centering
        \includegraphics[width = \linewidth]{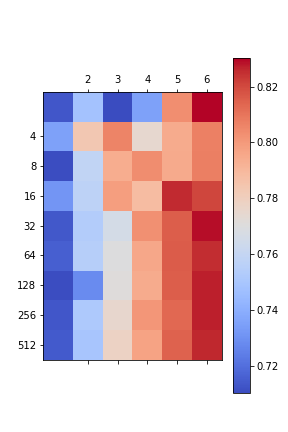}
        \caption{734}
    \end{subfigure}
    \begin{subfigure}[b]{0.24\textwidth} 
        \centering
        \includegraphics[width = \linewidth]{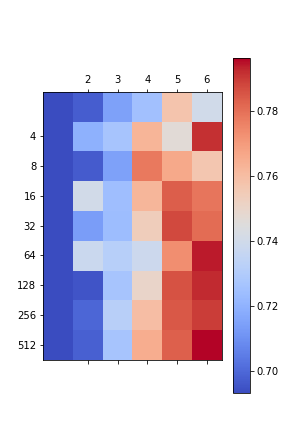}
        \caption{833}
    \end{subfigure}
    \begin{subfigure}[b]{0.24\textwidth} 
        \centering
        \includegraphics[width = \linewidth]{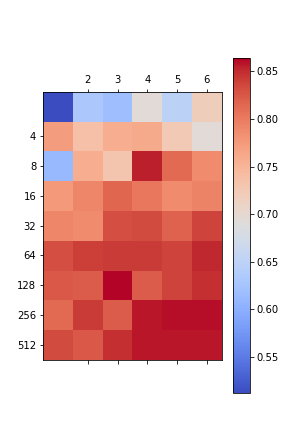}
        \caption{904}
    \end{subfigure}
    
    \centering
    \begin{subfigure}[b]{0.24\textwidth} 
        \centering
        \includegraphics[width = \linewidth]{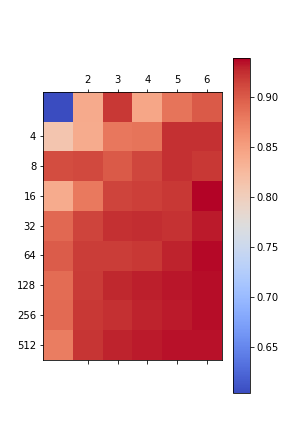}
        \caption{4534}
    \end{subfigure}
    \begin{subfigure}[b]{0.24\textwidth}
        \centering
        \includegraphics[width = \linewidth]{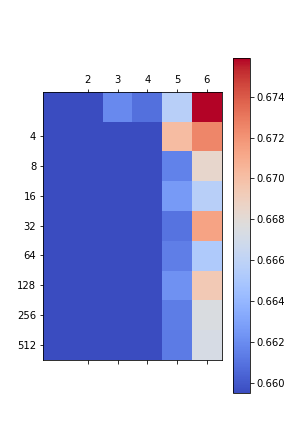}
        \caption{40668}
    \end{subfigure}
    \begin{subfigure}[b]{0.24\textwidth} 
        \centering
        \includegraphics[width = \linewidth]{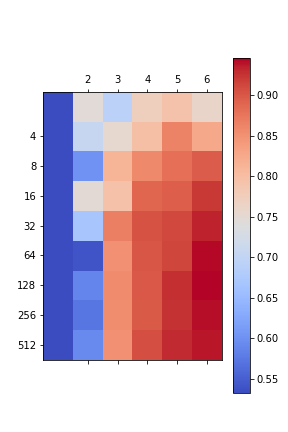}
        \caption{40670}
    \end{subfigure}
    \begin{subfigure}[b]{0.24\textwidth} 
        \centering
        \includegraphics[width = \linewidth]{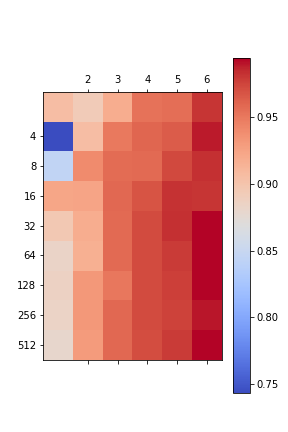}
        \caption{40997}
    \end{subfigure}
    
    \centering
    \begin{subfigure}[b]{0.24\textwidth} 
        \centering
        \includegraphics[width = \linewidth]{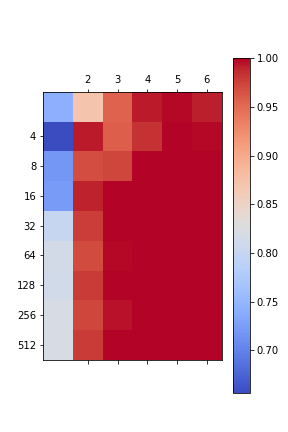}
        \caption{40999}
    \end{subfigure}
    \begin{subfigure}[b]{0.24\textwidth}
        \centering
        \includegraphics[width = \linewidth]{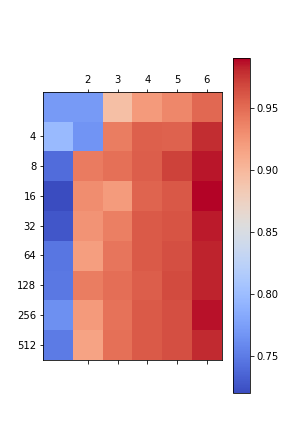}
        \caption{41000}
    \end{subfigure}
    \begin{subfigure}[b]{0.24\textwidth} 
        \centering
        \includegraphics[width = \linewidth]{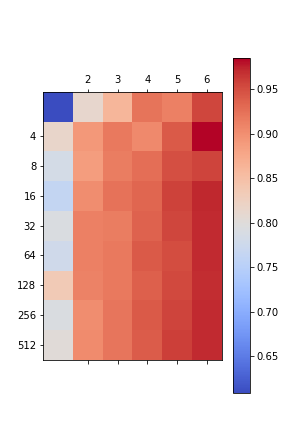}
        \caption{41004}
    \end{subfigure}
    \begin{subfigure}[b]{0.24\textwidth} 
        \centering
        \includegraphics[width = \linewidth]{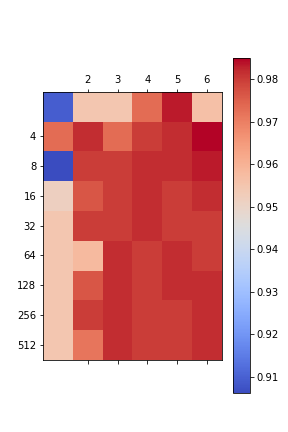}
        \caption{41048}
    \end{subfigure}
    
    \caption{Part of results of random forest}
    \label{fig:heatmap_forest2}
\end{figure}

After experimenting with amounts of datasets, partly shown in figure \ref{fig:heatmap_forest1} and \ref{fig:heatmap_forest2}, the performance can be found improved with increasing the max depth of each tree and the number of trees, usually more influenced by the former. 
Such a pattern is found to be common, that is for 43 out of 54 datasets showing this pattern, which can be found in the table \ref{tab:pattern_sum_mlp} and \ref{tab:pattern_sum_xgb_rf}. 
With this extensively existed reasonable tendency, the predictability is proved good enough so that, for most datasets, increasing the max depth of each tree and/or the number of trees in random forest models can improve the performance.

%------------------------------------------------------------------
%------------------------------------------------------------------

%------------------------------------------------------------------
%------------------------------------------------------------------
\subsubsection{XGboost}
\label{subsec:heatmap_xgb}
\begin{figure}[ht]
    \centering
    \includegraphics[width=0.3\linewidth]{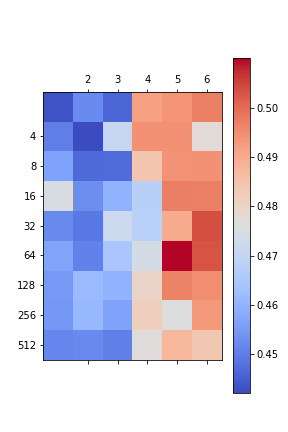}
    \caption{The performance of XGboost on dataset 4538}
    \label{fig:xgb_4538}
\end{figure}

The pattern appearing in random forest models that is the performance is increased with adding up the number of trees and/or the max depth of each tree. 
For the same dataset 4538, an XGBoost model with only two trees, when the max depth is 4, the performance is $0.492$, which is better than that of the model with 512 trees but the max depth only is equal to 1, which is $0.452$. 
It means the phenomenon that performance with a few but deep trees can be pretty well can be found in XGBoost models as well. 
The overall pattern can also be observed from the heatmap \ref{fig:xgb_4538}, where the x-axis represents the max depth of each tree and the y-axis represents the number of trees.

The experiments are using the function \texttt{tf.estimator.BoostedTreesClassifier} from the library TensorFlow, where other hyperparameters are set as default that is only \texttt{n\_trees}, representing the number of trees, and \texttt{max\_depth}, namely the max depth of each tree, are changed during implementation\cite{tensorflow2015-whitepaper}. 
Besides, the hyperparameter \texttt{n\_classes} is always set equal to the number of different labels in the dataset. 
To look into the detail of other hyperparameters, please check the reference and \hyperlink{https://tensorflow.google.cn/api_docs/python/tf/estimator/BoostedTreesClassifier?hl=en}{this link}.

\begin{figure}[H]
    \centering
    \begin{subfigure}[b]{0.24\textwidth} 
        \centering
        \includegraphics[width = \linewidth]{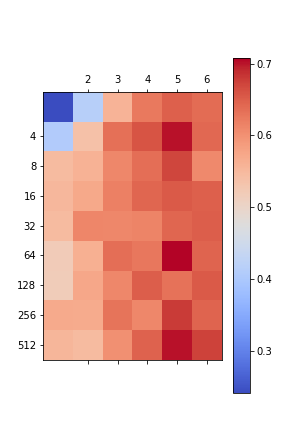}
        \caption{22}
    \end{subfigure}
    \begin{subfigure}[b]{0.24\textwidth} 
        \centering
        \includegraphics[width = \linewidth]{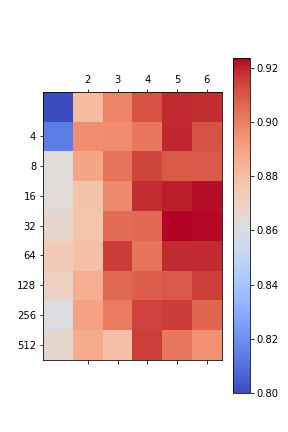}
        \caption{44}
    \end{subfigure}
    \begin{subfigure}[b]{0.24\textwidth}
        \centering
        \includegraphics[width = \linewidth]{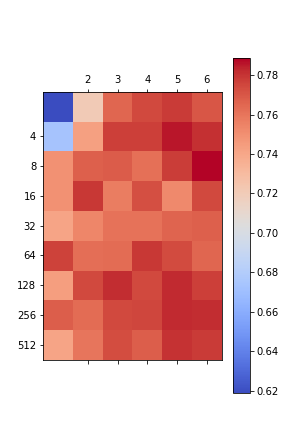}
        \caption{60}
    \end{subfigure}
    \begin{subfigure}[b]{0.24\textwidth} 
        \centering
        \includegraphics[width = \linewidth]{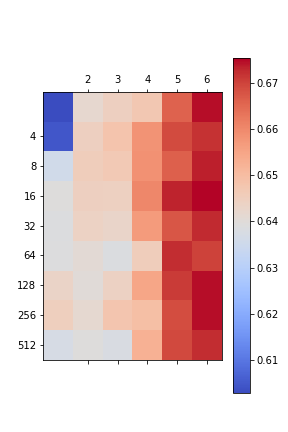}
        \caption{180}
    \end{subfigure}
    
    \caption{Part of results of XGBoost}
    \label{fig:heatmap_xgb1}
\end{figure}

\begin{figure}[H]
    \centering
    \begin{subfigure}[b]{0.24\textwidth} 
        \centering
        \includegraphics[width = \linewidth]{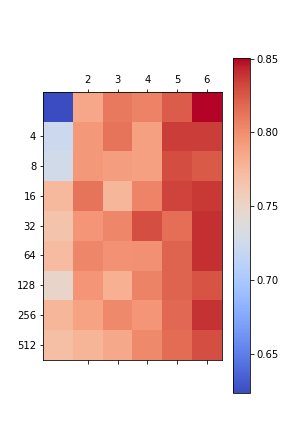}
        \caption{182}
    \end{subfigure}
    \begin{subfigure}[b]{0.24\textwidth} 
        \centering
        \includegraphics[width = \linewidth]{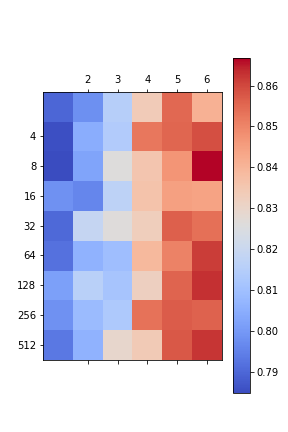}
        \caption{734}
    \end{subfigure}
    \begin{subfigure}[b]{0.24\textwidth} 
        \centering
        \includegraphics[width = \linewidth]{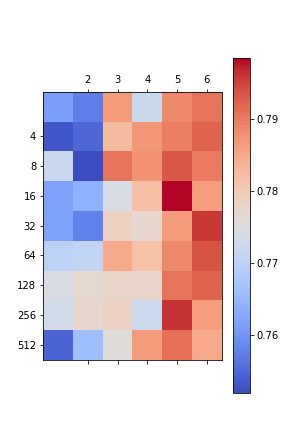}
        \caption{833}
    \end{subfigure}
    \begin{subfigure}[b]{0.24\textwidth}
        \centering
        \includegraphics[width = \linewidth]{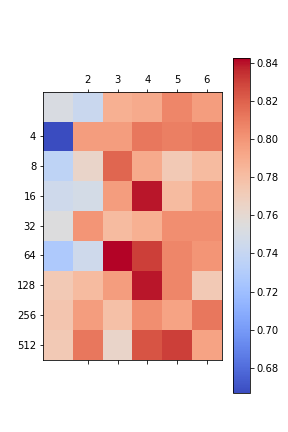}
        \caption{904}
    \end{subfigure}
    
    \centering
    \begin{subfigure}[b]{0.24\textwidth} 
        \centering
        \includegraphics[width = \linewidth]{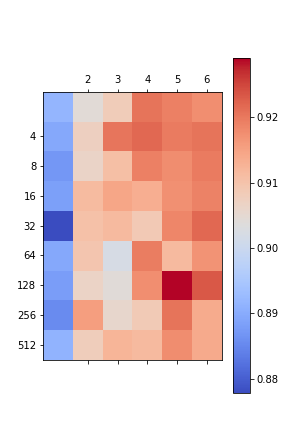}
        \caption{4534}
    \end{subfigure}
    \begin{subfigure}[b]{0.24\textwidth} 
        \centering
        \includegraphics[width = \linewidth]{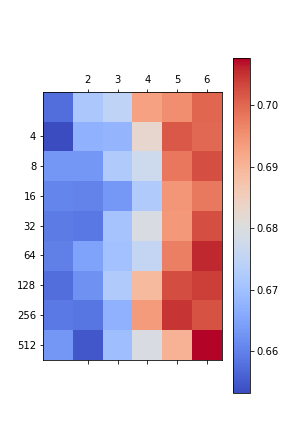}
        \caption{40668}
    \end{subfigure}
    \begin{subfigure}[b]{0.24\textwidth} 
        \centering
        \includegraphics[width = \linewidth]{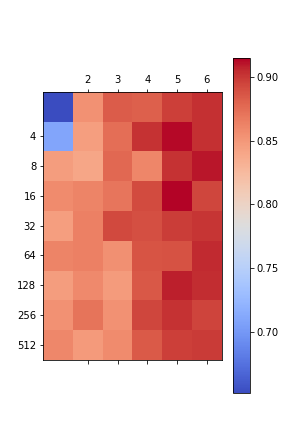}
        \caption{40670}
    \end{subfigure}
    \begin{subfigure}[b]{0.24\textwidth} 
        \centering
        \includegraphics[width = \linewidth]{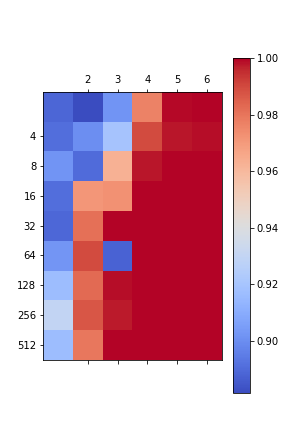}
        \caption{40997}
    \end{subfigure}
    
    \centering
    \begin{subfigure}[b]{0.24\textwidth} 
        \centering
        \includegraphics[width = \linewidth]{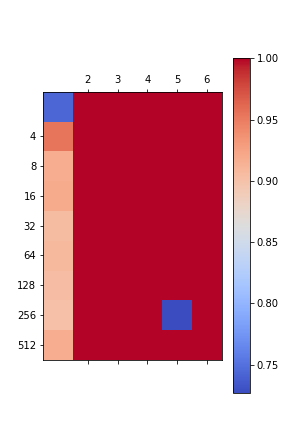}
        \caption{40999}
    \end{subfigure}
    \begin{subfigure}[b]{0.24\textwidth} 
        \centering
        \includegraphics[width = \linewidth]{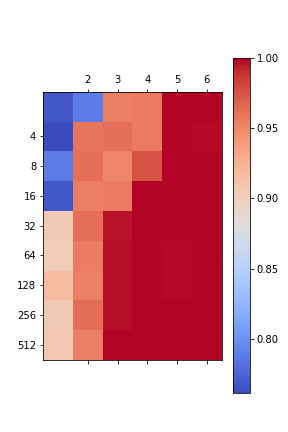}
        \caption{41000}
    \end{subfigure}
    \begin{subfigure}[b]{0.24\textwidth}
        \centering
        \includegraphics[width = \linewidth]{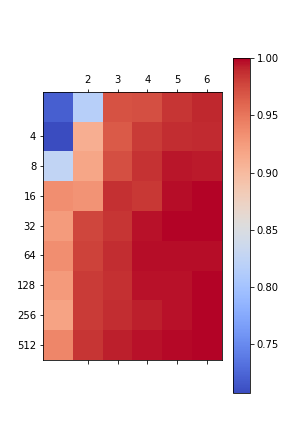}
        \caption{41004}
    \end{subfigure}
    \begin{subfigure}[b]{0.24\textwidth} 
        \centering
        \includegraphics[width = \linewidth]{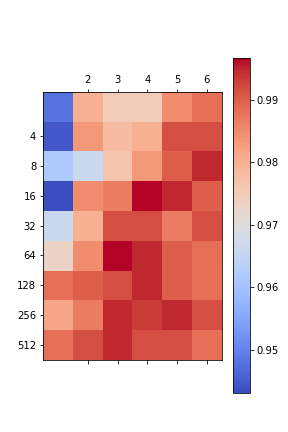}
        \caption{41048}
    \end{subfigure}
    \caption{Part of results of XGBoost}
    \label{fig:heatmap_xgb2}
\end{figure}

After experiments, the pattern can be found over half of the datasets that are for 29 out of 54 datasets, except 6 datasets cannot run XGBoost because of the limit of machine, showing performance enhanced with the growth of the max depth of each tree and/or the number of trees, which can be found in the table \ref{tab:pattern_sum_mlp} and \ref{tab:pattern_sum_xgb_rf}. 
Such patterns are shown in figure \ref{fig:heatmap_xgb1} and \ref{fig:heatmap_xgb2}. 
The vertical bound that divides the heatmap into left and right parts can be seen in some figures as well. 
With this existed reasonable tendency, which unfortunately is not an overwhelming majority, XGBoost can be considered to have the predictability but not as prominent as random forest. 
Hence, it is not reliable to anticipate the performance of XGBoost via these pattern.

%------------------------------------------------------------------
%------------------------------------------------------------------

%------------------------------------------------------------------
%------------------------------------------------------------------
\subsubsection{Multi-Layer Perceptron}
\label{subsec:heatmap_mlp}
\begin{figure}[ht]
    \centering
    \includegraphics[width=0.28\linewidth]{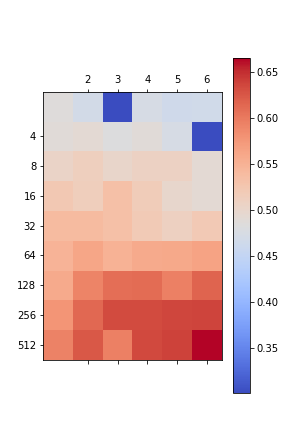}
    \caption{The performance of multi-layer perceptron on dataset 4538}
    \label{fig:mlp_4538}
\end{figure}

In random forest and XGBoost experiments, the performance is increased by adding up the number of trees and/or the max depth of each tree. 
In the experiment of multi-layer perceptron(MLP), two hyperparameters are mainly focused on, the number of layers and the number of nodes per layer. 
Similarly, first taking dataset 4538 as an example, it can be found that accuracy also improves with the increase of the number of nodes in each layer, which means that even the shallow network with enough neural nodes can perform as well as the deep network or even better. 
In specific, in an MLP model with only 1 layer, when the number of nodes per layer is 8, the performance is $0.504$, which is better than that of the model with 6 layers but the number of nodes per layer is only equal to 2, which is $0.4607$. 
The overall performance pattern of multi-layer perceptron on dataset 4538 can also be observed from the heatmap \ref{fig:mlp_4538}, where the x-axis represents the number of layers, namely the depth of the network, and the y-axis represents the number of nodes in each layer.

The experiments are using the function \texttt{tf.keras.layers.Dense} from the library TensorFlow, where hyperparameters \texttt{units} representing the number of nodes in each layer is changed during experiments, 
\texttt{activation} namely the activation function to use is set to \textit{relu}, and \texttt{kernel\_initializer} namely the initializer for the kernel weights matrix is set to \textit{he\_normal} and other hyperparameters are set as default except for the last layer of the model. 
In the last layer, if there are only $2$ classes in the dataset, \texttt{units} is set to $1$, \texttt{activation} is set to \textit{sigmoid}, 
otherwise, \texttt{units} is equal to the number of different labels and \texttt{activation} is assigned \textit{softmax}. 
Besides, during the implementation, the function \texttt{tf.keras.Sequential} is also used to construct MLP models. 
When using \texttt{tf.keras.Sequential.compile}, \texttt{optimizer} is always settled to \textit{adam}, 
if the dataset has only $2$ classes, \texttt{loss} is settled to \textit{binary\_crossentropy}, 
otherwise \textit{sparse\_categorical\_crossentropy}\cite{tensorflow2015-whitepaper}. 
Moreover, MLP models are all trained 150 epochs. 
To look into the detail of other hyperparameters, please check the reference, \hyperlink{https://tensorflow.google.cn/api\_docs/python/tf/keras/layers/Dense?hl=en}{the link for the first function}, and \hyperlink{https://www.tensorflow.org/api\_docs/python/tf/keras/Sequential\#used-in-the-notebooks\_1}{the link for the second function}.

\begin{figure}[H]
    \centering
    \begin{subfigure}[b]{0.22\textwidth}
        \centering
        \includegraphics[width = \linewidth]{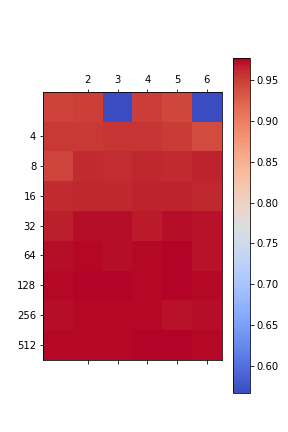}
        \caption{4534}
    \end{subfigure}
    \begin{subfigure}[b]{0.22\textwidth}
        \centering
        \includegraphics[width = \linewidth]{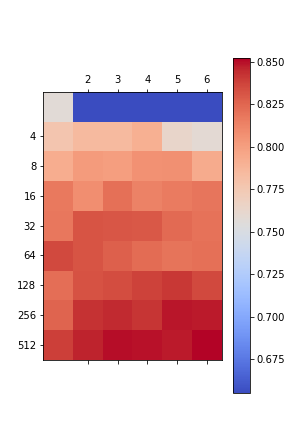}
        \caption{40668}
    \end{subfigure}
    \begin{subfigure}[b]{0.22\textwidth} 
        \centering
        \includegraphics[width = \linewidth]{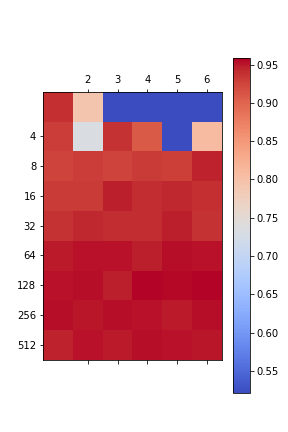}
        \caption{40670}
    \end{subfigure}
    \begin{subfigure}[b]{0.22\textwidth}
        \centering
        \includegraphics[width = \linewidth]{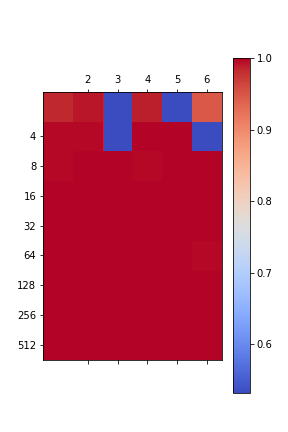}
        \caption{40997}
    \end{subfigure}
    
    % \centering
    % \begin{subfigure}[b]{0.24\textwidth} 
    %     \centering
    %     \includegraphics[width = \linewidth]{Figures/heatmap_mlp/40999_mlp_rescale.png}
    %     \caption{40999}
    % \end{subfigure}
    % \begin{subfigure}[b]{0.24\textwidth}
    %     \centering
    %     \includegraphics[width = \linewidth]{Figures/heatmap_mlp/41000_mlp_rescale.png}
    %     \caption{41000}
    % \end{subfigure}
    % \begin{subfigure}[b]{0.24\textwidth} 
    %     \centering
    %     \includegraphics[width = \linewidth]{Figures/heatmap_mlp/41004_mlp_rescale.png}
    %     \caption{41004}
    % \end{subfigure}
    % \begin{subfigure}[b]{0.24\textwidth} 
    %     \centering
    %     \includegraphics[width = \linewidth]{Figures/heatmap_mlp/41048_mlp_rescale.png}
    %     \caption{41048}
    % \end{subfigure}
    \caption{Part of results of multi-layer perceptron}
    \label{fig:heatmap_mlp2}
\end{figure}

\begin{figure}[H]
    \centering
    \begin{subfigure}[b]{0.24\textwidth} 
        \centering
        \includegraphics[width = \linewidth]{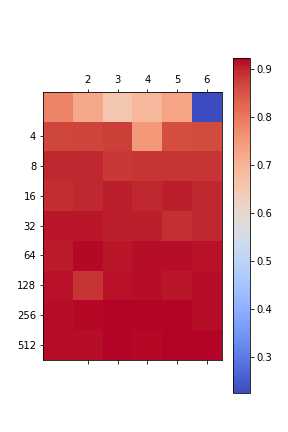}
        \caption{182}
    \end{subfigure}
    \begin{subfigure}[b]{0.24\textwidth} 
        \centering
        \includegraphics[width = \linewidth]{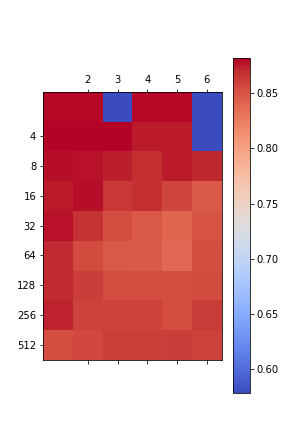}
        \caption{734}
    \end{subfigure}
    \begin{subfigure}[b]{0.24\textwidth} 
        \centering
        \includegraphics[width = \linewidth]{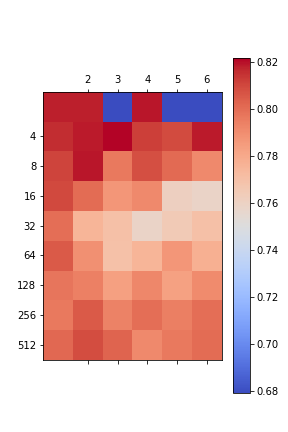}
        \caption{833}
    \end{subfigure}
    \begin{subfigure}[b]{0.24\textwidth} 
        \centering
        \includegraphics[width = \linewidth]{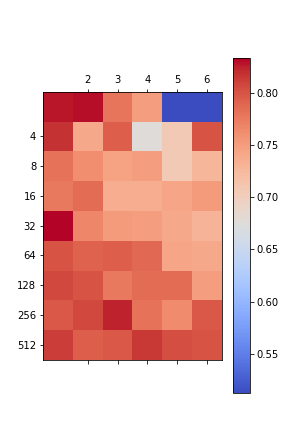}
        \caption{904}
    \end{subfigure}
    
    \centering
    \begin{subfigure}[b]{0.24\textwidth} 
        \centering
        \includegraphics[width = \linewidth]{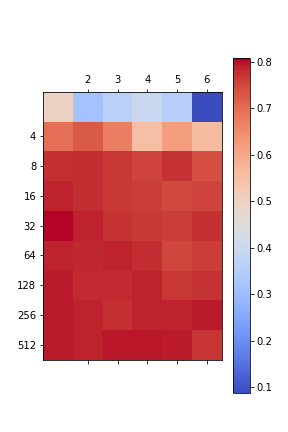}
        \caption{22}
    \end{subfigure}
    \begin{subfigure}[b]{0.24\textwidth} 
        \centering
        \includegraphics[width = \linewidth]{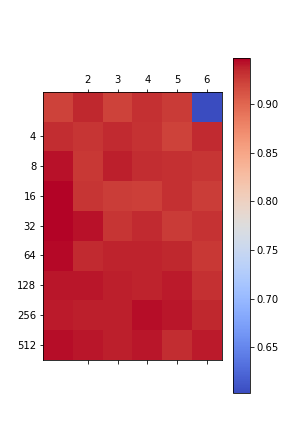}
        \caption{44}
    \end{subfigure}
    \begin{subfigure}[b]{0.24\textwidth}
        \centering
        \includegraphics[width = \linewidth]{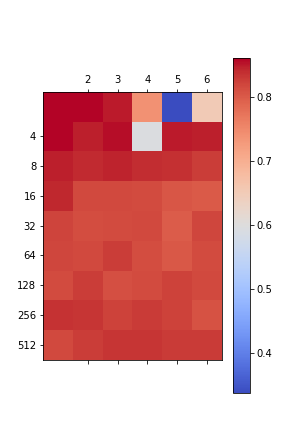}
        \caption{60}
    \end{subfigure}
    \begin{subfigure}[b]{0.24\textwidth} 
        \centering
        \includegraphics[width = \linewidth]{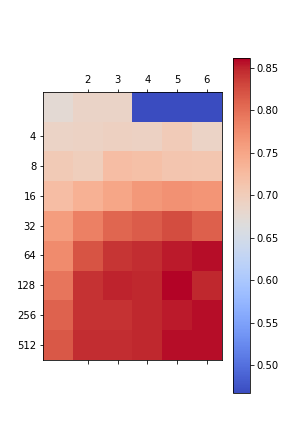}
        \caption{180}
    \end{subfigure}
    
    \caption{Part of results of multi-layer perceptron}
    \label{fig:heatmap_mlp1}
\end{figure}

Through plenty of experiments, it is pervasive that shallow networks can perform as well as complicated networks or even better. 
About the influence of two hyperparameters, the number of layers and the number of nodes per layer, accuracy changes with these two changes in 42 of the 54 datasets as shown in figures \ref{fig:heatmap_mlp1} and \ref{fig:heatmap_mlp2}. 
In some data sets, a horizontal line can be vaguely observed dividing the heatmap into upper and lower parts, 
which also indicates that the number of nodes per layer has a greater impact on accuracy. 

However, in addition to this discovery, many patterns are not as reasonable as random forest and XGBoost, namely the performance is foreseen increasing with the number of nodes and layers increasing. 
As shown in Figure \ref{fig:heatmap_mlp1}, accuracy decreases with the increase of two hyperparameters, especially the number of layers, on many data sets. 
Such counterintuitive patterns accounted for 25 of the 42 datasets with patterns, while the pattern shown in Figure \ref{fig:heatmap_mlp2} is relatively consistent with the ideal situation, accounting for only 17 datasets. 
See table \ref{tab:pattern_sum_mlp} and \ref{tab:pattern_sum_xgb_rf} for detailed results, where the expected patterns are denoted by 1 and unexpected patters are denoted by 0. 
This may be related to the problem of overfitting, which needs further discussion. 
But in any case, should we assume that MLP hardly has predictability due to the inconsistent and mostly counterintuitive patterns. 
Therefore, it seems not feasible to estimate the performance of MLP via these patterns.

%------------------------------------------------------------------
%------------------------------------------------------------------
\subsubsection{Summary}
\label{subsubsec:heatmap_summary}
Here the summary tables are given. 
The entire summary is divided into two tables based on the classifier with the highest accuracy on the dataset. 
In the table \ref{tab:pattern_sum_mlp} and \ref{tab:pattern_sum_xgb_rf}, not only the occurrence of pattern is summarized, but also the optimal performance corresponding to each algorithm, so as to compare which algorithm may have the best performance. 
Tables are arranged in ascending order by the \textit{ID} column within the partition.

The first column \textit{ID} is the id of datasets, which can be found in OpenML using the data id\cite{OpenML2013}, and the corresponding attributes can be viewed in table \ref{tab:binary_datasets} and \ref{tab:multi_datasets}. 
The whole row is various results for the dataset. 

The second column \textit{RF} corresponds to the occurrence of pattern about random forest (RF) on the dataset, and $1$ indicates that patter exists in the corresponding dataset. 
Similarly, the third column \textit{XGB} represents XGBoost(XGB) case, where labeled $null$ means that the XGBoost cannot run on that dataset because of the machine limitation. 
In the fourth column \textit{MLP}, in order to distinguish different patterns, $1$ is used to indicate that the existing pattern is prospective, that is, accuracy increases with the increase of node number and layer number, and on the contrary, $0$ is used to indicate the opposite pattern. 
Besides, if there is no pattern, the entry will be empty like other two columns.

The fifth, sixth, seventh, and eighth columns indicate the highest accuracy of linear classifier, random forest, XGBoost, and MLP on each dataset, respectively. 
The last column indicates the best performing algorithm on this dataset. 
As shown in table \ref{tab:pattern_sum_xgb_rf}, there are only 3 datasets, on which random forest has the best performance, including the dataset $313$ which should be seen as the abnormal one because all other three classifiers cannot work properly on it. 
Then, dataset $40999$ and $41005$ can be seen as simple datasets since, on these two datasets, the accuracy of three classifiers, random forest, XGBoost, and MLP, is equal to 1. 
Similarly, dataset $40997$, $41000$, and $41004$ are easy to classify as well. 
Apart from these, XGBoost and MLP show comparable excellent performance, each achieving the highest accuracy on 23 out of 54 datasets, respectively. 
However, it is worth mentioning that there are 5 datasets that XGBoost cannot run on because of the limited machine and it is on these 5 datasets that MLP achieves the best performance. 
\begin{table}[H]
  \centering
  \caption{Summary of performance patterns only with datasets MLP performs best}
    \begin{tabular}{l|l|l|l|l|l|l|l|l}
    \hline \hline
    ID    & RF    & XGB   & MLP   & Linear & Random Forest    &  XGBoost   & MLP   & Best \\
    \hline
    14    & 1     & 1     & 0     & 0.195455 & 0.821212 & 0.757576 & 0.830303 & mlp \\
    22    & 1     & 1     & 0     & 0.166667 & 0.769697 & 0.707576 & 0.807576 & mlp \\
    44    & 1     & 1     & 0     & 0.85451 & 0.931534 & 0.923634 & 0.946675 & mlp \\
    60    & 1     & 1     & 0     & 0.38  & 0.858182 & 0.788485 & 0.86  & mlp \\
    180   & 1     & 1     & 0     & 0.011419 & 0.666676 & 0.67535 & 0.861351 & mlp \\
    182   & 1     & 1     & 0     & 0.215834 & 0.880302 & 0.850141 & 0.921772 & mlp \\
    734   & 1     & 1     & 0     & 0.860952 & 0.830101 & 0.866681 & 0.881666 & mlp \\
    833   & 1     & 1     & 0     & 0.796228 & 0.796228 & 0.798447 & 0.821746 & mlp \\
    979   & 1     & 1     & 0     & 0.816364 & 0.867273 & 0.846061 & 0.893939 & mlp \\
    1479  &       &       &       & 0.5925 & 0.5575 & 0.5625 & 0.705 & mlp \\
    1491  & 1     & null  & 0     & 0.005682 & 0.537879 & null  & 0.831439 & mlp \\
    1492  & 1     & null  & 0     & 0.011364 & 0.416667 & null  & 0.704545 & mlp \\
    1493  & 1     & null  & 0     & 0.017045 & 0.642045 & null  & 0.829545 & mlp \\
    1494  & 1     & 1     & 0     & 0.802292 & 0.876791 & 0.853868 & 0.896848 & mlp \\
    1501  & 1     & 1     &       & 0.117871 & 0.901141 & 0.735741 & 0.948669 & mlp \\
    4534  & 1     & 1     & 1     & 0.890107 & 0.939161 & 0.929022 & 0.976706 & mlp \\
    4538  & 1     & 1     & 1     & 0.101565 & 0.540657 & 0.509972 & 0.664314 & mlp \\
    40668 & 1     & 1     & 1     & 0.245896 & 0.675877 & 0.70759 & 0.852023 & mlp \\
    40670 & 1     & 1     & 1     & 0.398289 & 0.945817 & 0.914449 & 0.958175 & mlp \\
    40705 & 1     &       & 0     & 0.905363 & 0.914826 & 0.946372 & 0.952681 & mlp \\
    40996 & 1     & null  & 1     & 0.120866 & 0.799957 & null  & 0.89697 & mlp \\
    41014 & 1     & null  & 1     & 0.010091 & 0.866567 & null  & 0.935239 & mlp \\
    41027 & 1     & 1     & 1     & 0.226219 & 0.748158 & 0.742411 & 0.979582 & mlp \\
    \toprule
    \end{tabular}%
  \label{tab:pattern_sum_mlp}%
\end{table}%

%------------------------------------------------------------------
%------------------------------------------------------------------
\begin{table}[H]
  \centering
  \caption{Summary of performance patterns and the best classifier for each dataset excluding MLP only}
    \begin{tabular}{l|l|l|l|l|l|l|l|l}
    \hline \hline
    ID    & RF    & XGB   & MLP   & Linear & Random Forest    &  XGBoost   & MLP   & Best Classifier\\
    \hline
    313   & 1     & null  &       & 0     & 0.551136 & null  & 0     & rf \\
    904   & 1     & 1     & 0     & 0.812121 & 0.863636 & 0.842424 & 0.833333 & rf \\
    4154  & 1     &       &       & 0.999362 & 0.999787 & 0.999787 & 0.999362 & rf \\
    \hline
    316   &       &       &       & 0.983709 & 0.989975 & 0.991228 & 0.986216 & xgb \\
    718   & 1     &       & 0     & 0.666667 & 0.8   & 0.89697 & 0.657576 & xgb \\
    1049  & 1     &       &       & 0.906639 & 0.894191 & 0.919087 & 0.906639 & xgb \\
    1050  &       &       & 0     & 0.895349 & 0.901163 & 0.932171 & 0.893411 & xgb \\
    1056  & 1     &       & 1     & 0.993598 & 0.995519 & 0.997119 & 0.995839 & xgb \\
    1069  &       &       & 0     & 0.996206 & 0.995122 & 0.997832 & 0.996206 & xgb \\
    1443  &       &       & 1     & 0.917808 & 0.936073 & 0.945205 & 0.936073 & xgb \\
    1444  &       &       & 0     & 0.843478 & 0.881159 & 0.913043 & 0.866667 & xgb \\
    1451  &       &       & 0     & 0.922747 & 0.914163 & 0.939914 & 0.922747 & xgb \\
    1452  &       &       & 0     & 0.971545 & 0.97561 & 0.987805 & 0.98374 & xgb \\
    1453  &       &       & 0     & 0.88764 & 0.879213 & 0.904494 & 0.882023 & xgb \\
    1487  &       &       &       & 0.92951 & 0.936679 & 0.949821 & 0.934289 & xgb \\
    1548  & 1     & 1     &       & 0.553939 & 0.595152 & 0.62303 & 0.552727 & xgb \\
    1549  & 1     &       & 0     & 0.104839 & 0.262097 & 0.33871 & 0.241935 & xgb \\
    1555  & 1     &       & 0     & 0.090909 & 0.266667 & 0.321212 & 0.221212 & xgb \\
    41007 & 1     & 1     & 1     & 0.861004 & 0.978121 & 1     & 0.993565 & xgb \\
    41025 &       &       & 0     & 0.6 & 0.8 & 0.9 & 0.6 & xgb \\
    41048 & 1     & 1     & 1     & 0.976549 & 0.984925 & 0.99665 & 0.994975 & xgb \\
    41049 & 1     &       &       & 0.984536 & 0.990979 & 0.993557 & 0.984536 & xgb \\
    41050 & 1     & 1     &       & 0.867672 & 0.889447 & 0.934673 & 0.909548 & xgb \\
    41051 & 1     & 1     &       & 0.896907 & 0.921392 & 0.947165 & 0.940722 & xgb \\
    41052 & 1     & 1     &       & 0.932998 & 0.860972 & 0.969849 & 0.954774 & xgb \\
    41053 & 1     & 1     & 1     & 0.904639 & 0.899485 & 0.926546 & 0.923969 & xgb \\
    \hline
    40999 & 1     & 1     & 1     & 0.841495 & 1     & 1     & 1     & rf, xgb, mlp \\
    41005 & 1     & 1     & 1     & 0.792219 & 1     & 1     & 1     & rf, xgb, mlp \\
    40997 & 1     & 1     & 1     & 0.423052 & 0.994205 & 1     & 1     & xgb, mlp \\
    41000 & 1     & 1     & 1     & 0.356729 & 0.990341 & 1     & 1     & xgb, mlp \\
    41004 & 1     & 1     & 1     & 0.507405 & 0.984546 & 1     & 1     & xgb, mlp \\
    \toprule
    \end{tabular}%
  \label{tab:pattern_sum_xgb_rf}%
\end{table}%

%------------------------------------------------------------------
%------------------------------------------------------------------
\subsection{Feature Extraction}
Firstly, it is essential to clarify what is the expectation for a well predictable technique of feature extraction. 
That is for a dataset $\mathcal{D}$, via some technology, $k$ features are selected to train a specific model $\mathcal{M}$ and based on the performance, it is possible to approximate the performance of the model $\mathcal{M}$ on the whole dataset $\mathcal{D}$ with all features. 
There is one reasonable situation for the feature extraction technique that the accuracy of the model will increase as the number of features that are selected by the technology increases. 
Hence it is fundamental to define this situation strictly. 
Before the formal definitions, it is essential to elaborate on some symbols:
\begin{itemize}
    \item $\mathcal{D}$ is a dataset that has $d$ features.
    \item $\mathcal{FE}$ is an algorithm of feature extraction that can select $k$ features of the dataset $\mathcal{D}$
    \item $\mathcal{M}$ is the machine learning classification model used on the dataset $\mathcal{D}$. 
    \item $\textit{Accuracy}(\mathcal{M},\mathcal{D})$ represents the test accuracy of the model $\mathcal{M}$ for the dataset $\mathcal{D}$. 
    \item $\mathcal{D}^i_{\mathcal{FE}}$ means $\mathcal{D}$ with only $k_i$ features that are selected at the $i^{th}$ time by the algorithm $\mathcal{FE}$.
\end{itemize}

Suppose $\mathcal{FE}$ extracts features $n$ times and each time the number of selected features increases $\delta$, sometimes $\delta = \frac{d}{n}$(rounding up), until it equals to $d$. 
Therefore, the number of features selected at the $i^{th}$ time is $k_i = k_{i-1} + \delta$ and $1\leq k_i\leq d$, where $i$ is an integer and $1\leq i\leq n$. 
On top of this, the following definition \ref{def:fea_ext}, \ref{def:fea_ext_rho} and \ref{def:fea_ext_alpha} are given.

\begin{definition}[Monotone Increasing Feature Extractor]
\label{def:fea_ext}
An algorithm $\mathcal{FE}$ for feature extraction is called a monotone increasing feature extractor for the model $\mathcal{M}$ and the dataset $\mathcal{D}$, 
if $Accuracy (\mathcal{M},\mathcal{D}^{i}_{\mathcal{FE}}) \ge Accuracy(\mathcal{M},\mathcal{D}^{i-1}_{\mathcal{FE}})$, $\forall\ i \in \{2,\cdots,n\}$, which is the number of times to run $\mathcal{FE}$.
\end{definition}

This definition may be too strict to allow most feature extraction algorithms to satisfy the conditions. 
Therefore, a loose definition is preferred. Beforehand, a new parameter $t_i$ is introduced to count the cases where the more features are selected and the higher accuracy is given, that is $\forall\ i \in \{2,\cdots,n\}$

$$ t_i=\left\{
\begin{array}{rcl}
1       &      & {,\ if\  Accuracy(\mathcal{M},\mathcal{D}^{i}_{\mathcal{FE}}) \ge Accuracy(\mathcal{M},\mathcal{D}^{i-1}_{\mathcal{FE}})}\\
0       &      & {,\ otherwise}
\end{array} \right. $$

Hereafter, an alternative definition is given.

\begin{definition}[$\rho$--Increasing Feature Extractor]
\label{def:fea_ext_rho}
An algorithm $\mathcal{FE}$ for feature extraction is called a $\rho$--increasing feature extractor for the model $\mathcal{M}$ and the dataset $\mathcal{D}$, for some $0< \rho < 1$, where 
$$\rho \leq \frac{\sum_{i=2}^n t_i}{n}$$ 
\end{definition}

Ideally, having a monotone increasing feature extractor or a $\rho$-monotonic feature extractor for which $\rho$ is close to one is helpful to estimate, with what features, the satisfactory accuracy can be achieved. 
Furthermore, in order to draw a comprehensive conclusion about a feature extraction technique, here is another definition.

\begin{definition}[$(\alpha,\rho)$--Increasing Feature Extractor]
\label{def:fea_ext_alpha}
Let $\mathcal{A}$ be a repository of datasets. 
$\mathcal{FE}$ is an $(\alpha,\rho)$--increasing feature extractor for the model $\mathcal{M}$ and repository $\mathcal{A}$ 
if $\mathcal{FE}$ is $\rho$--increasing feature extractor for more than $\alpha$ of all datasets in $\mathcal{A}$ with $\mathcal{M}$
that is for some $\alpha < 1$, 
$$\alpha = \frac{number\ of\ times\ \mathcal{FE}\ is\ \rho-increasing\ feature\ extractor}{number\ of\ datasets}$$
\end{definition}

With these definitions, especially definition \ref{def:fea_ext_rho} and \ref{def:fea_ext_alpha}, it is practicable to decide if a technology for feature extraction is predictable. 
That is, for a feature extraction algorithm, the closer ($\alpha$, $\rho$) is to (1, 1), the more predictable it is, and the easier it is to select features that are used to train the model achieving satisfactory accuracy, which can be used to answer question\ref{que:best_model} in \nameref{chap:intro}. 
All datasets are used  as $\mathcal{A}$ in experiments listed in Table \ref{tab:binary_datasets} and \ref{tab:multi_datasets}, and the classifier MLP with 4 layers and 128 nodes per layer is used as $\mathcal{M}$. 
Besides, there are table \ref{tab:fea_ext_sum_rho_rf_xgb} and \ref{tab:fea_ext_sum_rho_sp_pca} in section \ref{subsec:fea_ext_sum} to summarize $\rho$ values for all datasets. 
Another table \ref{tab:fea_ext_sum_obs} is also put in \nameref{subsec:fea_ext_sum} to show if there is an observed pattern.

%------------------------------------------------------------------
%------------------------------------------------------------------
\subsubsection{Permutation Importance of Random Forest}
\label{subsec:fea_rf}
\begin{figure}[ht]
    \centering
    \includegraphics[width=0.4\linewidth]{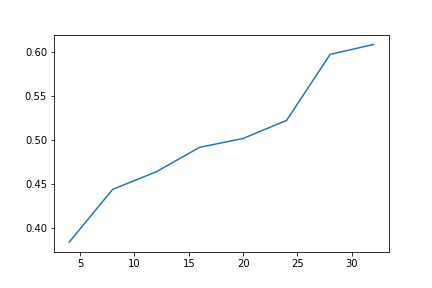}
    \caption{The performance of feature extraction based upon permutation importance on dataset 4538}
    \label{fig:fea_ext_rf_4538}
\end{figure}

To decide feature importance, a popular method is through permutation importance, which shows the drop in the score when the feature would be replaced with randomly permuted values. 
Firstly, still consider dataset 4538 as an instance shown in figure \ref{fig:fea_ext_rf_4538}, where x-axis is the number of selected features and y-axis is the accuracy of MLP with 4 layers and 128 nodes per layer trained and tested with the dataset 4538 with only selected features. 
According to the definition \ref{def:fea_ext_rho} for \nameref{def:fea_ext_rho}, feature extraction by permutation importance is a $\frac{7}{8}$--increasing extractor, which is the best it can be because the supremum of $\rho$ is $\frac{7}{8}$ under this situation. 
The given definition is consistent with what is observed. 

During the experiment, the function \texttt{sklearn.inspection.permutation\_importance} is used, where the parameter \texttt{n\_repeats} is set to 5 to guarantee the robustness of the results\cite{sklearn_api, scikit-learn}. 
Then, with the sorted importance, which is averaged, features are selected by increasing the $10\%$ of the total amount each time, namely for dataset 4538, there are $4, 8, 12, \cdots, 32$ features extracted in sequence. 
Through this method, the feature extractor is implemented and the MLP model is trained and tested over the dataset with different number of features.

\begin{figure}[H]
    \centering
    \begin{subfigure}[b]{0.24\textwidth} 
        \centering
        \includegraphics[width = \linewidth]{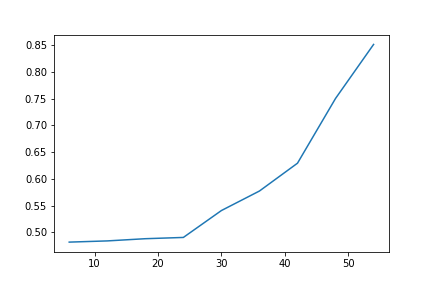}
        \caption{180}
    \end{subfigure}
    % \begin{subfigure}[b]{0.24\textwidth} 
    %     \centering
    %     \includegraphics[width = \linewidth]{Figures/imp_per_rf/734_5_forest_mlp_4_128.png}
    %     \caption{734}
    % \end{subfigure}
    \begin{subfigure}[b]{0.24\textwidth} 
        \centering
        \includegraphics[width = \linewidth]{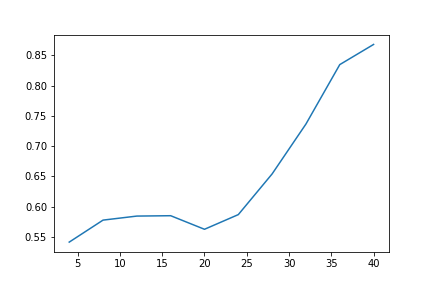}
        \caption{979}
    \end{subfigure}
    % \begin{subfigure}[b]{0.24\textwidth} 
    %     \centering
    %     \includegraphics[width = \linewidth]{Figures/imp_per_rf/1443_5_forest_mlp_4_128.png}
    %     \caption{1443}
    % \end{subfigure}
    \begin{subfigure}[b]{0.24\textwidth} 
        \centering
        \includegraphics[width = \linewidth]{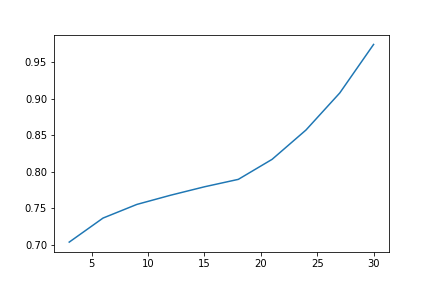}
        \caption{4534}
    \end{subfigure}
    
    \begin{subfigure}[b]{0.24\textwidth}
        \centering
        \includegraphics[width = \linewidth]{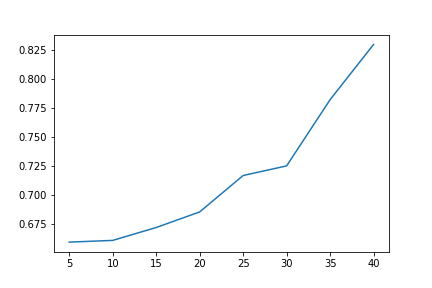}
        \caption{40668}
    \end{subfigure}
    \begin{subfigure}[b]{0.24\textwidth}
        \centering
        \includegraphics[width = \linewidth]{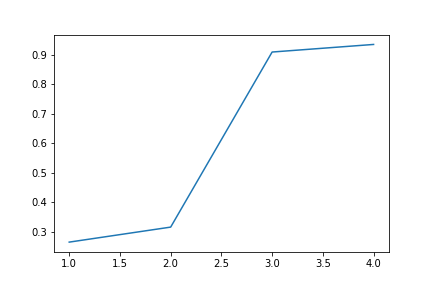}
        \caption{41014}
    \end{subfigure}
    \begin{subfigure}[b]{0.24\textwidth} 
        \centering
        \includegraphics[width = \linewidth]{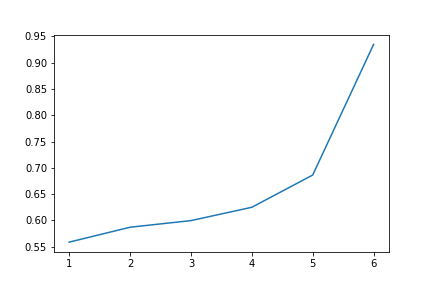}
        \caption{41027}
    \end{subfigure}
    \caption{Part of increasing results for extractor with permutation importance and highest $\rho$ values}
    \label{fig:imp_rf_both}
\end{figure}

\begin{figure}[H]
    \centering
    \begin{subfigure}[b]{0.24\textwidth} 
        \centering
        \includegraphics[width = \linewidth]{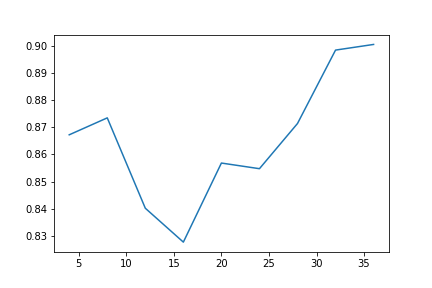}
        \caption{1049}
    \end{subfigure}
    \begin{subfigure}[b]{0.24\textwidth}
        \centering
        \includegraphics[width = \linewidth]{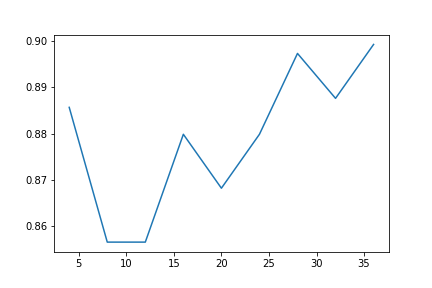}
        \caption{1050}
    \end{subfigure}
    \begin{subfigure}[b]{0.24\textwidth}
        \centering
        \includegraphics[width = \linewidth]{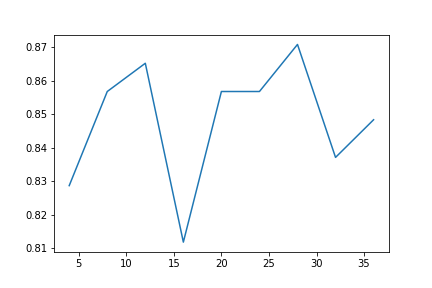}
        \caption{1453}
    \end{subfigure}
    
    \begin{subfigure}[b]{0.24\textwidth} 
        \centering
        \includegraphics[width = \linewidth]{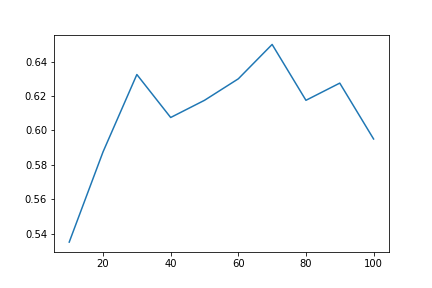}
        \caption{1479}
    \end{subfigure}
    \begin{subfigure}[b]{0.24\textwidth}
        \centering
        \includegraphics[width = \linewidth]{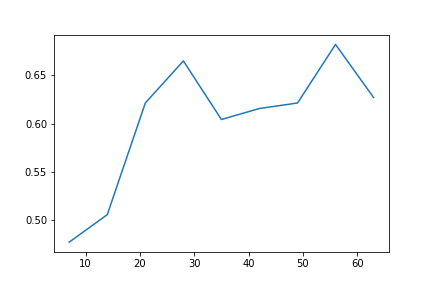}
        \caption{1492}
    \end{subfigure}
    \begin{subfigure}[b]{0.24\textwidth}
        \centering
        \includegraphics[width = \linewidth]{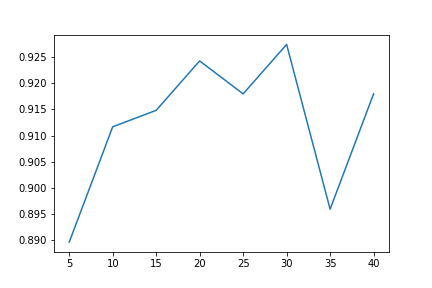}
        \caption{40705}
    \end{subfigure}
    \caption{All non-increasing results of extractor with permutation importance and highest $\rho$ values}
    \label{fig:imp_rf_def}
\end{figure}

This ranking method works fairly well that is out of 54 datasets, there are 23 datasets, on which this extractor has highest $\rho$ value, part of them are shown in figure \ref{fig:imp_rf_both} and \ref{fig:imp_rf_def}. 
Combined with direct observation of the experimental results, there are both times when the observation conforms to the definition and times when it does not.

%------------------------------------------------------------------
%------------------------------------------------------------------

\begin{figure}[H]
    \centering
    \begin{subfigure}[b]{0.24\textwidth} 
        \centering
        \includegraphics[width = \linewidth]{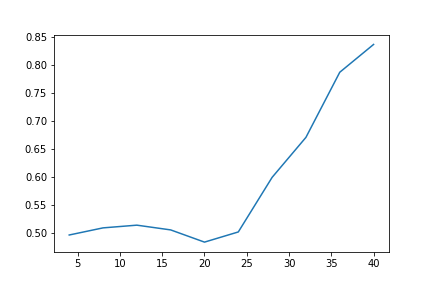}
        \caption{60}
    \end{subfigure}
    \begin{subfigure}[b]{0.24\textwidth} 
        \centering
        \includegraphics[width = \linewidth]{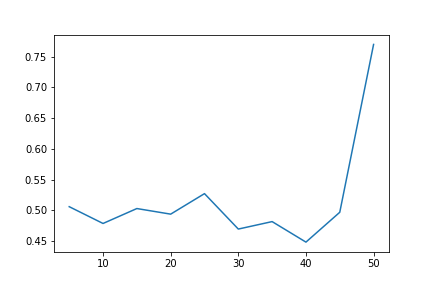}
        \caption{904}
    \end{subfigure}
    \begin{subfigure}[b]{0.24\textwidth}
        \centering
        \includegraphics[width = \linewidth]{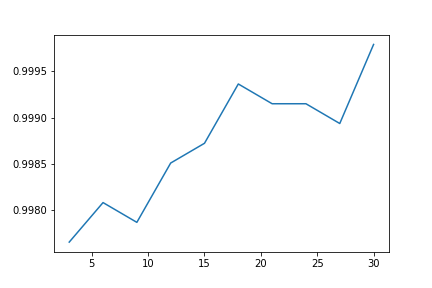}
        \caption{4154}
    \end{subfigure}
    \begin{subfigure}[b]{0.24\textwidth} 
        \centering
        \includegraphics[width = \linewidth]{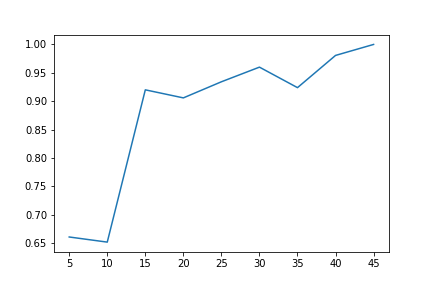}
        \caption{40999}
    \end{subfigure}
    \caption{Part of increasing results of extractor with permutation importance but not highest $\rho$ values}
    \label{fig:imp_rf_obs}
\end{figure}

Generally speaking, if the feature extractor has the highest $\rho$ value on a data set, the monotonically increasing pattern will be discovered by observing the behavior of it on this dataset like results shown in figure \ref{fig:imp_rf_both}. 
However, there are some exceptions, including datasets 1049, 1050, 1453, 1479, 1492, and 40705, shown in figure \ref{fig:imp_rf_def}, where the pattern can be hardly observed even though the technique has the highest $\rho$ value on them. 
Vice versa, on 13 datasets, part of which are shown in figure \ref{fig:imp_rf_obs}, the pattern can be perceived even though the highest $\rho$ value on the dataset is not achieved by this method. 

Overall, we observed a total of 30 datasets on which the feature extractor using permutation importance has monotonically increasing pattern, which can be inferred that this extraction technology is reasonable. 
Other than that, datasets that neither have significant $\rho$ value nor can be observed in increasing pattern are out of the scope of discussion in the paper.

%------------------------------------------------------------------
%------------------------------------------------------------------
\subsubsection{Gain-Based Feature Importance of XGBoost}
\label{subsec:fea_xgb}
\begin{figure}[ht]
    \centering
    \includegraphics[width=0.5\linewidth]{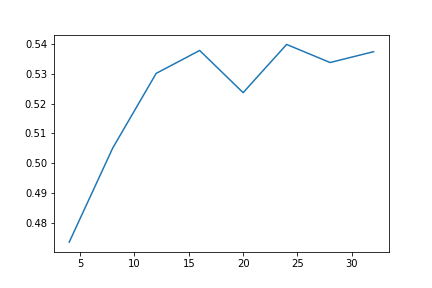}
    \caption{The performance of feature extraction based upon gain-based importance on dataset 4538}
    \label{fig:fea_ext_xgb_4538}
\end{figure}

Another method to decide feature importance is using gain-based feature importance based on XGBoost algorithm. 
Take dataset 4538 as an illustration shown in figure \ref{fig:fea_ext_xgb_4538}, which is considered as an increasing pattern but does not have the highest $\rho$ value. 
Due to the definition\ref{def:fea_ext_rho} of \nameref{def:fea_ext_rho}, feature extraction with gain-based importance is a $\frac{4}{7}$--increasing extractor.

In the course of experiments, what is utilized to select features is the method from TensorFlow, \texttt{experimental\_feature\_importances} of the function \texttt{tf.estimator.BoostedTreesClassifier}, where the parameter \texttt{normalize} is set to \textit{True} to normalize the feature importance\cite{tensorflow2015-whitepaper}. 
For detailed information, please check \hyperlink{https://www.tensorflow.org/api\_docs/python/tf/estimator/BoostedTreesClassifier\#experimental\_feature\_importances}{this link}. 
The following steps are as same as experiments of the extractor using permutation features, stated in section \ref{subsec:fea_rf} \nameref{subsec:fea_rf}.

\begin{figure}[ht]
    \centering
    \includegraphics[width=0.5\linewidth]{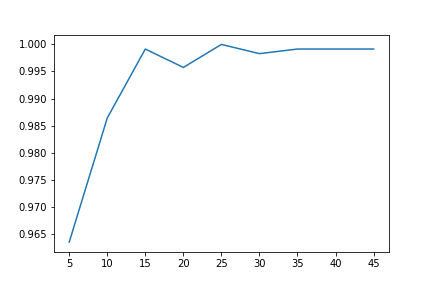}
    \caption{Only dataset 41000 increasing for extractor with gain-based importance and highest $\rho$ value}
    \label{fig:fea_ext_xgb_both}
\end{figure}

\begin{figure}[H]
    \centering
    \begin{subfigure}[b]{0.24\textwidth} 
        \centering
        \includegraphics[width = \linewidth]{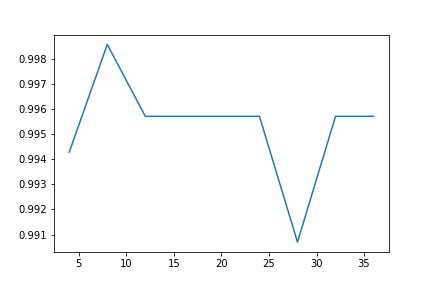}
        \caption{1069}
    \end{subfigure}
    \begin{subfigure}[b]{0.24\textwidth} 
        \centering
        \includegraphics[width = \linewidth]{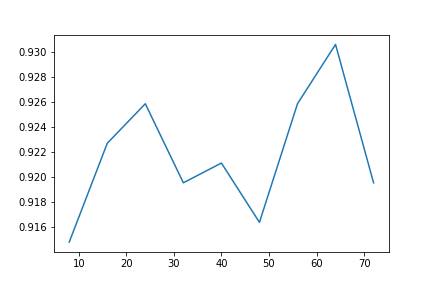}
        \caption{1487}
    \end{subfigure}
    \begin{subfigure}[b]{0.24\textwidth}
        \centering
        \includegraphics[width = \linewidth]{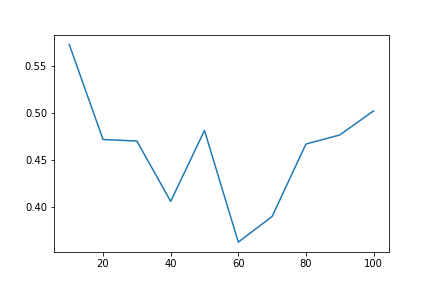}
        \caption{1548}
    \end{subfigure}
    
    \begin{subfigure}[b]{0.24\textwidth} 
        \centering
        \includegraphics[width = \linewidth]{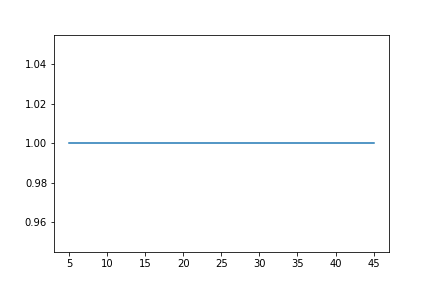}
        \caption{40999}
    \end{subfigure}
    \centering
    \begin{subfigure}[b]{0.24\textwidth} 
        \centering
        \includegraphics[width = \linewidth]{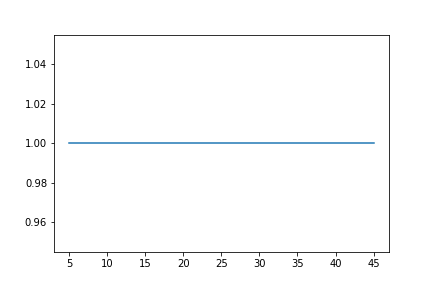}
        \caption{41005}
    \end{subfigure}
    \caption{All non-increasing results of extractor with gain-based importance and highest $\rho$ values}
    \label{fig:fea_ext_xgb_def}
\end{figure}

According to empirical results, this method is far less interpretable than method with permutation importance as a kernel because there is only one dataset, 41000 illustrated in figure \ref{fig:fea_ext_xgb_both}, on which the extractor has the highest $\rho$ value and can be observed having a increasing pattern. 
Besides, except 41000, there are only 6 other datasets 1069, 1487, 1548, 40999, and 41005 having the highest $\rho$ value, which however cannot be regarded as having reasonable pattern because the lines are either strongly wavy or deadly waveless. 
These results are shown in figure \ref{fig:fea_ext_xgb_def}.

%------------------------------------------------------------------
%------------------------------------------------------------------

\begin{figure}[H]
    \centering
    \begin{subfigure}[b]{0.24\textwidth} 
        \centering
        \includegraphics[width = \linewidth]{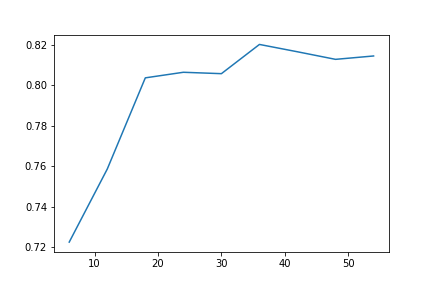}
        \caption{180}
    \end{subfigure}
    \begin{subfigure}[b]{0.24\textwidth} 
        \centering
        \includegraphics[width = \linewidth]{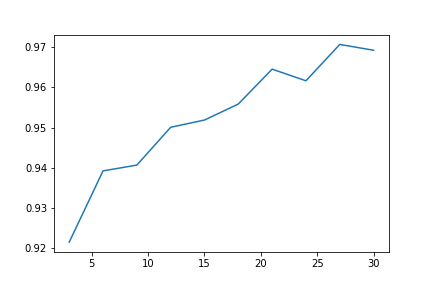}
        \caption{4534}
    \end{subfigure}
    \begin{subfigure}[b]{0.24\textwidth}
        \centering
        \includegraphics[width = \linewidth]{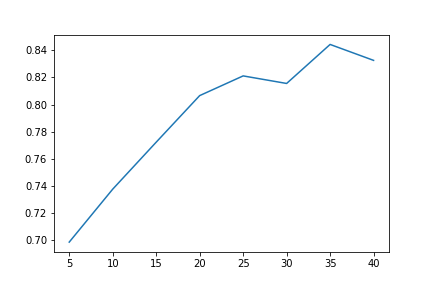}
        \caption{40668} 
    \end{subfigure}
    
    \begin{subfigure}[b]{0.24\textwidth} 
        \centering
        \includegraphics[width = \linewidth]{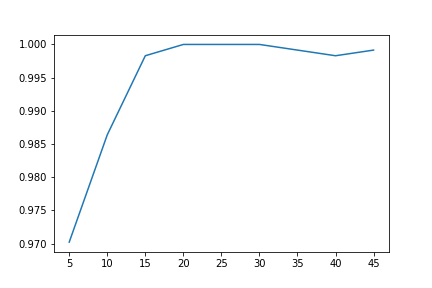}
        \caption{40997}
    \end{subfigure}
    \begin{subfigure}[b]{0.24\textwidth} 
        \centering
        \includegraphics[width = \linewidth]{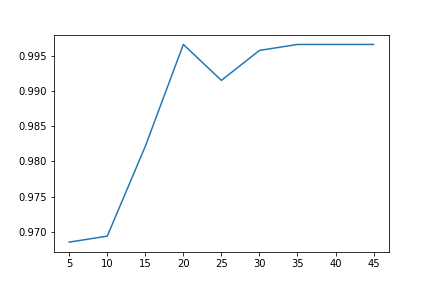}
        \caption{41004}
    \end{subfigure}
    \begin{subfigure}[b]{0.24\textwidth} 
        \centering
        \includegraphics[width = \linewidth]{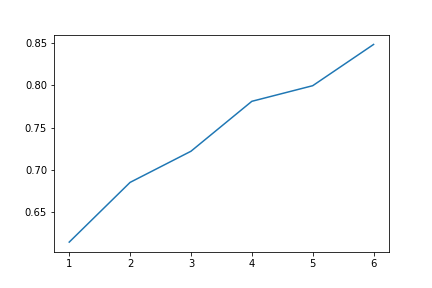}
        \caption{41027}
    \end{subfigure}
    \caption{All increasing results of extractor with gain-based importance but not highest $\rho$ values}
    \label{fig:fea_ext_xgb_obs}
\end{figure}

Even if it is not judged by the number of highest $\rho$ values it has, only 8 out of 54 datasets, including 180, 4534, 4538, 40668, 40997, 41000, 41004, and 41027, are observed increasing performance with the increase of feature number, all revealed in figure \ref{fig:fea_ext_xgb_4538}, \ref{fig:fea_ext_xgb_both}, and \ref{fig:fea_ext_xgb_obs}.

In general, the feature extraction technique with gain-based feature importance of XGBoost algorithm is very weak in the aspect to predictability and interpretability because most extractors are not with high $\rho$ values and cannot cause an increasing tendency when increasing the number of features.

%------------------------------------------------------------------
%------------------------------------------------------------------
\subsubsection{Hierarchical Clustering Based on Spearman Correlation}
\label{subsec:fea_spearman}
\begin{figure}[H]
    \centering
    \begin{subfigure}[b]{0.55\textwidth} 
        \centering
        \includegraphics[width= \linewidth]{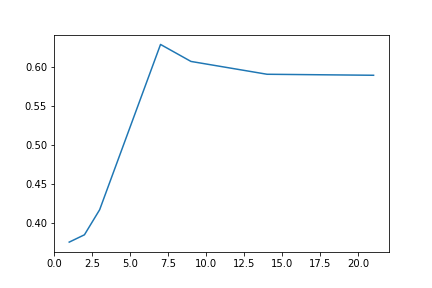}
        \caption{Feature extraction}
    \end{subfigure}
    
    \begin{subfigure}[b]{0.45\textwidth} 
        \centering
        \includegraphics[width = \linewidth]{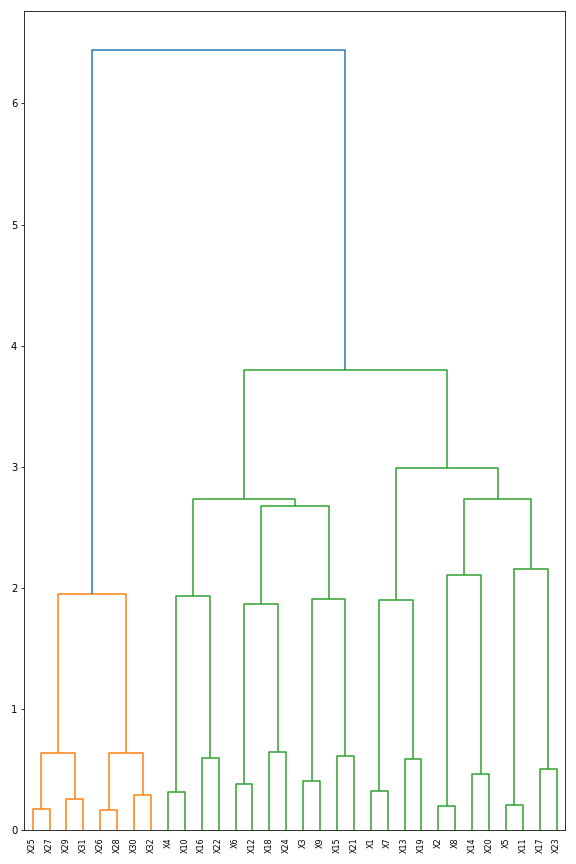}
        \caption{Dendrogram}
    \end{subfigure}
    \caption{The performance of feature extraction based upon Spearman correlation on dataset 4538}
    \label{fig:fea_ext_sp_4538}
\end{figure}

In addition to two methods mentioned in \ref{subsec:fea_rf} \nameref{subsec:fea_rf} and \ref{subsec:fea_xgb} \nameref{subsec:fea_xgb}, there is a method to extract features decided by hierarchical clustering based upon Spearman correlation. 
As a convention, 4538 is still introduced as an example, shown in figure \ref{fig:fea_ext_sp_4538}, which can be regarded as a predictable pattern though with only $\rho = \frac{3}{7}$, which means that in less than half the time, increasing the number of features can increase the accuracy.

% \begin{figure}[ht]
%     \centering
%     \includegraphics[width=0.5\linewidth]{Figures/imp_spearman/4538_cluster5_mlp.png}
%     \caption{The performance of feature extraction based upon Spearman correlation on dataset 4538}
%     \label{fig:fea_ext_sp_4538}
% \end{figure}

\begin{figure}[H]
    \centering
    \begin{subfigure}[b]{0.3\textwidth} 
        \centering
        \includegraphics[width = \linewidth]{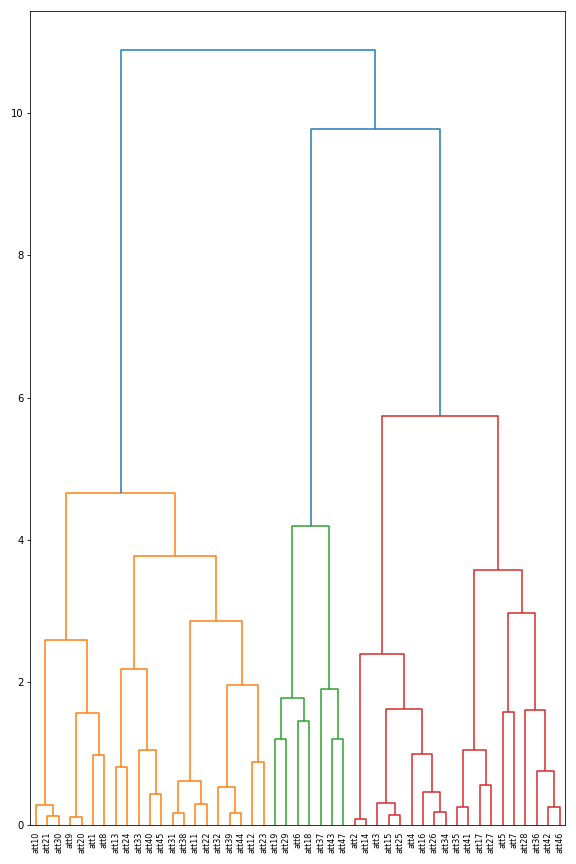}
        \caption{22}
    \end{subfigure}
    \begin{subfigure}[b]{0.3\textwidth} 
        \centering
        \includegraphics[width = \linewidth]{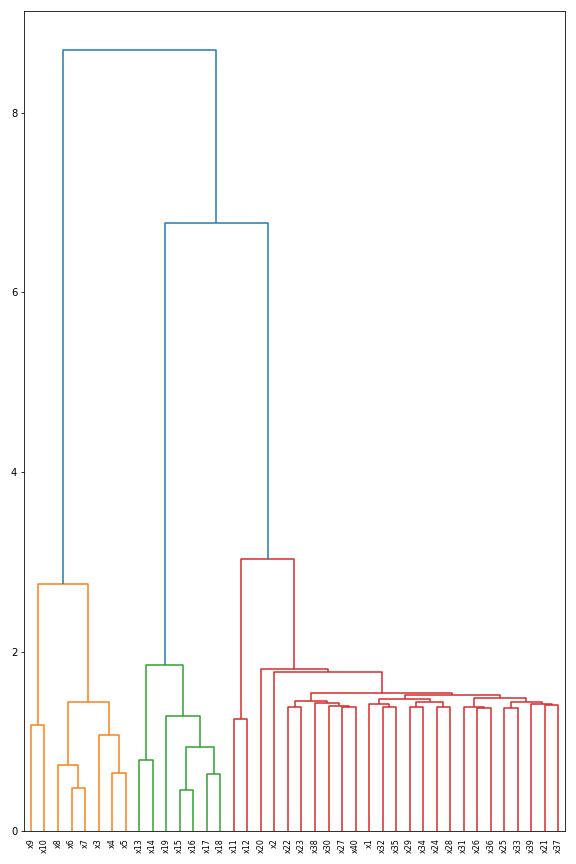}
        \caption{60}
    \end{subfigure}
    \begin{subfigure}[b]{0.3\textwidth}
        \centering
        \includegraphics[width = \linewidth]{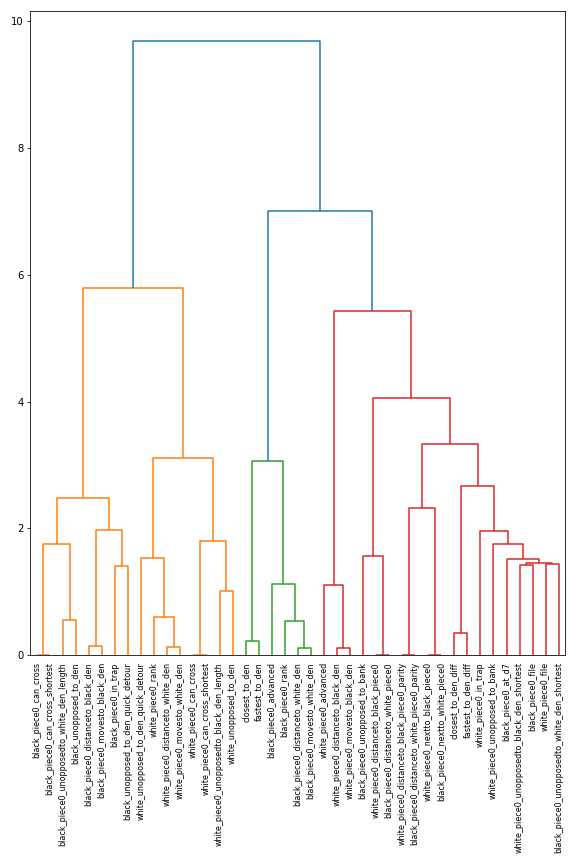}
        \caption{41007}
    \end{subfigure}
    \caption{Three examples of dendrogram}
    \label{fig:fea_ext_sp_dendrogram}
\end{figure}

Through experiments, function \texttt{scipy.stats.spearmanr} and \texttt{scipy.cluster.hierarchy.ward} from \texttt{scipy} are utilized for the calculation of Spearman correlation and the generation of hierarchical clustering\cite{2020SciPy-NMeth}. 
The parameters of these functions are not changed. 
Usually, to select features by hierarchical clustering, it is essential to look through related dendrograms and decide which features should be selected with the level of dendrograms, some of which are exampled in figure \ref{fig:fea_ext_sp_dendrogram}, where x-axis is the name of features, y-axis is the level, and colours symbolize different clusters. 
However, for automatic processing, going through the level from $0.5$ to $3$ plus $0.5$ each time and from $3$ to $12$ plus $1$ each time is enough to cover most possible level, that is selecting suitable numbers in each step. 
with this method the result of clustering is stable so that it is not necessary to repeat. 
After selection, experimental steps are as same as mentioned in section \ref{subsec:fea_rf}.

\begin{figure}[H]
    \centering
    \begin{subfigure}[b]{0.24\textwidth} 
        \centering
        \includegraphics[width = \linewidth]{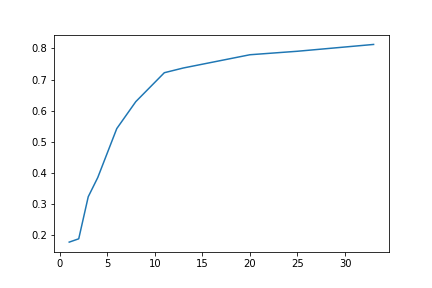}
        \caption{22}
    \end{subfigure}
    \begin{subfigure}[b]{0.24\textwidth} 
        \centering
        \includegraphics[width = \linewidth]{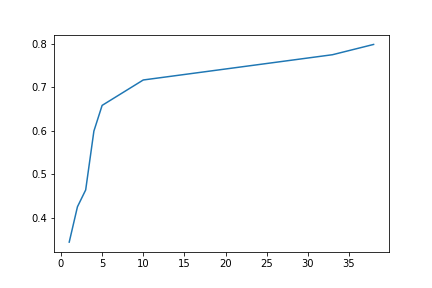}
        \caption{60}
    \end{subfigure}
    \begin{subfigure}[b]{0.24\textwidth} 
        \centering
        \includegraphics[width = \linewidth]{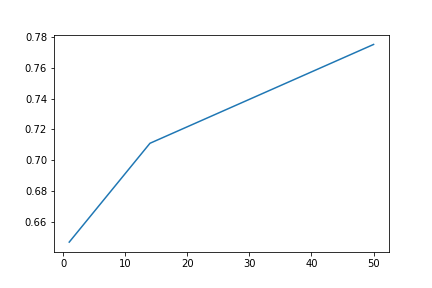}
        \caption{904}
    \end{subfigure}
    \begin{subfigure}[b]{0.24\textwidth}
        \centering
        \includegraphics[width = \linewidth]{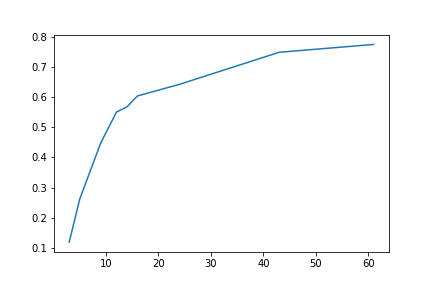}
        \caption{1491}
    \end{subfigure}
    
    \centering
    \begin{subfigure}[b]{0.24\textwidth} 
        \centering
        \includegraphics[width = \linewidth]{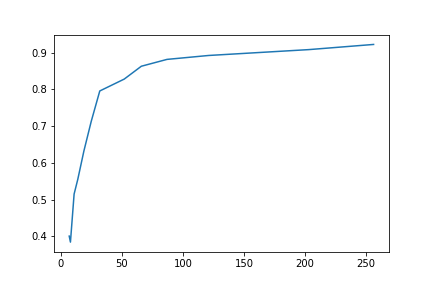}
        \caption{1501}
    \end{subfigure}
    \begin{subfigure}[b]{0.24\textwidth}
        \centering
        \includegraphics[width = \linewidth]{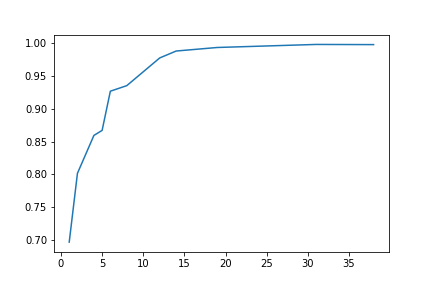}
        \caption{40997}
    \end{subfigure}
    \begin{subfigure}[b]{0.24\textwidth} 
        \centering
        \includegraphics[width = \linewidth]{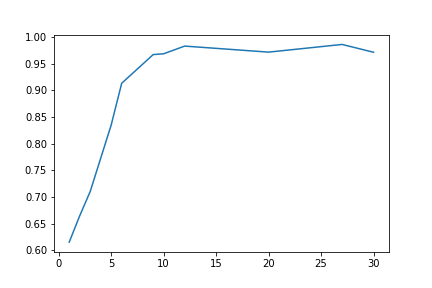}
        \caption{41007}
    \end{subfigure}
    \begin{subfigure}[b]{0.24\textwidth} 
        \centering
        \includegraphics[width = \linewidth]{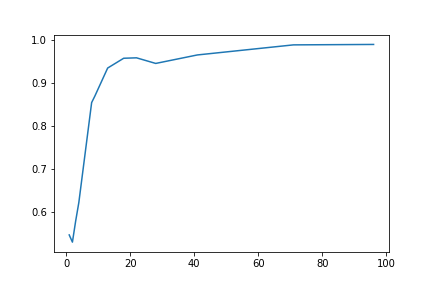}
        \caption{41048}
    \end{subfigure}
    \caption{Part of increasing results for extractor with Spearman correlation and highest $\rho$ values}
    \label{fig:fea_ext_sp_both}
\end{figure}

\begin{figure}[H]
    \centering
    \begin{subfigure}[b]{0.24\textwidth} 
        \centering
        \includegraphics[width = \linewidth]{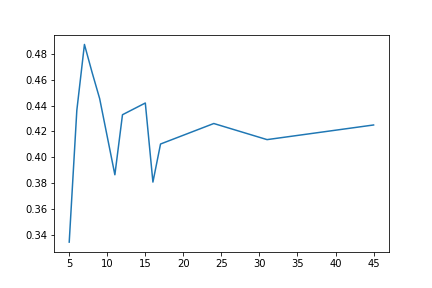}
        \caption{313}
    \end{subfigure}
    \begin{subfigure}[b]{0.24\textwidth} 
        \centering
        \includegraphics[width = \linewidth]{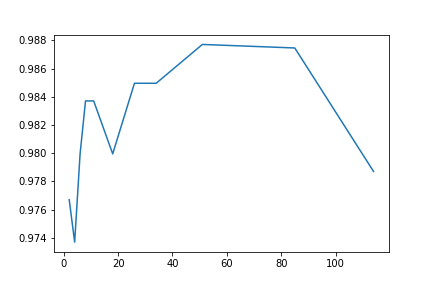}
        \caption{316}
    \end{subfigure}
    \begin{subfigure}[b]{0.24\textwidth}
        \centering
        \includegraphics[width = \linewidth]{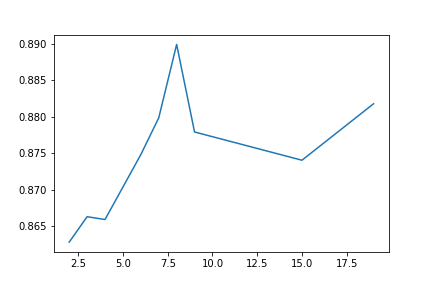}
        \caption{1050}
    \end{subfigure}
    
    \centering
    \begin{subfigure}[b]{0.24\textwidth} 
        \centering
        \includegraphics[width = \linewidth]{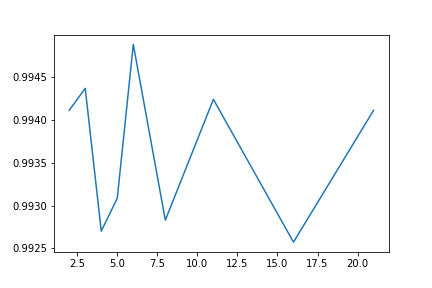}
        \caption{1056}
    \end{subfigure}
    \begin{subfigure}[b]{0.24\textwidth}
        \centering
        \includegraphics[width = \linewidth]{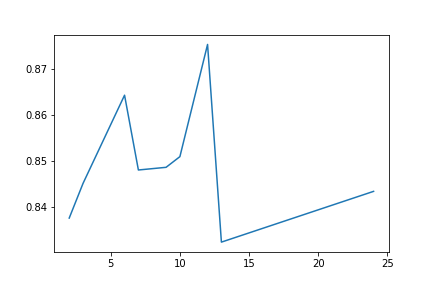}
        \caption{1444}
    \end{subfigure}
    \begin{subfigure}[b]{0.24\textwidth} 
        \centering
        \includegraphics[width = \linewidth]{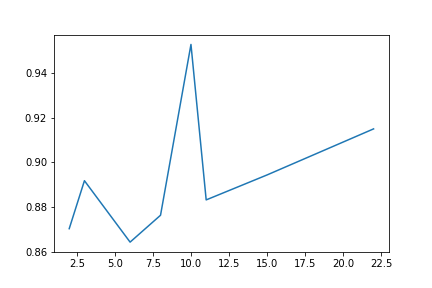}
        \caption{1451}
    \end{subfigure}
    \caption{All non-increasing results for extractor with Spearman correlation and highest $\rho$ values}
    \label{fig:fea_ext_sp_def}
\end{figure}

From the view of definition \ref{def:fea_ext_rho}, the extractor with Spearman correlation is the second best, which achieves the highest $\rho$--value on 21 out of 54 datasets. 
On dataset 22, it is a $\frac{10}{11}$--increasing extractor, the best one among all extractors no matter with permutation importance, gain-based importance, or PCA, which is illustrated in section \ref{subsec:fea_ext_pca}. 
Hence, it can be seen as the smoothest increasing pattern as well. 
Other results with highest $\rho$ values are partly shown in figure \ref{fig:fea_ext_sp_both} and \ref{fig:fea_ext_sp_def}.

In figure \ref{fig:fea_ext_sp_def}, there are only 6 datasets including 313, 316, 1050, 1056, 1444, and 1451 that cannot be observed as increasing pattern but with the highest $\rho$ value, namely that it is hardly to see the observation opposing to the definition.

\begin{figure}[H]
    \centering
    \begin{subfigure}[b]{0.24\textwidth} 
        \centering
        \includegraphics[width = \linewidth]{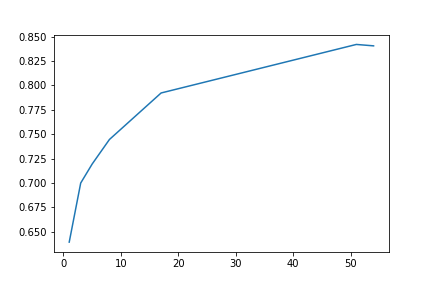}
        \caption{180}
    \end{subfigure}
    \begin{subfigure}[b]{0.24\textwidth} 
        \centering
        \includegraphics[width = \linewidth]{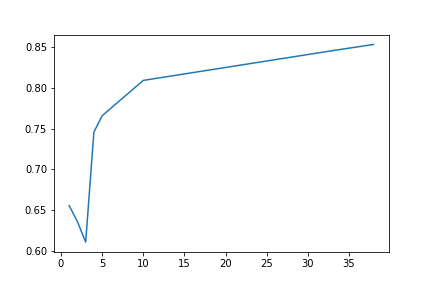}
        \caption{979}
    \end{subfigure}
    \begin{subfigure}[b]{0.24\textwidth}
        \centering
        \includegraphics[width = \linewidth]{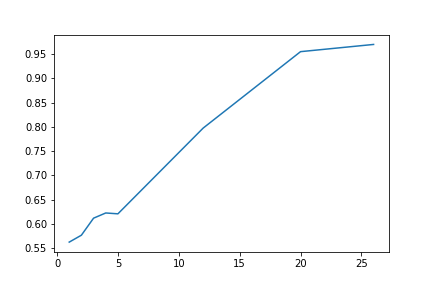}
        \caption{4534}
    \end{subfigure}
    \begin{subfigure}[b]{0.24\textwidth} 
        \centering
        \includegraphics[width = \linewidth]{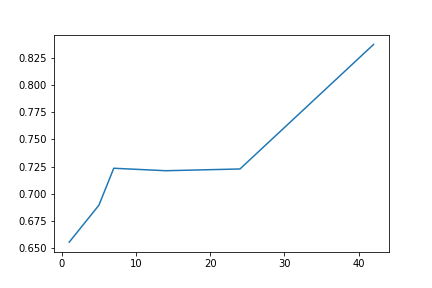}
        \caption{40668}
    \end{subfigure}
    
    \centering
    \begin{subfigure}[b]{0.24\textwidth} 
        \centering
        \includegraphics[width = \linewidth]{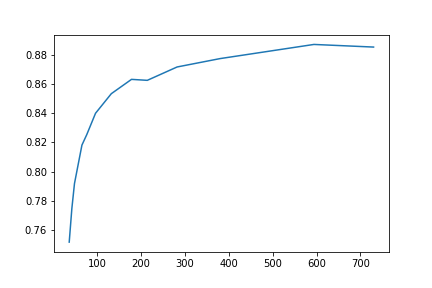}
        \caption{40996}
    \end{subfigure}
    \begin{subfigure}[b]{0.24\textwidth}
        \centering
        \includegraphics[width = \linewidth]{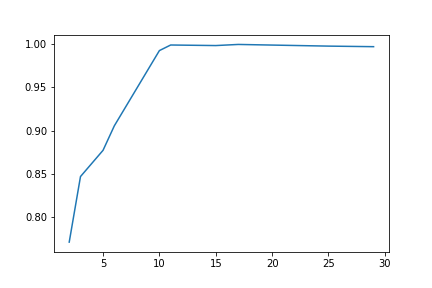}
        \caption{41000}
    \end{subfigure}
    \begin{subfigure}[b]{0.24\textwidth} 
        \centering
        \includegraphics[width = \linewidth]{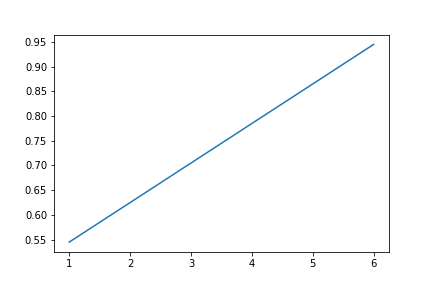}
        \caption{41027}
    \end{subfigure}
    \begin{subfigure}[b]{0.24\textwidth} 
        \centering
        \includegraphics[width = \linewidth]{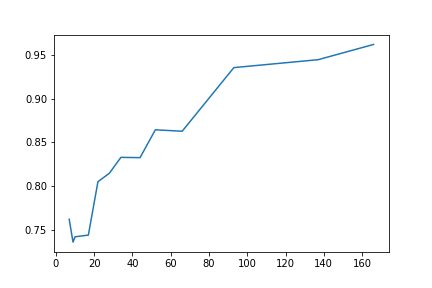}
        \caption{41052}
    \end{subfigure}
    \caption{Part of increasing results for extractor with Spearman correlation but not highest $\rho$ values}
    \label{fig:fea_ext_sp_obs}
\end{figure}

Additionally, it can be detected that the accuracy of the MLP model on 37 out of 54 datasets has an increasing pattern with the increase of the number of features used in the model, which are extracted by the extractor utilizing hierarchical clustering based on Spearman correlation. 
Hence, regarding the observation, this extractor is the best one with over $68\%$ of datasets showing an increasing pattern even though it does not have the highest $\rho$ value on these datasets, some of which can be scrutinized in figure \ref{fig:fea_ext_sp_obs}.

To sum up, although feature extractor with Spearman correlation is not as satisfying as that with permutation importance in terms of definition \ref{def:fea_ext_rho} \nameref{def:fea_ext_rho}, it cannot be inferred that the former is worse than the latter. 
There are two main reasons. 
Firstly, even if the $\rho$ values of extractor with Spearman correlation and permutation importance are very similar, there can only be one highest value, which is always unique. 
Secondly, the overall trend should be considered as well, for example, it is possible that technique of permutation importance has fewer declined segments but the drop is large, while that with Spearman correlation is just the opposite, so that the observation makes the latter better than the former. 
This will be analyzed in detail in section \ref{chap:discussion} \nameref{chap:discussion}. 
In a word, feature extraction with hierarchical clustering by Spearman correlation has the best interpretability among all studied techniques according to the results of experiments

%------------------------------------------------------------------
%------------------------------------------------------------------
\subsubsection{Principal Component Analysis}
\label{subsec:fea_ext_pca}
\begin{figure}[ht]
    \centering
    \includegraphics[width=0.5\linewidth]{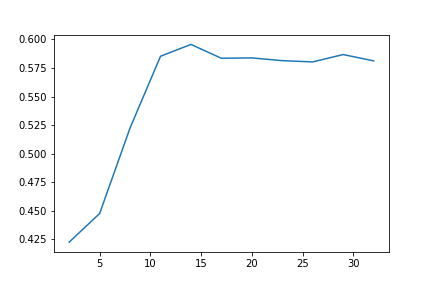}
    \caption{The performance of PCA on dataset 4538}
    \label{fig:fea_ext_pca_4538}
\end{figure}

Moreover, principal component analysis(PCA), a popular dimension reduction method, is also introduced into comparison in experimental implementation. 
Eventually, dataset 4538 has its final use, to first show how this last method PCA behaves in figure \ref{fig:fea_ext_pca_4538}. 
The performance of PCA on this dataset can be supposed with a interpretable pattern though with only $\rho = \frac{6}{11}$, which still means that there are increasing segments during the experimental process in more than half the time.

It is \texttt{sklearn.decomposition.PCA} from the library \texttt{sklearn} that the core function for PCA is deployed before training and testing the MLP model\cite{sklearn_api, scikit-learn}. 
The only parameter that needs to be changed is n\_components, which is determined by the number of features to keep each time. 
Hence, the dataset is projected into the specified number of dimensions, which increases $10\%$ of total amount as well, then train and test the classifier upon the new dataset, as described in section \ref{subsec:fea_rf}. 
It should be especially emphasized that PCA, as an 'extractor', unlike the previous three schemes can maintain the original features, it will map the entire dataset to new dimensions as introduced in section \ref{subsec:model_pca_intro}. 
However, considering the end result, they are highly similar in function, which is thought of regarding to dimensionality reduction. 
This is the reason why they are compared with each other.

\begin{figure}[ht]
    \centering
    \includegraphics[width=0.5\linewidth]{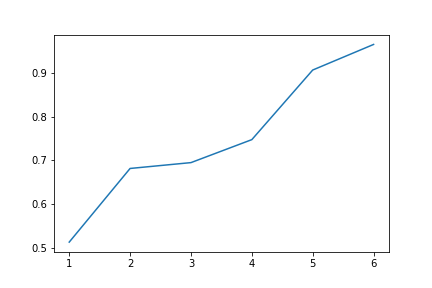}
    \caption{PCA with the highest $\rho$ value only increasing on dataset 41027}
    \label{fig:fea_ext_pca_both}
\end{figure}

According to the experiments, PCA is only better than extractor using gain-based feature importance based on XGBoost. 
Corresponding to the latter, PCA only once shows consistency of definition and observation on dataset 41027 presented in figure \ref{fig:fea_ext_pca_both}, where both PCA has $\rho = \frac{5}{6}$ reaching the supremum at the same time with the extractor based upon permutation importance.

\begin{figure}[H]
    \centering
    \begin{subfigure}[b]{0.24\textwidth} 
        \centering
        \includegraphics[width = \linewidth]{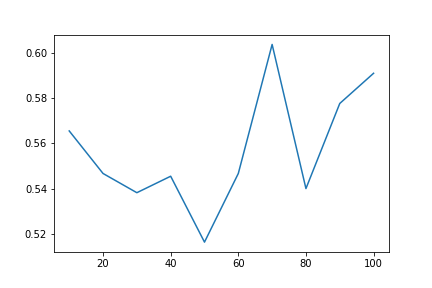}
        \caption{718}
    \end{subfigure}
    \begin{subfigure}[b]{0.24\textwidth} 
        \centering
        \includegraphics[width = \linewidth]{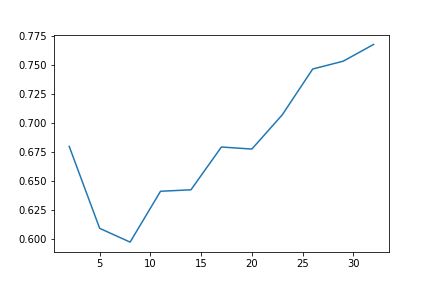}
        \caption{833}
    \end{subfigure}
    \begin{subfigure}[b]{0.24\textwidth}
        \centering
        \includegraphics[width = \linewidth]{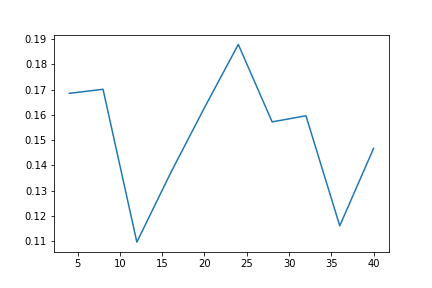}
        \caption{1549}
    \end{subfigure}
    
    \begin{subfigure}[b]{0.24\textwidth} 
        \centering
        \includegraphics[width = \linewidth]{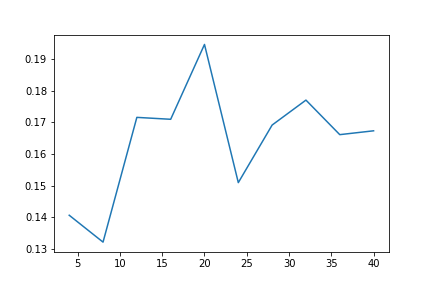}
        \caption{1555}
    \end{subfigure}
    \centering
    \begin{subfigure}[b]{0.24\textwidth} 
        \centering
        \includegraphics[width = \linewidth]{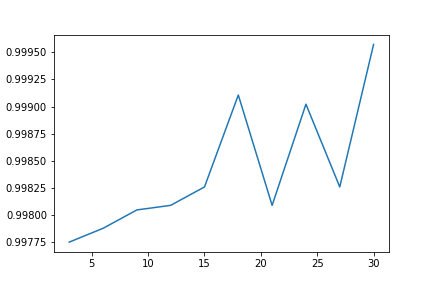}
        \caption{4154}
    \end{subfigure}
    \caption{All non-increasing results for PCA with highest $\rho$ values}
    \label{fig:fea_ext_pca_def}
\end{figure}

Similarly, except dataset 41027, there are also other 5 datasets 718, 833, 1549, 1555, and 4154 having the highest $\rho$ values but no stable increasing patterns, shown in figure \ref{fig:fea_ext_pca_def}.

\begin{figure}[H]
    \centering
    \begin{subfigure}[b]{0.24\textwidth} 
        \centering
        \includegraphics[width = \linewidth]{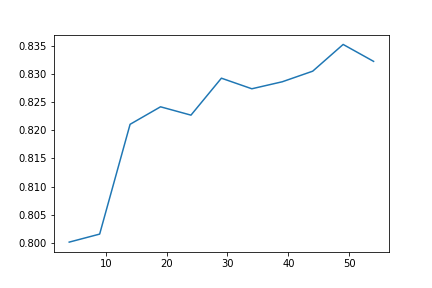}
        \caption{180}
    \end{subfigure}
    \begin{subfigure}[b]{0.24\textwidth} 
        \centering
        \includegraphics[width = \linewidth]{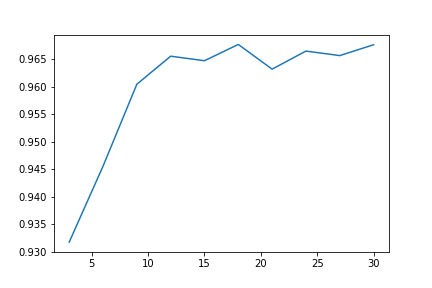}
        \caption{4534}
    \end{subfigure}
    \begin{subfigure}[b]{0.24\textwidth} 
        \centering
        \includegraphics[width = \linewidth]{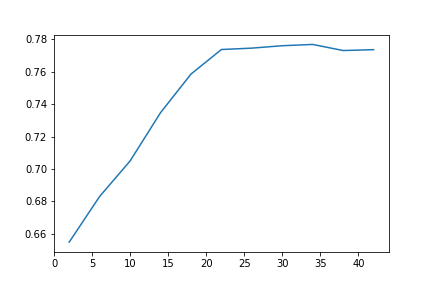}
        \caption{40668}
    \end{subfigure}
     \begin{subfigure}[b]{0.24\textwidth}
        \centering
        \includegraphics[width = \linewidth]{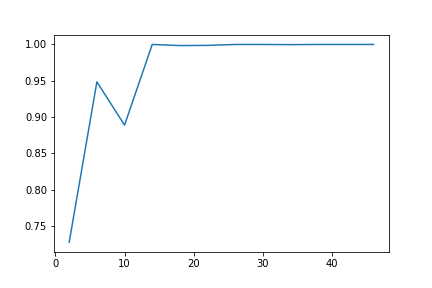}
        \caption{40999} 
    \end{subfigure}
    \caption{Part of increasing results for PCA not having highest $\rho$ values}
    \label{fig:fea_ext_pca_obs}
\end{figure}

Although by strict definition the two techniques are almost identical, through the investigation of the experimental results, PCA is actually slightly stronger than the extractor with gain-based feature importance because on 16 our of 54 datasets a general upward trend can be discovered, which is better than the case of the extractor with gain-based feature importance with only 8 datasets. 
Part of results are shown in figure \ref{fig:fea_ext_pca_obs}. 
Hence, it is advisable to assume that interpretibility of PCA is slightly stronger than the extractor based on XGBoost algorithm.

%------------------------------------------------------------------
%------------------------------------------------------------------
\subsubsection{Summary}
\label{subsec:fea_ext_sum}
To summarize the experimental results, three tables are given here. 
At first, table \ref{tab:fea_ext_sum_rho_sp_pca} and \ref{tab:fea_ext_sum_rho_rf_xgb} recapitulate the performance of different feature extraction techniques on all datasets and highlight the best one, which are divided by the best technology and ranked ascendingly by data id in each partition. 
Besides remarking the best extractor for each dataset, two tables summarize not only the $\rho$ value but also the number of trials for each dataset and extractor. 
Concretely speaking, the first column \textit{ID} still denotes the id of each dataset that can be exploited to download the dataset from OpenML and the corresponding attributes are showing in table \ref{tab:binary_datasets} and \ref{tab:multi_datasets}. 
Since similar abbreviations are used, it is necessary to specify their meanings once again:
\begin{itemize}
    \item \textit{RF} means feature extraction technique exploiting permutation importance of random forest;
    \item \textit{XGB} indicates feature extractor based on gain-based feature importance from XGBoost;
    \item \textit{SP} stands for feature extraction with hierarchical clustering based upon Spearman correlation;
    \item \textit{PCA} represents principal component analysis;
    \item $\rho$ implies the $\rho$ value in accordance with the definition; \ref{def:fea_ext_rho} that is calculated by the number of accuracy growth each time features are selected divided by total number of selections and is set to decimal for the convenience of comparison;
    \item $n$ suggests the total number of selections for a dataset.
\end{itemize}
To enhance the interpretation of each piece in the row, dataset 22 is taken as an instance, which is listed in the first position of table \ref{tab:fea_ext_sum_rho_sp_pca}. 
It can be read from it that extractor with permutation importance of random forest is experimented on dataset 22 RF\_ n times, which equals to 9, and the $\rho$ value of this extractor is $RF\_\ \rho = 0.667$, namely $\frac{6}{9}$ experiments demonstrates rising with increasing the number of features on dataset 22. 
Likewise, the extractor with gain-based feature importance from XGBoost is a $0.625$--increasing extractor, because XGB\_ $\rho = 0.625$ namely $\frac{5}{8}$, on dataset 22 after 8 experimental feature selections namely XGB\_ n $= 8$. 
The feature extraction technique with hierarchical clustering based on Spearman correlation is a $0.909$--feature extractor with extracting features 11 times and PCA is a $\frac{1}{2}$-feature extractor with 12 experiments. 
Therefore, the best extractor that is with the highest $\rho$ value is the extractor with Spearman correlation and is shown in the last column \textit{Best}.

According to the definition, that is deciding on the number of highest $\rho$ values for each extractor, the sequence is RF $>$ SP $>$ XGB $\ge$ PCA due to that RF has the highest $\rho$ value on 23 out of 54 dataset, 21 of which are exclusive. 
In contrast, SP performs best only on 21 dataset including a shared one. 
XGB and PCA accounted for 6 datasets, though PCA had one that tied for the highest $\rho$ value. 
It is also worth mentioning that there are still two datasets 40996 and 41014 cannot run XGBoost for feature extraction hence they are set to \textit{null}.

\begin{table}[H]
  \centering
  \caption{Summary of $\rho$ for techniques on datasets with Clustering and PCA scoring highest}
  \begin{adjustbox}{max width=\textwidth}
    \begin{tabular}{l|l|l|l|l|l|l|l|l|l}
    \hline \hline
    ID & RF\_\ $\rho$ & RF\_\ n & XGB\_\ $\rho$ & XGB\_\ n & SP\_\ $\rho$ & SP\_\ n & PCA\_\ $\rho$ & PCA\_\ n & Highest\_ $\rho$ \\
    \hline
    22    & 0.667 & 9     & 0.625 & 8    & 0.909 & 11     & 0.5   & 12    & SP\_\ $\rho$ \\
    60    & 0.7   & 10    & 0.556 & 9    & 0.875 & 8      & 0.2   & 10    & SP\_\ $\rho$ \\
    313   & 0.333 & 9     & 0.375 & 8    & 0.538 & 13     & 0.364 & 11    & SP\_\ $\rho$ \\
    316   & 0.444 & 9     & 0.5   & 8    & 0.545 & 11     & 0.455 & 11    & SP\_\ $\rho$ \\
    904   & 0.5   & 10    & 0.222 & 9    & 0.667 & 3      & 0.4   & 10    & SP\_\ $\rho$ \\
    1056  & 0.444 & 9     & 0.375 & 8    & 0.556 & 9      & 0.462 & 13    & SP\_\ $\rho$ \\
    1444  & 0.556 & 9     & 0.375 & 8    & 0.667 & 9      & 0.385 & 13    & SP\_\ $\rho$ \\
    1451  & 0.444 & 9     & 0.5   & 8    & 0.625 & 8      & 0.538 & 13    & SP\_\ $\rho$ \\
    1491  & 0.667 & 9     & 0.375 & 8    & 0.889 & 9      & 0.455 & 11    & SP\_\ $\rho$ \\
    1493  & 0.667 & 9     & 0.25  & 8    & 0.833 & 12     & 0.455 & 11    & SP\_\ $\rho$ \\
    1494  & 0.625 & 8     & 0.571 & 7    & 0.7   & 10     & 0.455 & 11    & SP\_\ $\rho$ \\
    1501  & 0.778 & 9     & 0.375 & 8    & 0.846 & 13     & 0.182 & 11    & SP\_\ $\rho$ \\
    40997 & 0.556 & 9     & 0.625 & 8    & 0.818 & 11     & 0.667 & 12    & SP\_\ $\rho$ \\
    41007 & 0.667 & 9     & 0.5   & 8    & 0.727 & 11     & 0.667 & 12    & SP\_\ $\rho$ \\
    41048 & 0.444 & 9     & 0.5   & 8    & 0.769 & 13     & 0.545 & 11    & SP\_\ $\rho$ \\
    41049 & 0.556 & 9     & 0.5   & 8    & 0.75  & 12     & 0.273 & 11    & SP\_\ $\rho$ \\
    41050 & 0.667 & 9     & 0.375 & 8    & 0.786 & 14     & 0.636 & 11    & SP\_\ $\rho$ \\
    41051 & 0.556 & 9     & 0.375 & 8    & 0.75  & 12     & 0.182 & 11    & SP\_\ $\rho$ \\
    41052 & 0.556 & 9     & 0.625 & 8    & 0.643 & 14     & 0.545 & 11    & SP\_\ $\rho$ \\
    41053 & 0.667 & 9     & 0.5   & 8    & 0.846 & 13     & 0.455 & 11    & SP\_\ $\rho$ \\
    \hline
    1050  & 0.556 & 9     & 0.5   & 8     & 0.462 & 13    & 0.556 & 9     & RF\_\ $\rho$, SP\_\ $\rho$ \\
    41027 & 0.833 & 6     & 0.8   & 5     & 0.833 & 6     & 0.5   & 2     & RF\_\ $\rho$, PCA\_\ $\rho$ \\
    \hline
    718   & 0.4   & 10    & 0.222 & 9     & 0.5   & 10    & 0.167 & 6     & PCA\_\ $\rho$ \\
    833   & 0.5   & 8     & 0.429 & 7     & 0.636 & 11    & 0.4   & 5     & PCA\_\ $\rho$ \\
    1549  & 0.5   & 10    & 0.444 & 9     & 0.6   & 10    & 0.333 & 3     & PCA\_\ $\rho$ \\
    1555  & 0.4   & 10    & 0.444 & 9     & 0.5   & 10    & 0.333 & 3     & PCA\_\ $\rho$ \\
    4154  & 0.6   & 10    & 0.444 & 9     & 0.7   & 10    & 0.333 & 6     & PCA\_\ $\rho$ \\
    \hline
    \end{tabular}%
  \end{adjustbox}
  \label{tab:fea_ext_sum_rho_sp_pca}%
\end{table}%

%------------------------------------------------------------------
%------------------------------------------------------------------
\begin{table}[H]
  \centering
  \caption{Summary of techniques with permutation importance and gain-based importance scoring highest}
    \begin{tabular}{l|l|l|l|l|l|l|l|l|l}
    \hline \hline
    ID & RF\_ $\rho$ & RF\_ n & XGB\_ $\rho$ & XGB\_ n & SP\_ $\rho$ & SP\_ n & PCA\_ $\rho$ & PCA\_ n & Highest\_ $\rho$ \\
    \hline
    14    & 0.778 & 9     & 0.5   & 8     & 0.556 & 9     & 0.545 & 11    & RF\_\ $\rho$ \\
    44    & 0.889 & 9     & 0.625 & 8     & 0.75  & 8     & 0.667 & 12    & RF\_\ $\rho$ \\
    180   & 0.889 & 9     & 0.5   & 8     & 0.714 & 7     & 0.636 & 11    & RF\_\ $\rho$ \\
    182   & 0.667 & 9     & 0.625 & 8     & 0.5   & 4     & 0.25  & 12    & RF\_\ $\rho$ \\
    734   & 0.6   & 10    & 0.444 & 9     & 0.571 & 7     & 0.5   & 10    & RF\_\ $\rho$ \\
    979   & 0.8   & 10    & 0.556 & 9     & 0.625 & 8     & 0.4   & 10    & RF\_\ $\rho$ \\
    1049  & 0.556 & 9     & 0.375 & 8     & 0.444 & 9     & 0.462 & 13    & RF\_\ $\rho$ \\
    1443  & 0.667 & 9     & 0.375 & 8     & 0.5   & 8     & 0.538 & 13    & RF\_\ $\rho$ \\
    1452  & 0.556 & 9     & 0.5   & 8     & 0.375 & 8     & 0.5   & 12    & RF\_\ $\rho$ \\
    1453  & 0.667 & 9     & 0.5   & 8     & 0.5   & 8     & 0.462 & 13    & RF\_\ $\rho$ \\
    1479  & 0.6   & 10    & 0.333 & 9     & 0     & 1     & 0.3   & 10    & RF\_\ $\rho$ \\
    1492  & 0.667 & 9     & 0.5   & 8     & 0.571 & 7     & 0.273 & 11    & RF\_\ $\rho$ \\
    4534  & 0.9   & 10    & 0.778 & 9     & 0.75  & 8     & 0.6   & 10    & RF\_\ $\rho$ \\
    4538  & 0.875 & 8     & 0.571 & 7     & 0.429 & 7     & 0.545 & 11    & RF\_\ $\rho$ \\
    40668 & 0.875 & 8     & 0.714 & 7     & 0.667 & 6     & 0.818 & 11    & RF\_\ $\rho$ \\
    40670 & 0.7   & 10    & 0.333 & 9     & 0.571 & 7     & 0.5   & 10    & RF\_\ $\rho$ \\
    40705 & 0.625 & 8     & 0.571 & 7     & 0.5   & 10    & 0.364 & 11    & RF\_\ $\rho$ \\
    40996 & 0.889 & 9     & null & null & 0.786 & 14    & 0.182 & 11    & RF\_\ $\rho$ \\
    41004 & 0.889 & 9     & 0.75  & 8     & 0.818 & 11    & 0.667 & 12    & RF\_\ $\rho$ \\
    41014 & 0.75  & 4     & null & null & 0     & 2     & 0.5   & 4     & RF\_\ $\rho$ \\
    41025 & 0.667 & 3     & 0.5   & 2     & 0.5   & 2     & 0.333 & 3     & RF\_\ $\rho$ \\
    \hline
    1069  & 0.556 & 9     & 0.625 & 8     & 0.5   & 8     & 0.417 & 12    & XGB\_ $\rho$ \\
    1487  & 0.444 & 9     & 0.625 & 8     & 0.3   & 10    & 0.455 & 11    & XGB\_ $\rho$ \\
    1548  & 0.4   & 10    & 0.444 & 9     & 0.333 & 3     & 0.4   & 10    & XGB\_ $\rho$ \\
    40999 & 0.556 & 9     & 0.875 & 8     & 0.8   & 10    & 0.667 & 12    & XGB\_ $\rho$ \\
    41000 & 0.222 & 9     & 0.625 & 8     & 0.6   & 10    & 0.5   & 12    & XGB\_ $\rho$ \\
    41005 & 0.556 & 9     & 0.875 & 8     & 0.7   & 10    & 0.583 & 12    & XGB\_ $\rho$ \\
    \hline
    \end{tabular}%
  \label{tab:fea_ext_sum_rho_rf_xgb}%
\end{table}%

%------------------------------------------------------------------
%------------------------------------------------------------------
\begin{table}[H]
  \centering
  \caption{Summary of observations}
  \begin{adjustbox}{max width=\textwidth}
    \begin{tabular}{l|l|l|l|l||l|l|l|l|l||l|l|l|l|l}
    \hline \hline
    ID   & RF    & XGB    & SP    & PCA   & ID   & RF    & XGB    & SP    & PCA   & ID   & RF    & XGB    & SP    & PCA \\
    \Xhline{3\arrayrulewidth}
    14    & 1     &       & 1     &       & 1444  &       &       &       &       & 40670 & 1     &       & 1     &  \\
    \hline
    22    & 1     &       & 1     & 1     & 1451  &       &       &       &       & 40705 &       &       & 1     &  \\
    \hline
    44    & 1     &       & 1     & 1     & 1452  & 1     &       &       &       & 40996 & 1     & null & 1     &  \\
    \hline
    60    & 1     &       & 1     &       & 1453  &       &       &       &       & 40997 & 1     & 1     & 1     & 1 \\
    \hline
    180   & 1     & 1     & 1     & 1     & 1479  &       &       &       &       & 40999 & 1     &       & 1     & 1 \\
    \hline
    182   & 1     &       & 1     &       & 1487  &       &       &       &       & 41000 &       & 1     & 1     &  \\
    \hline
    313   &       &       &       &       & 1491  & 1     &       & 1     &       & 41004 & 1     & 1     & 1     & 1 \\
    \hline
    316   &       &       &       &       & 1492  &       &       & 1     &       & 41005 & 1     &       & 1     & 1 \\
    \hline
    718   &       &       &       &       & 1493  & 1     &       & 1     &       & 41007 &       &       & 1     & 1 \\
    \hline
    734   & 1     &       & 1     & 1     & 1494  & 1     &       & 1     &       & 41014 & 1     & null &       & 1 \\
    \hline
    833   & 1     &       & 1     &       & 1501  & 1     &       & 1     &       & 41025 & 1     &       & 1     & 1 \\
    \hline
    904   & 1     &       & 1     & 1     & 1548  &       &       & 1     &       & 41027 & 1     & 1     & 1     & 1 \\
    \hline
    979   & 1     &       & 1     &       & 1549  &       &       &       &       & 41048 &       &       & 1     &  \\
    \hline
    1049  &       &       &       &       & 1555  &       &       &       &       & 41049 &       &       & 1     &  \\
    \hline
    1050  &       &       &       &       & 4154  & 1     &       &       &       & 41050 &       &       & 1     &  \\
    \hline
    1056  &       &       &       &       & 4534  & 1     & 1     & 1     & 1     & 41051 &       &       & 1     &  \\
    \hline
    1069  &       &       &       &       & 4538  & 1     & 1     & 1     & 1     & 41052 &       &       & 1     &  \\
    \hline
    1443  & 1     &       & 1     &       & 40668 & 1     & 1     & 1     & 1     & 41053 & 1     &       & 1     &  \\
    \hline
    \end{tabular}
  \end{adjustbox}
  \label{tab:fea_ext_sum_obs}%
\end{table}

Apart from the definition, slightly different conclusion can be drawn from the visual observation. 
Table \ref{tab:fea_ext_sum_obs} summarizes all results from observation, where same abbreviations with same meanings are used. 
For each entry in the table, if there is an rising pattern can be detected, a number $1$ will be filled in. 
Still, to denote the two datasets 40996 and 41014 that XGBoost cannot run on, \textit{null} is filled. 
Follow the table, a new sequence is produced that SP $>$ RF $>$ PCA $>$ XGB. 
On 37 out of 54 datasets, the pattern of SP can be discovered and RF closely follows, with 30 observed patterns on different datasets while PCA and XGB have rather bad results with 16 and 8 datasets seen patterns, respectively.

From what has been mentioned above, on the whole, the observed results are different from what is defined, and often the observed results are a little bit looser than the definition. 
This may be due to several reasons. 
One is the criteria that only considers extractor with the highest $\rho$ value, which seems unfair to the other candidates. 
Other possible reasons for this situation will be analyzed in section \ref{chap:discussion} \nameref{chap:discussion}.

%------------------------------------------------------------------
%------------------------------------------------------------------
\subsection{Prediction Based on Matrix Factorization}
\label{subsec:pred_mf}
A recommendation/prediction system is built with matrix factorization as the kernel, whose rows are machine learning models with different parameters and columns are datasets, where the entry $(i,j)$ of this matrix corresponds to the accuracy of a specific model $i$ for a dataset $j$. 
For example, a matrix is given in table \ref{tab:example_mf_max1} and \ref{tab:example_mf_max2}, 
where the different accuracy of linear classifier, random forest(RF), XGBoost(XGB), 
and multi-layer perceptron(MLP) on 48 datasets are given. 
To make it easier to read the table, a line is drawn every 10 rows and the whole table is ascending with the data id.

RF\_max, XGB\_max, and MLP\_max mean the best performance of model RF, XGB, and MLP, severally. 
RF\_6\_512 and XGB\_6\_512 symbolize the accuracy of model RF and XGB with 512 trees and max depth 6 for each tree in the model, which can be seen as the most complex model. 
Similarly, MLP\_7\_512\_150 stands for the accuracy of the model MLP trained for 150 epochs with 7 layers and 512 nodes per layer. 

\begin{table}[H]
  \centering
  \caption{A matrix of all models with the highest accuracy and the most complex structure}
    \begin{tabular}{l|l|l|l|l|l|l|l}
    \hline \hline
    ID & Linear & RF\_6\_512 & RF\_max & XGB\_6\_512 & XGB\_max & MLP\_7\_512\_150 & MLP\_max \\
    \hline
    14    & 0.195455 & 0.813636 & 0.821212 & 0.710606 & 0.757576 & 0.828788 & 0.830303 \\
    22    & 0.166667 & 0.760606 & 0.769697 & 0.674242 & 0.707576 & 0.769697 & 0.807576 \\
    44    & 0.85451 & 0.929559 & 0.931534 & 0.895326 & 0.923634 & 0.939434 & 0.946675 \\
    60    & 0.38  & 0.853939 & 0.858182 & 0.778182 & 0.788485 & 0.829091 & 0.86 \\
    180   & 0.011419 & 0.662586 & 0.666676 & 0.67244 & 0.67535 & 0.857535 & 0.861351 \\
    182   & 0.215834 & 0.877945 & 0.880302 & 0.828464 & 0.850141 & 0.921772 & 0.921772 \\
    316   & 0.983709 & 0.989975 & 0.989975 & 0.987469 & 0.991228 & 0.984962 & 0.986216 \\
    718   & 0.666667 & 0.790909 & 0.8   & 0.845455 & 0.89697 & 0.618182 & 0.657576 \\
    734   & 0.860952 & 0.825915 & 0.830101 & 0.862494 & 0.866681 & 0.859189 & 0.881666 \\
    833   & 0.796228 & 0.796228 & 0.796228 & 0.784763 & 0.798447 & 0.800296 & 0.821746 \\
    \hline
    904   & 0.812121 & 0.857576 & 0.863636 & 0.793939 & 0.842424 & 0.8   & 0.833333 \\
    979   & 0.816364 & 0.866667 & 0.867273 & 0.836364 & 0.846061 & 0.867879 & 0.893939 \\
    1049  & 0.906639 & 0.887967 & 0.894191 & 0.873444 & 0.919087 & 0.892116 & 0.906639 \\
    1050  & 0.895349 & 0.895349 & 0.901163 & 0.877907 & 0.932171 & 0.858527 & 0.893411 \\
    1056  & 0.993598 & 0.995519 & 0.995519 & 0.994238 & 0.997119 & 0.995519 & 0.995839 \\
    1069  & 0.996206 & 0.995122 & 0.995122 & 0.99729 & 0.997832 & 0.99458 & 0.996206 \\
    1443  & 0.917808 & 0.926941 & 0.936073 & 0.945205 & 0.945205 & 0.917808 & 0.936073 \\
    1444  & 0.843478 & 0.869565 & 0.881159 & 0.837681 & 0.913043 & 0.826087 & 0.866667 \\
    1451  & 0.922747 & 0.892704 & 0.914163 & 0.935622 & 0.939914 & 0.866953 & 0.922747 \\
    1452  & 0.971545 & 0.97561 & 0.97561 & 0.987805 & 0.987805 & 0.97561 & 0.98374 \\
    \hline
    \end{tabular}%
  \label{tab:example_mf_max1}%
\end{table}%

%------------------------------------------------------------------
%------------------------------------------------------------------
\begin{table}[H]
  \centering
  \caption{A matrix of all models with the highest accuracy and the most complex structure}
    \begin{tabular}{l|l|l|l|l|l|l|l}
    \hline \hline
    ID & Linear & RF\_6\_512 & RF\_max & XGB\_6\_512 & XGB\_max & MLP\_7\_512\_150 & MLP\_max \\
    \hline
    1453  & 0.88764 & 0.870787 & 0.879213 & 0.848315 & 0.904494 & 0.851124 & 0.882023 \\
    1479  & 0.5925 & 0.5175 & 0.5575 & 0.4675 & 0.5625 & 0.4825 & 0.705 \\
    1487  & 0.92951 & 0.934289 & 0.936679 & 0.949821 & 0.949821 & 0.930705 & 0.934289 \\
    1494  & 0.802292 & 0.87106 & 0.876791 & 0.816619 & 0.853868 & 0.882522 & 0.896848 \\
    1501  & 0.117871 & 0.889734 & 0.901141 & 0.69962 & 0.735741 & 0.93346 & 0.948669 \\
    1548  & 0.553939 & 0.595152 & 0.595152 & 0.596364 & 0.62303 & 0.506667 & 0.552727 \\
    1549  & 0.104839 & 0.245968 & 0.262097 & 0.217742 & 0.33871 & 0.173387 & 0.241935 \\
    1555  & 0.090909 & 0.263636 & 0.266667 & 0.3   & 0.321212 & 0.20303 & 0.221212 \\
    4154  & 0.999362 & 0.999574 & 0.999787 & 0.999362 & 0.999787 & 0.999149 & 0.999362 \\
    4534  & 0.890107 & 0.934229 & 0.939161 & 0.913949 & 0.929022 & 0.975062 & 0.976706 \\
    \hline
    4538  & 0.101565 & 0.536668 & 0.540657 & 0.483584 & 0.509972 & 0.664314 & 0.664314 \\
    40668 & 0.245896 & 0.667265 & 0.675877 & 0.70759 & 0.70759 & 0.852023 & 0.852023 \\
    40670 & 0.398289 & 0.939163 & 0.945817 & 0.898289 & 0.914449 & 0.95057 & 0.958175 \\
    40705 & 0.905363 & 0.905363 & 0.914826 & 0.927445 & 0.946372 & 0.899054 & 0.952681 \\
    40997 & 0.423052 & 0.993561 & 0.994205 & 1     & 1     & 0.999356 & 1 \\
    40999 & 0.841495 & 1     & 1     & 0.755155 & 1     & 1     & 1 \\
    41000 & 0.356729 & 0.980039 & 0.990341 & 1     & 1     & 1     & 1 \\
    41004 & 0.507405 & 0.971024 & 0.984546 & 0.996136 & 1     & 0.995493 & 1 \\
    41005 & 0.792219 & 1     & 1     & 1     & 1     & 1     & 1 \\
    41007 & 0.861004 & 0.97426 & 0.978121 & 0.998713 & 1     & 0.989704 & 0.993565 \\
    \hline
    41025 & 0.6   & 0.6   & 0.8   & 0.5   & 0.9   & 0.3   & 0.6 \\
    41027 & 0.226219 & 0.74667 & 0.748158 & 0.738895 & 0.742411 & 0.977081 & 0.979582 \\
    41048 & 0.976549 & 0.981575 & 0.984925 & 0.991625 & 0.99665 & 0.994975 & 0.994975 \\
    41049 & 0.984536 & 0.990979 & 0.990979 & 0.987113 & 0.993557 & 0.940722 & 0.984536 \\
    41050 & 0.867672 & 0.889447 & 0.889447 & 0.921273 & 0.934673 & 0.907873 & 0.909548 \\
    41051 & 0.896907 & 0.916237 & 0.921392 & 0.917526 & 0.947165 & 0.938144 & 0.940722 \\
    41052 & 0.932998 & 0.812395 & 0.860972 & 0.953099 & 0.969849 & 0.954774 & 0.954774 \\
    41053 & 0.904639 & 0.898196 & 0.899485 & 0.916237 & 0.926546 & 0.916237 & 0.923969 \\
    \hline
    \end{tabular}%
  \label{tab:example_mf_max2}%
\end{table}%

In experiments, function \texttt{BaselineModel} of library \texttt{matrix\_factorization} is employed to implement matrix factorization with alternating least squares(ALS), which alternately fix one of the desired matrices and solve another one\cite{rendle2008online}. 
Hence, parameter \texttt{method} is settled to \textit{ALS} and \texttt{reg}, that is the lambda parameter for L2 regularization, is set to $0.5$. 
Besides, \texttt{n\_epochs} is equal to $1000$ namely the model is trained one thousand times. 

\begin{figure}[H]
    \centering
    \begin{subfigure}[b]{0.45\textwidth} 
        \centering
        \includegraphics[width = \linewidth]{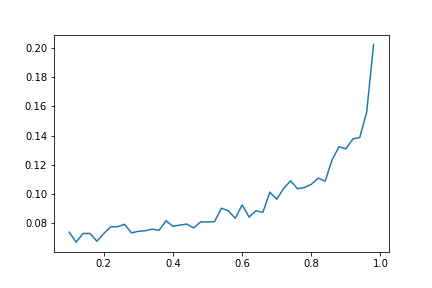}
        \caption{Mean Absolute Error}
    \end{subfigure}
    \begin{subfigure}[b]{0.45\textwidth} 
        \centering
        \includegraphics[width = \linewidth]{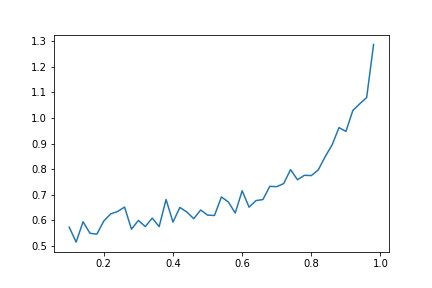}
        \caption{Root-Mean-Square Error}
    \end{subfigure}
    \caption{The performance of prediction with matrix factorization}
    \label{fig:mf_performance}
\end{figure}

Then, both mean absolute error(MAE) and root-mean-square error(RMSE) are the measurement of the performance, that it the less error the better performance. 
To guarantee the robustness of this system, models are trained and tested $5$ times and the results are averaged. 
To examine the accomplishment of the system, the ratio of test data is changed from $0.1$ to $1$ increasing $0.2$ each time, which means that a variety of scenarios are went through, from a few missing values in the matrix to most of the data being unknown.

Results are in figure \ref{fig:mf_performance}, where MAE is at least $0.067$ when the ratio of test data is $12\%$. 
That is, when the data to be predicted is about $12\%$ of the obtained data, the prediction error is less than $7\%$ on average. 
Moreover, the MAE is kept less than $10\%$ until the ratio of test data is more than $70\%$ as well as less than $15\%$ until the ratio reaches $95\%$. 
The result with RMSE shows similar regularity.

Furthermore, what is recognized is that there is also an explicable inclination that, with training data expanding, the prediction accuracy is growing. 
This is a perspective by analogy with definition \ref{def:fea_ext_rho}, from which the system scaling ability is tested.

\subsubsection{Distance Between Different Algorithms and Datasets}
\label{subsec:dist_alg_data}

\begin{figure}[H]
    \centering
    \includegraphics[width=0.8\linewidth]{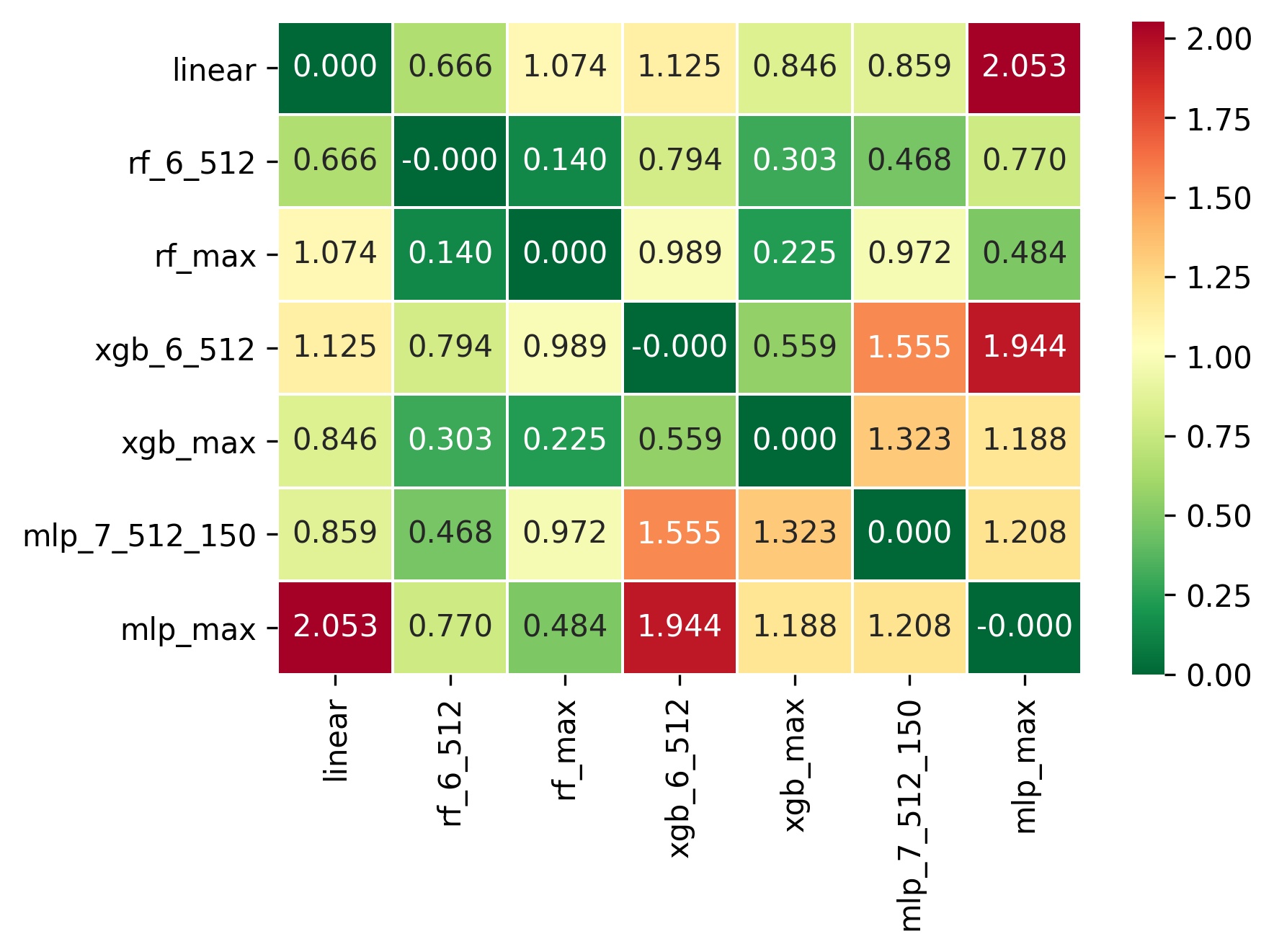}
    \caption{The distance between different models according to matrix factorization}
    \label{fig:algorithm_dist}
\end{figure}

\begin{figure}[H]
    \centering
    \begin{subfigure}[b]{0.45\textwidth} 
        \centering
        \includegraphics[width = \linewidth]{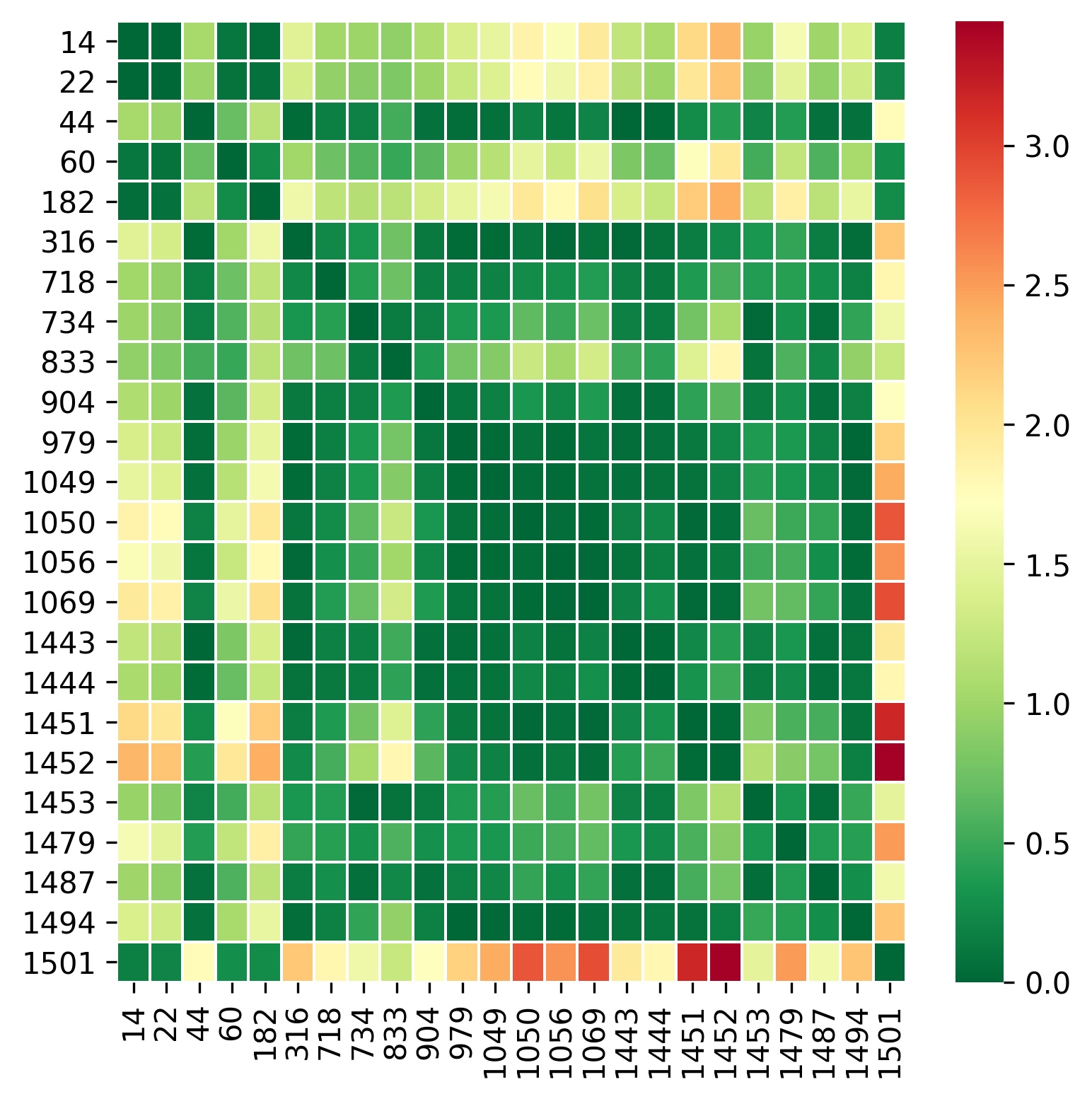}
        \caption{Detail of Left-up Corner}
    \end{subfigure}
    \begin{subfigure}[b]{0.45\textwidth} 
        \centering
        \includegraphics[width = \linewidth]{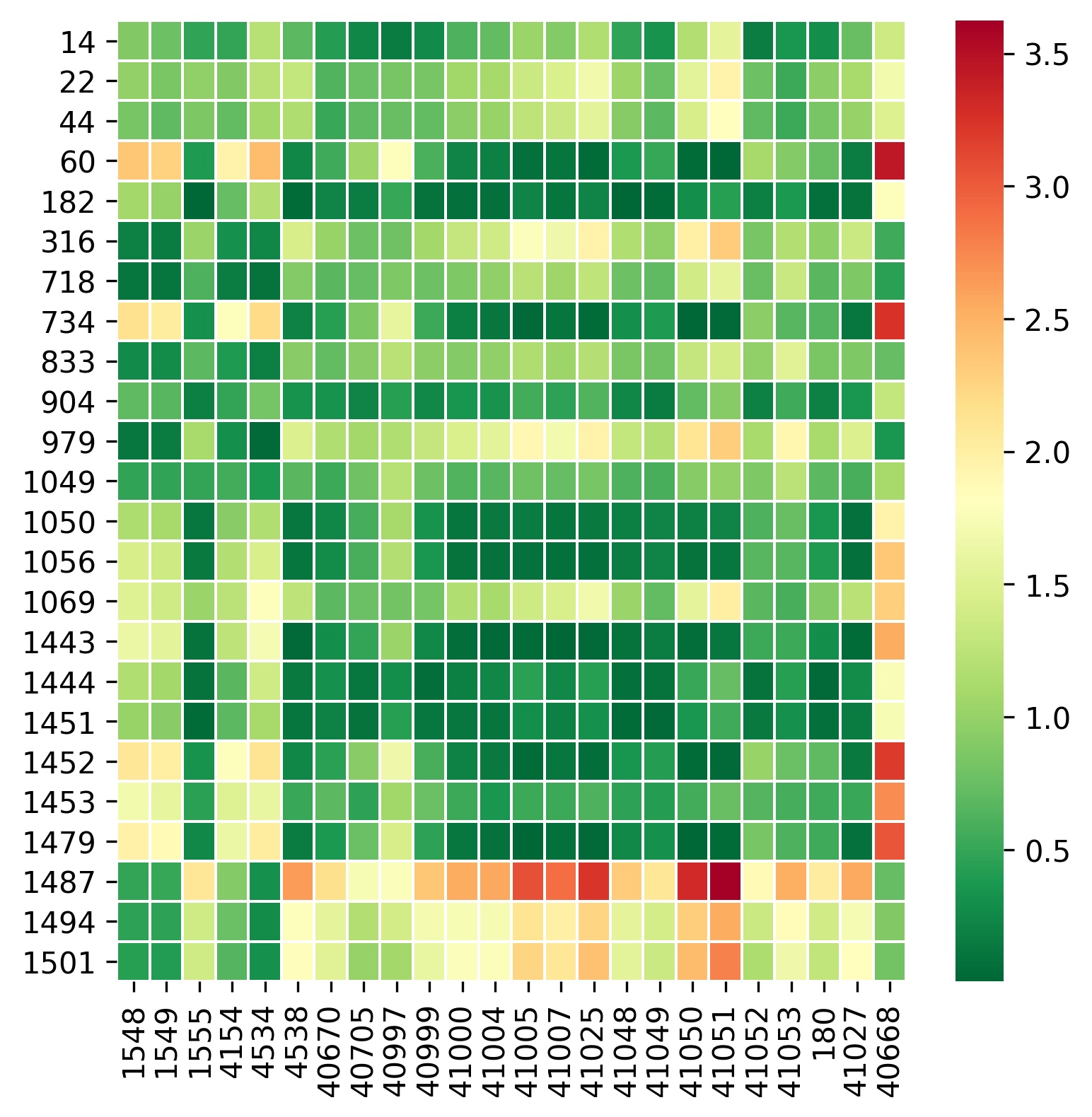}
        \caption{Detail of Left-bottom Corner}
    \end{subfigure}
    
    \begin{subfigure}[b]{0.45\textwidth} 
        \centering
        \includegraphics[width = \linewidth]{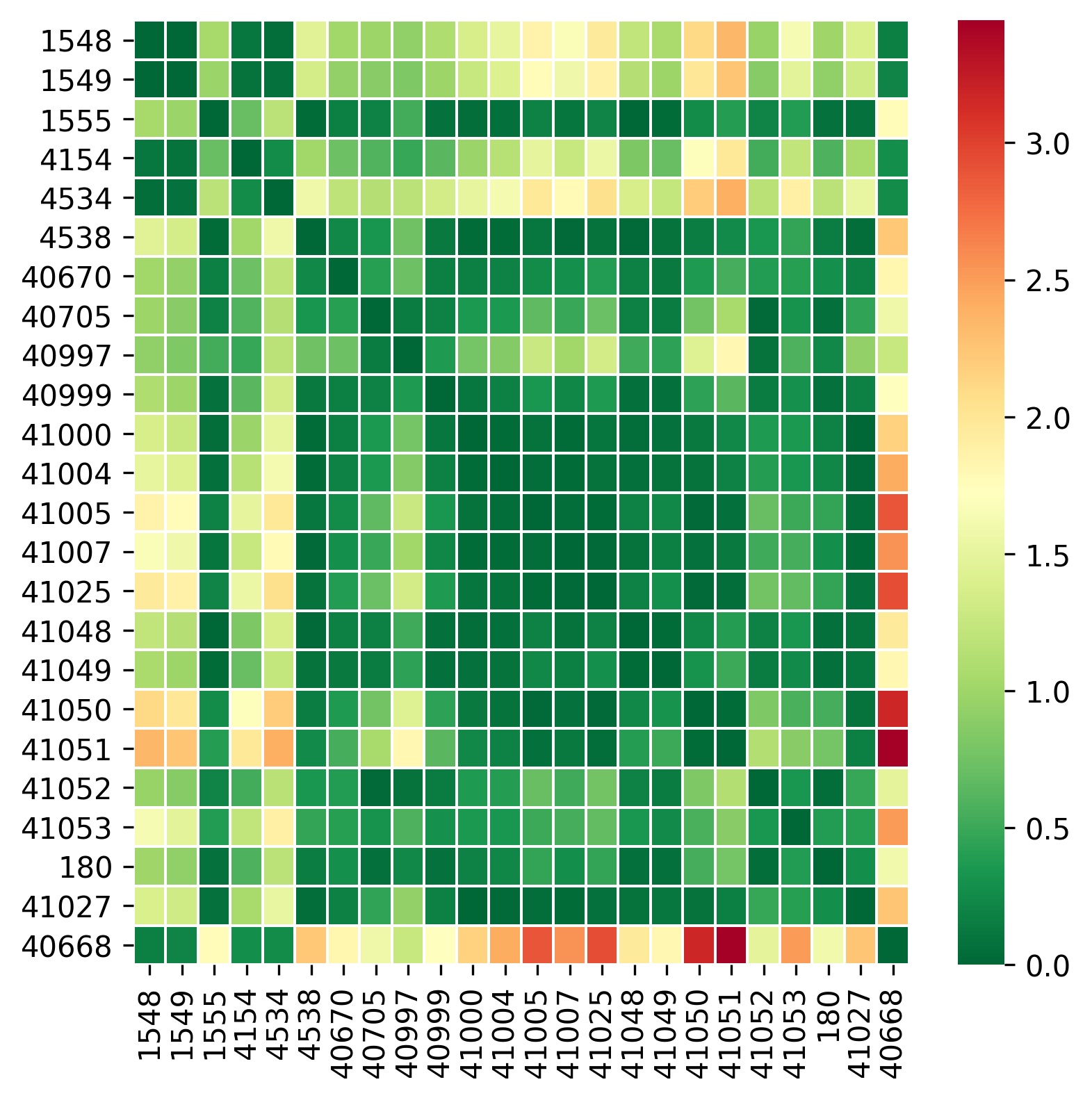}
        \caption{Detail of Right-bottom Corner}
    \end{subfigure}
    
    \caption{The detail of the distance between different datasets according to matrix factorization}
    \label{fig:dataset_dist_detail}
\end{figure}

\begin{figure}[H]
    \centering
    \includegraphics[width=0.8\linewidth]{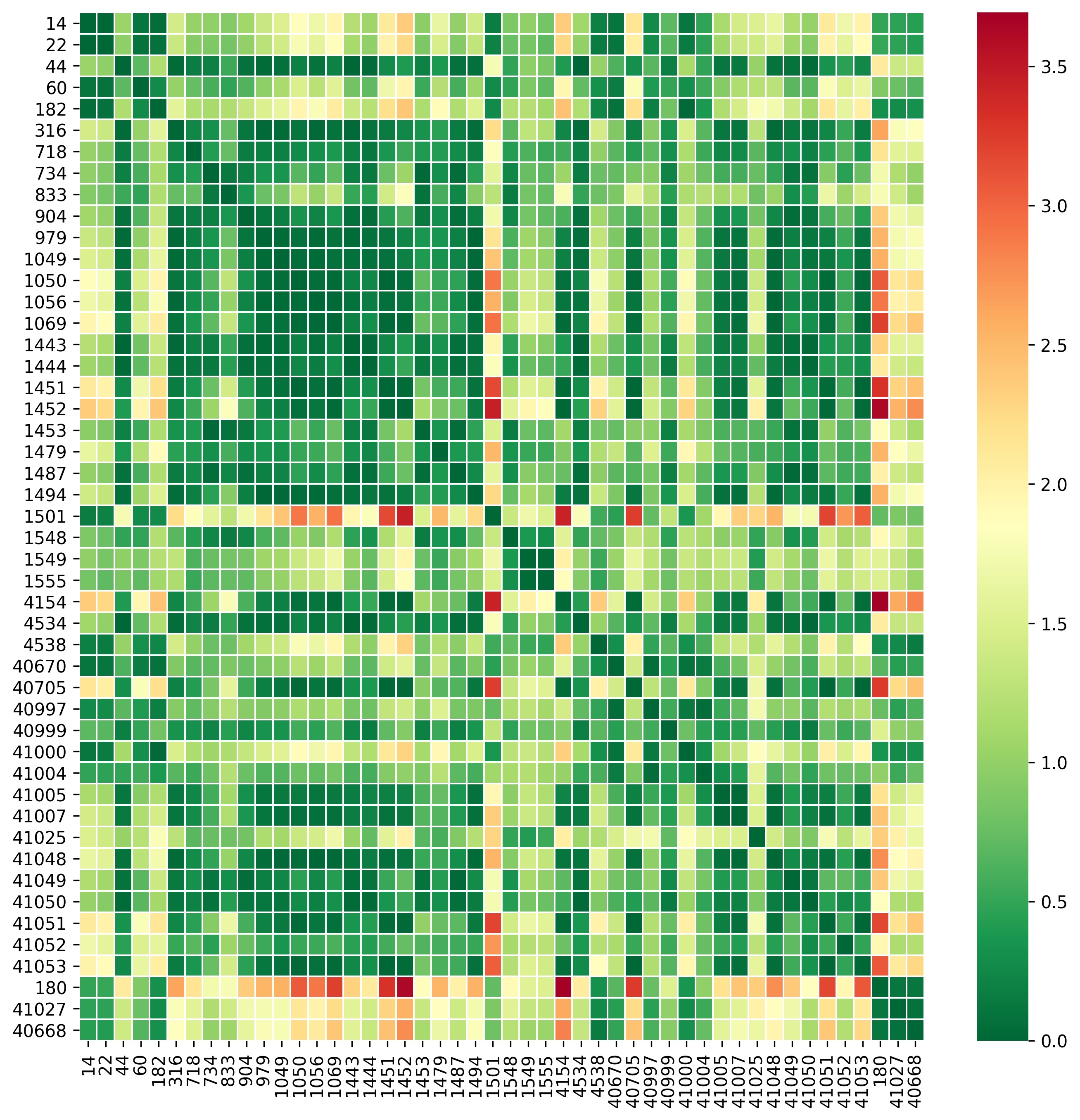}
    \caption{The overall results of distance between different dataset according to matrix factorization}
    \label{fig:dataset_dist}
\end{figure}

Along with the prediction of accuracy from the system, another application will be useful that is to compare the similarity of specific models or two datasets as pointed in figure \ref{fig:algorithm_dist}, \ref{fig:dataset_dist_detail}, and \ref{fig:dataset_dist}. 
Figure \ref{fig:algorithm_dist} displays an instance using the matrix decomposition to calculate the irrelevance or distance between different models, where both x-axis and y-axis are the abbreviations of different models. 

As same as the abbreviations used in table \ref{tab:example_mf_max1} in section \ref{subsec:pred_mf}, RF\_max, XGB\_max, and MLP\_max indicate the best performance of model RF, XGB, and MLP as RF\_6\_512 and XGB\_6\_512 symbolize the accuracy of model RF and XGB with 512 trees and max depth 6 for each tree in the model. 
Similarly, MLP\_7\_512\_150 stands for the accuracy of the model MLP trained for 150 epochs with 7 layers and 512 nodes per layer. 
Furthermore, color shows the level of correlation between models, where dark green is for the high correlation, namely short distance, while dark red is on the contrary. 

Obviously, the correlation within random forest and XGBoost algorithm is very strong that the distance between RF\_max and RF\_6\_512 is $0.140$, the closest one as well as RF\_max and XGB\_max plus XGB\_max and RF\_6\_512 are less than $0.5$. 
This is logical because the principal of these two algorithms are comparable like introduced in section \ref{subsec:model_rf} and  \ref{subsec:model_xgb}. 
At the same time, it also confirms previous experimental results in section \ref{subsec:heatmap_rf} and \ref{subsec:heatmap_xgb}, that is, performance will be improved after models become complicated in many cases, equivalently, the most complex model is similar to the model with the largest accuracy. 
It is the same explanation that can interpret why MLP\_max and MLP\_7\_512\_150 are rather far away, which is consistent to results in section \ref{subsec:heatmap_mlp}. 
Moreover, figure \ref{fig:alg_coordinate} visualizes the example of vectors produced by the system, based upon which the distance between models is calculated.

The results of datasets produced by the system are shown in figure \ref{fig:dataset_dist_detail} and \ref{fig:dataset_dist}. 
Figure \ref{fig:dataset_dist_detail} provides the detail of figure \ref{fig:dataset_dist}, which does not include the detail of right-up corner because it has the same content as the left-bottom corner due to the symmetry. 
Color can also be utilized to quickly determine the similarity between different datasets. 
Theoretically, the closer two data ids are, the more similar these two datasets will be. 
In other words, there should be more green alongside the diagonal, which is obvious before id 1548 and after id 4538.

% \subsection{Practice of the System and Other Findings}
% \todo{add plots for all classifiers (linear, rf, xgb, mlp), x-axis is percent of instances, y-axis is acc} 

% Among all models of linear classifiers, random forest, XGboost, and MLP
% \begin{enumerate}
%     \item If MLP performs very well, linear classifier will perform well also; vice versa
%     \item Linear classifiers performs poorest
%     \item Linear classifiers always are the fastest ones, the second fast model is random forest
%     \item MLP always has the highest accuracy(add the table)
%     \item RF, XGB, and SP have an advantage that they can analyze/rank the original features, but PCA cannot.
%     \todo{this is left to the end because we can just delete this subsection}
% \end{enumerate}

\newpage\chapter{Further Discussion}
\label{chap:discussion}
In the section \ref{chap:result} \nameref{chap:result}, the experimental process and relevant results obtained are shown and summarized. 
This chapter is mainly about the further discussion about the above content. 
First, it is a fundamental task to supplement the statement of the previous experimental results with further analysis about the possible causes of these results. 
Then, the limiting factors are explained, which exist in the whole research process, both for the experimental process and the research results. 
Finally, possible future directions for the research are proposed. 

\subsection{Results Analysis}
First of all, it is important to explain why the MLP sometimes has a counterintuitive behavior, that is, as the number of neural layers and the number of ganglion nodes in each layer increases, the accuracy of classification decreases. 
This may be the result of over-fitting. 
Overfitting means that it performs well on the training dataset and becomes worse on the validation dataset. 
The overfitting model may even learn noise during training. 
This is why MLP needs regularization to constrain the entire neural network during training. 
In addition, in more complex models, the neural network avoids overfitting by adding dropout layers\cite{srivastava2014dropout}. 
Besides, 
% from the final result of accuracy, MLP always performs better because 
heatmaps of MLP are dominated by red areas, which accentuates blue areas that may be occasional, giving the impression that performance is declining and exaggerating the situation of accuracy drop. 
This involves the limitation of the research method, that is, subjective observation.

Second, it is possible to consider different evaluation criteria for feature extraction technique. 
In this experiment, what is focused on is the predictability and interpretability of extractor according to its performance, but in practical applications, perhaps extractors with the poor predictability is what engineers actually want to use. 
Contrary to the definition of \nameref{def:fea_ext_rho}, there is a possibility that the fewer features are used, the more accurate the prediction is, which means the better performance can be achieved by consuming fewer resources, which is of course what is expected in practical application. 
However, with technology like this, it is demanded of verification if this is a fluke or if it's always the case. 

In these two cases, no matter increasing the number of features increases or decreases the accuracy of model prediction, which may be caused by the data itself because, when features actually related the response rather than noise features are added, the performance should be improved and vice versa\cite{archer2008empirical}. 
This means that most of the time considering the attributes of the data features themselves is fundamental, and if features are correlated, which is quite common, extra attention and caution is necessary. 
Therefore, adding features related to labels usually leads to more accurate model prediction, while adding features related to noise degrades the model performance. 
This is why the conclusion is drawn that perhaps feature extraction with hierarchical clustering based on Spearman correlations is better than with permutation importance, although by definition the latter is a little better.

Finally, it is still the discussion about feature extraction. 
Although the same growth pattern exists, perhaps an image similar to convex function or \textit{log} function is more preferable than that of concave function for the same reason mentioned above. 
The patter is increasing like a convex function or a \textit{log} function means that the performance can reach the up-bound with fewer features than a concave function. 
From this perspective, according to Figure \ref{fig:imp_rf_both} and \ref{fig:fea_ext_sp_both} in section \ref{subsec:fea_rf} and \ref{subsec:fea_spearman}, feature extraction with hierarchical clustering can also be considered better than with permutation importance.

\subsection{Limitation}
In the course of the study, there are also some limiting factors. 
First of all, the limitation of machine for experiments has the greatest impact. 
Experiments are mainly run on the personal computer, which uses a 2070 graphics card with 8 gigabytes of GPU memory, plus 16 gigabytes of computer memory. 
This size of memory is still relatively limited for experiments, especially when dealing with XGBoost tasks, there can be an error message for \textit{resourcehaustederror} says \textit{OOM when allocating tensor \dots}, which means out of memory. 
This is the reason why there are some \textit{null} in the results for XGBoost.

Another confine is that there are too many influence factors to fully study in one project. 
When dealing with a machine learning task, it is usually divided into the following steps: data acquisition, data preprocessing, model construction, hyperparameter optimization, and application. 
Therefore, it is not only required to consider the problem of the model architecture and feature extraction, but also necessary to carry out other work such as feature engineering from the data itself. 
Due to only thinking about the problem within these two areas, this project is regarded as the study of predictability rather than AutoML. 
Even so, there are still a lot of hyperparameters that haven't been researched yet, such as learning rate and initializer for the bias vector for MLP and regularization for XGBoost. 
As for the study of feature extraction, though there are not so many hyperparameters as model construction, performance can also be more detailed classified and investigated.

The last point is that human observation is too subjective. 
Just as the explanation of why MLP has counterintuitive patterns, one possible reason is that perception of viewing images may be misleading. 
Similar problems will also be encountered by observing the patterns of feature extraction. 
In the absence of definition, when can a pattern be considered as increasing despite slight fluctuations? 
That is why definitions are introduced to help decide.

\subsection{Future Work}
First of all, combined with the analysis of results and limitation, there are several directions can be drawn for further research. 
One is to incorporate more models and parameters into the study by improving the computational power or crawling existing results from platform like OpenML. 
The second is to establish a more rigorous definition for model predictability, as what have done for feature extraction. 
For this, it is also worthy to further consider whether there are other judging criteria besides the monotonic increase of accuracy. 
For example, for feature extraction, the definition \ref{def:fea_ext_rho} \nameref{def:fea_ext_rho} can be further refined by adding a consideration of the variation range of accuracy so that if accuracy decreases but within an acceptable range, the $t_i$ will still equal to $1$. 
This can be considered as a relaxation of the definition of $\rho$.

Additionally, we can summarize a criterion for the $\rho$ value, for instance, the predictability of the extractor can be regarded as acceptable when the $\rho$ value is greater than a specific value, like $0.5$. 
Further, it is feasible to fit a function to the pattern of the changing accuracy value for relatively predictive techniques and decide how many features should be extracted such that the performance of the model will maintain acceptable. 
Once the criterion for the $\rho$ value is established, the definition \ref{def:fea_ext_alpha} \nameref{def:fea_ext_alpha} has a purpose. 
This definition can, on the one hand, cover a specific family of datasets to determine whether the technique performs well for that class of datasets, such as image classification, speech recognition, nature language processing, and other areas. 
On the other hand, the definition can be used to visualize how the $\rho$ and $\alpha$ values change and in what way they interact with each other, such as log growth or exponential growth. 

Finally the study results can be integrated into the autoML process, where the system based on matrix factorization, which is used to predict accuracy, can be the alternative method for deciding models and parameters, and the study results of feature extraction can partially complete the function of feature engineering. 
In particular, extractor with hierarchical clustering based upon Spearman correlation can be combined with permutation importance, sorting the features within the same cluster according to permutation importance, and selecting features from each cluster in accordance with the rank. 
The regression function of the variation of accuracy and the number of features can determine how many clusters are selected.

% \begin{itemize}
%     \item if there is any other possibility to measure the predictability, for example, considering more factors? for $\rho$, it is to add amplitude of variation.
%     \item it is possible to visualize different feature extraction technique by $(\alpha,\ \rho)$ in a plan
%     \item set a bar for $\rho$ value rather than only consider the highest one
%     \item combine sp with rf
%     \item how many clusters should be selected?
% \end{itemize}
% For future research, this matrix factorization technique can be integrated in to the process of autoML, which can be the alternative method of selecting the particular algorithm.

\newpage
\chapter{Conclusions}
\label{chap:conclu}
In the whole article, the following accomplishments are trying to answer the questions posed in section\ref{chap:intro} \nameref{chap:intro}. 
First, a platform is built to mainly study the performance of three algorithms, random forest, XGBoost and MLP, in terms of both the highest test accuracy and the predictability of classification performance. 
It is believed that random forest is a good choice when time and computational power are very limited because it is fast and the accuracy is usually in an acceptable range. 
But when two main limitations are loosed, XGBoost and MLP can be considered, especially MLP, because some shallow networks can achieve good results. 
This is our answer to question \ref{que:best_model} of how to choose a best algorithm.

For the second question, from the point of view of predictability, both random forest and XGBoost are significantly better than MLP because of the overfitting problem that exists in MLP with models becoming more complex. 
Thus for random forest or XGBoost, it is possible to predict complex learners directly from simple learners, whereas for MLP, this is infeasible or at least requires more considerations.

Then, the predictability of four dimension reduction techniques is compared, including that based on permutation importance of random forest, gain-based feature importance of XGBoost, hierarchical clustering based upon Spearman correlation, and PCA. 
It is proposed to find efficient methods for dimensionality reduction of datasets by highly predictable feature extraction techniques, namely to achieve the highest possible performance with the smallest possible number of features. 
If a technique is highly predictable, it should come with a growth pattern that is pervasive in multiple datasets such that a regression function can match it. 
It is found that extractors with permutation importance and hierarchical clustering have high predictability, and features can be extracted efficiently by these two schemes afterward. 
Thus for question \ref{que:extract_fea}, a specific method can be derived by continuing to study these two methods in depth.

As for the last problem, a novel prediction system is offered, by which either a new dataset or an new algorithm can be predicted the final result with an average error of at least $6.7\%$ from the true value. As long as the tested data does not exceed $70\%$ of the total data, the system can guarantee that the average error is less than $10\%$. 
Hence, if there is a new algorithm on the hand, it is easy and low-time-consumption to train and test some simple models of the algorithms on a family of datasets, then the highest accuracy of the algorithm can be given by the system with known results of all models and datasets. 
Similarly, if there is a new dataset, after training and testing with all simple models, the best performance and the corresponding model can be given.

In conclusion, the main achievements of this project are to examine the predictability of three classification algorithms and four feature extraction techniques as well as design a system to predict the performance of a specific model on a particular dataset.

\clearemptydoublepage

%Choose a good bibliography style, plain would do often, but these might be nice too
%\bibliographystyle{these}
\bibliographystyle{plain}
\bibliography{thesis}

\clearemptydoublepage

\appendix
\addcontentsline{toc}{chapter}{Appendix}

\chapter{Heatmaps}
\label{app:heatmaps}

% In this file (appendices/main.tex) you can add appendix chapters, just as you did in the thesis.tex file for the `normal' chapters.
% You can also choose to include everything in this single file, whatever you prefer.

\section{Random Forest}
\label{subapp:rf_heat}
\begin{figure}[H]
    \centering
    \begin{subfigure}[b]{0.24\textwidth} 
        \centering
        \includegraphics[width = \linewidth]{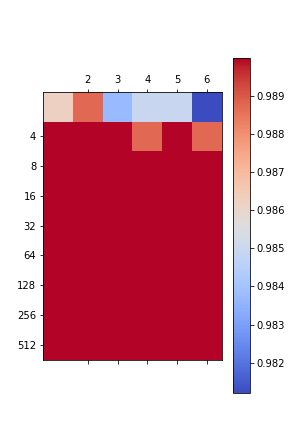}
        \caption{316}
    \end{subfigure}
    \begin{subfigure}[b]{0.24\textwidth}
        \centering
        \includegraphics[width = \linewidth]{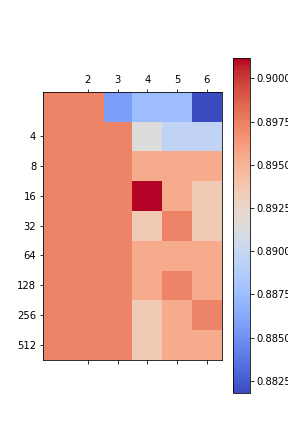}
        \caption{1050}
    \end{subfigure}
    \begin{subfigure}[b]{0.24\textwidth} 
        \centering
        \includegraphics[width = \linewidth]{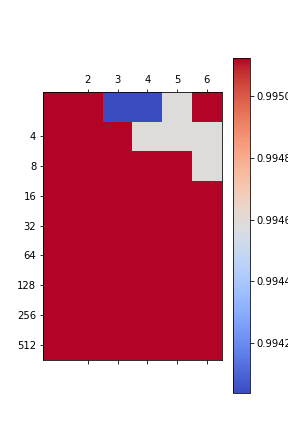}
        \caption{1069}
    \end{subfigure}
    \begin{subfigure}[b]{0.24\textwidth} 
        \centering
        \includegraphics[width = \linewidth]{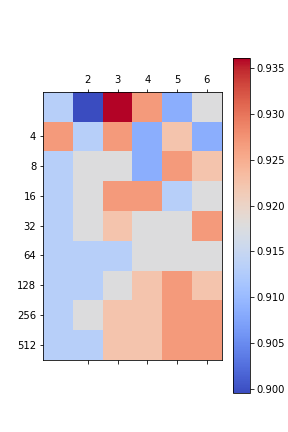}
        \caption{1443}
    \end{subfigure}
    
    \centering
    \begin{subfigure}[b]{0.24\textwidth} 
        \centering
        \includegraphics[width = \linewidth]{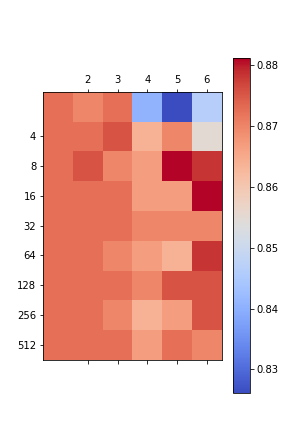}
        \caption{1444}
    \end{subfigure}
    \begin{subfigure}[b]{0.24\textwidth} 
        \centering
        \includegraphics[width = \linewidth]{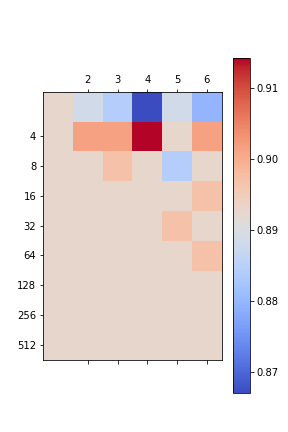}
        \caption{1451}
    \end{subfigure}
    \begin{subfigure}[b]{0.24\textwidth} 
        \centering
        \includegraphics[width = \linewidth]{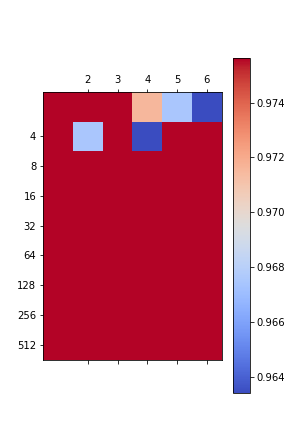}
        \caption{1452}
    \end{subfigure}
    \begin{subfigure}[b]{0.24\textwidth} 
        \centering
        \includegraphics[width = \linewidth]{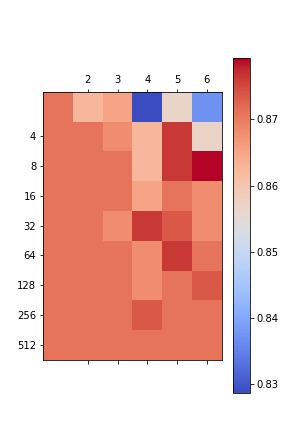}
        \caption{1453}
    \end{subfigure}
    \caption{Part of results of random forest without obvious pattern}
    \label{app:fig:heatmap_rf_no1}
\end{figure}
%------------------------------------------------------------------
%------------------------------------------------------------------
\begin{figure}[H]
    \centering
    \begin{subfigure}[b]{0.24\textwidth} 
        \centering
        \includegraphics[width = \linewidth]{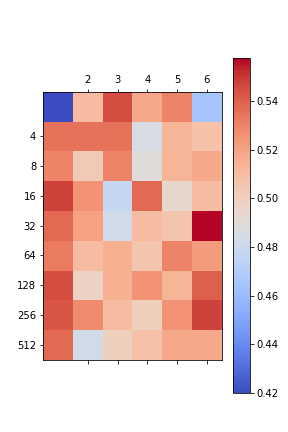}
        \caption{1479}
    \end{subfigure}
    \begin{subfigure}[b]{0.24\textwidth} 
        \centering
        \includegraphics[width = \linewidth]{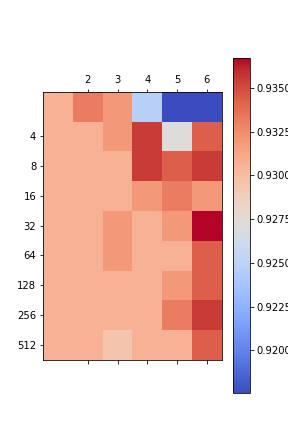}
        \caption{1487}
    \end{subfigure}
    \begin{subfigure}[b]{0.24\textwidth} 
        \centering
        \includegraphics[width = \linewidth]{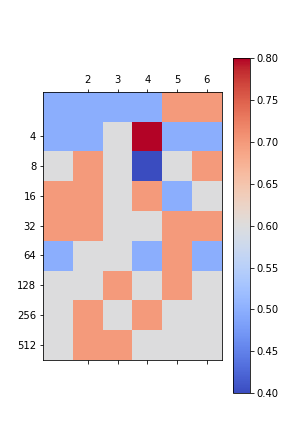}
        \caption{41025}
    \end{subfigure}
    
    \caption{Part of results of random forest without obvious pattern}
    \label{app:fig:heatmap_rf_no2}
\end{figure}
%------------------------------------------------------------------
%------------------------------------------------------------------
\begin{figure}[H]
    \centering
    \begin{subfigure}[b]{0.24\textwidth} 
        \centering
        \includegraphics[width = \linewidth]{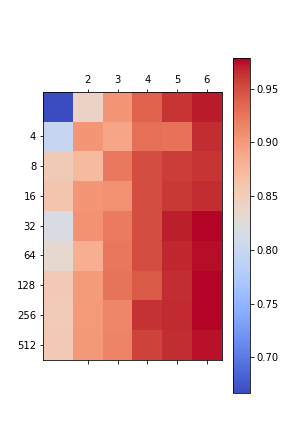}
        \caption{41007}
    \end{subfigure}
    \begin{subfigure}[b]{0.24\textwidth} 
        \centering
        \includegraphics[width = \linewidth]{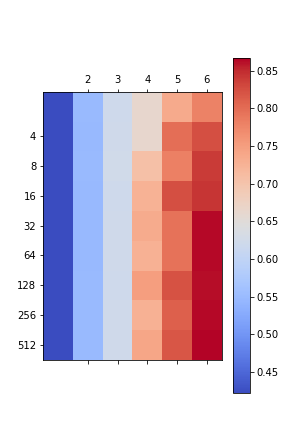}
        \caption{41014}
    \end{subfigure}
    \begin{subfigure}[b]{0.24\textwidth} 
        \centering
        \includegraphics[width = \linewidth]{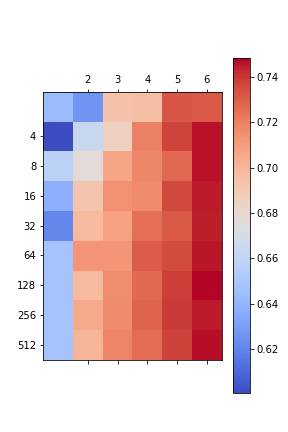}
        \caption{41027}
    \end{subfigure}
    \begin{subfigure}[b]{0.24\textwidth} 
        \centering
        \includegraphics[width = \linewidth]{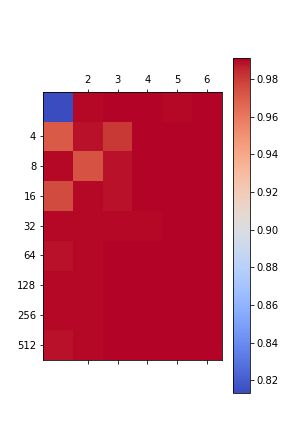}
        \caption{41049}
    \end{subfigure}
    
    \centering
    \begin{subfigure}[b]{0.24\textwidth} 
        \centering
        \includegraphics[width = \linewidth]{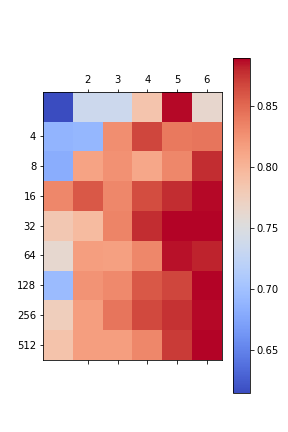}
        \caption{41050}
    \end{subfigure}
    \begin{subfigure}[b]{0.24\textwidth} 
        \centering
        \includegraphics[width = \linewidth]{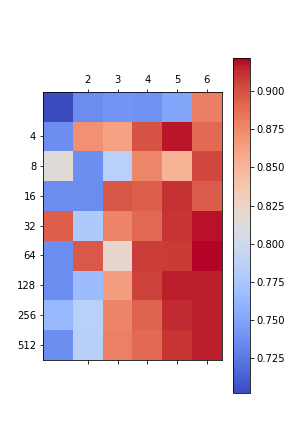}
        \caption{41501}
    \end{subfigure}
    \begin{subfigure}[b]{0.24\textwidth} 
        \centering
        \includegraphics[width = \linewidth]{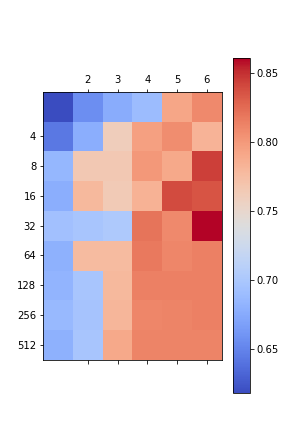}
        \caption{41052}
    \end{subfigure}
    \begin{subfigure}[b]{0.24\textwidth}
        \centering
        \includegraphics[width = \linewidth]{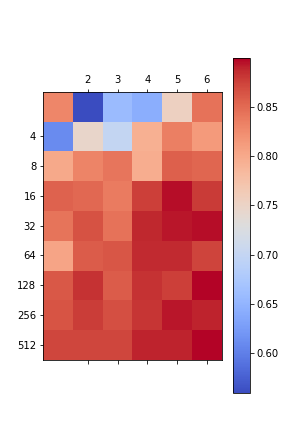}
        \caption{41053}
    \end{subfigure}
    \caption{Part of results of random forest with a pattern}
    \label{app:fig:heatmap_rf_pattern3}
\end{figure}
%------------------------------------------------------------------
%------------------------------------------------------------------
\begin{figure}[H]
    \centering
    \begin{subfigure}[b]{0.24\textwidth} 
        \centering
        \includegraphics[width = \linewidth]{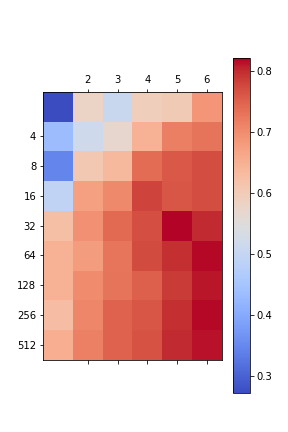}
        \caption{14}
    \end{subfigure}
    \begin{subfigure}[b]{0.24\textwidth}
        \centering
        \includegraphics[width = \linewidth]{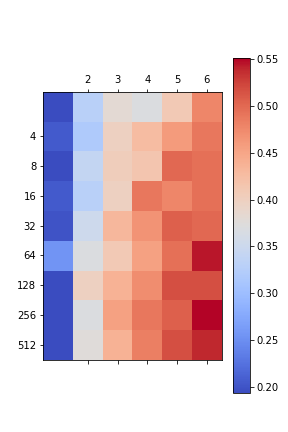}
        \caption{313}
    \end{subfigure}
    \begin{subfigure}[b]{0.24\textwidth} 
        \centering
        \includegraphics[width = \linewidth]{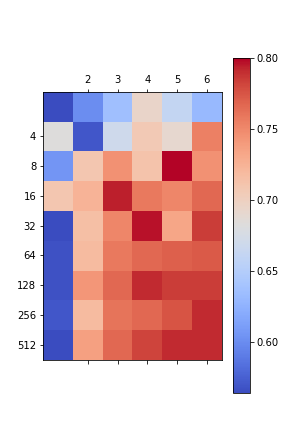}
        \caption{718}
    \end{subfigure}
    \begin{subfigure}[b]{0.24\textwidth} 
        \centering
        \includegraphics[width = \linewidth]{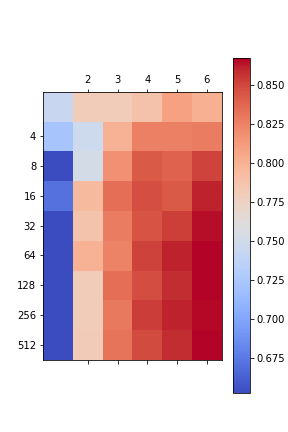}
        \caption{979}
    \end{subfigure}
    
    \centering
    \begin{subfigure}[b]{0.24\textwidth} 
        \centering
        \includegraphics[width = \linewidth]{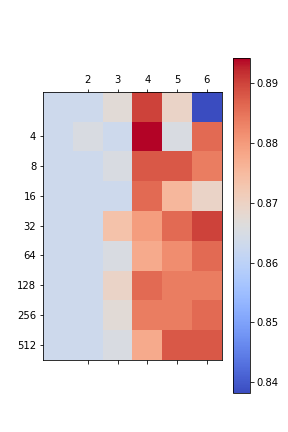}
        \caption{1049}
    \end{subfigure}
    \begin{subfigure}[b]{0.24\textwidth} 
        \centering
        \includegraphics[width = \linewidth]{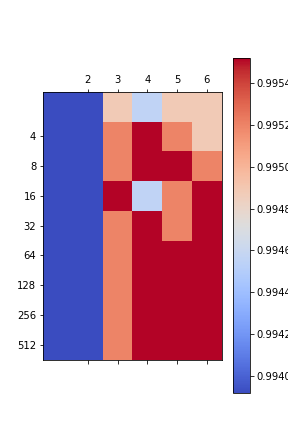}
        \caption{1056}
    \end{subfigure}
    \begin{subfigure}[b]{0.24\textwidth} 
        \centering
        \includegraphics[width = \linewidth]{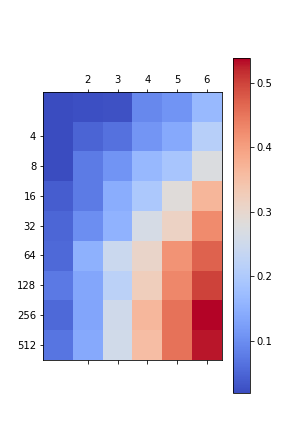}
        \caption{1491}
    \end{subfigure}
    \begin{subfigure}[b]{0.24\textwidth} 
        \centering
        \includegraphics[width = \linewidth]{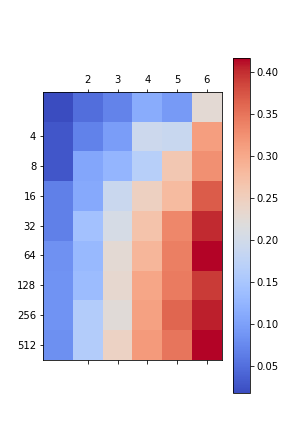}
        \caption{1492}
    \end{subfigure}
    
    \centering
    \begin{subfigure}[b]{0.24\textwidth} 
        \centering
        \includegraphics[width = \linewidth]{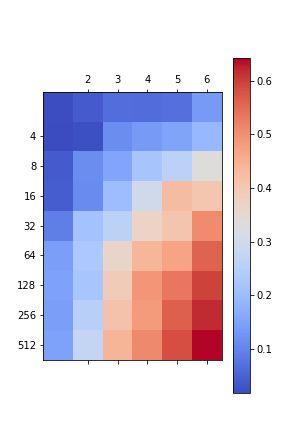}
        \caption{1493}
    \end{subfigure}
    \begin{subfigure}[b]{0.24\textwidth} 
        \centering
        \includegraphics[width = \linewidth]{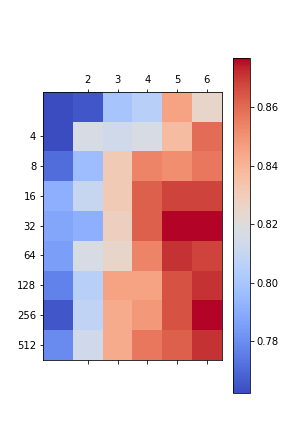}
        \caption{1494}
    \end{subfigure}
    \begin{subfigure}[b]{0.24\textwidth} 
        \centering
        \includegraphics[width = \linewidth]{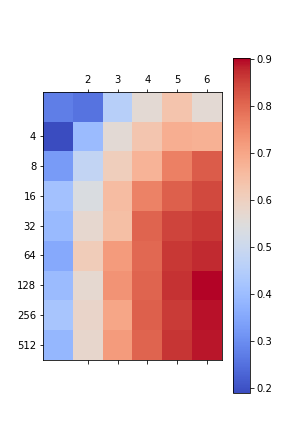}
        \caption{1501}
    \end{subfigure}
    \begin{subfigure}[b]{0.24\textwidth} 
        \centering
        \includegraphics[width = \linewidth]{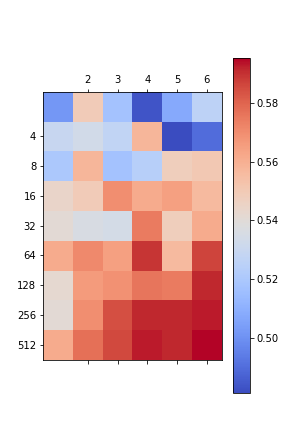}
        \caption{1548}
    \end{subfigure}
    
    \caption{Part of results of random forest with a pattern}
    \label{app:fig:heatmap_rf_pattern1}
\end{figure}
%------------------------------------------------------------------
%------------------------------------------------------------------
\begin{figure}[H]
    \centering
    \begin{subfigure}[b]{0.24\textwidth} 
        \centering
        \includegraphics[width = \linewidth]{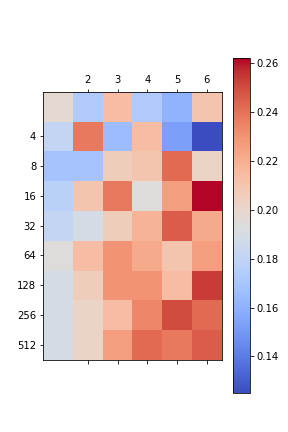}
        \caption{1549}
    \end{subfigure}
    \begin{subfigure}[b]{0.24\textwidth}
        \centering
        \includegraphics[width = \linewidth]{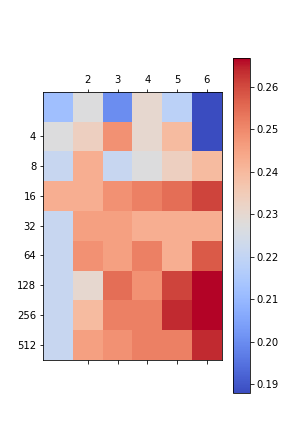}
        \caption{1555}
    \end{subfigure}
    \begin{subfigure}[b]{0.24\textwidth} 
        \centering
        \includegraphics[width = \linewidth]{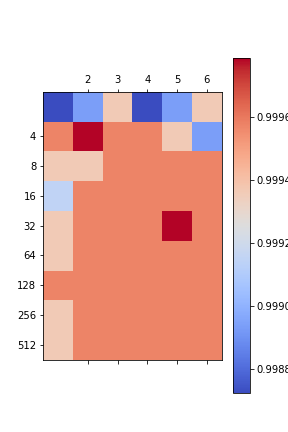}
        \caption{4154}
    \end{subfigure}

    \centering
    \begin{subfigure}[b]{0.24\textwidth} 
        \centering
        \includegraphics[width = \linewidth]{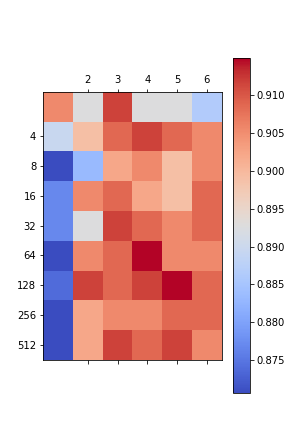}
        \caption{40705}
    \end{subfigure}
    \begin{subfigure}[b]{0.24\textwidth} 
        \centering
        \includegraphics[width = \linewidth]{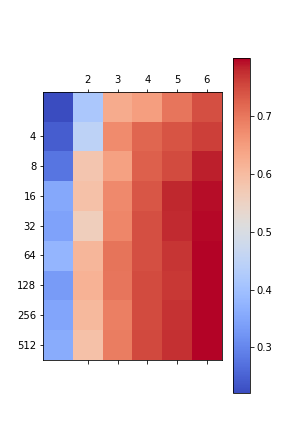}
        \caption{40996}
    \end{subfigure}
    \begin{subfigure}[b]{0.24\textwidth} 
        \centering
        \includegraphics[width = \linewidth]{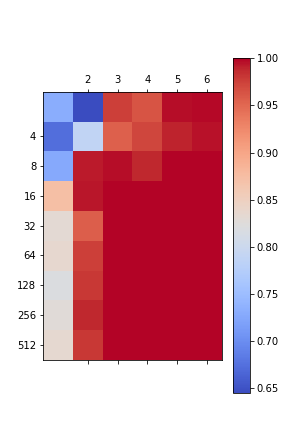}
        \caption{41005}
    \end{subfigure}
    \caption{Part of results of random forest with a pattern}
    \label{app:fig:heatmap_rf_pattern2}
\end{figure}

\section{XGBoost}
\label{subapp:xgb_heat}
\begin{figure}[H]
    \centering
    \begin{subfigure}[b]{0.24\textwidth} 
        \centering
        \includegraphics[width = \linewidth]{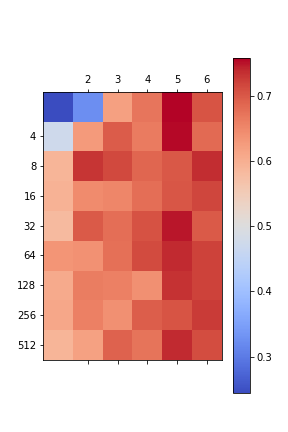}
        \caption{14}
    \end{subfigure}
    \begin{subfigure}[b]{0.24\textwidth} 
        \centering
        \includegraphics[width = \linewidth]{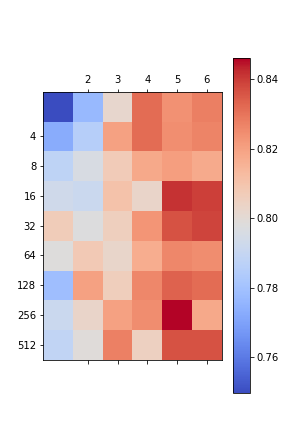}
        \caption{979}
    \end{subfigure}
    \begin{subfigure}[b]{0.24\textwidth} 
        \centering
        \includegraphics[width = \linewidth]{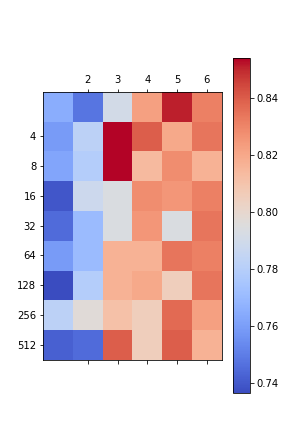}
        \caption{1494}
    \end{subfigure}
    \begin{subfigure}[b]{0.24\textwidth} 
        \centering
        \includegraphics[width = \linewidth]{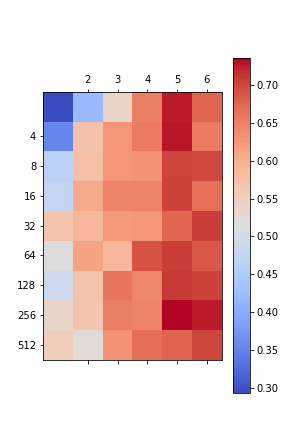}
        \caption{1501}
    \end{subfigure}
    
    \centering
    \begin{subfigure}[b]{0.24\textwidth} 
        \centering
        \includegraphics[width = \linewidth]{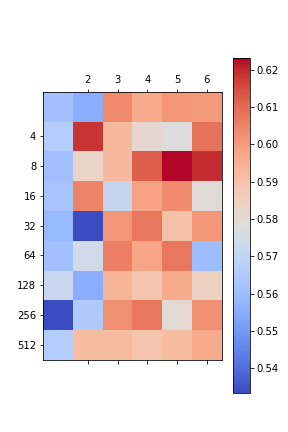}
        \caption{1548}
    \end{subfigure}
    \begin{subfigure}[b]{0.24\textwidth} 
        \centering
        \includegraphics[width = \linewidth]{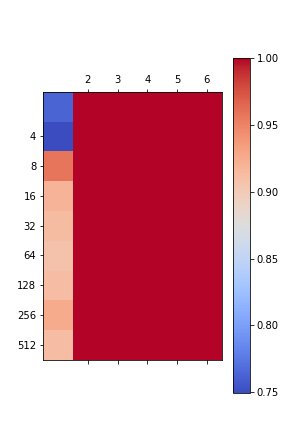}
        \caption{41005}
    \end{subfigure}
    \begin{subfigure}[b]{0.24\textwidth}
        \centering
        \includegraphics[width = \linewidth]{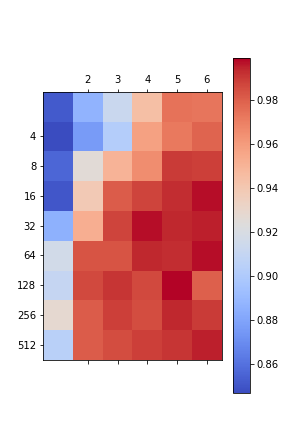}
        \caption{41007}
    \end{subfigure}
    \begin{subfigure}[b]{0.24\textwidth} 
        \centering
        \includegraphics[width = \linewidth]{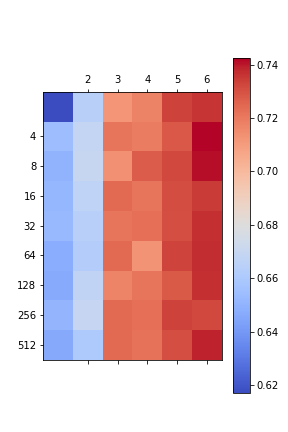}
        \caption{41027}
    \end{subfigure}
    
    \centering
    \begin{subfigure}[b]{0.24\textwidth} 
        \centering
        \includegraphics[width = \linewidth]{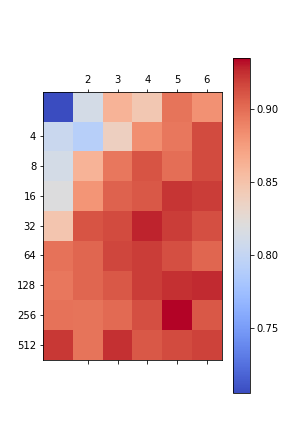}
        \caption{41050}
    \end{subfigure}
    \begin{subfigure}[b]{0.24\textwidth} 
        \centering
        \includegraphics[width = \linewidth]{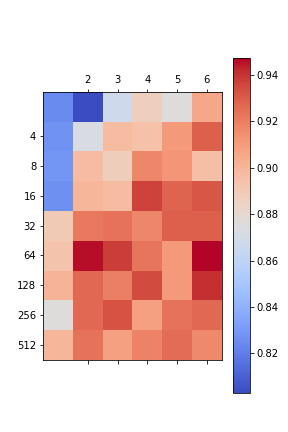}
        \caption{41051}
    \end{subfigure}
    \begin{subfigure}[b]{0.24\textwidth}
        \centering
        \includegraphics[width = \linewidth]{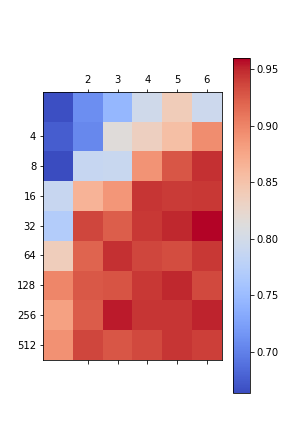}
        \caption{41052}
    \end{subfigure}
    \begin{subfigure}[b]{0.24\textwidth} 
        \centering
        \includegraphics[width = \linewidth]{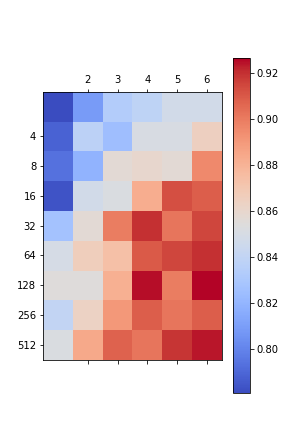}
        \caption{41053}
    \end{subfigure}
    
    \caption{Results of XGBoost with patterns}
    \label{app:fig:heatmap_xgb_pattern}
\end{figure}
%------------------------------------------------------------------
%------------------------------------------------------------------
\begin{figure}[H]
    \centering
    \begin{subfigure}[b]{0.24\textwidth} 
        \centering
        \includegraphics[width = \linewidth]{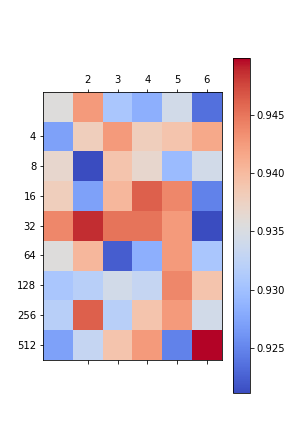}
        \caption{1487}
    \end{subfigure}
    \begin{subfigure}[b]{0.24\textwidth} 
        \centering
        \includegraphics[width = \linewidth]{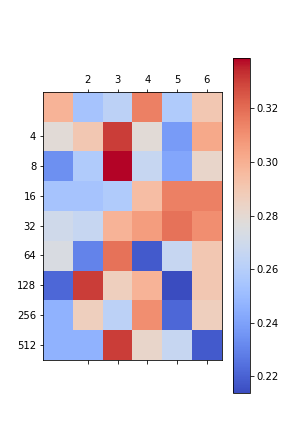}
        \caption{1549}
    \end{subfigure}
    \begin{subfigure}[b]{0.24\textwidth} 
        \centering
        \includegraphics[width = \linewidth]{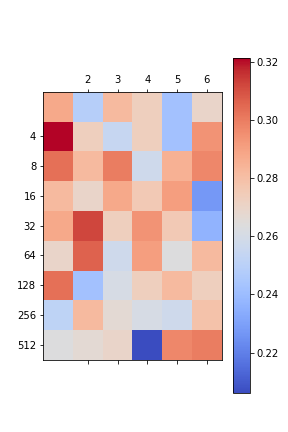}
        \caption{1555}
    \end{subfigure}
    \begin{subfigure}[b]{0.24\textwidth} 
        \centering
        \includegraphics[width = \linewidth]{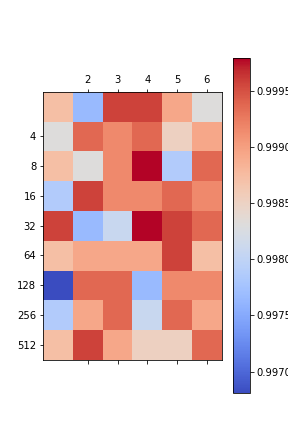}
        \caption{4154}
    \end{subfigure}
    
    \centering
    \begin{subfigure}[b]{0.24\textwidth} 
        \centering
        \includegraphics[width = \linewidth]{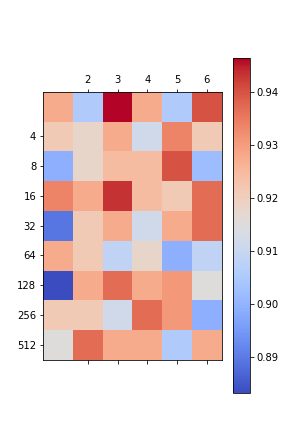}
        \caption{40705}
    \end{subfigure}
    \begin{subfigure}[b]{0.24\textwidth} 
        \centering
        \includegraphics[width = \linewidth]{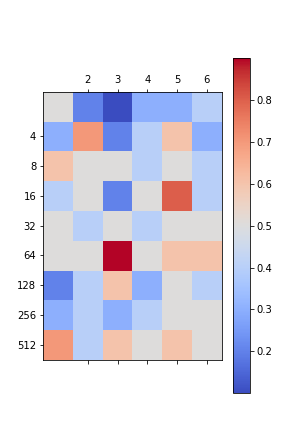}
        \caption{41025}
    \end{subfigure}
    \begin{subfigure}[b]{0.24\textwidth}
        \centering
        \includegraphics[width = \linewidth]{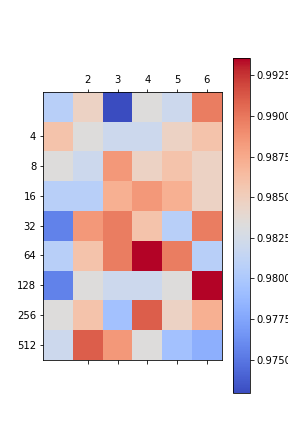}
        \caption{41049}
    \end{subfigure}
    
    \caption{Part of results of XGBoost without obvious pattern}
    \label{app:fig:heatmap_xgb_no2}
\end{figure}
%------------------------------------------------------------------
%------------------------------------------------------------------
\begin{figure}[H]
    \centering
    \begin{subfigure}[b]{0.24\textwidth} 
        \centering
        \includegraphics[width = \linewidth]{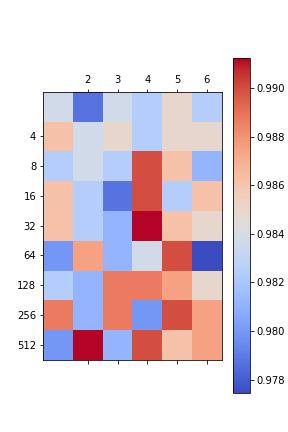}
        \caption{316}
    \end{subfigure}
    \begin{subfigure}[b]{0.24\textwidth} 
        \centering
        \includegraphics[width = \linewidth]{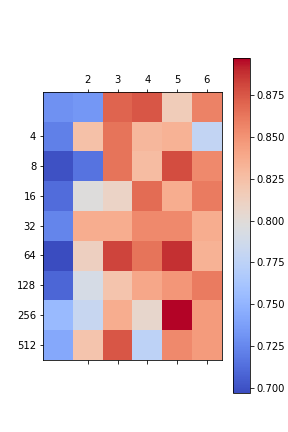}
        \caption{718}
    \end{subfigure}
    \begin{subfigure}[b]{0.24\textwidth} 
        \centering
        \includegraphics[width = \linewidth]{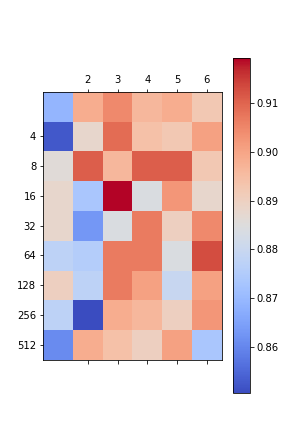}
        \caption{1049}
    \end{subfigure}
    \begin{subfigure}[b]{0.24\textwidth} 
        \centering
        \includegraphics[width = \linewidth]{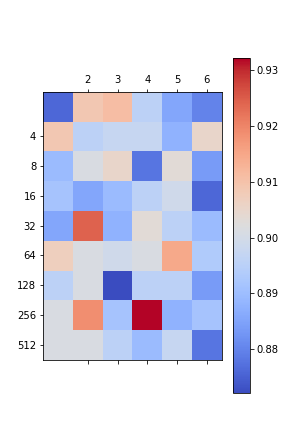}
        \caption{1050}
    \end{subfigure}
    
    \centering
    \begin{subfigure}[b]{0.24\textwidth} 
        \centering
        \includegraphics[width = \linewidth]{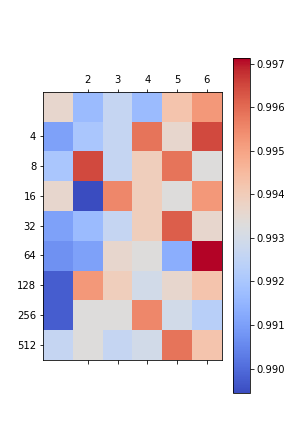}
        \caption{1056}
    \end{subfigure}
    \begin{subfigure}[b]{0.24\textwidth} 
        \centering
        \includegraphics[width = \linewidth]{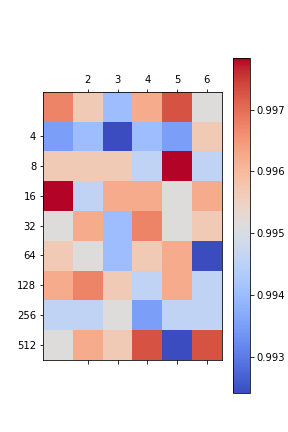}
        \caption{1069}
    \end{subfigure}
    \begin{subfigure}[b]{0.24\textwidth}
        \centering
        \includegraphics[width = \linewidth]{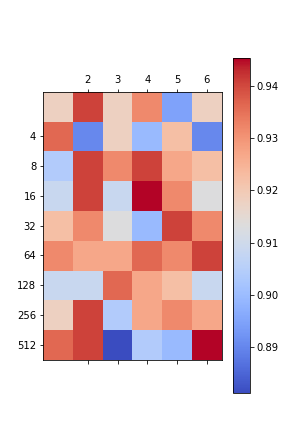}
        \caption{1443}
    \end{subfigure}
    \begin{subfigure}[b]{0.24\textwidth} 
        \centering
        \includegraphics[width = \linewidth]{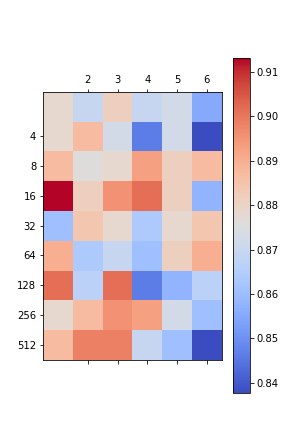}
        \caption{1444}
    \end{subfigure}
    
    \centering
    \begin{subfigure}[b]{0.24\textwidth} 
        \centering
        \includegraphics[width = \linewidth]{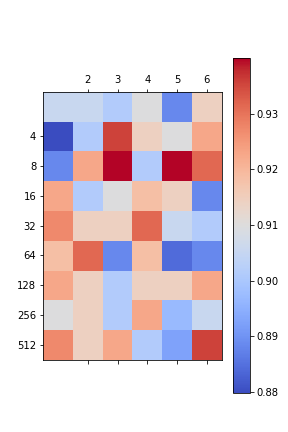}
        \caption{1451}
    \end{subfigure}
    \begin{subfigure}[b]{0.24\textwidth} 
        \centering
        \includegraphics[width = \linewidth]{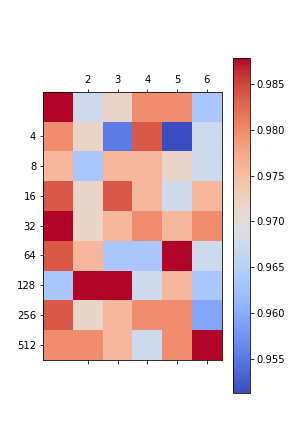}
        \caption{1452}
    \end{subfigure}
    \begin{subfigure}[b]{0.24\textwidth}
        \centering
        \includegraphics[width = \linewidth]{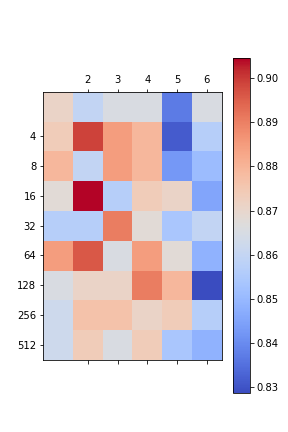}
        \caption{1453}
    \end{subfigure}
    \begin{subfigure}[b]{0.24\textwidth} 
        \centering
        \includegraphics[width = \linewidth]{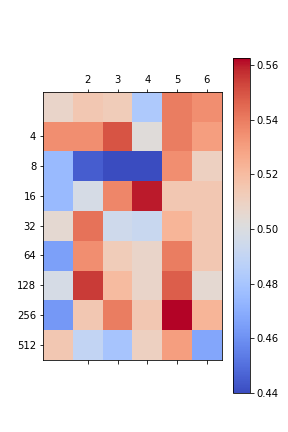}
        \caption{1479}
    \end{subfigure}
    
    \caption{Part of results of XGBoost without obvious pattern}
    \label{app:fig:heatmap_xgb_no1}
\end{figure}

\section{MLP}
\label{subapp:mlp_heat}
\begin{figure}[H]
    \centering
    \begin{subfigure}[b]{0.24\textwidth} 
        \centering
        \includegraphics[width = \linewidth]{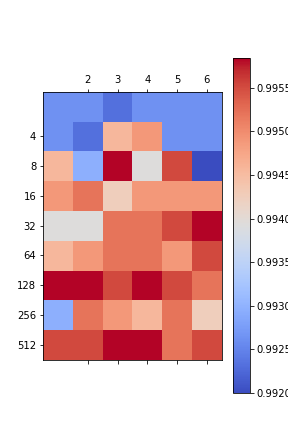}
        \caption{1056}
    \end{subfigure}
    \begin{subfigure}[b]{0.24\textwidth} 
        \centering
        \includegraphics[width = \linewidth]{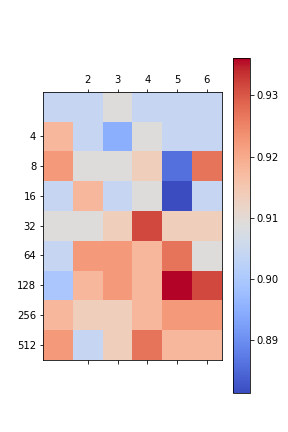}
        \caption{1443}
    \end{subfigure}
    \begin{subfigure}[b]{0.24\textwidth}
        \centering
        \includegraphics[width = \linewidth]{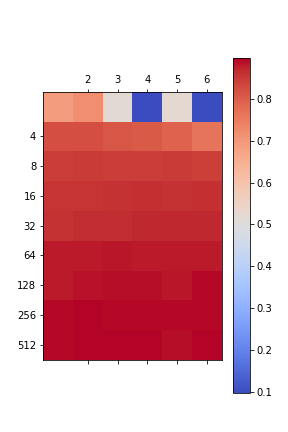}
        \caption{40996}
    \end{subfigure}
    \begin{subfigure}[b]{0.24\textwidth} 
        \centering
        \includegraphics[width = \linewidth]{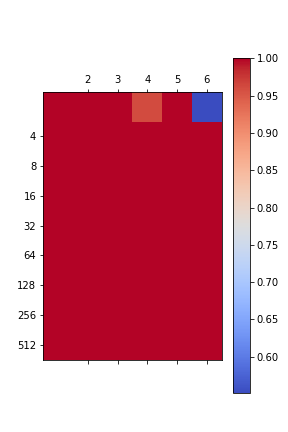}
        \caption{41005}
    \end{subfigure}
    
    \centering
    \begin{subfigure}[b]{0.24\textwidth} 
        \centering
        \includegraphics[width = \linewidth]{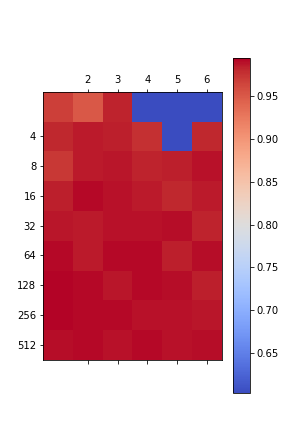}
        \caption{41007}
    \end{subfigure}
    \begin{subfigure}[b]{0.24\textwidth} 
        \centering
        \includegraphics[width = \linewidth]{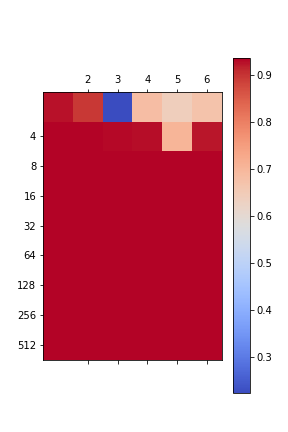}
        \caption{41014}
    \end{subfigure}
    \begin{subfigure}[b]{0.24\textwidth} 
        \centering
        \includegraphics[width = \linewidth]{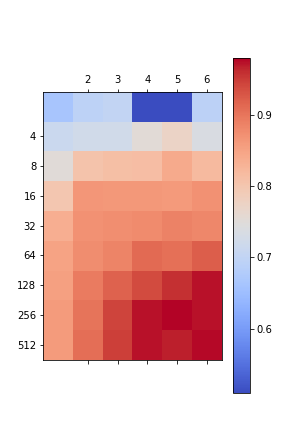}
        \caption{41027}
    \end{subfigure}
    \begin{subfigure}[b]{0.24\textwidth} 
        \centering
        \includegraphics[width = \linewidth]{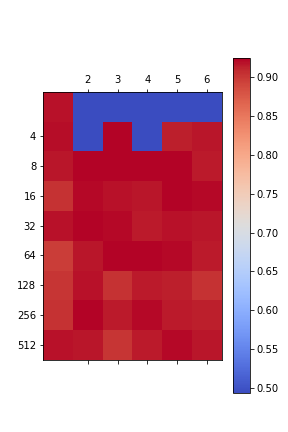}
        \caption{41053}
    \end{subfigure}  
    
    \caption{Results of multi-layer perceptron with expected patterns}
    \label{app:fig:heatmap_mlp1}
\end{figure}
%------------------------------------------------------------------
%------------------------------------------------------------------
\begin{figure}[H]
    \centering
    \begin{subfigure}[b]{0.24\textwidth} 
        \centering
        \includegraphics[width = \linewidth]{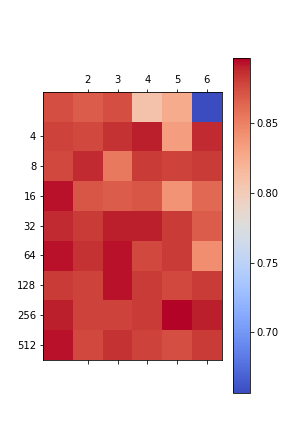}
        \caption{1494}
    \end{subfigure}
    \begin{subfigure}[b]{0.24\textwidth} 
        \centering
        \includegraphics[width = \linewidth]{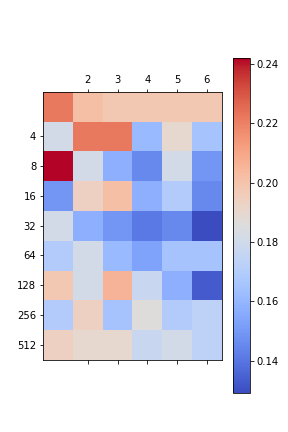}
        \caption{1549}
    \end{subfigure}
    \begin{subfigure}[b]{0.24\textwidth}
        \centering
        \includegraphics[width = \linewidth]{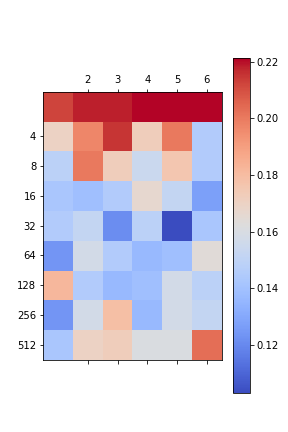}
        \caption{1555}
    \end{subfigure}
    \begin{subfigure}[b]{0.24\textwidth} 
        \centering
        \includegraphics[width = \linewidth]{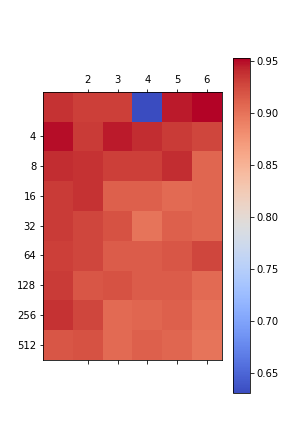}
        \caption{40705}
    \end{subfigure}
    
    \caption{Part of results of multi-layer perceptron with unexpected patterns}
    \label{app:fig:heatmap_mlp02}
\end{figure}
%------------------------------------------------------------------
%------------------------------------------------------------------
\begin{figure}[htbp]
    \centering
    \begin{subfigure}[b]{0.24\textwidth} 
        \centering
        \includegraphics[width = \linewidth]{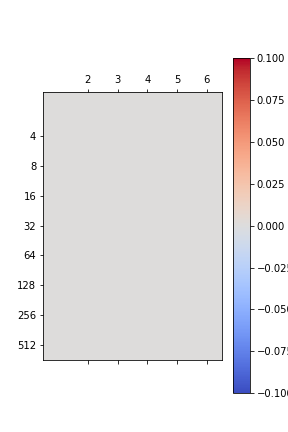}
        \caption{313}
    \end{subfigure}
    \begin{subfigure}[b]{0.24\textwidth} 
        \centering
        \includegraphics[width = \linewidth]{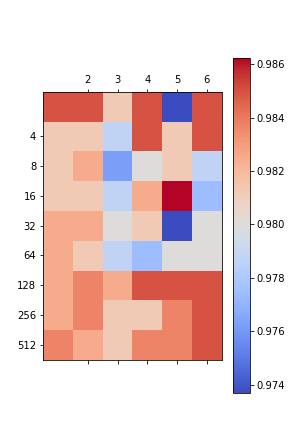}
        \caption{316}
    \end{subfigure}
    \begin{subfigure}[b]{0.24\textwidth}
        \centering
        \includegraphics[width = \linewidth]{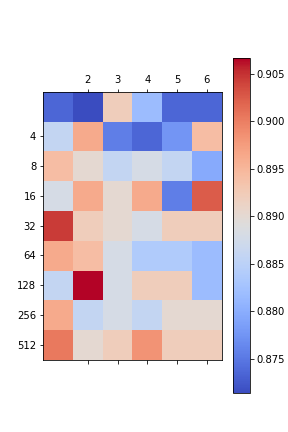}
        \caption{1049}
    \end{subfigure}
    \begin{subfigure}[b]{0.24\textwidth} 
        \centering
        \includegraphics[width = \linewidth]{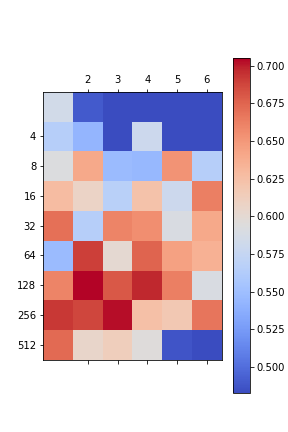}
        \caption{1479}
    \end{subfigure}
    
    \centering
    \begin{subfigure}[b]{0.24\textwidth} 
        \centering
        \includegraphics[width = \linewidth]{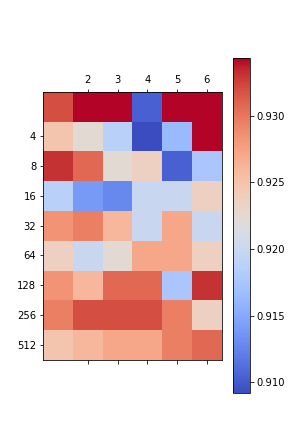}
        \caption{1487}
    \end{subfigure}
    \begin{subfigure}[b]{0.24\textwidth} 
        \centering
        \includegraphics[width = \linewidth]{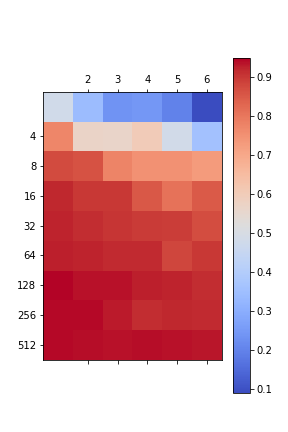}
        \caption{1501}
    \end{subfigure}
    \begin{subfigure}[b]{0.24\textwidth} 
        \centering
        \includegraphics[width = \linewidth]{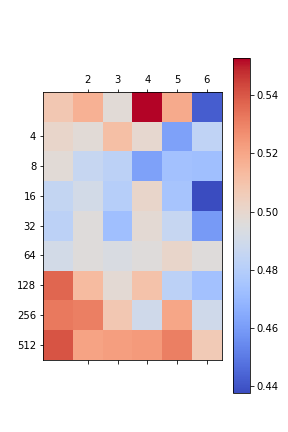}
        \caption{1548}
    \end{subfigure}
    \begin{subfigure}[b]{0.24\textwidth} 
        \centering
        \includegraphics[width = \linewidth]{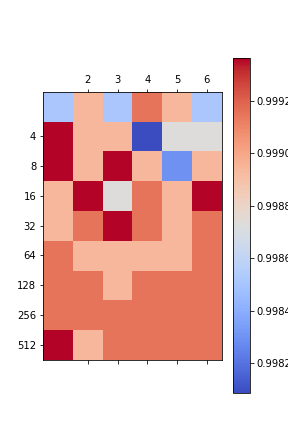}
        \caption{4154}
    \end{subfigure}  
    
    \centering
    \begin{subfigure}[b]{0.24\textwidth} 
        \centering
        \includegraphics[width = \linewidth]{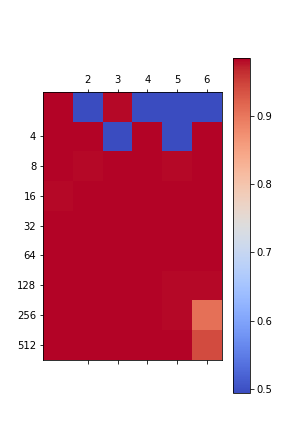}
        \caption{41049}
    \end{subfigure}
    \begin{subfigure}[b]{0.24\textwidth} 
        \centering
        \includegraphics[width = \linewidth]{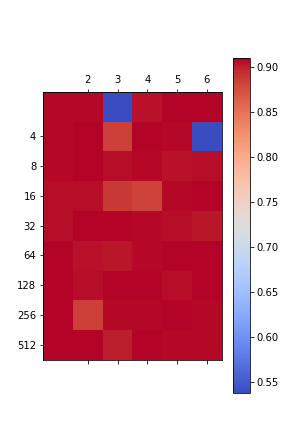}
        \caption{41050}
    \end{subfigure}
    \begin{subfigure}[b]{0.24\textwidth} 
        \centering
        \includegraphics[width = \linewidth]{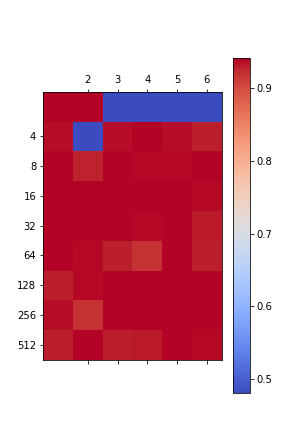}
        \caption{41051}
    \end{subfigure}
    \begin{subfigure}[b]{0.24\textwidth} 
        \centering
        \includegraphics[width = \linewidth]{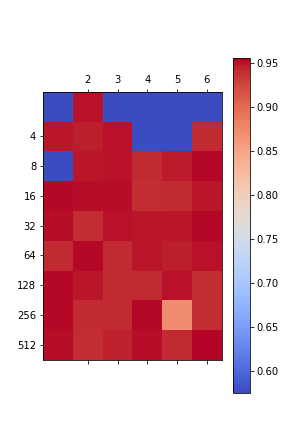}
        \caption{41052}
    \end{subfigure}
    
    \caption{Results of multi-layer perceptron without pattern}
    \label{app:fig:heatmap_mlp_no}
\end{figure}
%------------------------------------------------------------------
%------------------------------------------------------------------
\begin{figure}[htbp]
    \centering
    \begin{subfigure}[b]{0.24\textwidth} 
        \centering
        \includegraphics[width = \linewidth]{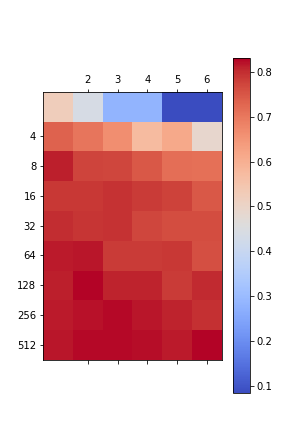}
        \caption{14}
    \end{subfigure}
    \begin{subfigure}[b]{0.24\textwidth} 
        \centering
        \includegraphics[width = \linewidth]{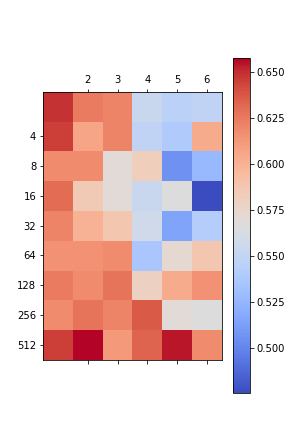}
        \caption{718}
    \end{subfigure}
    \begin{subfigure}[b]{0.24\textwidth}
        \centering
        \includegraphics[width = \linewidth]{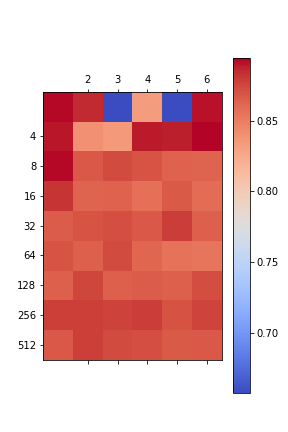}
        \caption{979}
    \end{subfigure}
    \begin{subfigure}[b]{0.24\textwidth} 
        \centering
        \includegraphics[width = \linewidth]{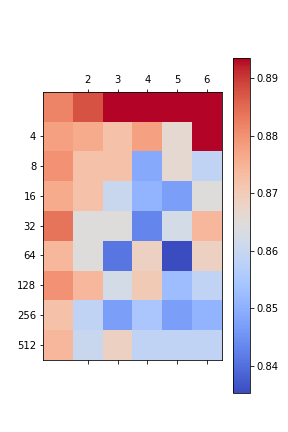}
        \caption{1050}
    \end{subfigure}
    
    \centering
    \begin{subfigure}[b]{0.24\textwidth} 
        \centering
        \includegraphics[width = \linewidth]{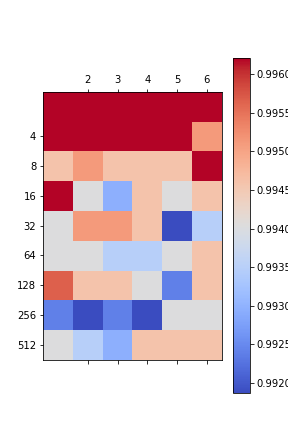}
        \caption{1069}
    \end{subfigure}
    \begin{subfigure}[b]{0.24\textwidth} 
        \centering
        \includegraphics[width = \linewidth]{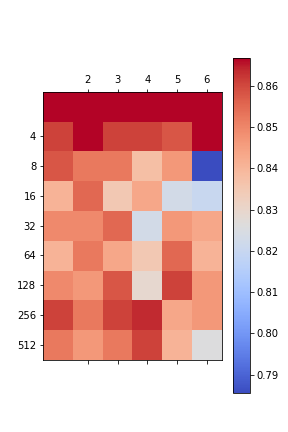}
        \caption{1444}
    \end{subfigure}
    \begin{subfigure}[b]{0.24\textwidth} 
        \centering
        \includegraphics[width = \linewidth]{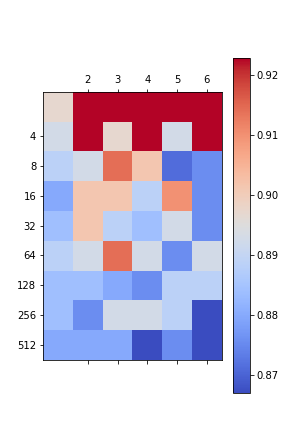}
        \caption{1451}
    \end{subfigure}
    \begin{subfigure}[b]{0.24\textwidth} 
        \centering
        \includegraphics[width = \linewidth]{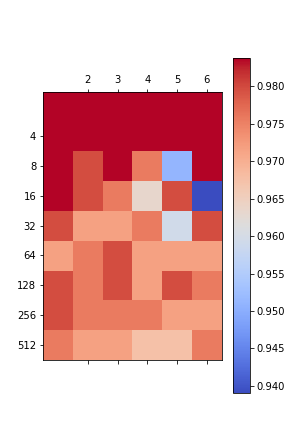}
        \caption{1452}
    \end{subfigure}  
    
    \centering
    \begin{subfigure}[b]{0.24\textwidth} 
        \centering
        \includegraphics[width = \linewidth]{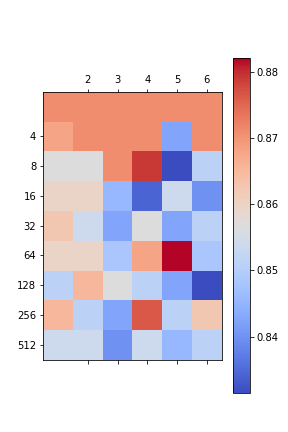}
        \caption{1453}
    \end{subfigure}
    \begin{subfigure}[b]{0.24\textwidth} 
        \centering
        \includegraphics[width = \linewidth]{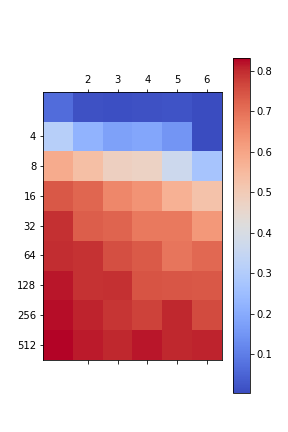}
        \caption{1491}
    \end{subfigure}
    \begin{subfigure}[b]{0.24\textwidth} 
        \centering
        \includegraphics[width = \linewidth]{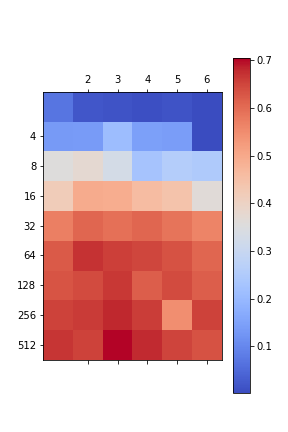}
        \caption{1492}
    \end{subfigure}
    \begin{subfigure}[b]{0.24\textwidth} 
        \centering
        \includegraphics[width = \linewidth]{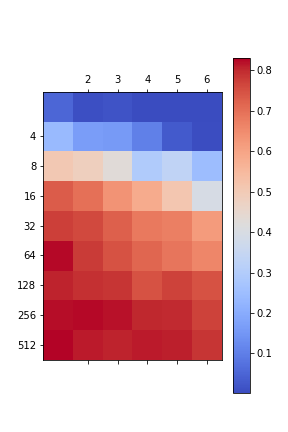}
        \caption{1493}
    \end{subfigure}
    
    \caption{Part of results of multi-layer perceptron with unexpected patterns}
    \label{app:fig:heatmap_mlp01}
\end{figure}
%------------------------------------------------------------------
%------------------------------------------------------------------
\begin{figure}[H]
    \centering
    \includegraphics[width=0.24\linewidth]{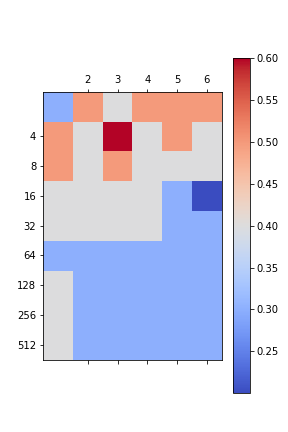}
    \caption{Multi-layer perceptron with unexpected pattern on dataset 41025}
    \label{fig:heatmap_mlp0_41025}
\end{figure}

\chapter{Feature Extraction}
\label{app:fea_ext}

\section{Permutation Importance}
\label{subapp:rf_fea_ext}
\begin{figure}[H]
    \centering
    \begin{subfigure}[b]{0.24\textwidth} 
        \centering
        \includegraphics[width = \linewidth]{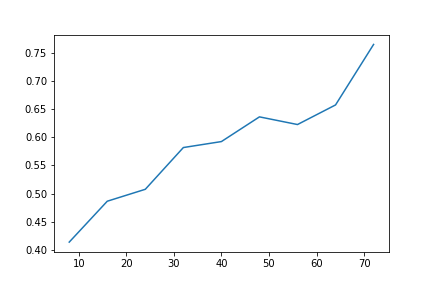}
        \caption{14}
    \end{subfigure}
    \begin{subfigure}[b]{0.24\textwidth} 
        \centering
        \includegraphics[width = \linewidth]{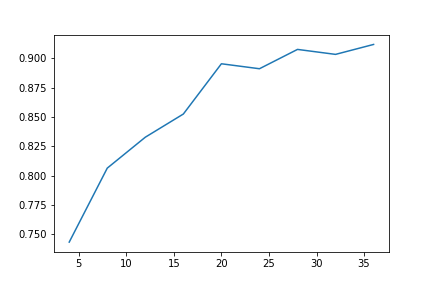}
        \caption{182}
    \end{subfigure}
    \begin{subfigure}[b]{0.24\textwidth} 
        \centering
        \includegraphics[width = \linewidth]{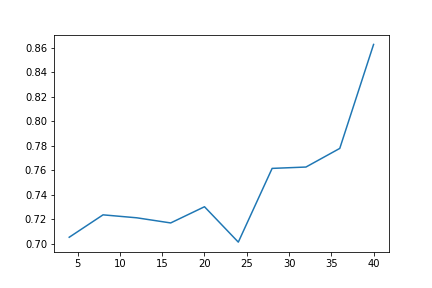}
        \caption{734}
    \end{subfigure}
    \begin{subfigure}[b]{0.24\textwidth}
        \centering
        \includegraphics[width = \linewidth]{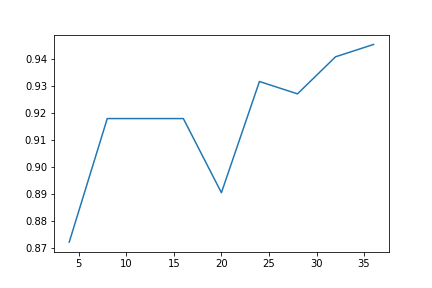}
        \caption{1443}
    \end{subfigure}
    
    \begin{subfigure}[b]{0.24\textwidth}
        \centering
        \includegraphics[width = \linewidth]{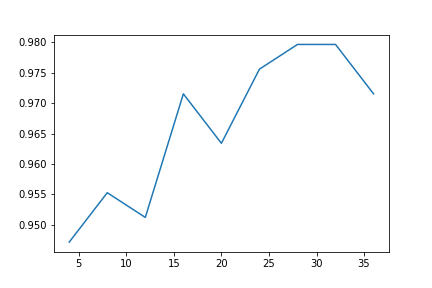}
        \caption{1452}
    \end{subfigure}
    \begin{subfigure}[b]{0.24\textwidth} 
        \centering
        \includegraphics[width = \linewidth]{Figures/imp_per_rf/4534_5_forest_mlp_4_128.png}
        \caption{4534}
    \end{subfigure}
    \begin{subfigure}[b]{0.24\textwidth}
        \centering
        \includegraphics[width = \linewidth]{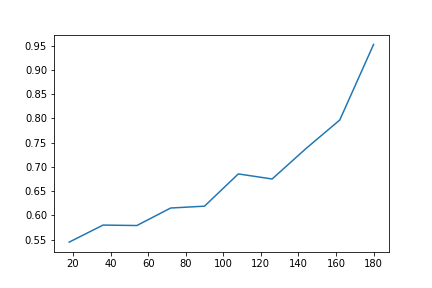}
        \caption{40670}
    \end{subfigure}
    \begin{subfigure}[b]{0.24\textwidth} 
        \centering
        \includegraphics[width = \linewidth]{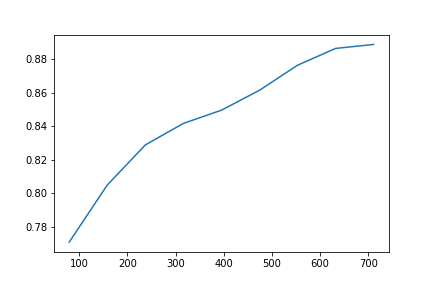}
        \caption{40996}
    \end{subfigure}

    \begin{subfigure}[b]{0.24\textwidth}
        \centering
        \includegraphics[width = \linewidth]{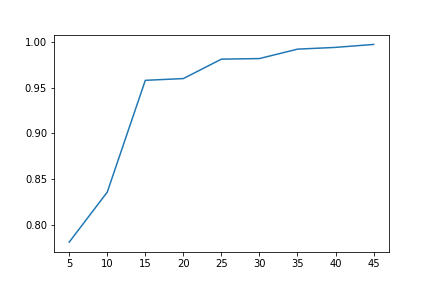}
        \caption{41004}
    \end{subfigure}
    \begin{subfigure}[b]{0.24\textwidth} 
        \centering
        \includegraphics[width = \linewidth]{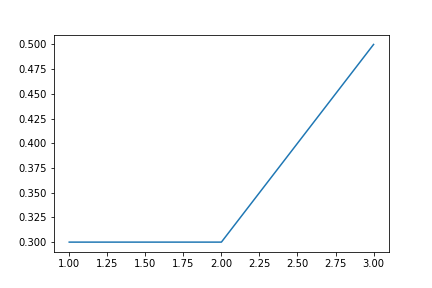}
        \caption{41025}
    \end{subfigure}
    \caption{Increasing results for extractor with permutation importance and highest $\rho$ values}
    \label{app:fig:imp_rf_both}
\end{figure}
%------------------------------------------------------------------
%------------------------------------------------------------------
\begin{figure}[ht]
    \centering
    \begin{subfigure}[b]{0.24\textwidth} 
        \centering
        \includegraphics[width = \linewidth]{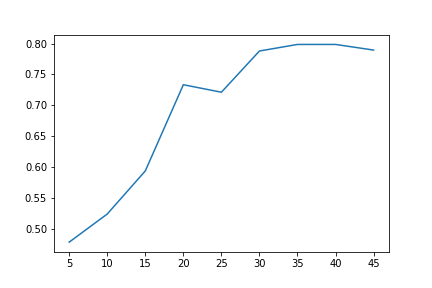}
        \caption{22}
    \end{subfigure}
    \begin{subfigure}[b]{0.24\textwidth} 
        \centering
        \includegraphics[width = \linewidth]{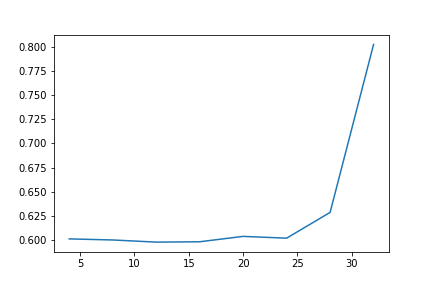}
        \caption{833}
    \end{subfigure}
    \begin{subfigure}[b]{0.24\textwidth} 
        \centering
        \includegraphics[width = \linewidth]{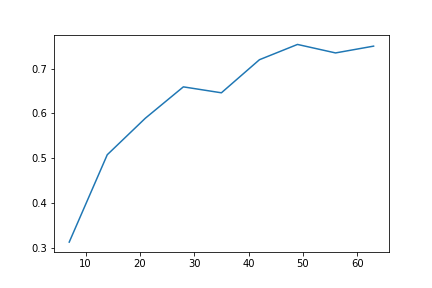}
        \caption{1491}
    \end{subfigure}
    
    \begin{subfigure}[b]{0.24\textwidth}
        \centering
        \includegraphics[width = \linewidth]{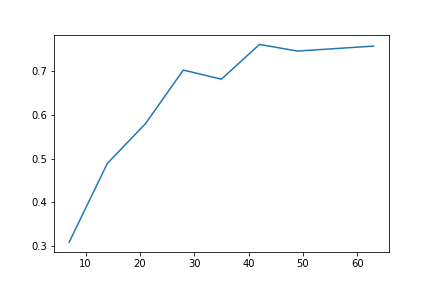}
        \caption{1493}
    \end{subfigure}
    \begin{subfigure}[b]{0.24\textwidth}
        \centering
        \includegraphics[width = \linewidth]{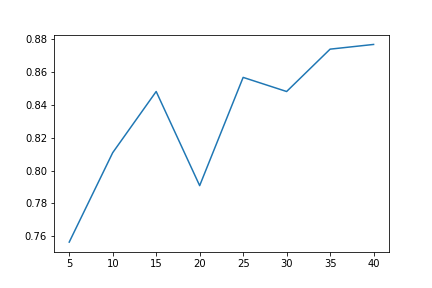}
        \caption{1494}
    \end{subfigure}
    \begin{subfigure}[b]{0.24\textwidth} 
        \centering
        \includegraphics[width = \linewidth]{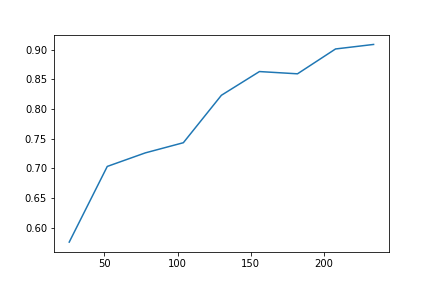}
        \caption{1501}
    \end{subfigure}
    
    \begin{subfigure}[b]{0.24\textwidth}
        \centering
        \includegraphics[width = \linewidth]{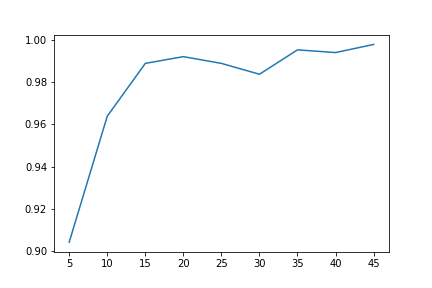}
        \caption{40997}
    \end{subfigure}
    \begin{subfigure}[b]{0.24\textwidth} 
        \centering
        \includegraphics[width = \linewidth]{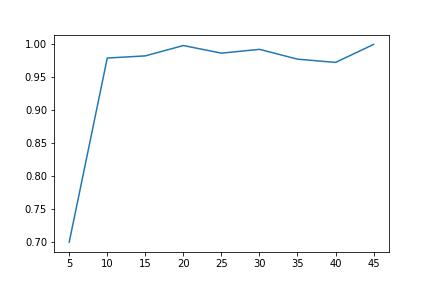}
        \caption{41005}
    \end{subfigure}
    \begin{subfigure}[b]{0.24\textwidth}
        \centering
        \includegraphics[width = \linewidth]{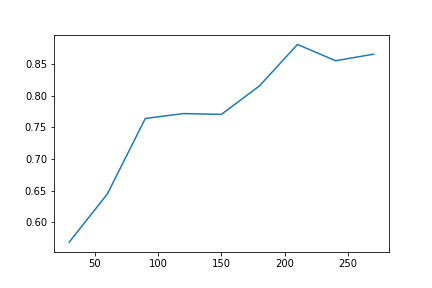}
        \caption{41053}
    \end{subfigure}
    \caption{Increasing results for extractor with permutation importance but not highest $\rho$ value}
    \label{app:fig:imp_rf_obs}
\end{figure}
%------------------------------------------------------------------
%------------------------------------------------------------------
\begin{figure}[ht]
    \centering
    \begin{subfigure}[b]{0.24\textwidth} 
        \centering
        \includegraphics[width = \linewidth]{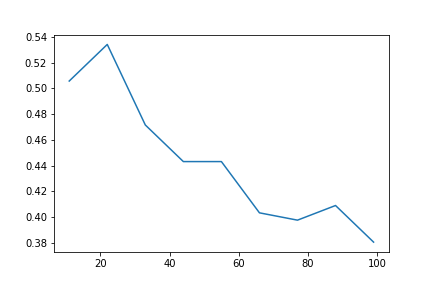}
        \caption{313}
    \end{subfigure}
    \begin{subfigure}[b]{0.24\textwidth} 
        \centering
        \includegraphics[width = \linewidth]{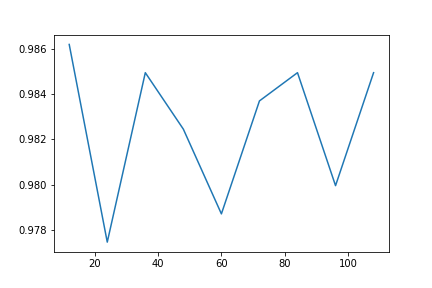}
        \caption{316}
    \end{subfigure}
    \begin{subfigure}[b]{0.24\textwidth} 
        \centering
        \includegraphics[width = \linewidth]{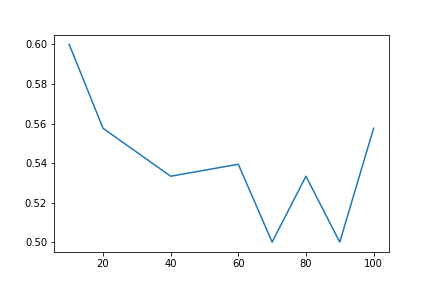}
        \caption{718}
    \end{subfigure}
    \begin{subfigure}[b]{0.24\textwidth}
        \centering
        \includegraphics[width = \linewidth]{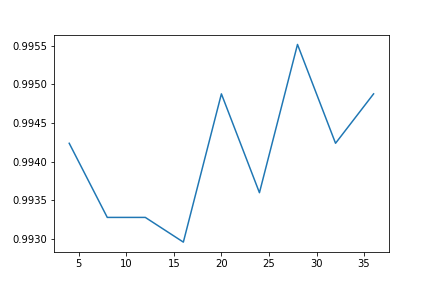}
        \caption{1056}
    \end{subfigure}
    
    \begin{subfigure}[b]{0.24\textwidth}
        \centering
        \includegraphics[width = \linewidth]{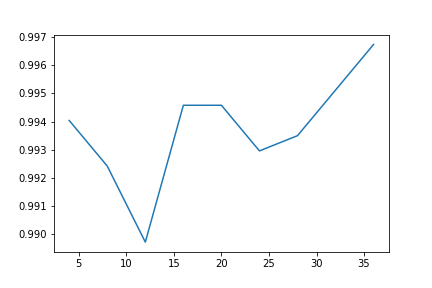}
        \caption{1069}
    \end{subfigure}
    \begin{subfigure}[b]{0.24\textwidth} 
        \centering
        \includegraphics[width = \linewidth]{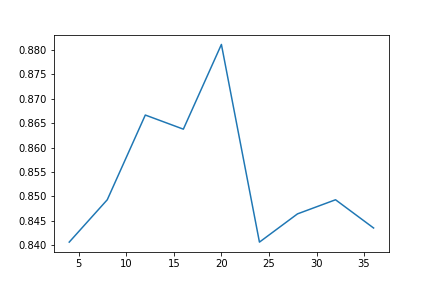}
        \caption{1444}
    \end{subfigure}
    \begin{subfigure}[b]{0.24\textwidth}
        \centering
        \includegraphics[width = \linewidth]{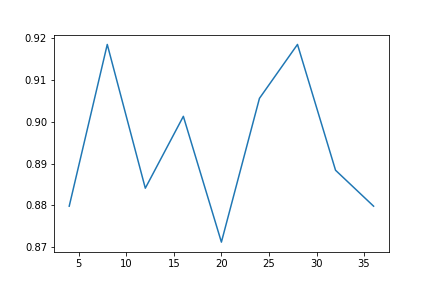}
        \caption{1451}
    \end{subfigure}
    \begin{subfigure}[b]{0.24\textwidth} 
        \centering
        \includegraphics[width = \linewidth]{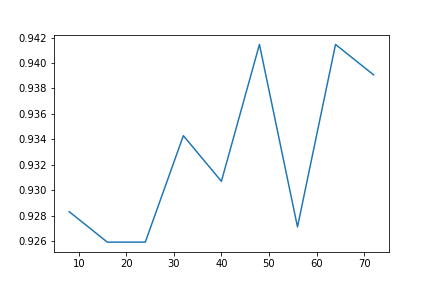}
        \caption{1487}
    \end{subfigure}
    
    \begin{subfigure}[b]{0.24\textwidth}
        \centering
        \includegraphics[width = \linewidth]{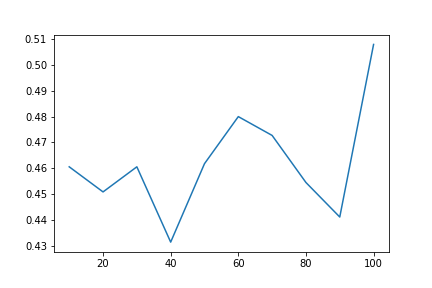}
        \caption{1548}
    \end{subfigure}
    \begin{subfigure}[b]{0.24\textwidth} 
        \centering
        \includegraphics[width = \linewidth]{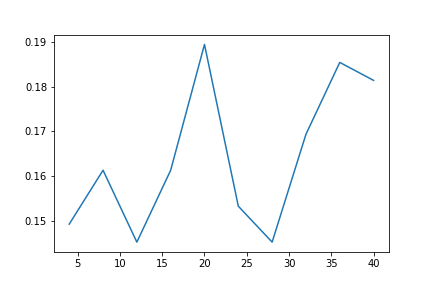}
        \caption{1549}
    \end{subfigure}
    \begin{subfigure}[b]{0.24\textwidth}
        \centering
        \includegraphics[width = \linewidth]{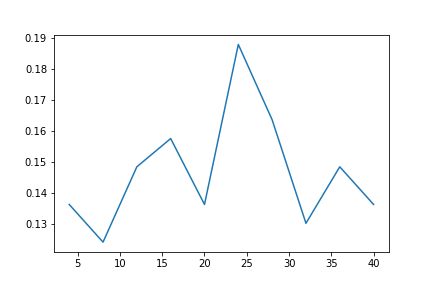}
        \caption{1555}
    \end{subfigure}
    \begin{subfigure}[b]{0.24\textwidth} 
        \centering
        \includegraphics[width = \linewidth]{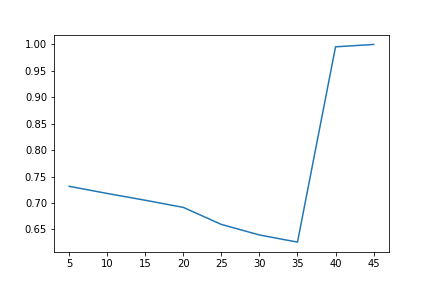}
        \caption{41000}
    \end{subfigure}
    \caption{Part of non-increasing results for extractor with permutation importance}
    \label{app:fig:imp_rf_no1}
\end{figure}
%------------------------------------------------------------------
%------------------------------------------------------------------
\begin{figure}[htb]
    \centering
    \begin{subfigure}[b]{0.24\textwidth} 
        \centering
        \includegraphics[width = \linewidth]{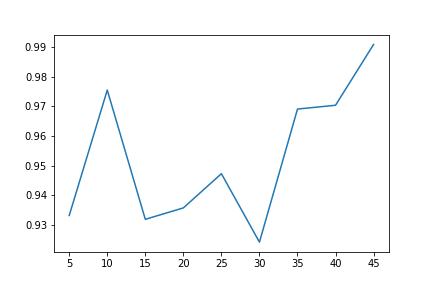}
        \caption{41007}
    \end{subfigure}
    \begin{subfigure}[b]{0.24\textwidth} 
        \centering
        \includegraphics[width = \linewidth]{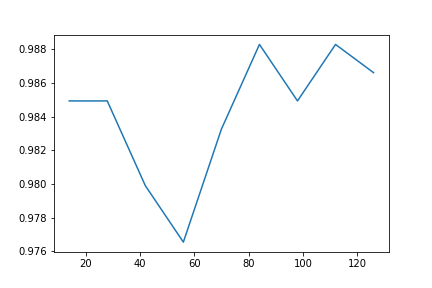}
        \caption{41048}
    \end{subfigure}
    \begin{subfigure}[b]{0.24\textwidth} 
        \centering
        \includegraphics[width = \linewidth]{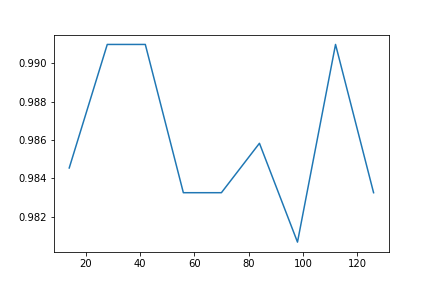}
        \caption{41049}
    \end{subfigure}
    
    \begin{subfigure}[b]{0.24\textwidth}
        \centering
        \includegraphics[width = \linewidth]{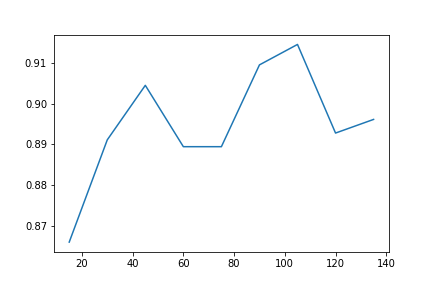}
        \caption{41050}
    \end{subfigure}
    \begin{subfigure}[b]{0.24\textwidth}
        \centering
        \includegraphics[width = \linewidth]{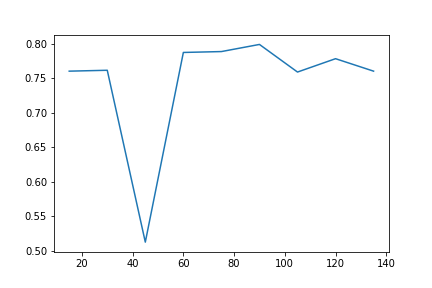}
        \caption{41051}
    \end{subfigure}
    \begin{subfigure}[b]{0.24\textwidth} 
        \centering
        \includegraphics[width = \linewidth]{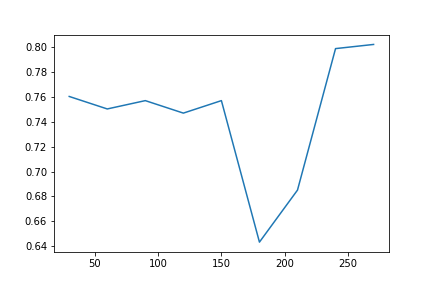}
        \caption{41052}
    \end{subfigure}
    \caption{Part of non-increasing results for extractor with permutation importance}
    \label{app:fig:imp_rf_no2}
\end{figure}

\section{Gain-Based Feature Importance}
\label{subapp:xgb_fea_ext}
\begin{figure}[H]
    \centering
    \begin{subfigure}[b]{0.24\textwidth} 
        \centering
        \includegraphics[width = \linewidth]{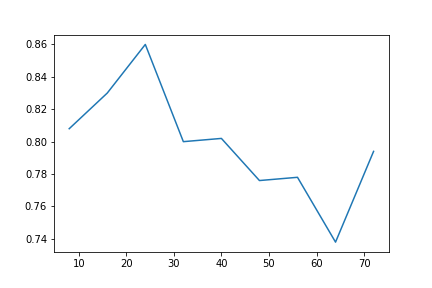}
        \caption{14}
    \end{subfigure}
    \begin{subfigure}[b]{0.24\textwidth} 
        \centering
        \includegraphics[width = \linewidth]{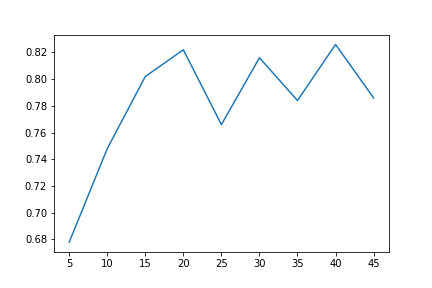}
        \caption{22}
    \end{subfigure}
    \begin{subfigure}[b]{0.24\textwidth}
        \centering
        \includegraphics[width = \linewidth]{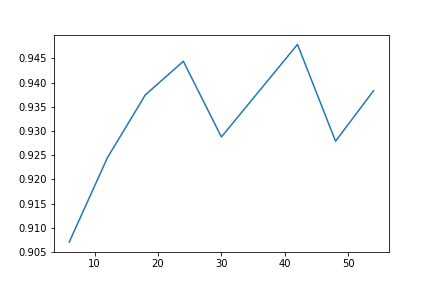}
        \caption{44}
    \end{subfigure}
    \begin{subfigure}[b]{0.24\textwidth} 
        \centering
        \includegraphics[width = \linewidth]{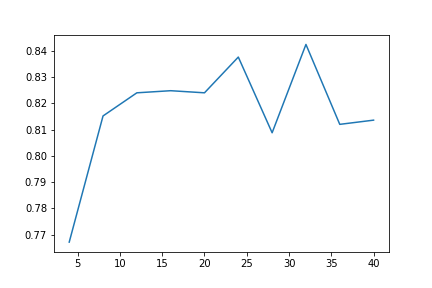}
        \caption{60}
    \end{subfigure}
    
    \centering
    \begin{subfigure}[b]{0.24\textwidth} 
        \centering
        \includegraphics[width = \linewidth]{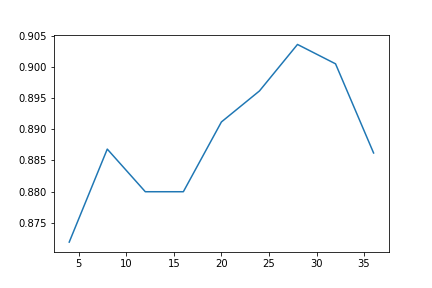}
        \caption{182}
    \end{subfigure}
    \begin{subfigure}[b]{0.24\textwidth} 
        \centering
        \includegraphics[width = \linewidth]{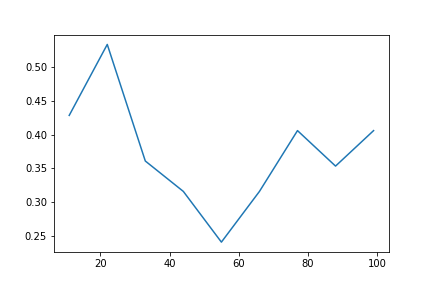}
        \caption{313}
    \end{subfigure}
    \begin{subfigure}[b]{0.24\textwidth} 
        \centering
        \includegraphics[width = \linewidth]{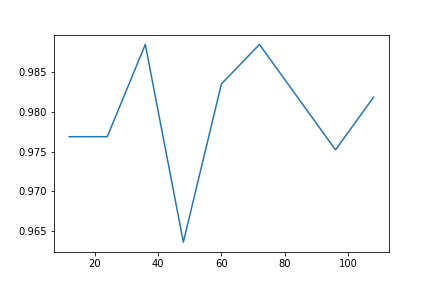}
        \caption{316}
    \end{subfigure}
    \begin{subfigure}[b]{0.24\textwidth}
        \centering
        \includegraphics[width = \linewidth]{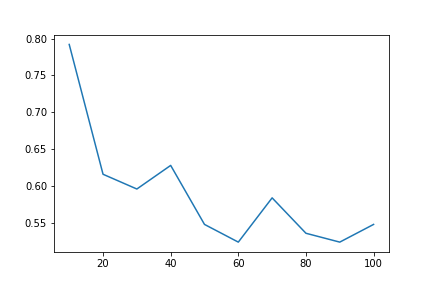}
        \caption{718}
    \end{subfigure}    
    
    \centering
    \begin{subfigure}[b]{0.24\textwidth} 
        \centering
        \includegraphics[width = \linewidth]{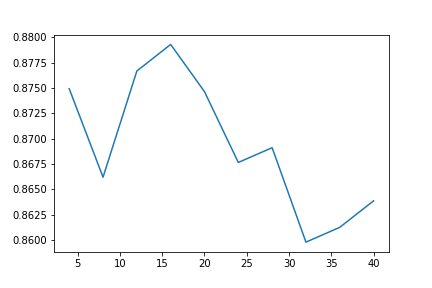}
        \caption{734}
    \end{subfigure}
    \begin{subfigure}[b]{0.24\textwidth} 
        \centering
        \includegraphics[width = \linewidth]{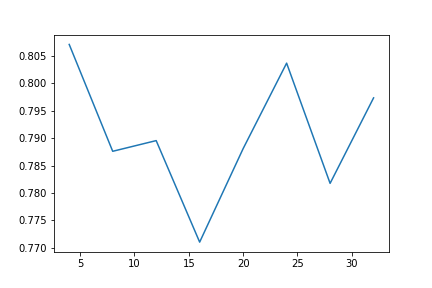}
        \caption{833}
    \end{subfigure}
    \begin{subfigure}[b]{0.24\textwidth} 
        \centering
        \includegraphics[width = \linewidth]{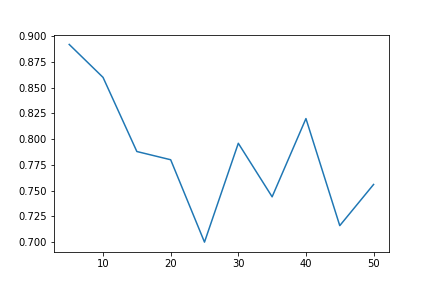}
        \caption{904}
    \end{subfigure}
    \begin{subfigure}[b]{0.24\textwidth}
        \centering
        \includegraphics[width = \linewidth]{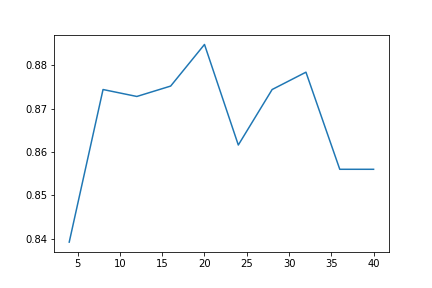}
        \caption{979}
    \end{subfigure}
    
    \centering
    \begin{subfigure}[b]{0.24\textwidth} 
        \centering
        \includegraphics[width = \linewidth]{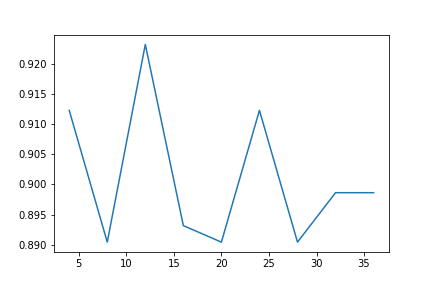}
        \caption{1049}
    \end{subfigure}
    \begin{subfigure}[b]{0.24\textwidth} 
        \centering
        \includegraphics[width = \linewidth]{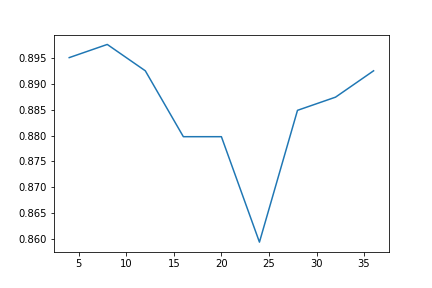}
        \caption{1050}
    \end{subfigure}
    \begin{subfigure}[b]{0.24\textwidth} 
        \centering
        \includegraphics[width = \linewidth]{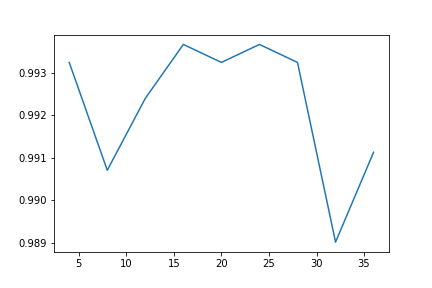}
        \caption{1056}
    \end{subfigure}
    \begin{subfigure}[b]{0.24\textwidth}
        \centering
        \includegraphics[width = \linewidth]{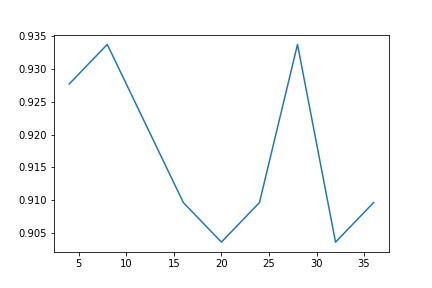}
        \caption{1443}
    \end{subfigure}
    
    \centering
    \begin{subfigure}[b]{0.24\textwidth} 
        \centering
        \includegraphics[width = \linewidth]{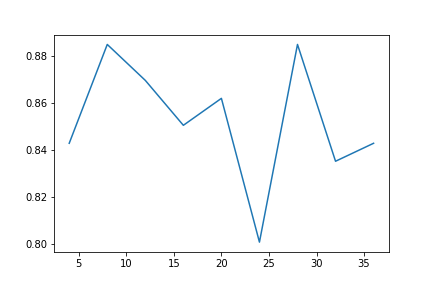}
        \caption{1444}
    \end{subfigure}
    \begin{subfigure}[b]{0.24\textwidth} 
        \centering
        \includegraphics[width = \linewidth]{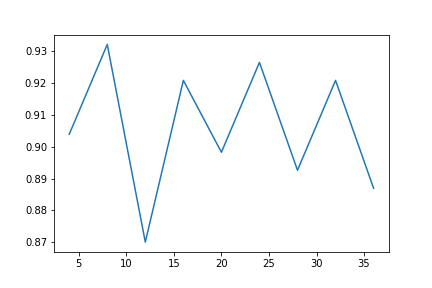}
        \caption{1451}
    \end{subfigure}
    \begin{subfigure}[b]{0.24\textwidth} 
        \centering
        \includegraphics[width = \linewidth]{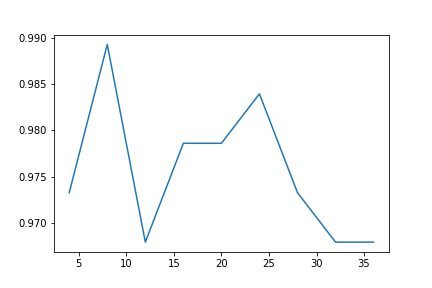}
        \caption{1452}
    \end{subfigure}
    \begin{subfigure}[b]{0.24\textwidth}
        \centering
        \includegraphics[width = \linewidth]{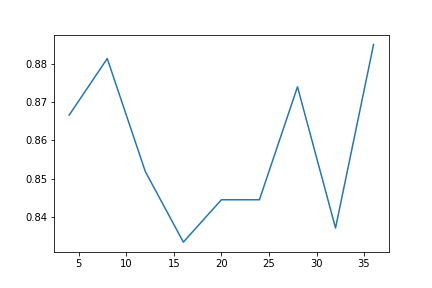}
        \caption{1453}
    \end{subfigure}
    
    \centering
    \begin{subfigure}[b]{0.24\textwidth} 
        \centering
        \includegraphics[width = \linewidth]{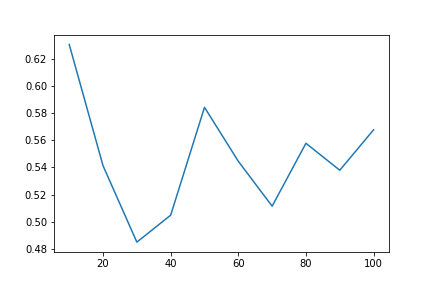}
        \caption{1479}
    \end{subfigure}
    \begin{subfigure}[b]{0.24\textwidth} 
        \centering
        \includegraphics[width = \linewidth]{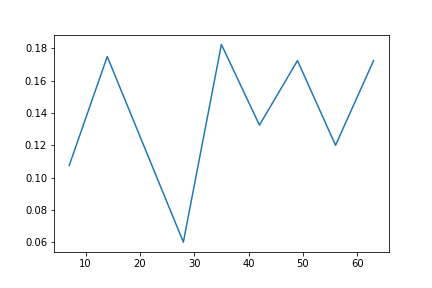}
        \caption{1491}
    \end{subfigure}
    \begin{subfigure}[b]{0.24\textwidth} 
        \centering
        \includegraphics[width = \linewidth]{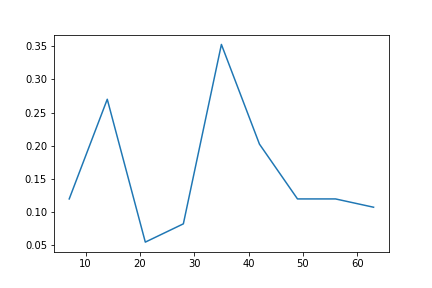}
        \caption{1492}
    \end{subfigure}
    \begin{subfigure}[b]{0.24\textwidth}
        \centering
        \includegraphics[width = \linewidth]{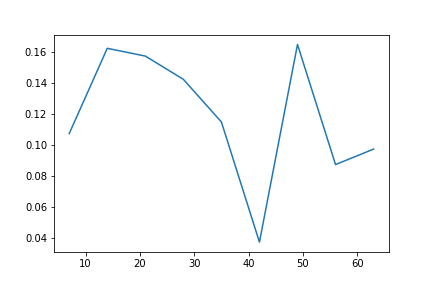}
        \caption{1493}
    \end{subfigure}
    \caption{Non-increasing results of extractor with gain-based importance}
    \label{app:fig:fea_ext_xgb_no1}
\end{figure}
%------------------------------------------------------------------
%------------------------------------------------------------------
\begin{figure}[htbp]
    \centering
    \begin{subfigure}[b]{0.24\textwidth} 
        \centering
        \includegraphics[width = \linewidth]{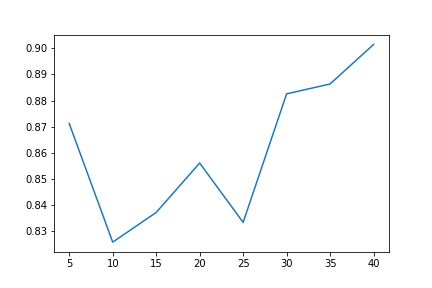}
        \caption{1494}
    \end{subfigure}
    \begin{subfigure}[b]{0.24\textwidth} 
        \centering
        \includegraphics[width = \linewidth]{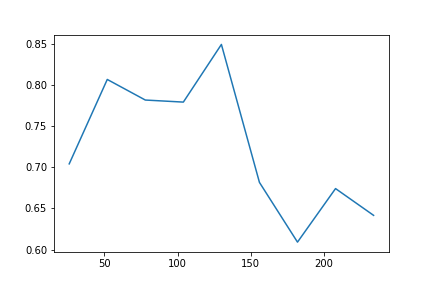}
        \caption{1501}
    \end{subfigure}
    \begin{subfigure}[b]{0.24\textwidth}
        \centering
        \includegraphics[width = \linewidth]{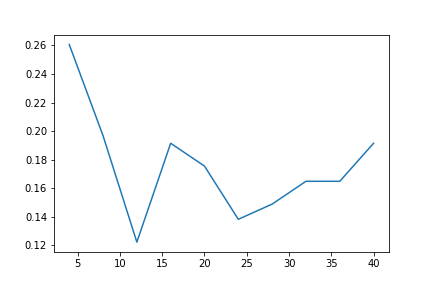}
        \caption{1549}
    \end{subfigure}
    \begin{subfigure}[b]{0.24\textwidth} 
        \centering
        \includegraphics[width = \linewidth]{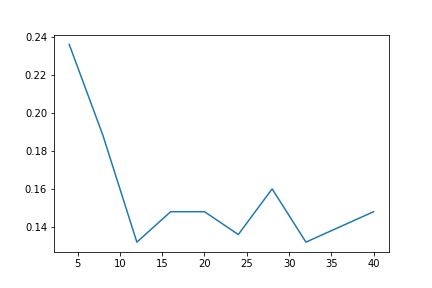}
        \caption{1555}
    \end{subfigure}
    
    \centering
    \begin{subfigure}[b]{0.24\textwidth} 
        \centering
        \includegraphics[width = \linewidth]{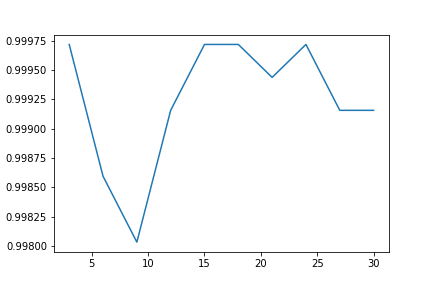}
        \caption{4154}
    \end{subfigure}
    \begin{subfigure}[b]{0.24\textwidth} 
        \centering
        \includegraphics[width = \linewidth]{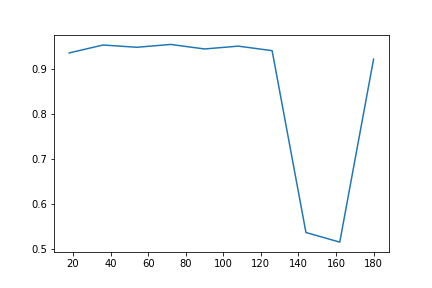}
        \caption{40670}
    \end{subfigure}
    \begin{subfigure}[b]{0.24\textwidth} 
        \centering
        \includegraphics[width = \linewidth]{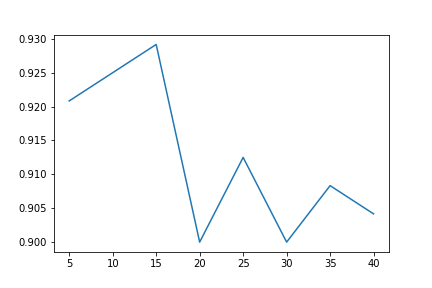}
        \caption{40705}
    \end{subfigure}
    \begin{subfigure}[b]{0.24\textwidth}
        \centering
        \includegraphics[width = \linewidth]{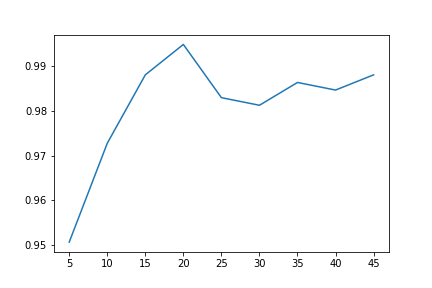}
        \caption{41007}
    \end{subfigure}    
    
    \centering
    \begin{subfigure}[b]{0.24\textwidth} 
        \centering
        \includegraphics[width = \linewidth]{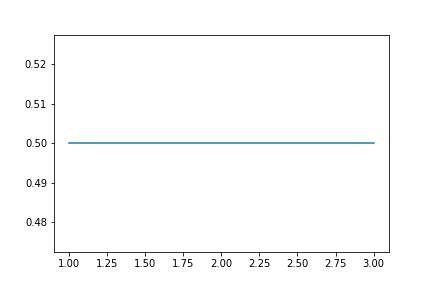}
        \caption{41025}
    \end{subfigure}
    \begin{subfigure}[b]{0.24\textwidth} 
        \centering
        \includegraphics[width = \linewidth]{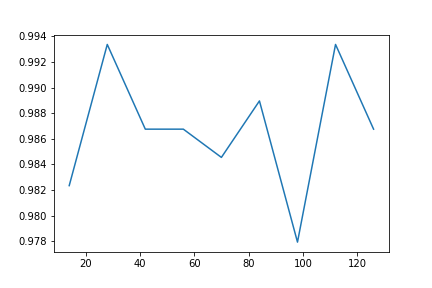}
        \caption{41048}
    \end{subfigure}
    \begin{subfigure}[b]{0.24\textwidth} 
        \centering
        \includegraphics[width = \linewidth]{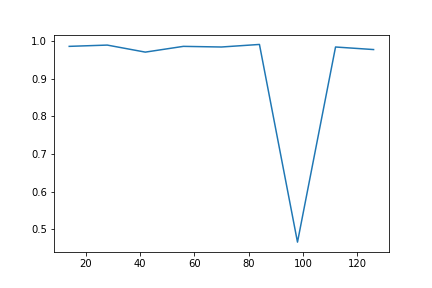}
        \caption{41049}
    \end{subfigure}
    \begin{subfigure}[b]{0.24\textwidth}
        \centering
        \includegraphics[width = \linewidth]{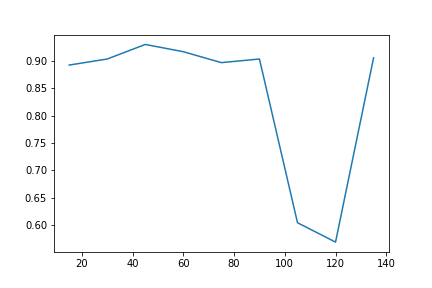}
        \caption{41050}
    \end{subfigure}
    
    \centering
    \begin{subfigure}[b]{0.24\textwidth} 
        \centering
        \includegraphics[width = \linewidth]{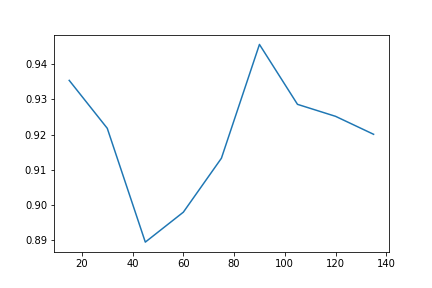}
        \caption{41051}
    \end{subfigure}
    \begin{subfigure}[b]{0.24\textwidth} 
        \centering
        \includegraphics[width = \linewidth]{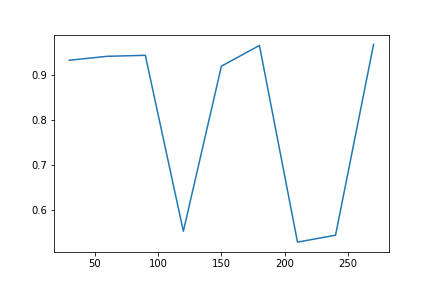}
        \caption{41052}
    \end{subfigure}
    \begin{subfigure}[b]{0.24\textwidth} 
        \centering
        \includegraphics[width = \linewidth]{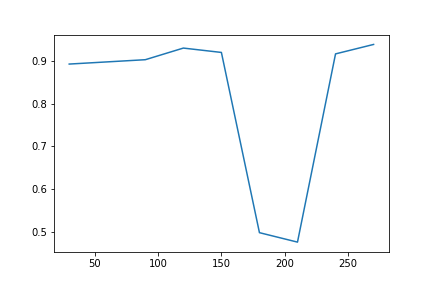}
        \caption{41053}
    \end{subfigure}
    \caption{Non-increasing results of extractor with gain-based importance}
    \label{app:fig:fea_ext_xgb_no2}
\end{figure}

\section{Hierarchical Clustering Based on Spearman Correlation}
\label{subapp:sp_fea_ext}
\subsection{Dendrogram}
\begin{figure}[H]
    \centering
    \begin{subfigure}[b]{0.24\textwidth} 
        \centering
        \includegraphics[width = \linewidth]{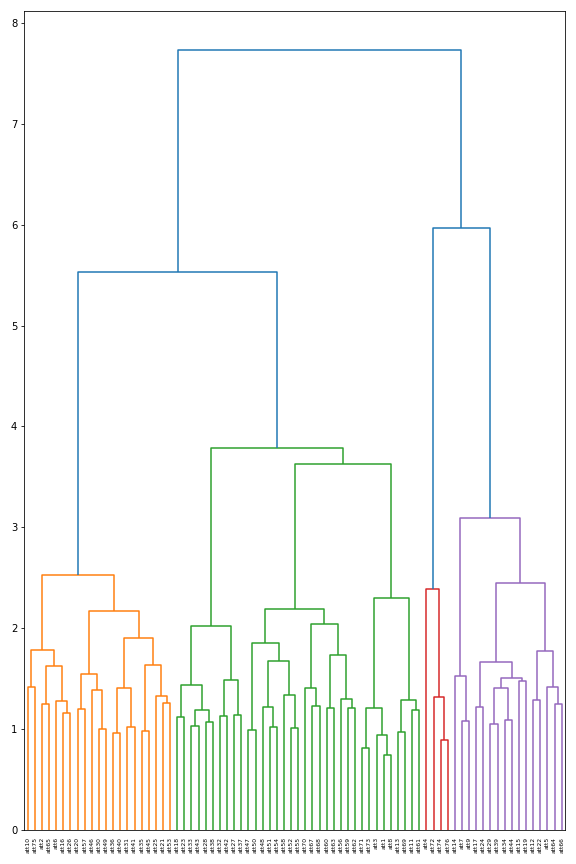}
        \caption{14}
    \end{subfigure}
    \begin{subfigure}[b]{0.24\textwidth} 
        \centering
        \includegraphics[width = \linewidth]{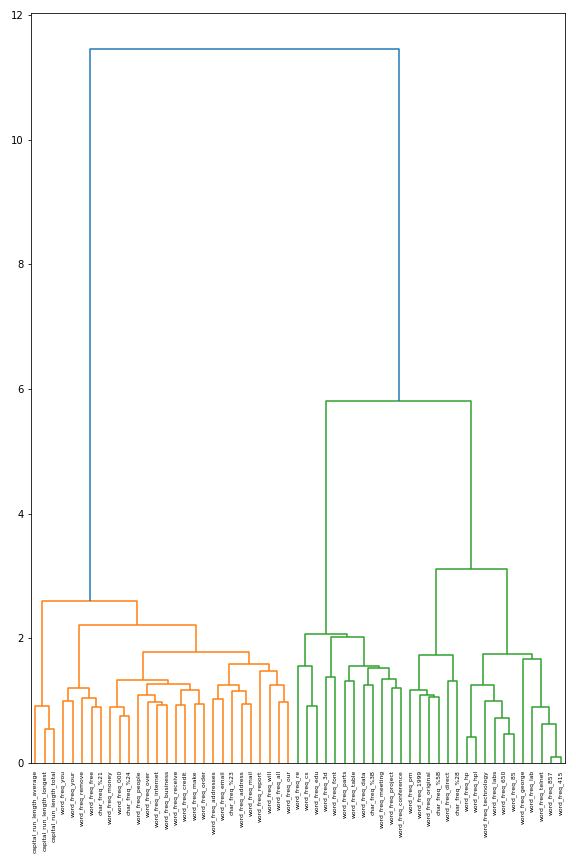}
        \caption{44}
    \end{subfigure}
    \begin{subfigure}[b]{0.24\textwidth}
        \centering
        \includegraphics[width = \linewidth]{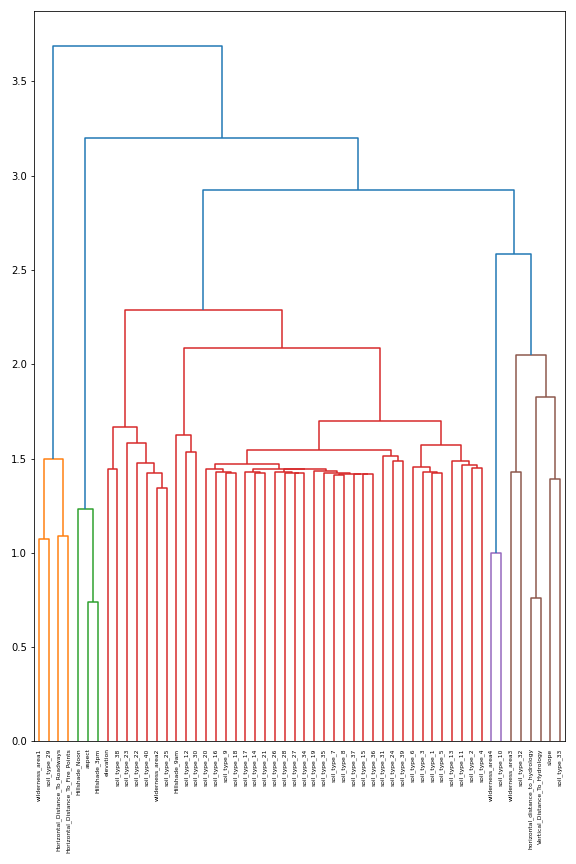}
        \caption{180}
    \end{subfigure}
    \begin{subfigure}[b]{0.24\textwidth}
        \centering
        \includegraphics[width = \linewidth]{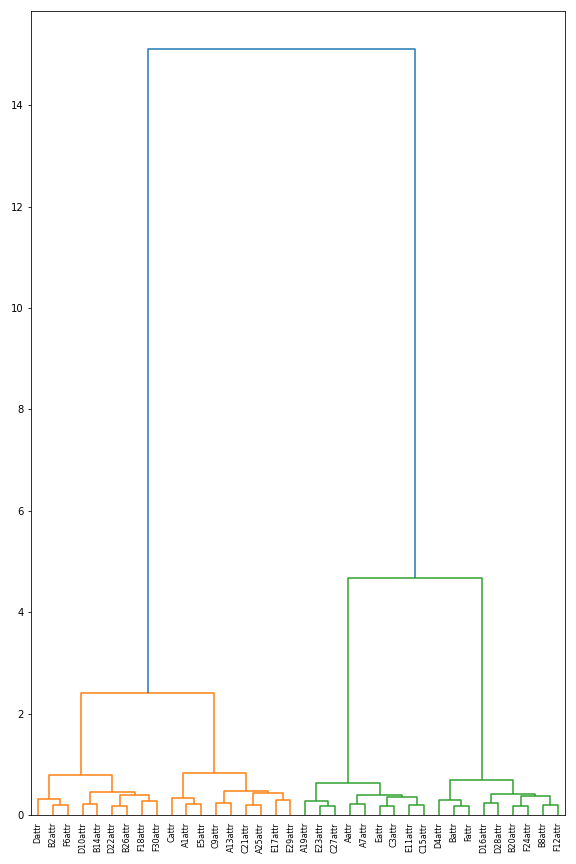}
        \caption{182}
    \end{subfigure}

    \centering
    \begin{subfigure}[b]{0.24\textwidth} 
        \centering
        \includegraphics[width = \linewidth]{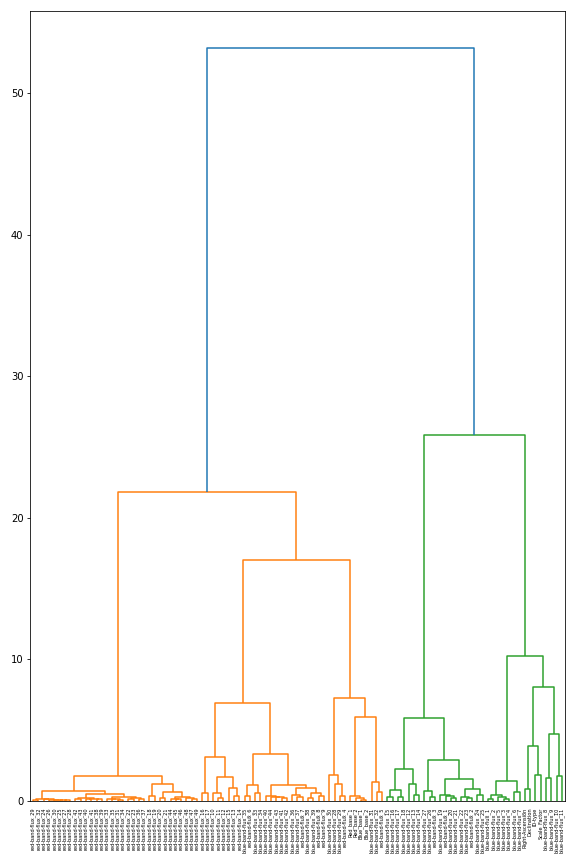}
        \caption{313}
    \end{subfigure}
    \begin{subfigure}[b]{0.24\textwidth} 
        \centering
        \includegraphics[width = \linewidth]{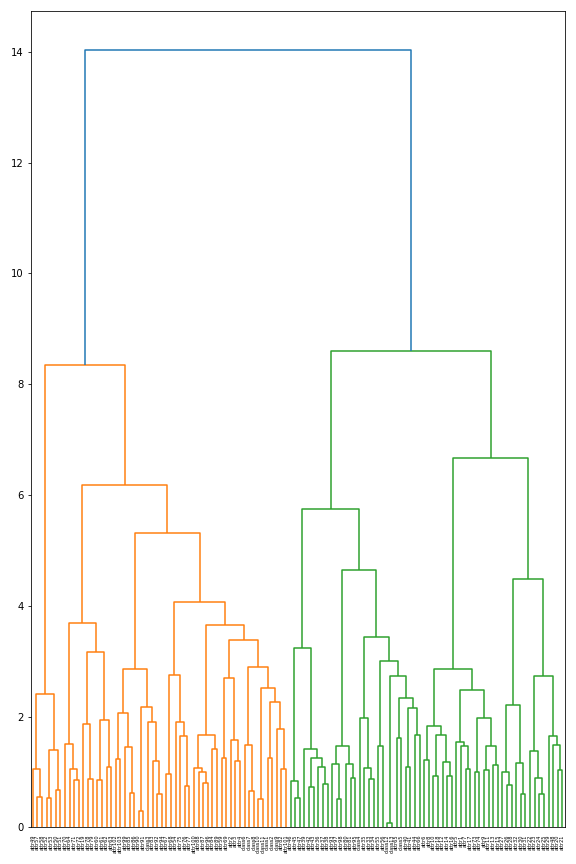}
        \caption{316}
    \end{subfigure}
    \begin{subfigure}[b]{0.24\textwidth}
        \centering
        \includegraphics[width = \linewidth]{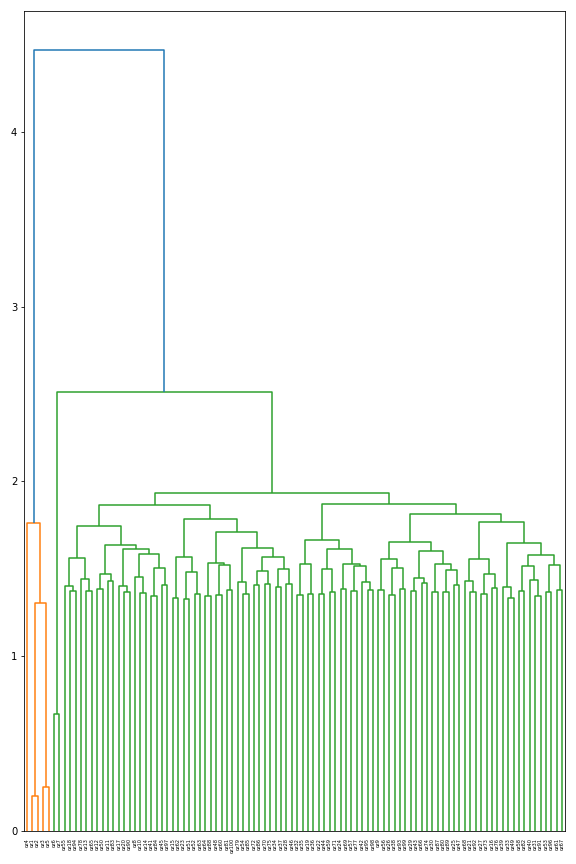}
        \caption{718}
    \end{subfigure}
    \begin{subfigure}[b]{0.24\textwidth}
        \centering
        \includegraphics[width = \linewidth]{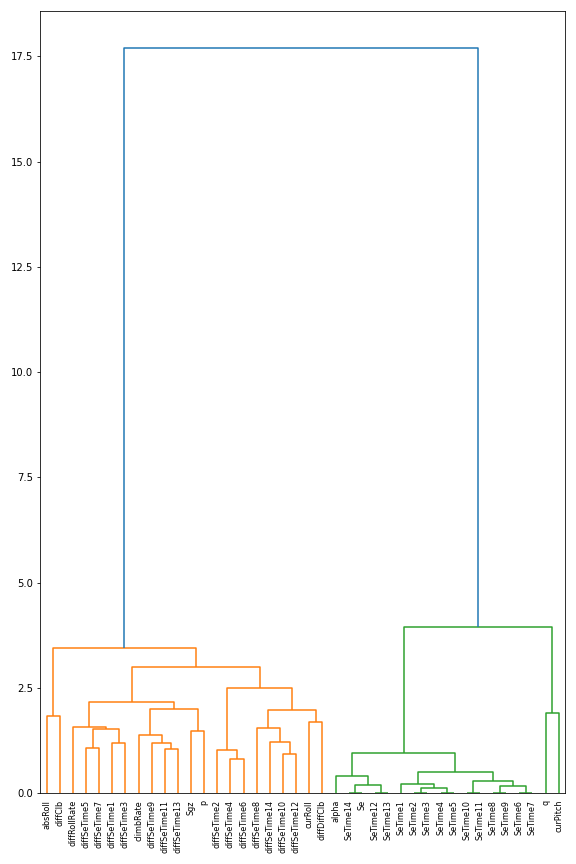}
        \caption{734}
    \end{subfigure}

    \centering
    \begin{subfigure}[b]{0.24\textwidth} 
        \centering
        \includegraphics[width = \linewidth]{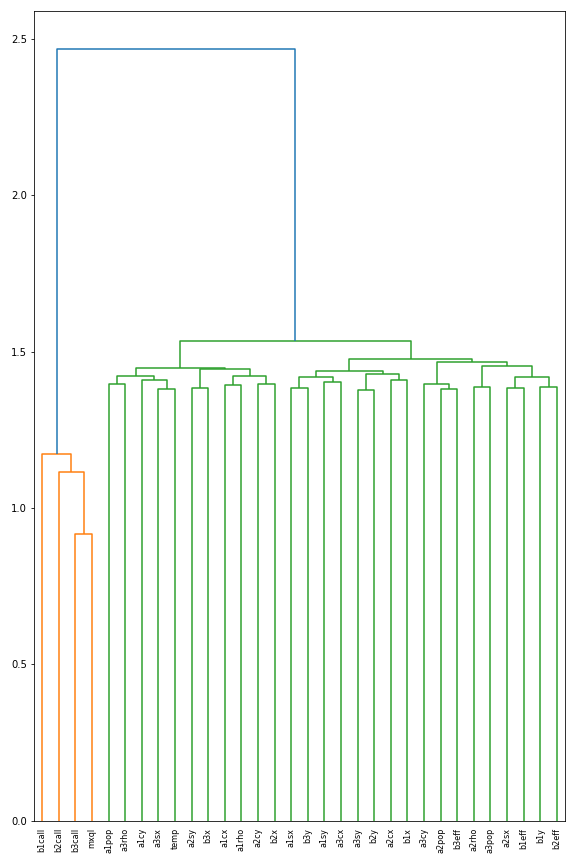}
        \caption{833}
    \end{subfigure}
    \begin{subfigure}[b]{0.24\textwidth} 
        \centering
        \includegraphics[width = \linewidth]{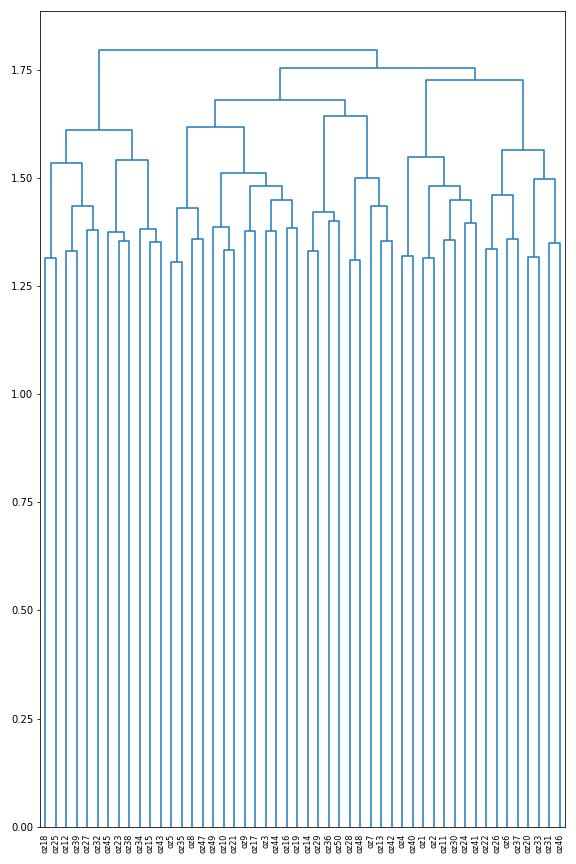}
        \caption{904}
    \end{subfigure}
    \begin{subfigure}[b]{0.24\textwidth}
        \centering
        \includegraphics[width = \linewidth]{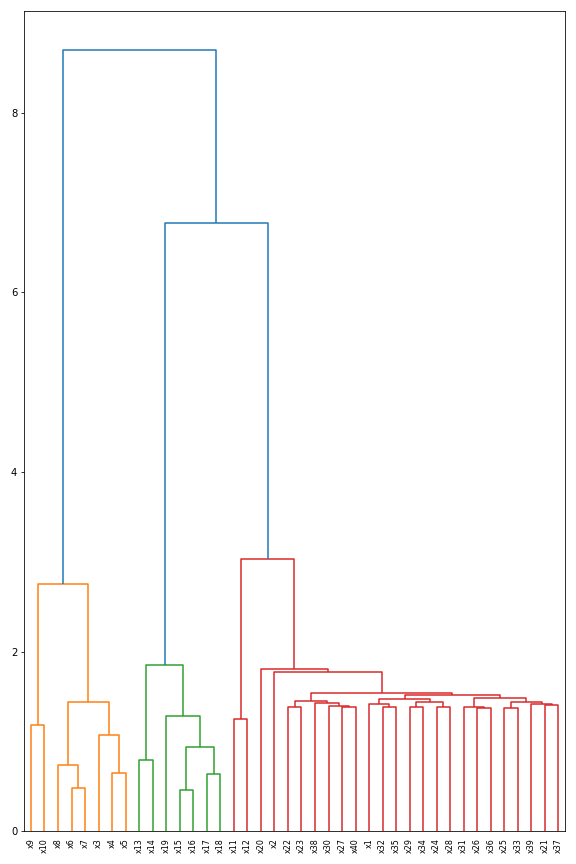}
        \caption{979}
    \end{subfigure}
    \begin{subfigure}[b]{0.24\textwidth}
        \centering
        \includegraphics[width = \linewidth]{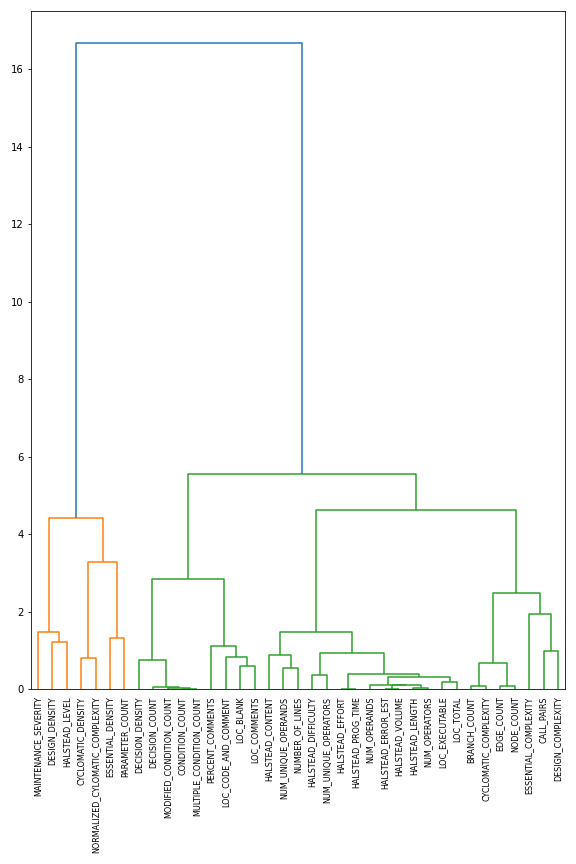}
        \caption{1049}
    \end{subfigure}
    \caption{Dendrograms}
    \label{app:fig:fea_ext_sp_dendrogram1}
\end{figure}
%------------------------------------------------------------------
%------------------------------------------------------------------
\begin{figure}[H]
    \centering
    \begin{subfigure}[b]{0.24\textwidth} 
        \centering
        \includegraphics[width = \linewidth]{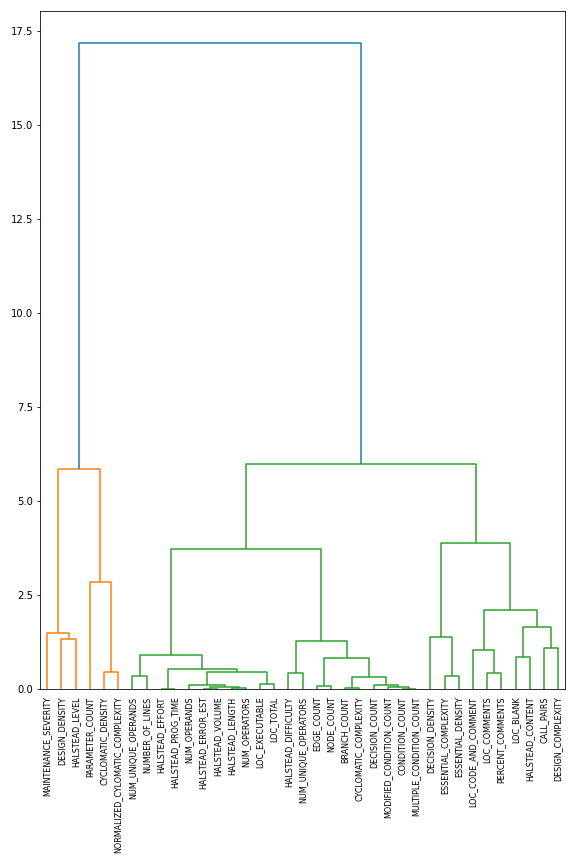}
        \caption{1050}
    \end{subfigure}
    \begin{subfigure}[b]{0.24\textwidth} 
        \centering
        \includegraphics[width = \linewidth]{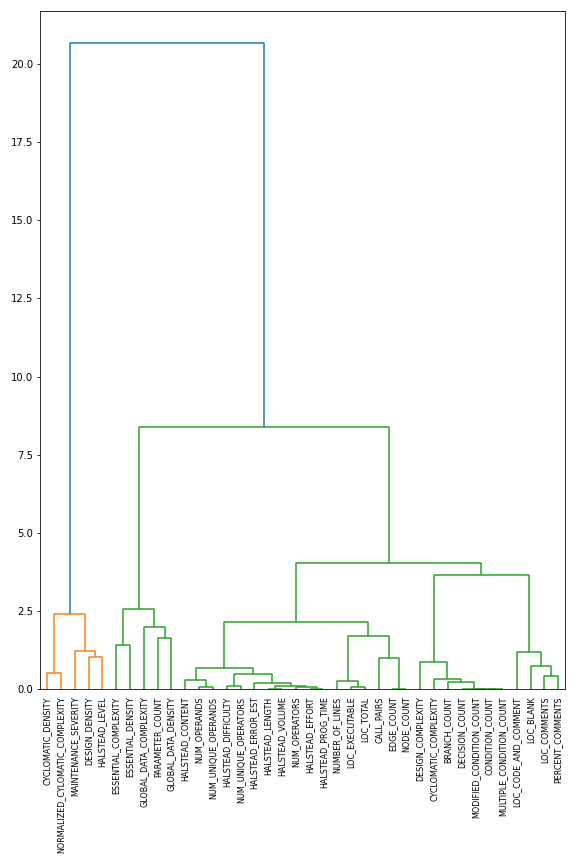}
        \caption{1056}
    \end{subfigure}
    \begin{subfigure}[b]{0.24\textwidth}
        \centering
        \includegraphics[width = \linewidth]{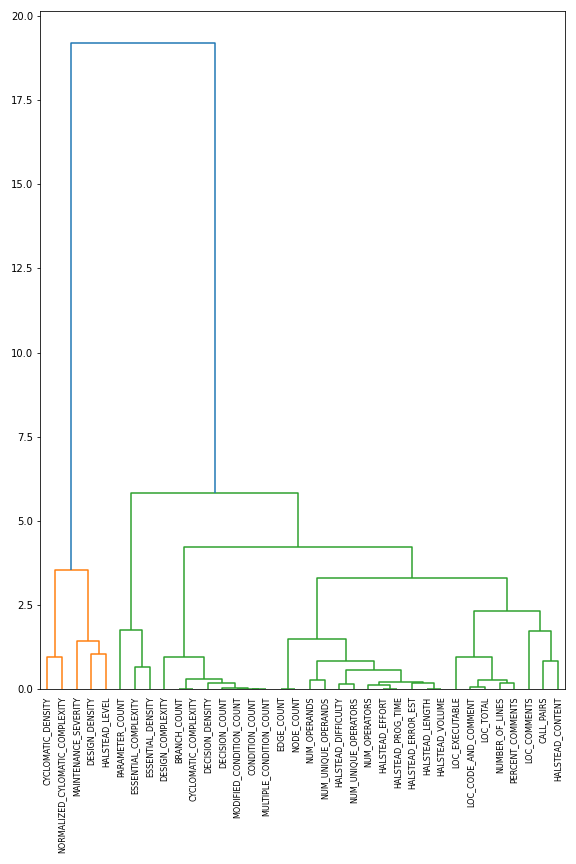}
        \caption{1069}
    \end{subfigure}
    \begin{subfigure}[b]{0.24\textwidth}
        \centering
        \includegraphics[width = \linewidth]{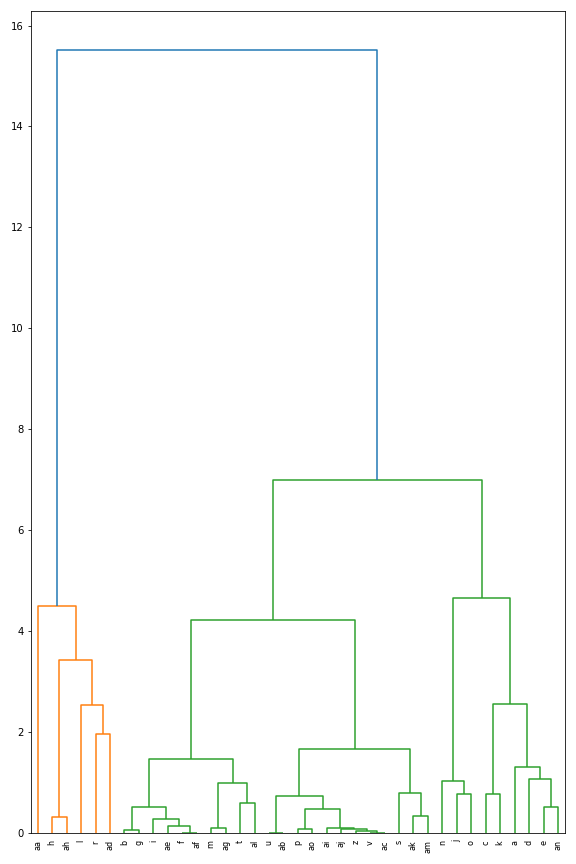}
        \caption{1443}
    \end{subfigure}

    \centering
    \begin{subfigure}[b]{0.24\textwidth} 
        \centering
        \includegraphics[width = \linewidth]{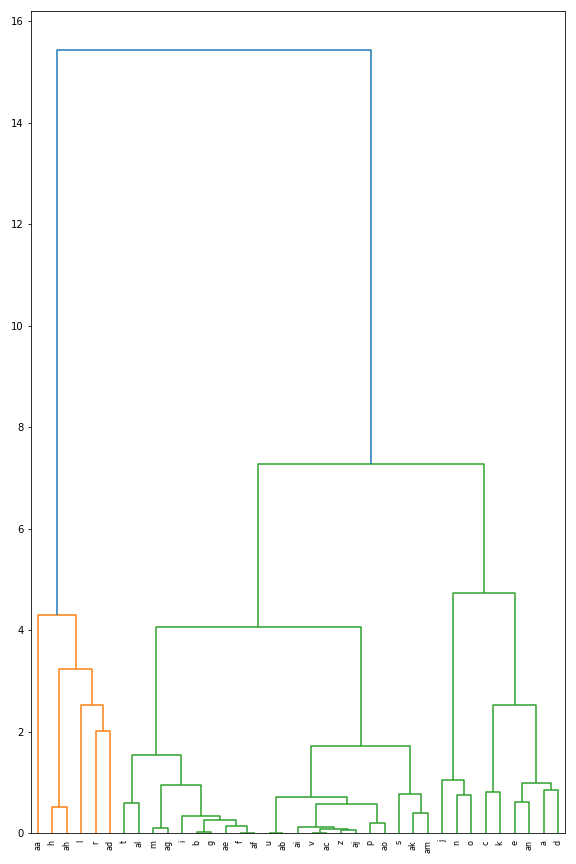}
        \caption{1444}
    \end{subfigure}
    \begin{subfigure}[b]{0.24\textwidth} 
        \centering
        \includegraphics[width = \linewidth]{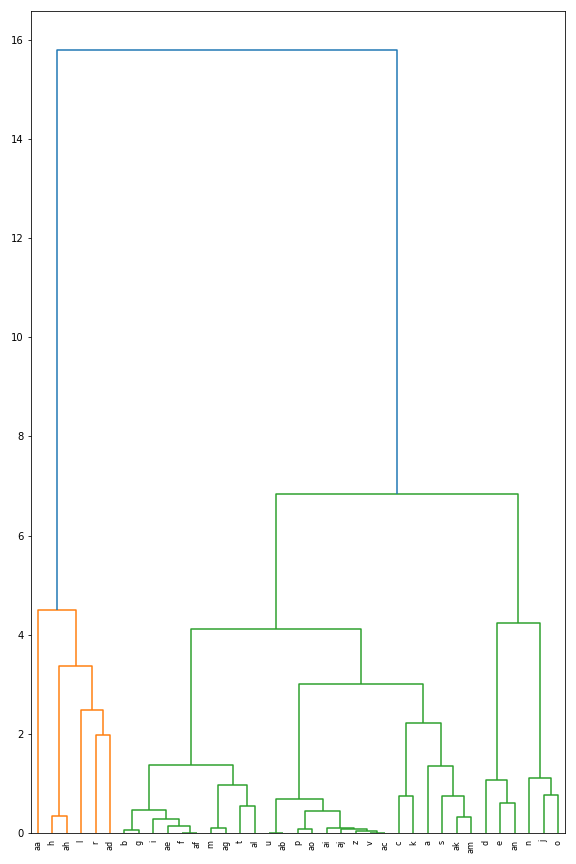}
        \caption{1451}
    \end{subfigure}
    \begin{subfigure}[b]{0.24\textwidth}
        \centering
        \includegraphics[width = \linewidth]{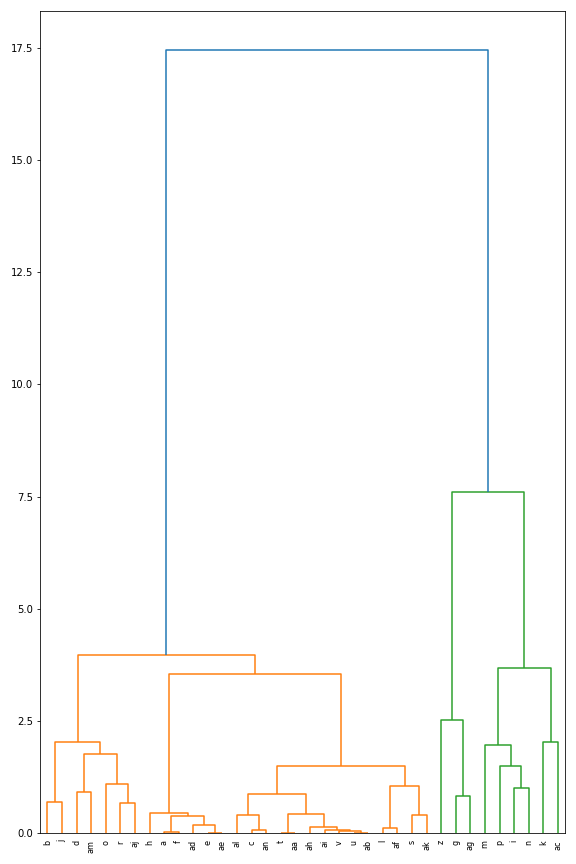}
        \caption{1452}
    \end{subfigure}
    \begin{subfigure}[b]{0.24\textwidth}
        \centering
        \includegraphics[width = \linewidth]{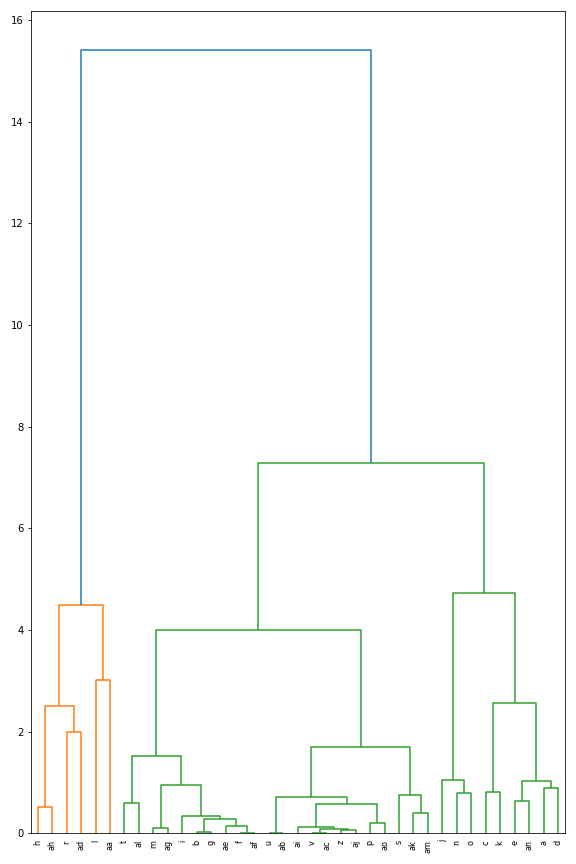}
        \caption{1453}
    \end{subfigure}

    \centering
    \begin{subfigure}[b]{0.24\textwidth} 
        \centering
        \includegraphics[width = \linewidth]{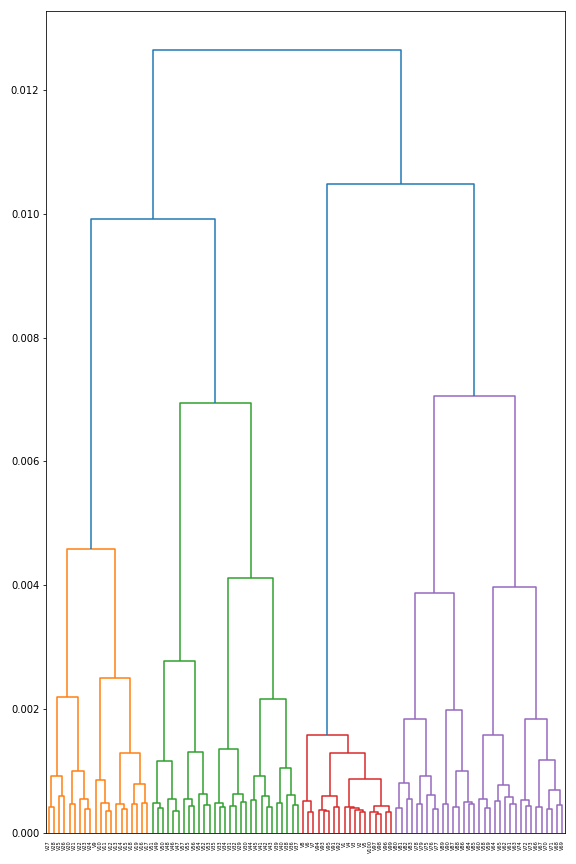}
        \caption{1479}
    \end{subfigure}
    \begin{subfigure}[b]{0.24\textwidth} 
        \centering
        \includegraphics[width = \linewidth]{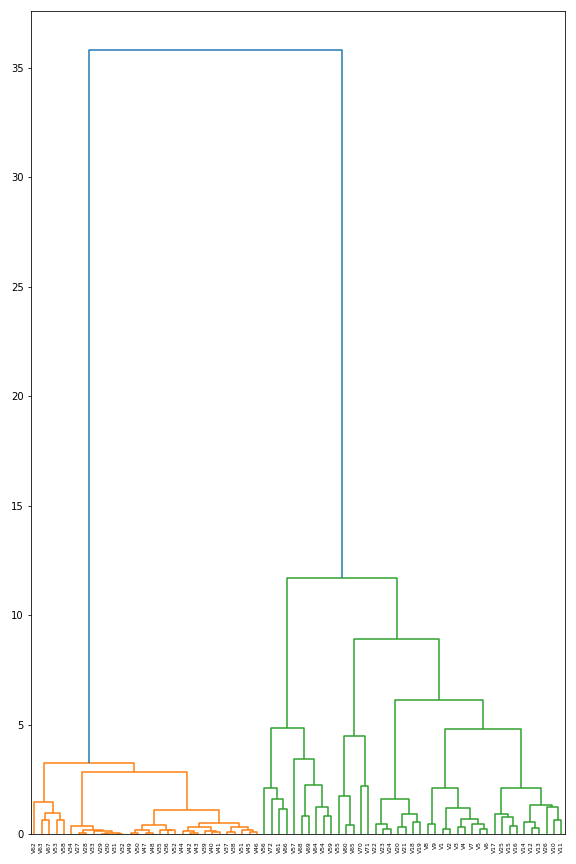}
        \caption{1487}
    \end{subfigure}
    \begin{subfigure}[b]{0.24\textwidth}
        \centering
        \includegraphics[width = \linewidth]{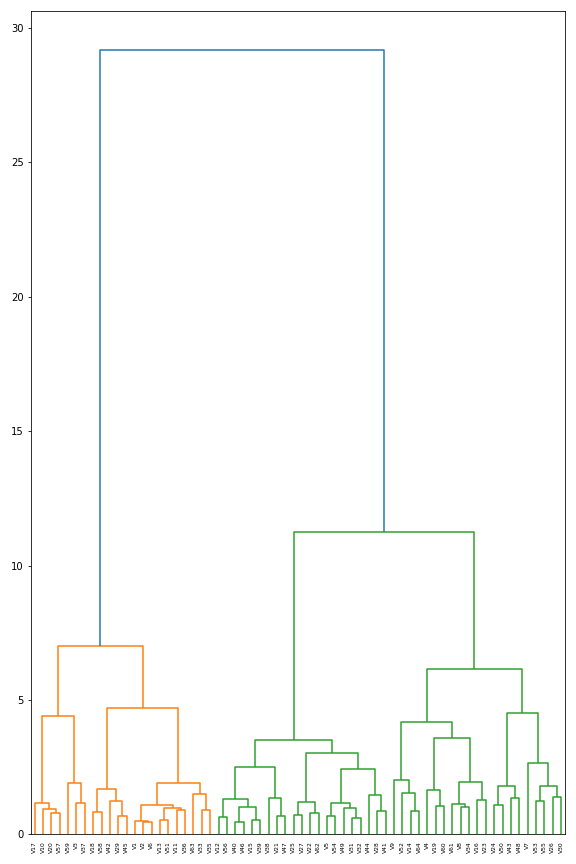}
        \caption{1491}
    \end{subfigure}
    \begin{subfigure}[b]{0.24\textwidth}
        \centering
        \includegraphics[width = \linewidth]{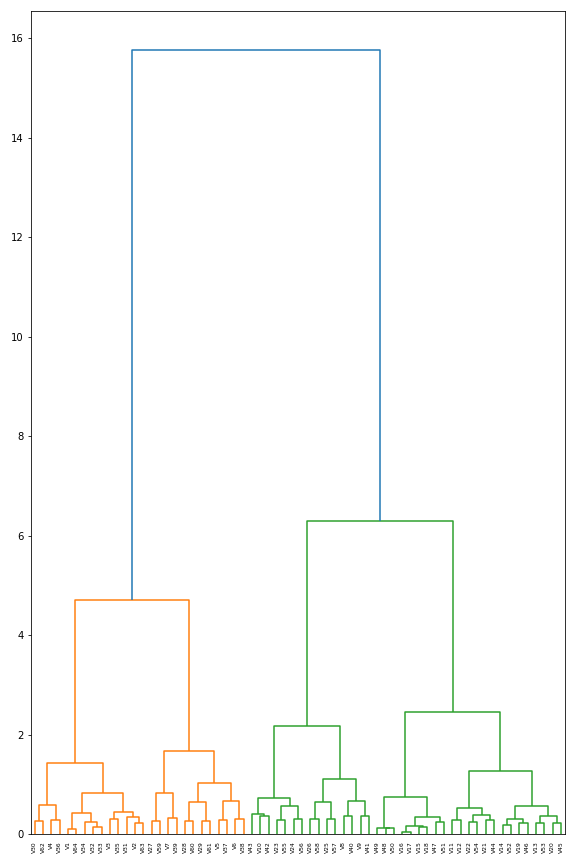}
        \caption{1492}
    \end{subfigure}
    \caption{Dendrograms}
    \label{app:fig:fea_ext_sp_dendrogram2}
\end{figure}
%------------------------------------------------------------------
%------------------------------------------------------------------
\begin{figure}[H]
    \centering
    \begin{subfigure}[b]{0.24\textwidth} 
        \centering
        \includegraphics[width = \linewidth]{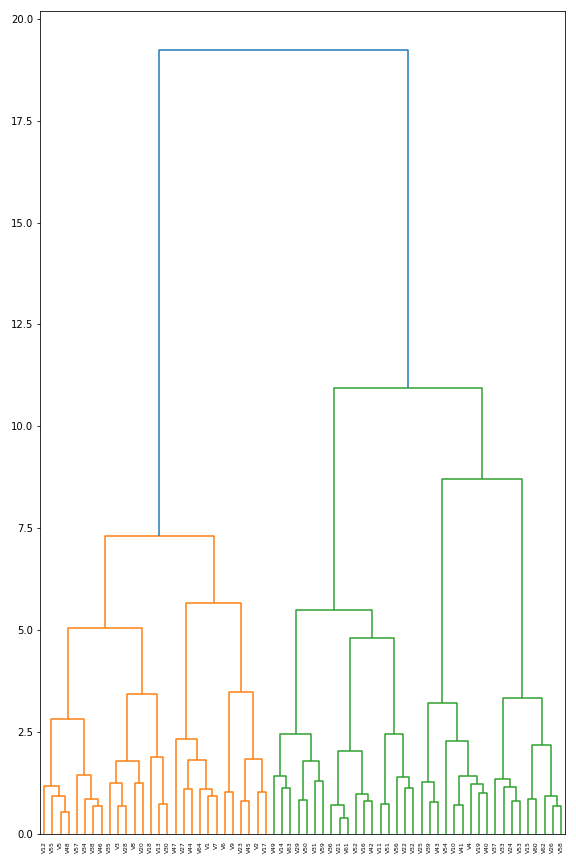}
        \caption{1493}
    \end{subfigure}
    \begin{subfigure}[b]{0.24\textwidth} 
        \centering
        \includegraphics[width = \linewidth]{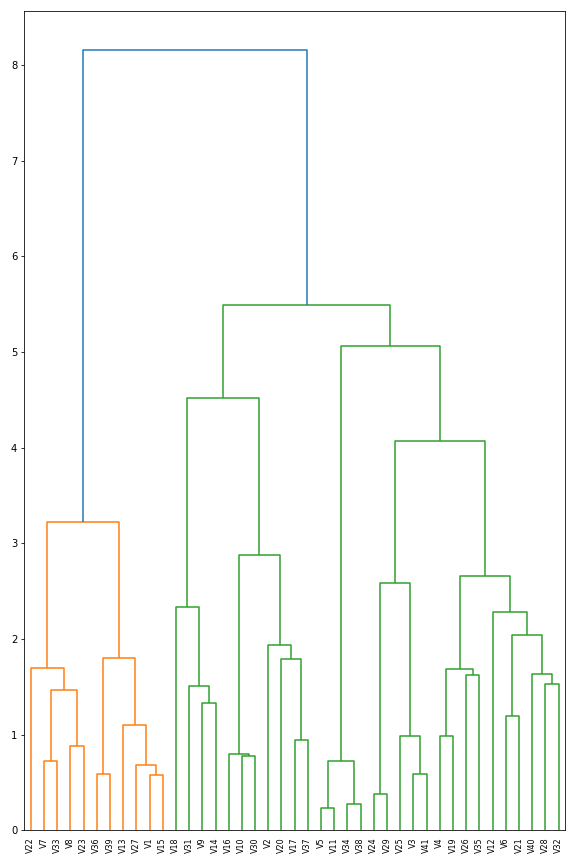}
        \caption{1494}
    \end{subfigure}
    \begin{subfigure}[b]{0.24\textwidth}
        \centering
        \includegraphics[width = \linewidth]{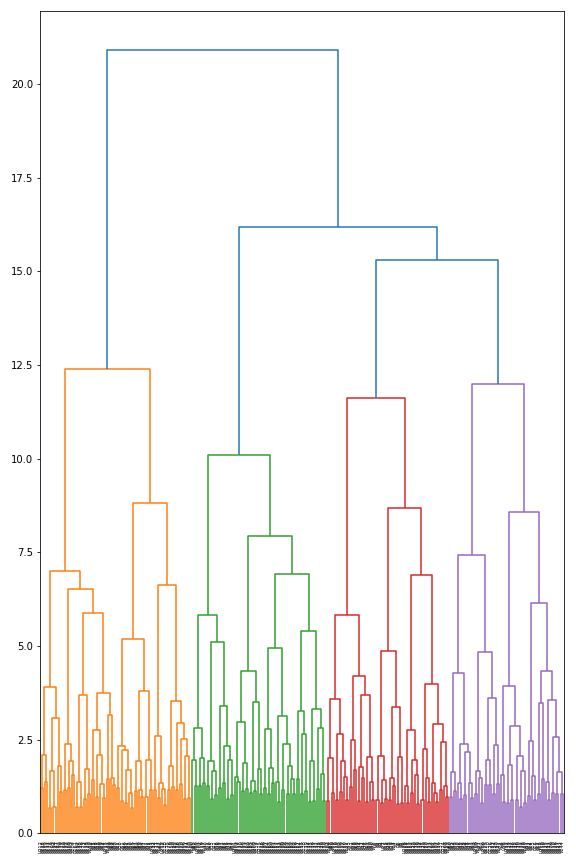}
        \caption{1501}
    \end{subfigure}
    \begin{subfigure}[b]{0.24\textwidth}
        \centering
        \includegraphics[width = \linewidth]{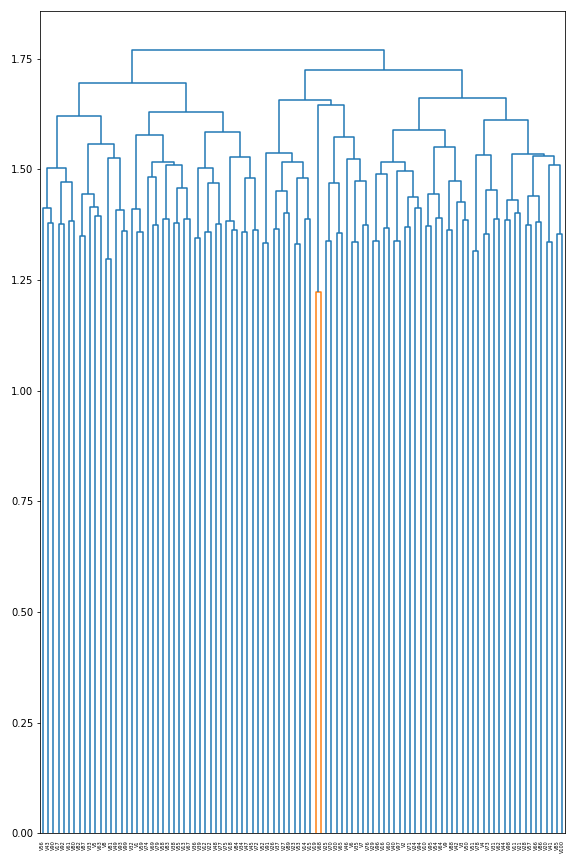}
        \caption{1548}
    \end{subfigure}

    \centering
    \begin{subfigure}[b]{0.24\textwidth} 
        \centering
        \includegraphics[width = \linewidth]{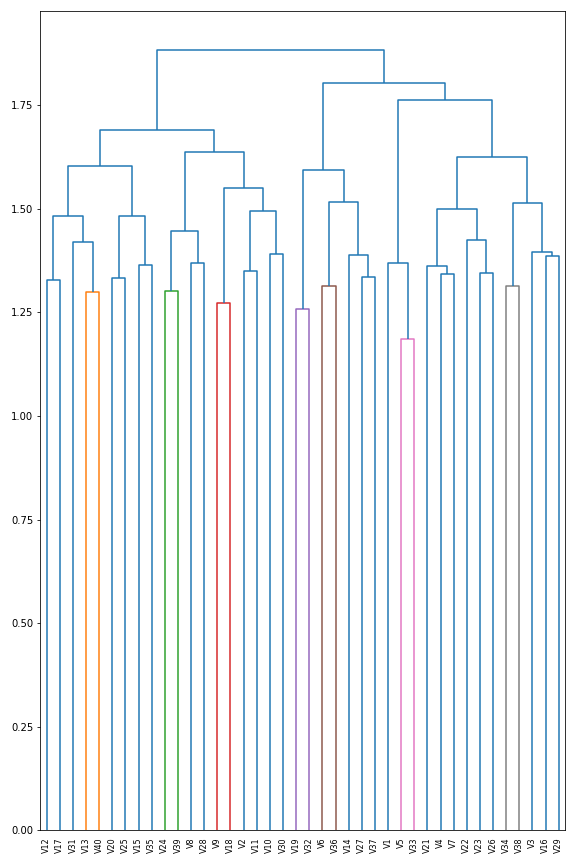}
        \caption{1549}
    \end{subfigure}
    \begin{subfigure}[b]{0.24\textwidth} 
        \centering
        \includegraphics[width = \linewidth]{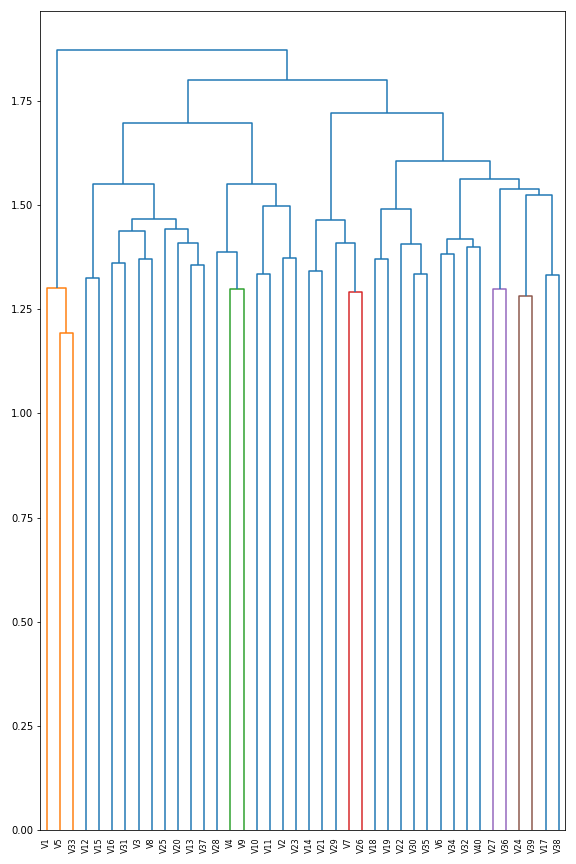}
        \caption{1555}
    \end{subfigure}
    \begin{subfigure}[b]{0.24\textwidth}
        \centering
        \includegraphics[width = \linewidth]{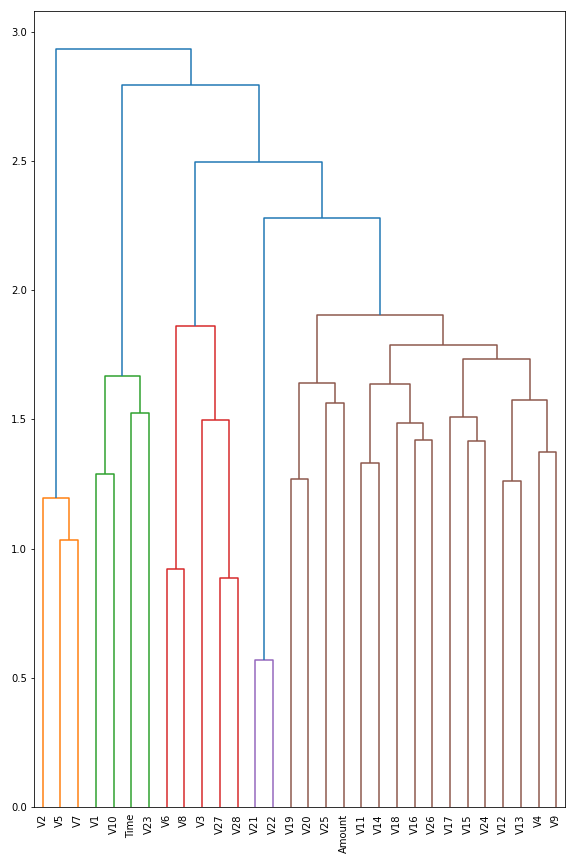}
        \caption{4154}
    \end{subfigure}
    \begin{subfigure}[b]{0.24\textwidth}
        \centering
        \includegraphics[width = \linewidth]{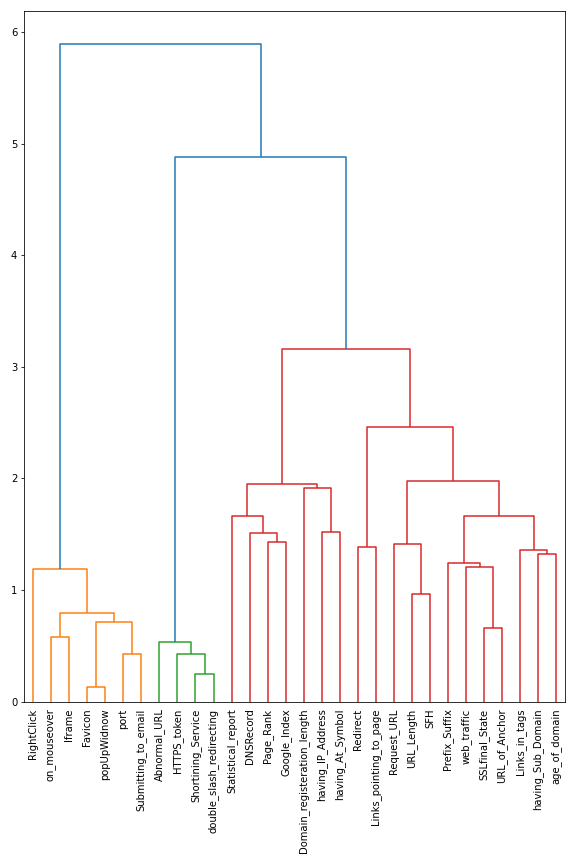}
        \caption{4534}
    \end{subfigure}

    \centering
    \begin{subfigure}[b]{0.24\textwidth} 
        \centering
        \includegraphics[width = \linewidth]{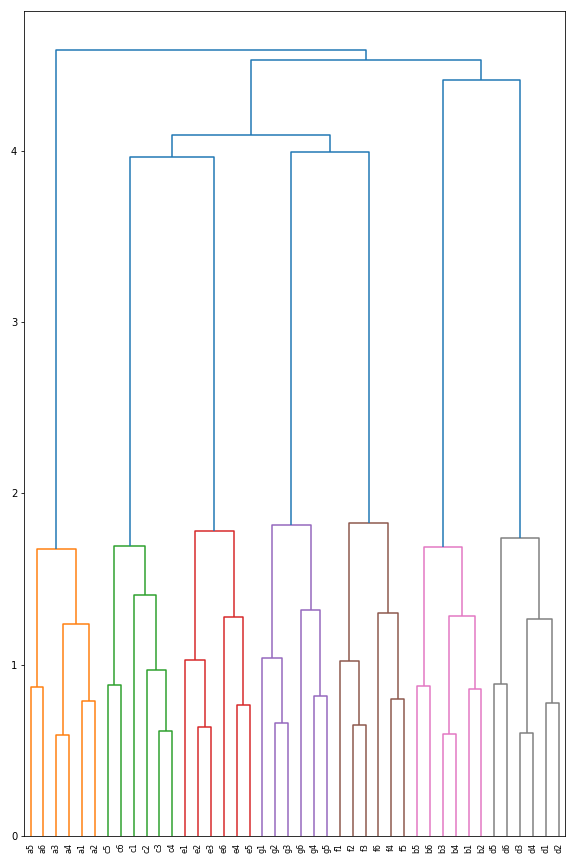}
        \caption{40668}
    \end{subfigure}
    \begin{subfigure}[b]{0.24\textwidth} 
        \centering
        \includegraphics[width = \linewidth]{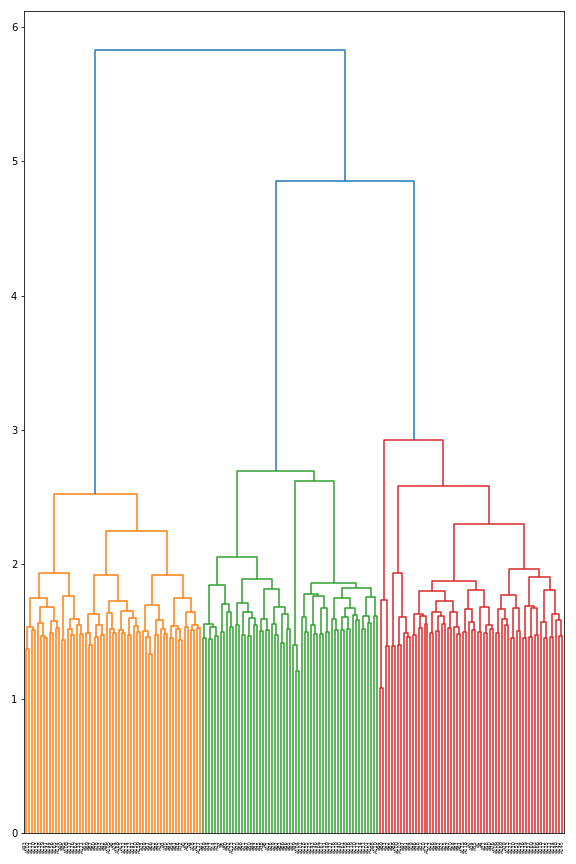}
        \caption{40670}
    \end{subfigure}
    \begin{subfigure}[b]{0.24\textwidth}
        \centering
        \includegraphics[width = \linewidth]{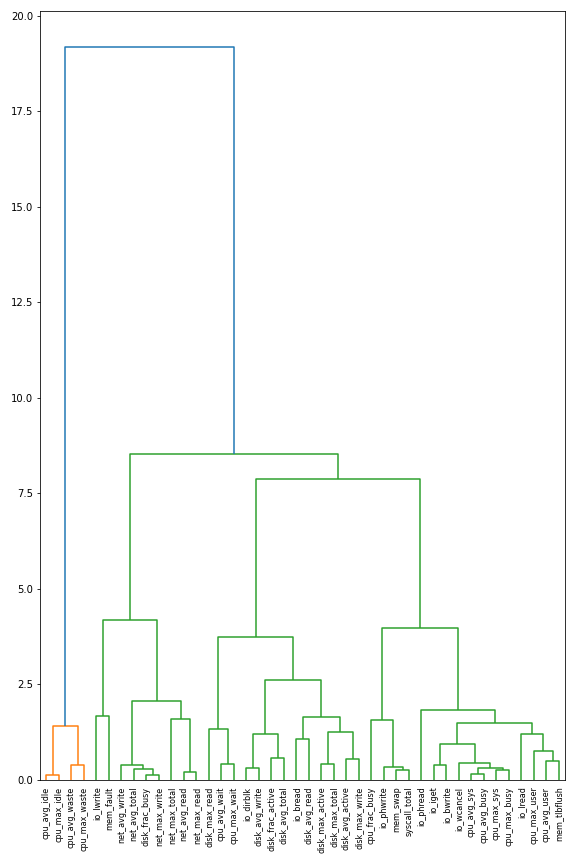}
        \caption{40705}
    \end{subfigure}
    \begin{subfigure}[b]{0.24\textwidth}
        \centering
        \includegraphics[width = \linewidth]{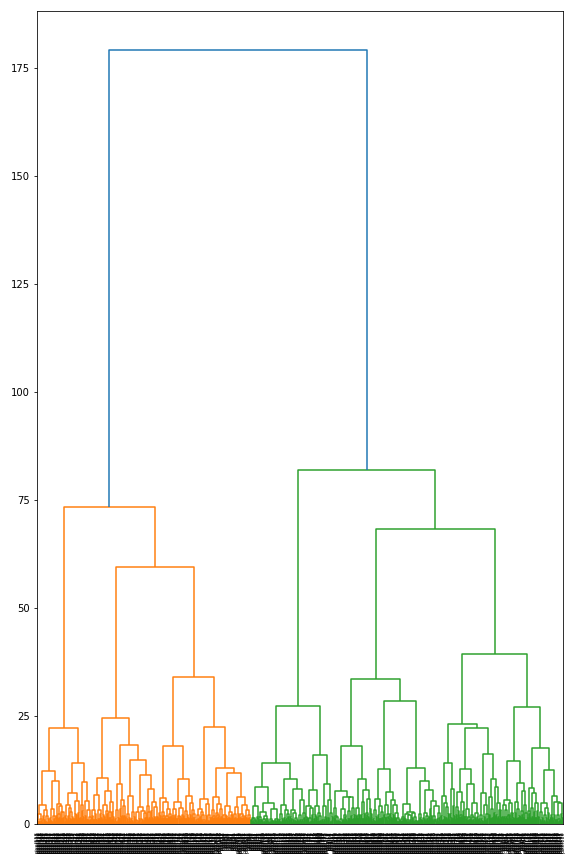}
        \caption{40996}
    \end{subfigure}
    \caption{Dendrograms}
    \label{app:fig:fea_ext_sp_dendrogram3}
\end{figure}
%------------------------------------------------------------------
%------------------------------------------------------------------
\begin{figure}[H]
    \centering
    \begin{subfigure}[b]{0.24\textwidth} 
        \centering
        \includegraphics[width = \linewidth]{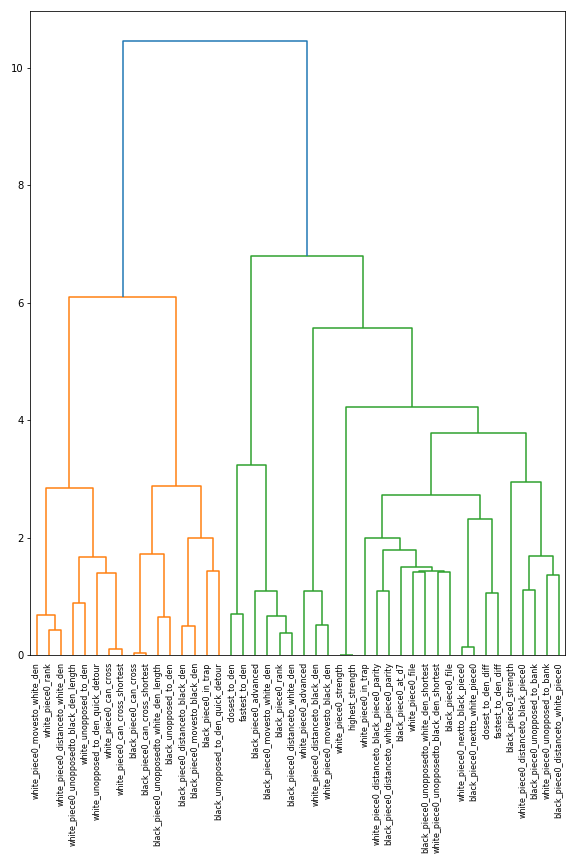}
        \caption{40997}
    \end{subfigure}
    \begin{subfigure}[b]{0.24\textwidth} 
        \centering
        \includegraphics[width = \linewidth]{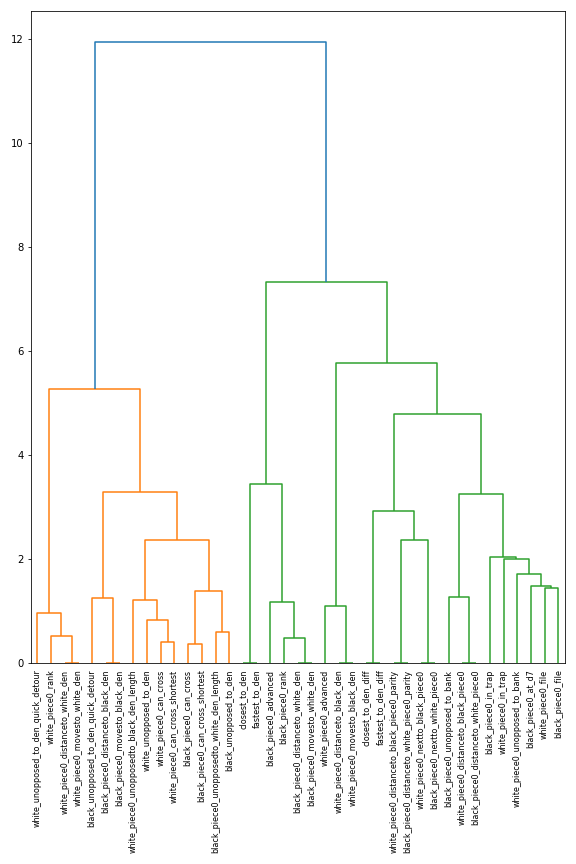}
        \caption{40999}
    \end{subfigure}
    \begin{subfigure}[b]{0.24\textwidth}
        \centering
        \includegraphics[width = \linewidth]{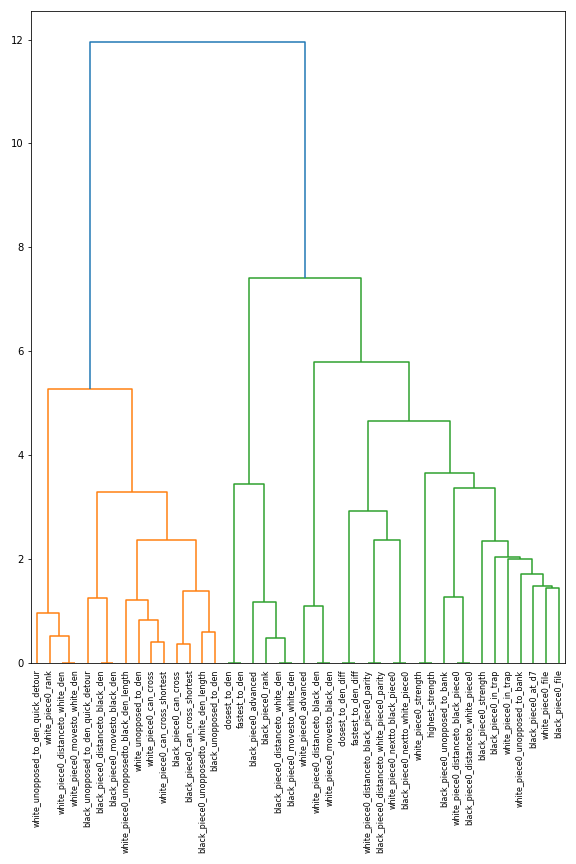}
        \caption{41000}
    \end{subfigure}
    \begin{subfigure}[b]{0.24\textwidth}
        \centering
        \includegraphics[width = \linewidth]{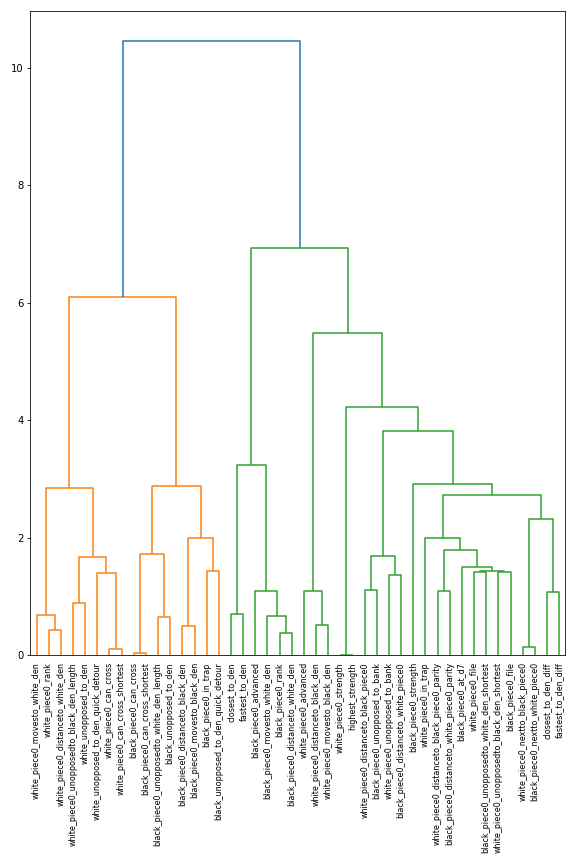}
        \caption{41004}
    \end{subfigure}

    \centering
    \begin{subfigure}[b]{0.24\textwidth} 
        \centering
        \includegraphics[width = \linewidth]{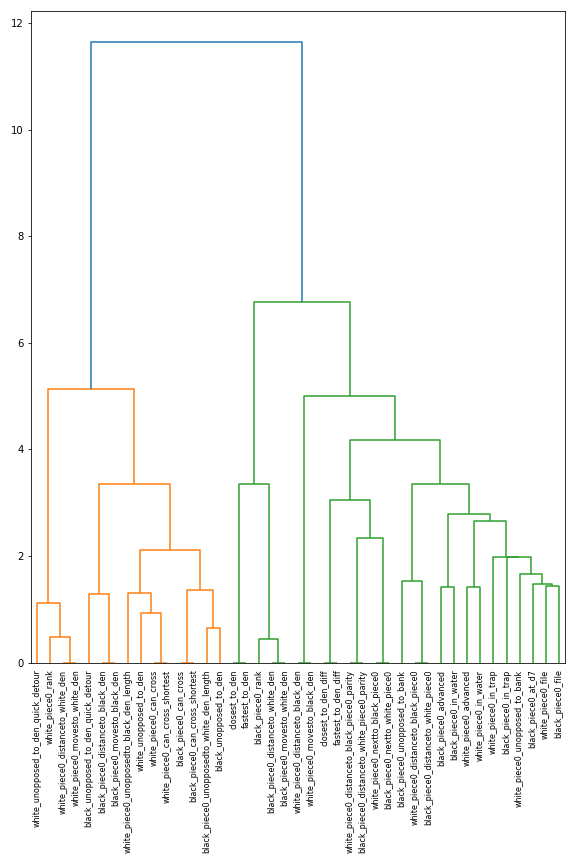}
        \caption{41005}
    \end{subfigure}
    \begin{subfigure}[b]{0.24\textwidth} 
        \centering
        \includegraphics[width = \linewidth]{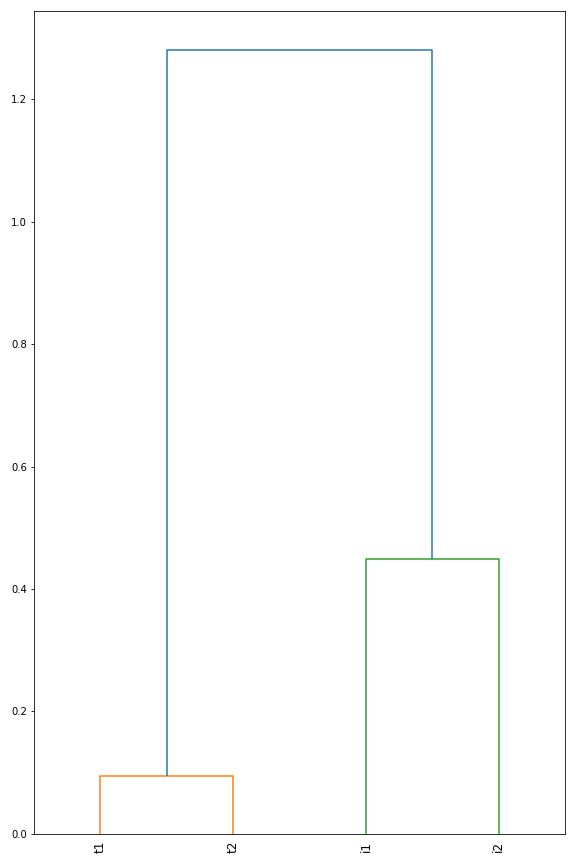}
        \caption{41014}
    \end{subfigure}
    \begin{subfigure}[b]{0.24\textwidth}
        \centering
        \includegraphics[width = \linewidth]{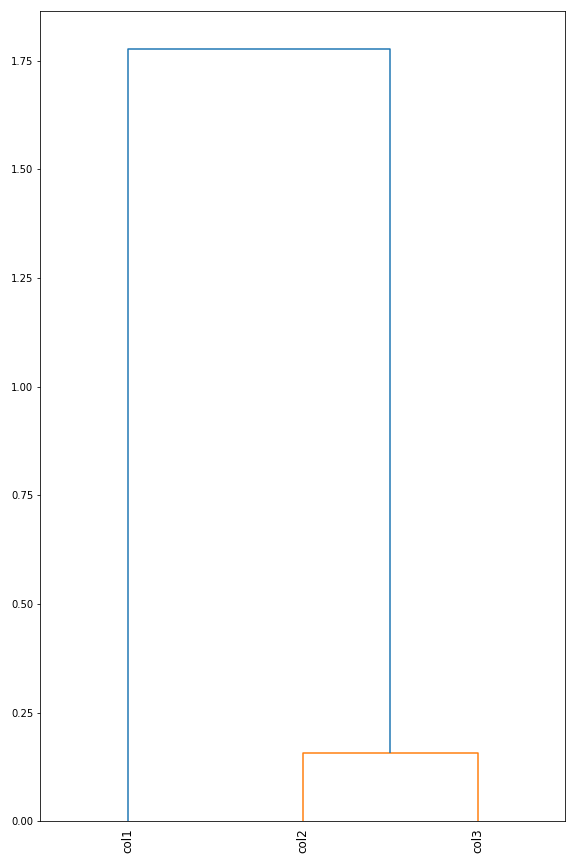}
        \caption{41025}
    \end{subfigure}
    \begin{subfigure}[b]{0.24\textwidth}
        \centering
        \includegraphics[width = \linewidth]{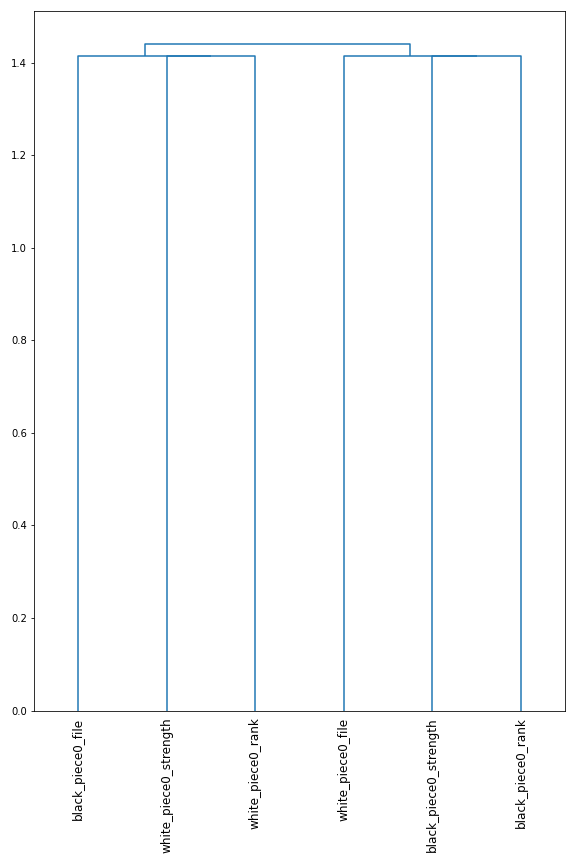}
        \caption{41027}
    \end{subfigure}

    \centering
    \begin{subfigure}[b]{0.24\textwidth} 
        \centering
        \includegraphics[width = \linewidth]{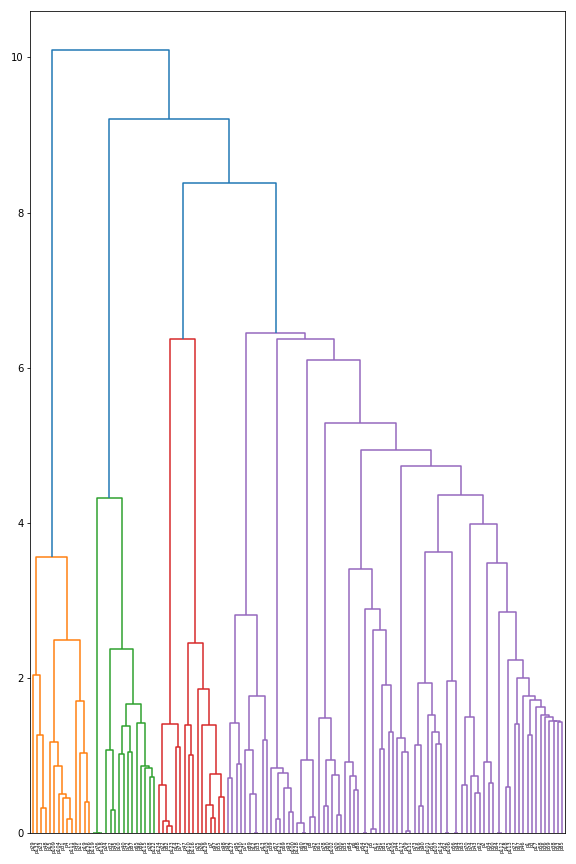}
        \caption{41048}
    \end{subfigure}
    \begin{subfigure}[b]{0.24\textwidth} 
        \centering
        \includegraphics[width = \linewidth]{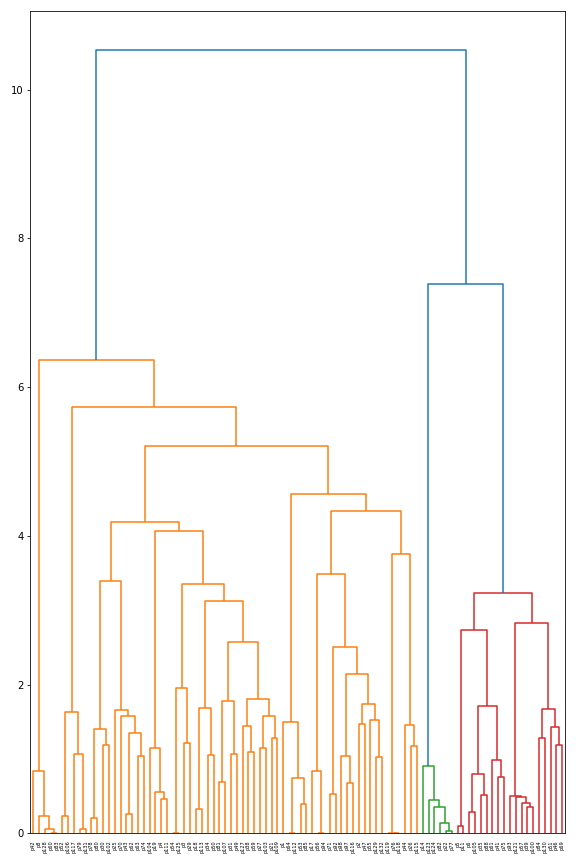}
        \caption{41049}
    \end{subfigure}
    \begin{subfigure}[b]{0.24\textwidth}
        \centering
        \includegraphics[width = \linewidth]{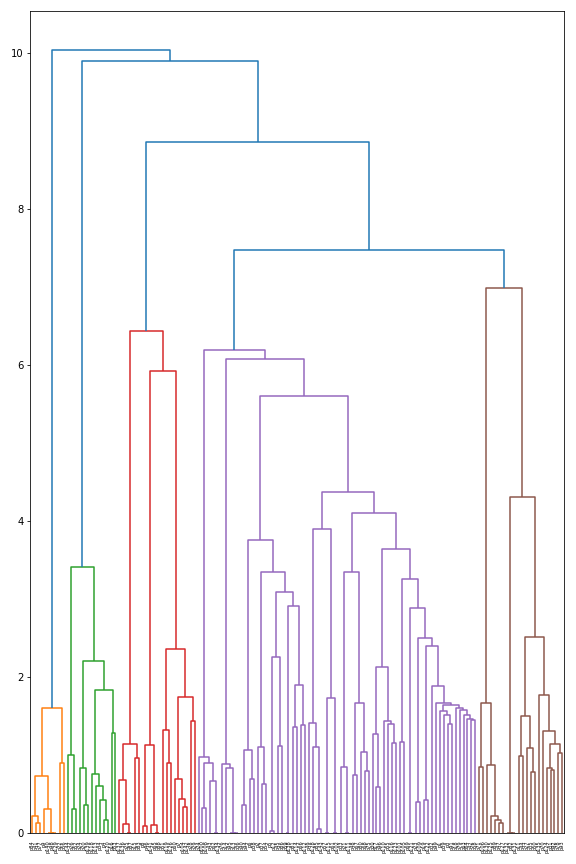}
        \caption{41050}
    \end{subfigure}
    \begin{subfigure}[b]{0.24\textwidth}
        \centering
        \includegraphics[width = \linewidth]{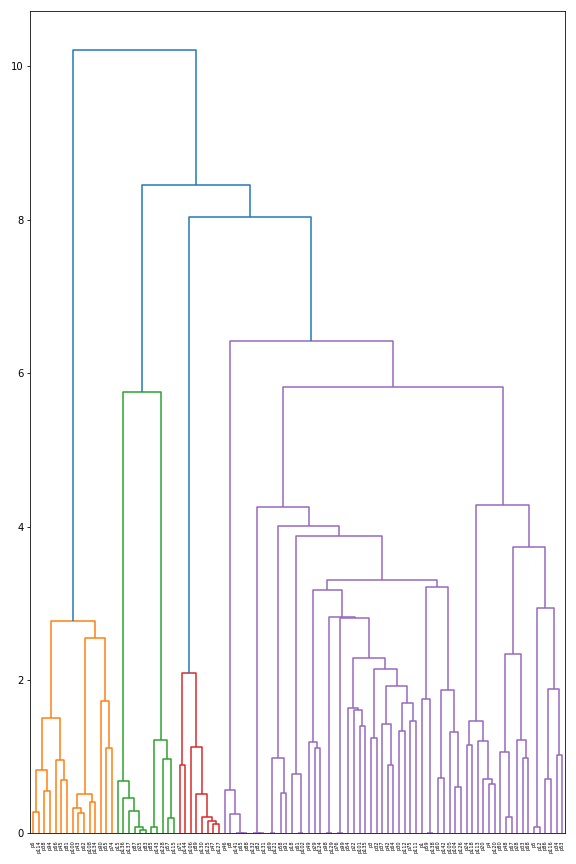}
        \caption{41051}
    \end{subfigure}
    \caption{Dendrograms}
    \label{app:fig:fea_ext_sp_dendrogram4}
\end{figure}
%------------------------------------------------------------------
%------------------------------------------------------------------
\begin{figure}[H]
    \centering
    \begin{subfigure}[b]{0.24\textwidth} 
        \centering
        \includegraphics[width = \linewidth]{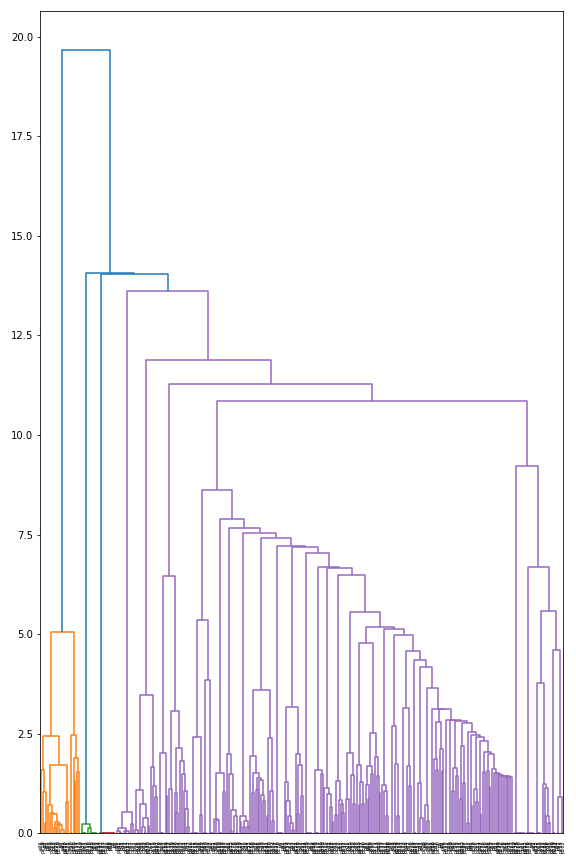}
        \caption{41052}
    \end{subfigure}
    \begin{subfigure}[b]{0.24\textwidth} 
        \centering
        \includegraphics[width = \linewidth]{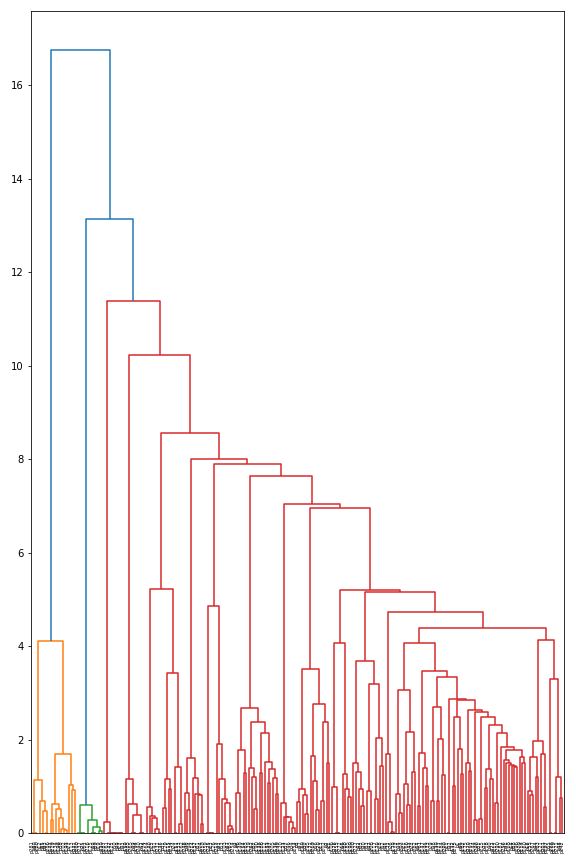}
        \caption{41053}
    \end{subfigure}
    \caption{Dendrograms}
    \label{app:fig:fea_ext_sp_dendrogram5}
\end{figure}

%------------------------------------------------------------------
%------------------------------------------------------------------
\subsection{Feature Extraction}
\begin{figure}[H]
    \centering
    \begin{subfigure}[b]{0.24\textwidth} 
        \centering
        \includegraphics[width = \linewidth]{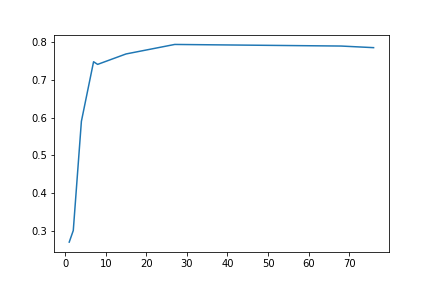}
        \caption{14}
    \end{subfigure}
    \begin{subfigure}[b]{0.24\textwidth} 
        \centering
        \includegraphics[width = \linewidth]{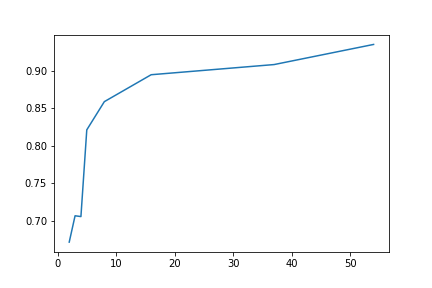}
        \caption{44}
    \end{subfigure}
    \begin{subfigure}[b]{0.24\textwidth} 
        \centering
        \includegraphics[width = \linewidth]{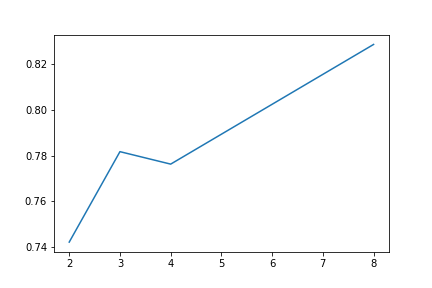}
        \caption{182}
    \end{subfigure}
    \begin{subfigure}[b]{0.24\textwidth}
        \centering
        \includegraphics[width = \linewidth]{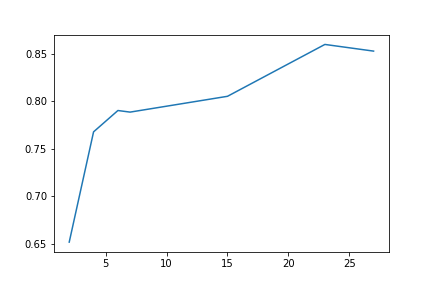}
        \caption{734}
    \end{subfigure}
    
    \centering
    \begin{subfigure}[b]{0.24\textwidth} 
        \centering
        \includegraphics[width = \linewidth]{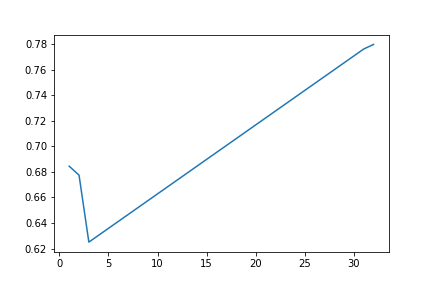}
        \caption{833}
    \end{subfigure}
    \begin{subfigure}[b]{0.24\textwidth}
        \centering
        \includegraphics[width = \linewidth]{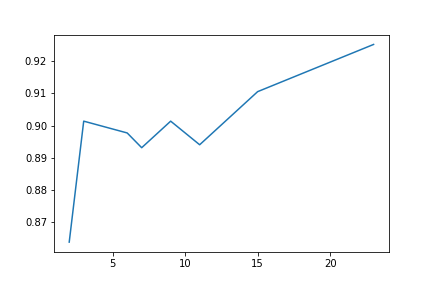}
        \caption{1443}
    \end{subfigure}
    \begin{subfigure}[b]{0.24\textwidth} 
        \centering
        \includegraphics[width = \linewidth]{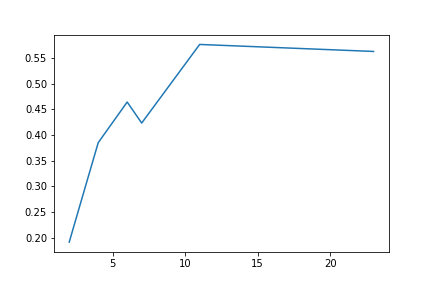}
        \caption{1492}
    \end{subfigure}
    \begin{subfigure}[b]{0.24\textwidth}
        \centering
        \includegraphics[width = \linewidth]{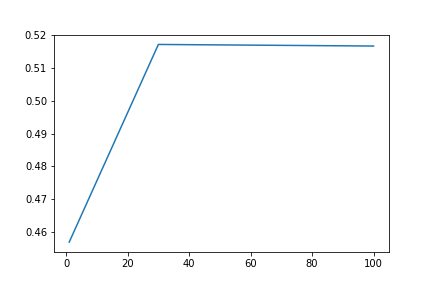}
        \caption{1548}
    \end{subfigure}
    
    \centering
    \begin{subfigure}[b]{0.24\textwidth} 
        \centering
        \includegraphics[width = \linewidth]{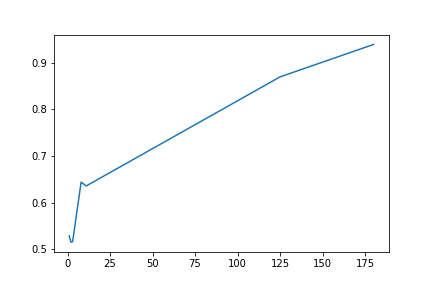}
        \caption{40670}
    \end{subfigure}
    \begin{subfigure}[b]{0.24\textwidth} 
        \centering
        \includegraphics[width = \linewidth]{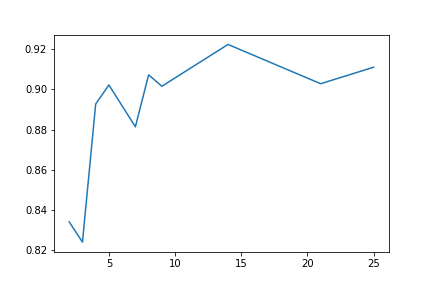}
        \caption{40705}
    \end{subfigure}
    \begin{subfigure}[b]{0.24\textwidth} 
        \centering
        \includegraphics[width = \linewidth]{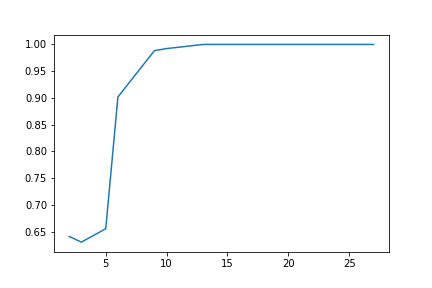}
        \caption{40999}
    \end{subfigure}
    \begin{subfigure}[b]{0.24\textwidth}
        \centering
        \includegraphics[width = \linewidth]{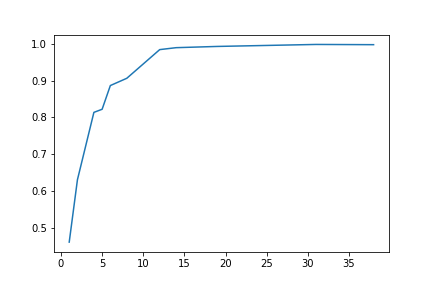}
        \caption{41004}
    \end{subfigure}
    
    \centering
    \begin{subfigure}[b]{0.24\textwidth} 
        \centering
        \includegraphics[width = \linewidth]{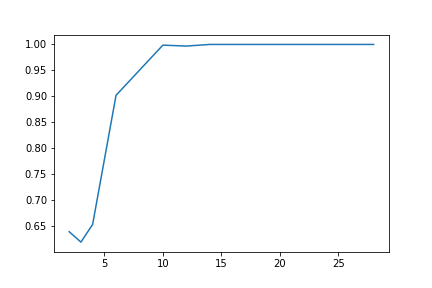}
        \caption{41005}
    \end{subfigure}
    \begin{subfigure}[b]{0.24\textwidth} 
        \centering
        \includegraphics[width = \linewidth]{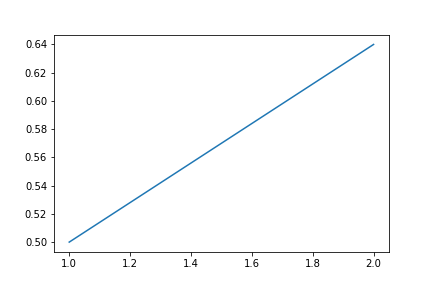}
        \caption{41025}
    \end{subfigure}
    \caption{All increasing results for extractor with Spearman correlation but not highest $\rho$ values}
    \label{app:fig:fea_ext_sp_obs}
\end{figure}
%------------------------------------------------------------------
%------------------------------------------------------------------
\begin{figure}[H]
    \centering
    \begin{subfigure}[b]{0.24\textwidth} 
        \centering
        \includegraphics[width = \linewidth]{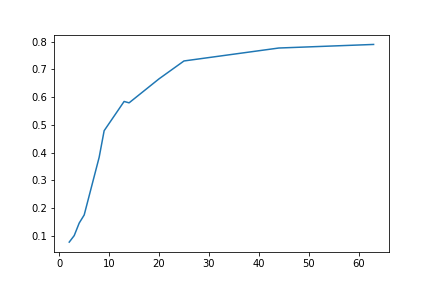}
        \caption{1493}
    \end{subfigure}
    \begin{subfigure}[b]{0.24\textwidth} 
        \centering
        \includegraphics[width = \linewidth]{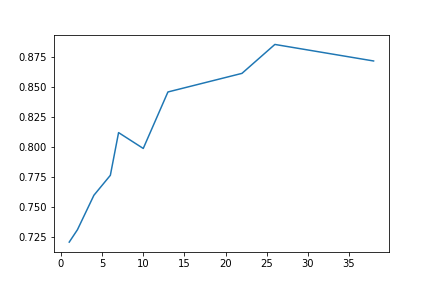}
        \caption{1494}
    \end{subfigure}
    \begin{subfigure}[b]{0.24\textwidth} 
        \centering
        \includegraphics[width = \linewidth]{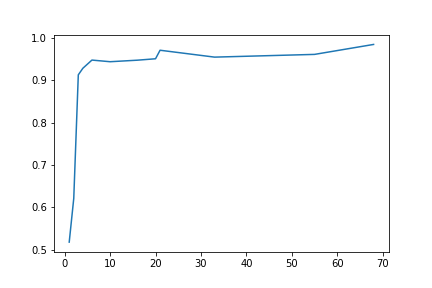}
        \caption{41049}
    \end{subfigure}
    \begin{subfigure}[b]{0.24\textwidth}
        \centering
        \includegraphics[width = \linewidth]{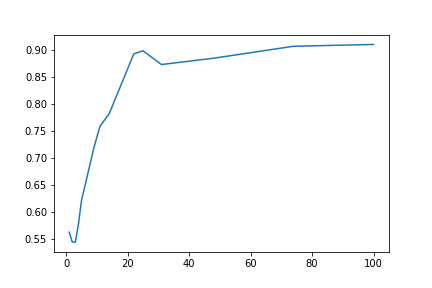}
        \caption{41050}
    \end{subfigure}

    \centering
    \begin{subfigure}[b]{0.24\textwidth} 
        \centering
        \includegraphics[width = \linewidth]{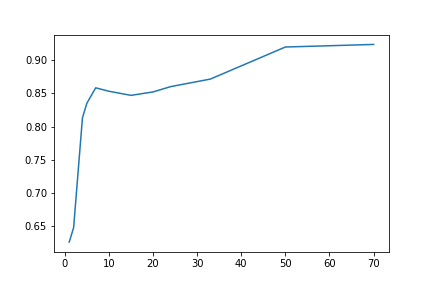}
        \caption{41051}
    \end{subfigure}
    \begin{subfigure}[b]{0.24\textwidth} 
        \centering
        \includegraphics[width = \linewidth]{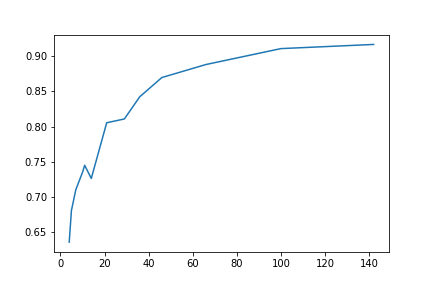}
        \caption{41053}
    \end{subfigure}
    \caption{All increasing results for extractor with Spearman correlation and highest $\rho$ values}
    \label{app:fig:fea_ext_sp_both}
\end{figure}
%------------------------------------------------------------------
%------------------------------------------------------------------
\begin{figure}[H]
    \centering
    \begin{subfigure}[b]{0.24\textwidth} 
        \centering
        \includegraphics[width = \linewidth]{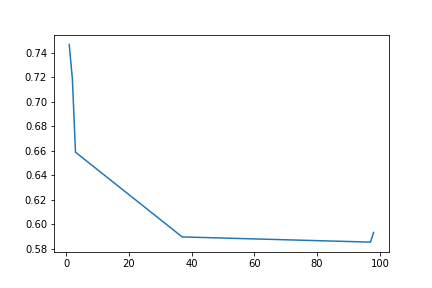}
        \caption{718}
    \end{subfigure}
    \begin{subfigure}[b]{0.24\textwidth} 
        \centering
        \includegraphics[width = \linewidth]{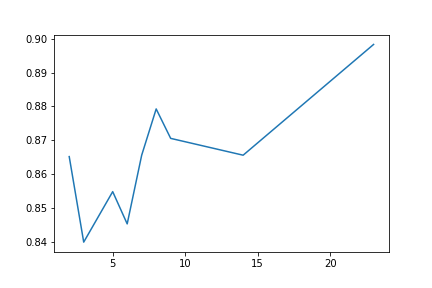}
        \caption{1049}
    \end{subfigure}
    \begin{subfigure}[b]{0.24\textwidth} 
        \centering
        \includegraphics[width = \linewidth]{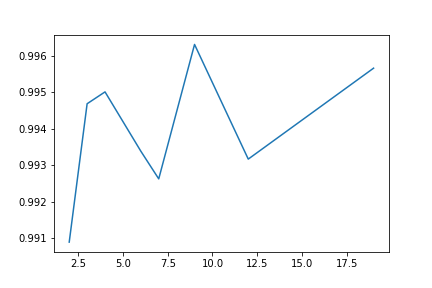}
        \caption{1069}
    \end{subfigure}
    \begin{subfigure}[b]{0.24\textwidth}
        \centering
        \includegraphics[width = \linewidth]{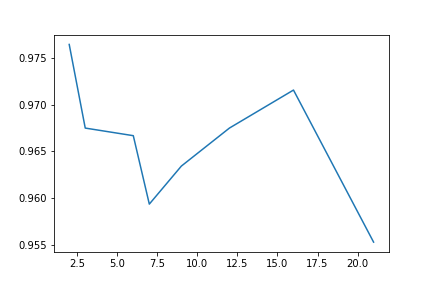}
        \caption{1452}
    \end{subfigure}

    \centering
    \begin{subfigure}[b]{0.24\textwidth} 
        \centering
        \includegraphics[width = \linewidth]{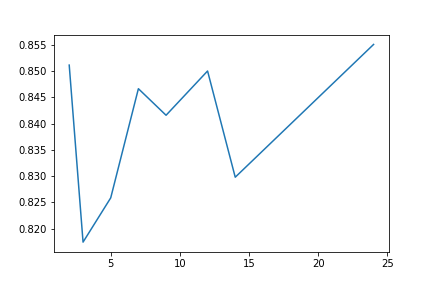}
        \caption{1453}
    \end{subfigure}
    \begin{subfigure}[b]{0.24\textwidth} 
        \centering
        \includegraphics[width = \linewidth]{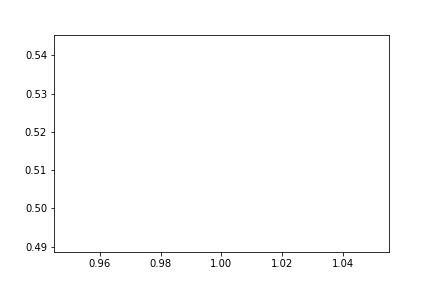}
        \caption{1479}
    \end{subfigure}
    \begin{subfigure}[b]{0.24\textwidth} 
        \centering
        \includegraphics[width = \linewidth]{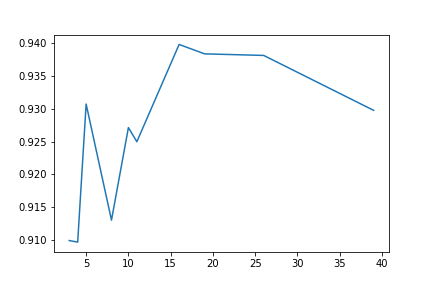}
        \caption{1487}
    \end{subfigure}
    \begin{subfigure}[b]{0.24\textwidth} 
        \centering
        \includegraphics[width = \linewidth]{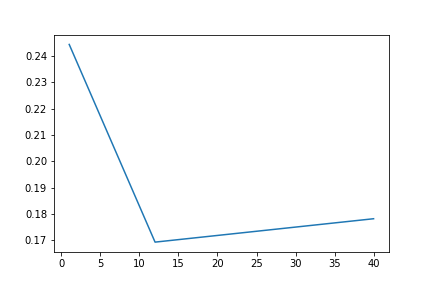}
        \caption{1549}
    \end{subfigure}

    \centering
    \begin{subfigure}[b]{0.24\textwidth} 
        \centering
        \includegraphics[width = \linewidth]{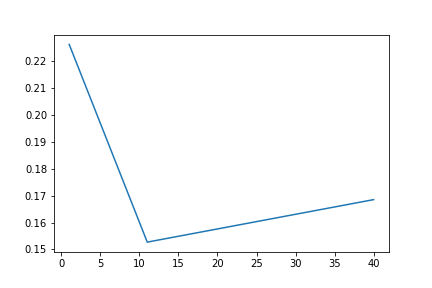}
        \caption{1555}
    \end{subfigure}
    \begin{subfigure}[b]{0.24\textwidth} 
        \centering
        \includegraphics[width = \linewidth]{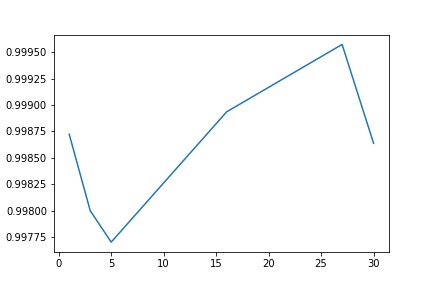}
        \caption{4154}
    \end{subfigure}
    \begin{subfigure}[b]{0.24\textwidth} 
        \centering
        \includegraphics[width = \linewidth]{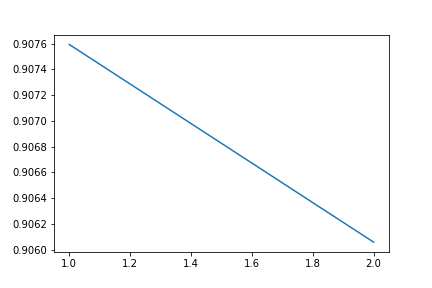}
        \caption{41014}
    \end{subfigure}
    \caption{Non-increasing results for extractor with Spearman correlation}
    \label{app:fig:fea_ext_sp_no}
\end{figure}

\section{Principal Component Analysis}
\label{subapp:pca_fea_ext}
\begin{figure}[H]
    \centering
    \begin{subfigure}[b]{0.24\textwidth} 
        \centering
        \includegraphics[width = \linewidth]{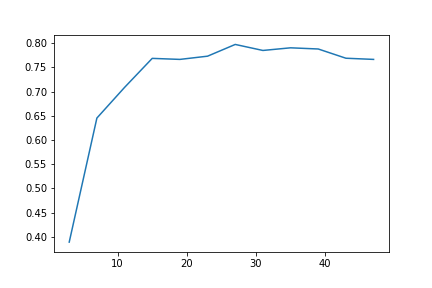}
        \caption{22}
    \end{subfigure}
    \begin{subfigure}[b]{0.24\textwidth} 
        \centering
        \includegraphics[width = \linewidth]{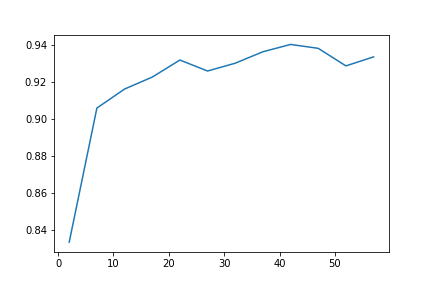}
        \caption{44}
    \end{subfigure}
    \begin{subfigure}[b]{0.24\textwidth}
        \centering
        \includegraphics[width = \linewidth]{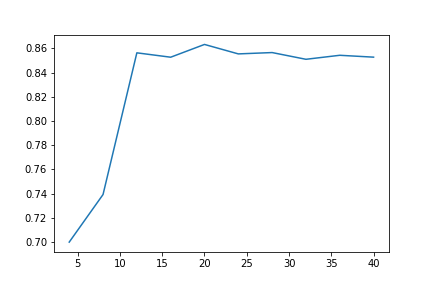}
        \caption{734}
    \end{subfigure}
    \begin{subfigure}[b]{0.24\textwidth} 
        \centering
        \includegraphics[width = \linewidth]{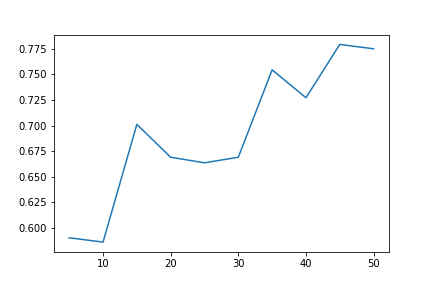}
        \caption{904}
    \end{subfigure}
    
    \centering
    \begin{subfigure}[b]{0.24\textwidth} 
        \centering
        \includegraphics[width = \linewidth]{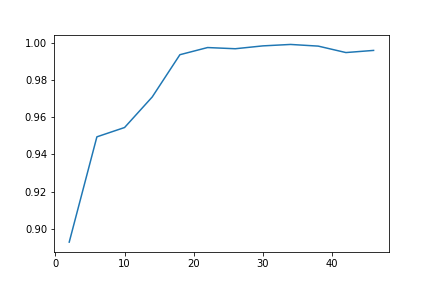}
        \caption{40997}
    \end{subfigure}
    \begin{subfigure}[b]{0.24\textwidth} 
        \centering
        \includegraphics[width = \linewidth]{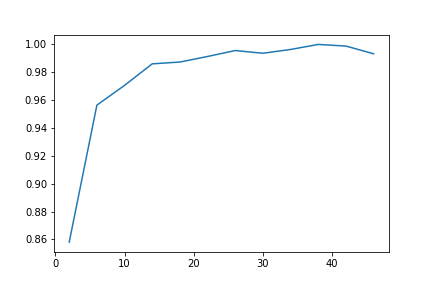}
        \caption{41004}
    \end{subfigure}
    \begin{subfigure}[b]{0.24\textwidth}
        \centering
        \includegraphics[width = \linewidth]{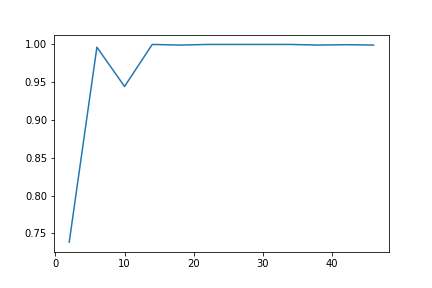}
        \caption{41005}
    \end{subfigure}
    \begin{subfigure}[b]{0.24\textwidth} 
        \centering
        \includegraphics[width = \linewidth]{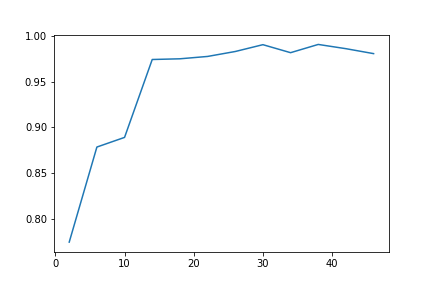}
        \caption{41007}
    \end{subfigure}
    
    \centering
    \begin{subfigure}[b]{0.24\textwidth} 
        \centering
        \includegraphics[width = \linewidth]{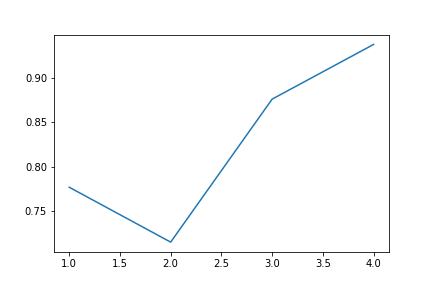}
        \caption{41014}
    \end{subfigure}
    \begin{subfigure}[b]{0.24\textwidth} 
        \centering
        \includegraphics[width = \linewidth]{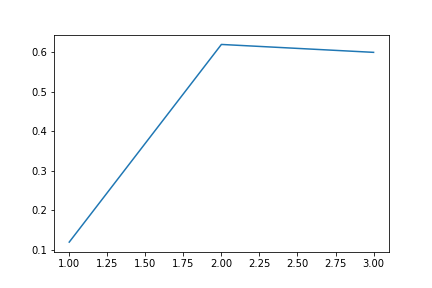}
        \caption{41025}
    \end{subfigure}
    \caption{All increasing results for PCA}
    \label{app:fig:fea_ext_pca_obs}
\end{figure}
%------------------------------------------------------------------
%------------------------------------------------------------------
\begin{figure}[H]
    \centering
    \begin{subfigure}[b]{0.24\textwidth} 
        \centering
        \includegraphics[width = \linewidth]{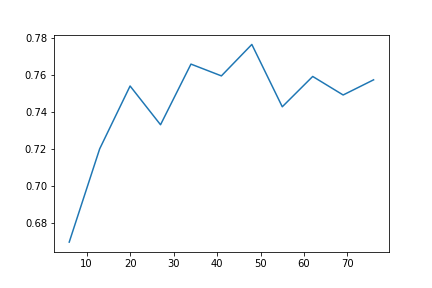}
        \caption{14}
    \end{subfigure}
    \begin{subfigure}[b]{0.24\textwidth} 
        \centering
        \includegraphics[width = \linewidth]{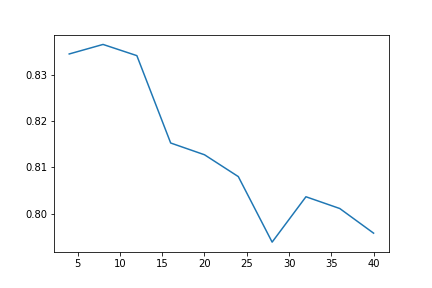}
        \caption{60}
    \end{subfigure}
    \begin{subfigure}[b]{0.24\textwidth}
        \centering
        \includegraphics[width = \linewidth]{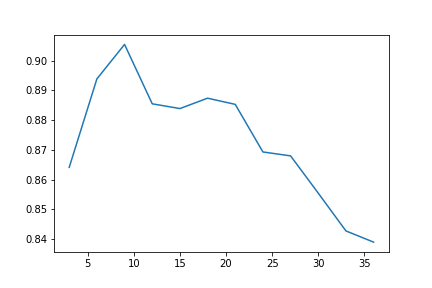}
        \caption{182}
    \end{subfigure}
    \begin{subfigure}[b]{0.24\textwidth} 
        \centering
        \includegraphics[width = \linewidth]{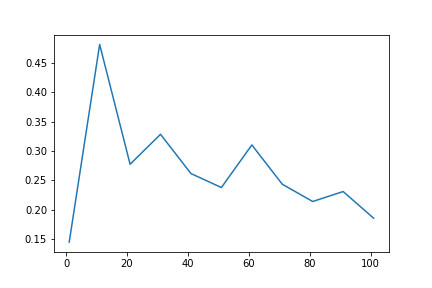}
        \caption{313}
    \end{subfigure}
    
    \centering
    \begin{subfigure}[b]{0.24\textwidth} 
        \centering
        \includegraphics[width = \linewidth]{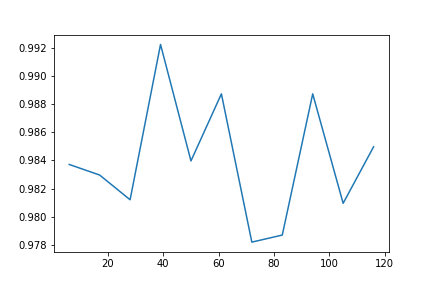}
        \caption{316}
    \end{subfigure}
    \begin{subfigure}[b]{0.24\textwidth} 
        \centering
        \includegraphics[width = \linewidth]{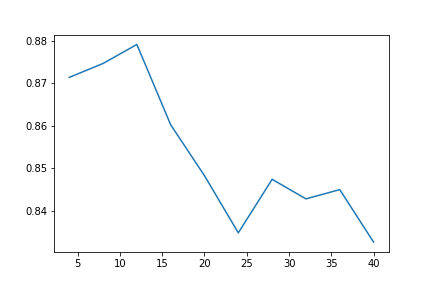}
        \caption{979}
    \end{subfigure}
    \begin{subfigure}[b]{0.24\textwidth}
        \centering
        \includegraphics[width = \linewidth]{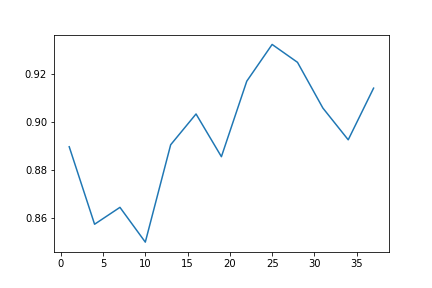}
        \caption{1049}
    \end{subfigure}
    \begin{subfigure}[b]{0.24\textwidth} 
        \centering
        \includegraphics[width = \linewidth]{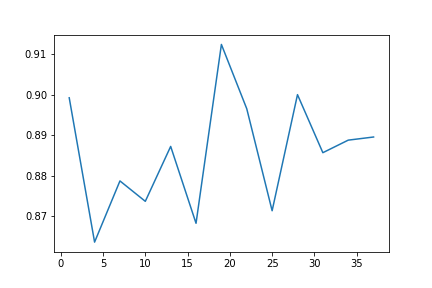}
        \caption{1050}
    \end{subfigure}
    
    \centering
    \begin{subfigure}[b]{0.24\textwidth} 
        \centering
        \includegraphics[width = \linewidth]{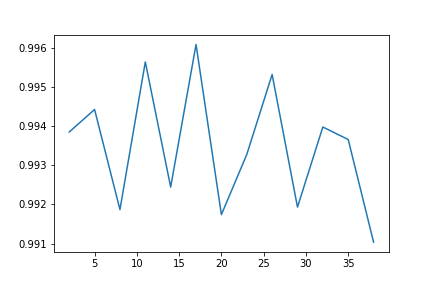}
        \caption{1056}
    \end{subfigure}
    \begin{subfigure}[b]{0.24\textwidth} 
        \centering
        \includegraphics[width = \linewidth]{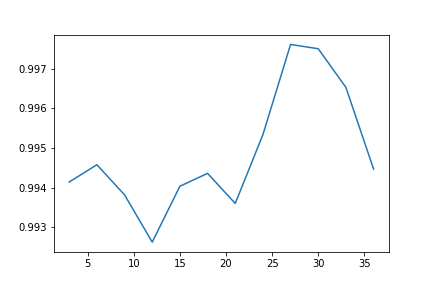}
        \caption{1069}
    \end{subfigure}
    \begin{subfigure}[b]{0.24\textwidth}
        \centering
        \includegraphics[width = \linewidth]{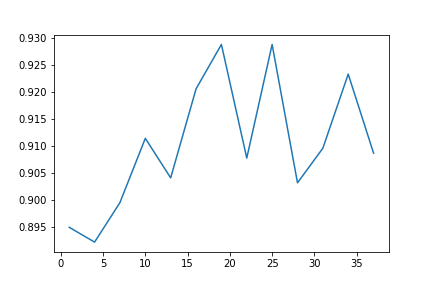}
        \caption{1443}
    \end{subfigure}
    \begin{subfigure}[b]{0.24\textwidth} 
        \centering
        \includegraphics[width = \linewidth]{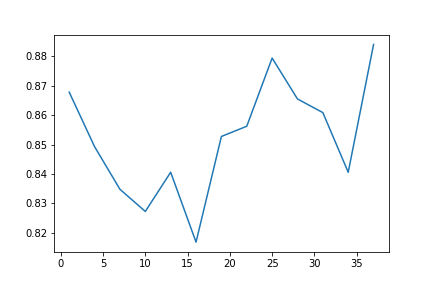}
        \caption{1444}
    \end{subfigure}
    \caption{Non-increasing results for PCA}
    \label{app:fig:fea_ext_pca_no1}
\end{figure}
%------------------------------------------------------------------
%------------------------------------------------------------------
\begin{figure}[H]
    \centering
    \begin{subfigure}[b]{0.24\textwidth} 
        \centering
        \includegraphics[width = \linewidth]{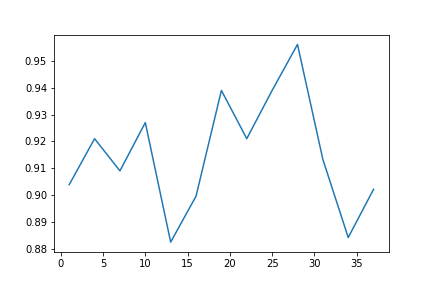}
        \caption{1451}
    \end{subfigure}
    \begin{subfigure}[b]{0.24\textwidth} 
        \centering
        \includegraphics[width = \linewidth]{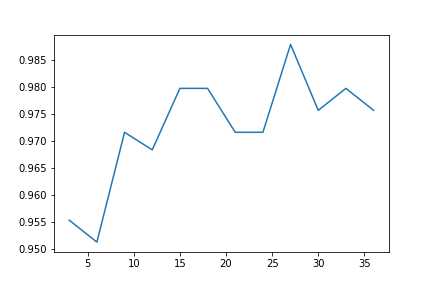}
        \caption{1452}
    \end{subfigure}
    \begin{subfigure}[b]{0.24\textwidth}
        \centering
        \includegraphics[width = \linewidth]{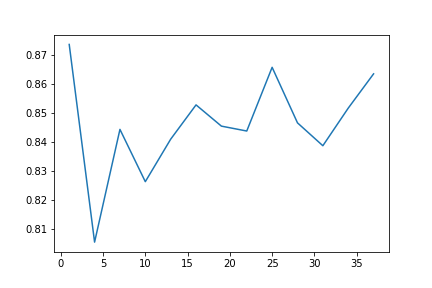}
        \caption{1453}
    \end{subfigure}
    \begin{subfigure}[b]{0.24\textwidth} 
        \centering
        \includegraphics[width = \linewidth]{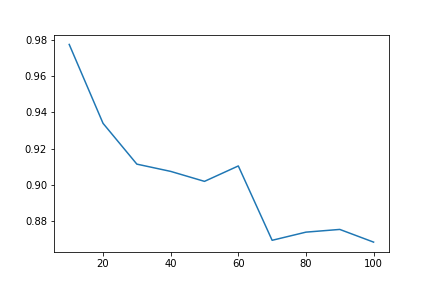}
        \caption{1479}
    \end{subfigure}
    
    \centering
    \begin{subfigure}[b]{0.24\textwidth} 
        \centering
        \includegraphics[width = \linewidth]{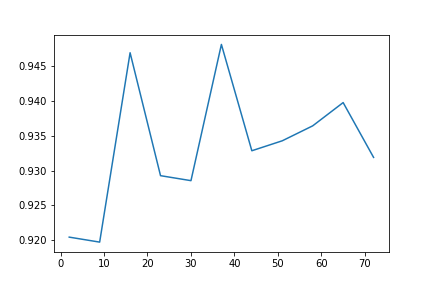}
        \caption{1487}
    \end{subfigure}
    \begin{subfigure}[b]{0.24\textwidth} 
        \centering
        \includegraphics[width = \linewidth]{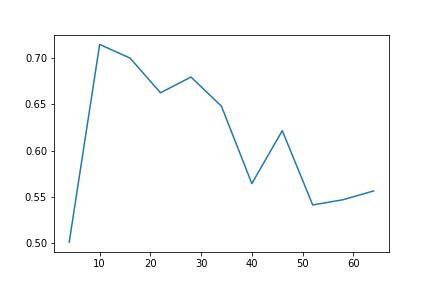}
        \caption{1491}
    \end{subfigure}
    \begin{subfigure}[b]{0.24\textwidth}
        \centering
        \includegraphics[width = \linewidth]{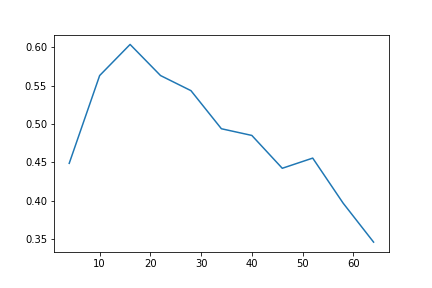}
        \caption{1492}
    \end{subfigure}
    \begin{subfigure}[b]{0.24\textwidth} 
        \centering
        \includegraphics[width = \linewidth]{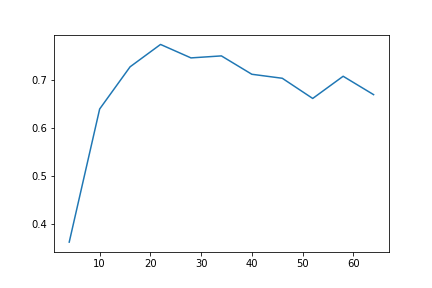}
        \caption{1493}
    \end{subfigure}
    
    \centering
    \begin{subfigure}[b]{0.24\textwidth} 
        \centering
        \includegraphics[width = \linewidth]{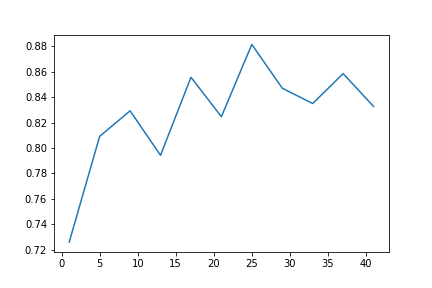}
        \caption{1494}
    \end{subfigure}
    \begin{subfigure}[b]{0.24\textwidth} 
        \centering
        \includegraphics[width = \linewidth]{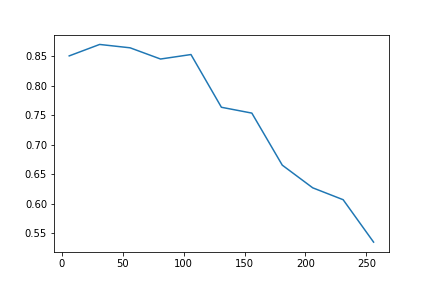}
        \caption{1501}
    \end{subfigure}
    \begin{subfigure}[b]{0.24\textwidth}
        \centering
        \includegraphics[width = \linewidth]{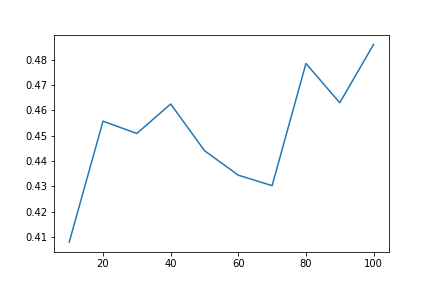}
        \caption{1548}
    \end{subfigure}
    \begin{subfigure}[b]{0.24\textwidth} 
        \centering
        \includegraphics[width = \linewidth]{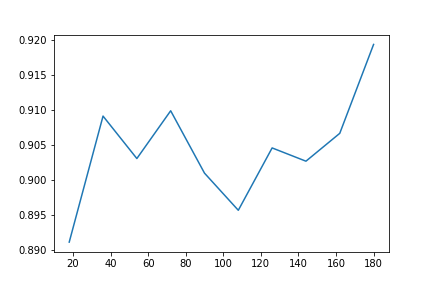}
        \caption{40670}
    \end{subfigure}
    
    \centering
    \begin{subfigure}[b]{0.24\textwidth} 
        \centering
        \includegraphics[width = \linewidth]{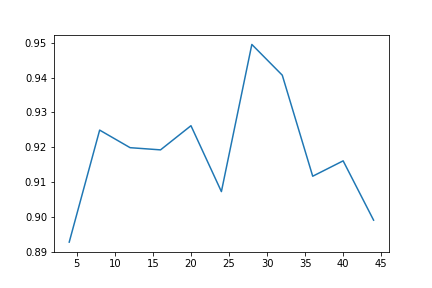}
        \caption{40705}
    \end{subfigure}
    \begin{subfigure}[b]{0.24\textwidth} 
        \centering
        \includegraphics[width = \linewidth]{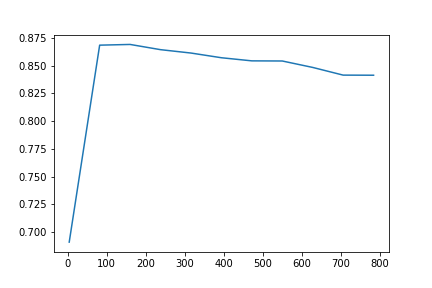}
        \caption{40996}
    \end{subfigure}
    \begin{subfigure}[b]{0.24\textwidth}
        \centering
        \includegraphics[width = \linewidth]{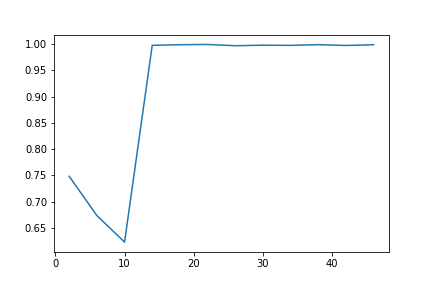}
        \caption{41000}
    \end{subfigure}
    \begin{subfigure}[b]{0.24\textwidth} 
        \centering
        \includegraphics[width = \linewidth]{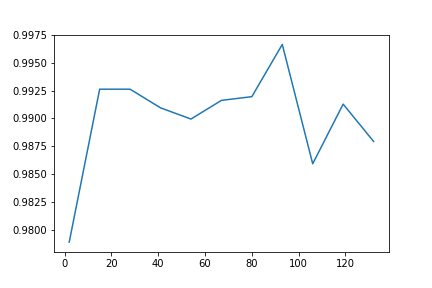}
        \caption{41048}
    \end{subfigure}
    
    \centering
    \begin{subfigure}[b]{0.24\textwidth} 
        \centering
        \includegraphics[width = \linewidth]{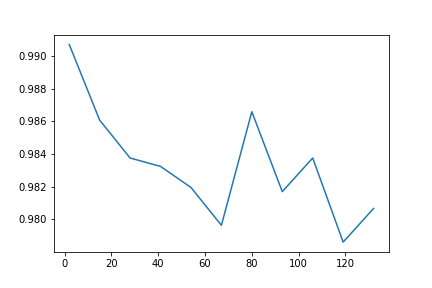}
        \caption{41049}
    \end{subfigure}
    \begin{subfigure}[b]{0.24\textwidth} 
        \centering
        \includegraphics[width = \linewidth]{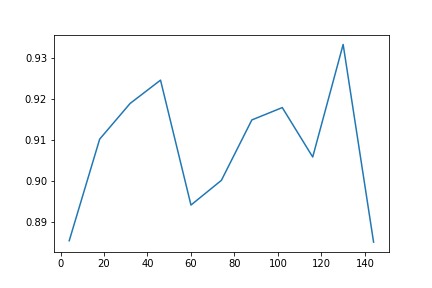}
        \caption{41050}
    \end{subfigure}
    \begin{subfigure}[b]{0.24\textwidth}
        \centering
        \includegraphics[width = \linewidth]{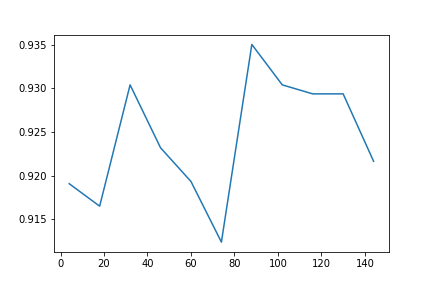}
        \caption{41051}
    \end{subfigure}
    \begin{subfigure}[b]{0.24\textwidth} 
        \centering
        \includegraphics[width = \linewidth]{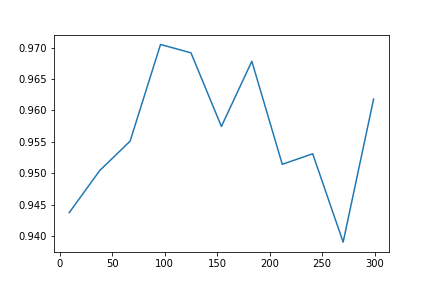}
        \caption{41052}
    \end{subfigure}
    
    \centering
    \begin{subfigure}[b]{0.24\textwidth} 
        \centering
        \includegraphics[width = \linewidth]{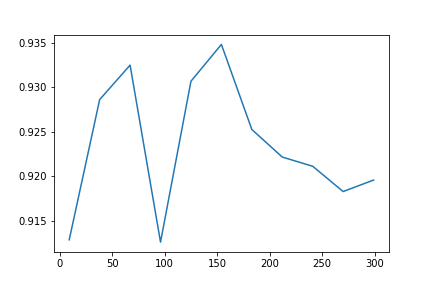}
        \caption{41053}
    \end{subfigure}
    \caption{Non-increasing results for PCA}
    \label{app:fig:fea_ext_pca_no2}
\end{figure}

\end{document}